%% file: main.tex
\documentclass[anonymous, journal]{IEEEtran}

\usepackage{caption}
\usepackage{url}
\usepackage[hidelinks]{hyperref}
\hypersetup{
    colorlinks=true,
    linkcolor=blue,
    citecolor=blue,      
    urlcolor=blue,
}
\usepackage{amsmath}
\usepackage{tikz}
\usetikzlibrary{shapes.geometric, arrows, positioning}
\usepackage{graphicx}
\usepackage{algorithm}
\usepackage{algpseudocode}
\usepackage{url}

\usepackage{lscape}
\usepackage{rotating}
\usepackage{colortbl}
\usepackage[table,xcdraw]{xcolor}
\usepackage{graphicx}
\usepackage{algorithm}
\usepackage{algpseudocode}
\usepackage{stfloats}
\usepackage{enumitem}
\usepackage{multirow}
\usepackage{float}
\usepackage{subcaption}
\usepackage{comment}
\usepackage{orcidlink}
\usepackage[numbers,sort&compress]{natbib}
\def\BibTeX{{\rm B\kern-.05em{\sc i\kern-.025em b}\kern-.08em
    T\kern-.1667em\lower.7ex\hbox{E}\kern-.125emX}}
    
\usepackage[section]{placeins}

\begin{document}

\title{Multi-population Ensemble Genetic Programming via Cooperative Coevolution and Multi-view Learning for Classification}

\author{Mohammad~Sadegh~Khorshidi\orcidlink{0000-0001-6556-2926}
 , Navid~Yazdanjue\orcidlink{0000-0001-9670-8422}%
 , Hassan~Gharoun\orcidlink{0000-0001-8298-7512}%
 , Mohammad~Reza~Nikoo\orcidlink{0000-0002-3740-4389}%
 , Fang~Chen\orcidlink{0000-0003-4971-8729}%
 , Amir~H.~Gandomi\orcidlink{0000-0002-2798-0104}
\thanks{Mohammad Sadegh Khorshidi, Navid Yazdanjue, Hassan Gharoun, Fang Chen, and Amir H. Gandomi are with the Faculty of Engineering \& Information Technology, University of Technology Sydney, Ultimo 2007, Australia. e-mails: msadegh.khorshidi.ak@gmail.com, navid.yazdanjue@gmail.com, hassan.gharoun@student.uts.edu.au, fang.chen@uts.edu.au, gandomi@uts.edu.au}
\thanks{Mohammad Reza Nikoo is with the Department of Civil and Architectural Engineering, Sultan Qaboos University, Muscat, Oman. (e-mail: m.reza@squ.edu.om}%
\thanks{Amir H. Gandomi is also with the University Research and Innovation Center (EKIK), Obuda University, Budapest 1034, Hungary.}%
\thanks{This work was supported by the Australian Government through the Australian Research Council under Project DE210101808.}
\thanks{Corresponding author: Amir H. Gandomi}}

\maketitle

\begin{abstract}
This paper introduces Multi-population Ensemble Genetic Programming (MEGP), a computational intelligence framework that integrates cooperative coevolution and the multi-view learning paradigm to address classification challenges in high-dimensional and heterogeneous feature spaces.
 MEGP decomposes the input space into conditionally independent feature subsets, enabling multiple subpopulations to evolve in parallel while interacting through a dynamic ensemble-based fitness mechanism.
Each individual encodes multiple genes whose outputs are aggregated via a differentiable softmax-based weighting layer, enhancing both model interpretability and adaptive decision fusion.
 A hybrid selection mechanism incorporating both isolated and ensemble-level fitness promotes inter-population cooperation while preserving intra-population diversity. 
This dual-level evolutionary dynamic facilitates structured search exploration and reduces premature convergence.
Experimental evaluations across eight benchmark datasets demonstrate that MEGP consistently outperforms a baseline GP model in terms of convergence behavior and generalization performance.
 Comprehensive statistical analyses validate significant improvements in Log-Loss, Precision, Recall, $F_1$ score, and AUC. MEGP also exhibits robust diversity retention and accelerated fitness gains throughout evolution, highlighting its effectiveness for scalable, ensemble-driven evolutionary learning.
By unifying population-based optimization, multi-view representation learning, and cooperative coevolution, MEGP contributes a structurally adaptive and interpretable framework that advances emerging directions in evolutionary machine learning.

\end{abstract}


\section{Introduction} \label{Section:Introduction}
\IEEEPARstart{T}{he} increasing reliance on data-driven insights across diverse domains necessitates scalable, interpretable, and robust models for complex datasets \cite{shen2022two}. Genetic Programming (GP), an evolutionary approach, autonomously derives solutions without predefined structural constraints \cite{koza1992programming}. Tree-based GP, leveraging hierarchical structures, is effective in classification and symbolic regression. However, its efficacy declines with increasing data dimensionality, encountering issues like computational inefficiency, feature redundancy, and overfitting \cite{chen2017feature, wang2024improving}.

Multi-Tree Genetic Programming (MTGP) mitigates scalability concerns by encoding multiple genes per individual, enhancing pattern discovery \cite{searson2010gptips}. Despite this, MTGP struggles with effective solution space exploration in high-dimensional settings, leading to premature convergence and increased computational overhead \cite{zhao2019improved}. While feature selection and extraction aid dimensionality reduction, they risk oversimplifying feature interactions or discarding informative attributes \cite{khorshidi2023filter, fan2024multi, gharoun2024shadow}.

Multi-Population Genetic Programming (MPGP) structures the search process by evolving multiple independent populations, promoting diversity and reducing premature convergence risks \cite{augusto2010coevolutionary}. However, conventional MPGP approaches do not explicitly assign feature subsets to populations, focusing instead on broad evolutionary diversification. Cooperative coevolution extends MPGP by facilitating inter-population information exchange, fostering solution adaptability \cite{shi2019prediction}. Strategies such as archive-based coevolution \cite{hong2024archive}, multiform gene expression \cite{zhong2025multiform}, and multi-level selection \cite{maua2017co} enhance robustness but often employ static configurations with limited adaptability to varying complexities \cite{sachindra2019machine, shi2020research}.

Ensemble learning improves robustness by integrating diverse model predictions, leveraging methods like bagging, boosting, and stacking \cite{virgolin2021genetic, zhang2023genetic, liao2024learning}. When combined with GP, ensemble learning enhances adaptability and interpretability, particularly in high-dimensional classification \cite{purohit2010construction}. However, balancing computational efficiency with interpretability remains a challenge, particularly for datasets with unbalanced classes or nonlinear feature relationships \cite{garcia2017multi, bi2022multitask}.

Multi-View Learning (MVL) aligns naturally with MPGP and ensemble strategies by partitioning data into distinct feature subsets, enabling models to extract complementary, conditionally independent information \cite{zhao2017multi}. MVL mitigates overfitting and enhances adaptability in applications like fault diagnosis \cite{peng2021multi, lu2023novel}, image classification \cite{zhang2023genetic}, and text analysis \cite{ju2021text}. MVL-GP frameworks highlight the advantages of integrating evolutionary techniques with multi-view paradigms. However, dynamic MVL integration with MPGP and ensemble-based strategies remains underexplored \cite{garcia2017multi, shi2019prediction}.

This study proposes the Multi-Population Ensemble Genetic Programming (MEGP) framework \cite{khorshidi2024enhancing}, which synthesizes MPGP, ensemble learning, and MVL principles to address high-dimensional classification challenges. MEGP introduces:

\begin{itemize}
	\item \textbf{Independent Evolution of Feature Subsets}: Populations evolve on distinct feature subsets, enhancing diversity and reducing redundancy \cite{xie2021two}.
	\item \textbf{Dynamic Ensemble Synthesis}: Gene outputs integrate via a Softmax-based function, balancing interpretability and classification accuracy \cite{zhang2023genetic}.
	\item \textbf{Adaptive Optimization Mechanisms}: Gradient-based optimization dynamically adjusts gene contributions, improving generalization \cite{shi2019prediction}.
	\item \textbf{Collaborative Fitness Evaluation}: Fitness incorporates both individual and ensemble-level performance, promoting inter-population cooperation \cite{hong2024archive}.
	\item \textbf{Enhanced Diversity Preservation}: Diversity maintenance strategies mitigate premature convergence and balance exploration-exploitation trade-offs \cite{liao2024learning}.
\end{itemize}

MEGP tackles feature interaction, redundancy, unbalanced distributions, and scalability by integrating cooperative coevolution, ensemble synthesis, and multi-view principles. 

The paper is structured as follows: Section \ref{Section:Related_Works} reviews multi-population GP and MVL in evolutionary algorithms. Section \ref{Section:Methodology} details the MEGP framework. Section \ref{Section:Experimental_Design} presents the experimental setup, and Section \ref{Section:Results_Discussion} discusses results. Finally, Section \ref{Section:Conclusion} concludes with future research directions.

\section{Related Works} \label{Section:Related_Works}
This section outlines the foundations of MEGP, focusing on multi-population strategies in GP and the role of multi-view learning in evolutionary computation.

\subsection{Multi-population Approaches in GP} \label{relevant_works_multipop_gp}
Multi-population GP enhances genetic diversity, balances exploration-exploitation, and improves search efficiency in complex, high-dimensional problems \cite{chen2017feature}. By evolving subpopulations independently, it mitigates premature convergence, though effectiveness depends on inter-population interactions and knowledge transfer mechanisms.

A basic form of MPGP maintains isolated populations without direct exchange, ensuring diversity but risking stagnation \cite{lin2007designing}. Migration-based strategies mitigate this by periodically transferring individuals across populations \cite{sachindra2019machine}. However, fixed migration rules \cite{sachindra2019machine, shi2019prediction} lack adaptability to dynamic landscapes, limiting optimal knowledge exchange.

Cooperative coevolutionary approaches improve adaptability by specializing populations in different solution components, later integrating their contributions \cite{shi2020research}. Layered population structures enhance robustness, as seen in Lin et al. \cite{lin2007designing} for medical diagnosis. Multi-objective MPGP further refines partitioning, balancing classification performance and feature selection \cite{li2024multi}.

Recent MPGP models introduce adaptive coevolution, dynamically regulating inter-population interactions. Hong et al. \cite{hong2024archive} proposed an archive-based model selectively sharing high-performing individuals, while Wang et al. \cite{wang2025semantics} leveraged semantics-guided crossover in multi-task GP to improve generalization. Adaptive subpopulation sizing further optimizes the exploration-exploitation trade-off \cite{zhao2019improved}.

Ensemble-based MPGP frameworks integrate decision trees and symbolic regression for classification \cite{zhang2023genetic}, leveraging structured diversity via guided mutation \cite{virgolin2021genetic}. Parallel MPGP approaches enhance generalization by reducing redundancy \cite{sachindra2019machine}. Despite these advances, existing MPGP frameworks often rely on static inter-population communication rules, limiting adaptability \cite{sachindra2019machine}. Additionally, few explicitly partition feature subsets across populations, leading to redundant solutions \cite{rokach2008genetic}. Structured feature partitioning could enhance specialization and mitigate competitive inefficiencies \cite{wang2024improving, rodriguez2025problem}.

\subsection{Multi-view Ensemble Learning in Genetic Programming}
Multi-view learning enhances generalization by leveraging conditionally independent feature subsets, aligning well with MPGP and ensemble-based evolutionary models \cite{yan2021deep}. By partitioning feature spaces, it fosters diversity, reduces redundancy, and reinforces ensemble learning principles.

GP-based ensemble learning has addressed classification challenges in high-dimensional and imbalanced datasets. Zhang and Bhattacharyya \cite{zhang2004genetic} demonstrated the effectiveness of evolving classifiers on distinct feature subsets aggregated through weighted voting, analogous to bagging. Kumar and Yadav \cite{kumar2023review} further highlighted structured decomposition’s role in optimizing multi-view ensemble performance, reinforcing its applicability in MPGP.

Coevolutionary multi-view GP frameworks have been explored for structured feature partitioning. García-Martínez and Ventura \cite{garcia2020multi} introduced a grammar-based multi-view GP approach that evolved interpretable rule-based classifiers across independent feature views, promoting cross-view consistency. Ceci et al. \cite{ceci2015semi} demonstrated how cross-view collaboration improves semi-supervised learning reliability.

Feature construction plays a crucial role in optimizing multi-view GP. Ain et al. \cite{ain2020genetic} proposed a multi-tree GP model where each tree specialized in extracting distinct feature representations, enhancing discrimination. Peng et al. \cite{peng2021multi} integrated classification accuracy with inter-class and intra-class distance metrics to optimize feature discrimination in fault diagnosis. Bi et al. \cite{bi2019automated, bi2020genetic} combined feature extraction and classifier selection into an integrated GP-based framework, improving ensemble diversity.

Multi-population GP must balance diversity and performance optimization. Bhowan et al. \cite{bhowan2012evolving} employed a multi-objective GP ensemble framework with Pareto-based optimization, negative correlation learning, and failure crediting to ensure complementary decision boundaries. Wen and Ting \cite{wen2016learning} introduced a multifactorial GP framework that concurrently evolved multiple classifiers with structured knowledge transfer, aligning with MPGP principles.

Heterogeneous GP-based ensemble models further highlight structured diversity’s role. Zhang et al. \cite{zhang2023sr} combined GP-based symbolic regression with decision trees, optimizing ensemble interactions via guided mutation. Li and Wong \cite{li2015financial} employed multi-objective GP for fraud detection, refining classifier composition via statistical selection learning. Bi et al. \cite{bi2019automated, bi2020genetic} demonstrated the potential of GP-based feature learning in structured ensembles.

Recent work emphasizes cooperative evolutionary strategies for flexible feature decomposition. Fan et al. \cite{fan2024multi} introduced a multi-tree GP ensemble model for image classification, leveraging AdaBoost-style weight assignments for classifier coordination. Dick \cite{dick2024ensemble} reinterpreted geometric semantic GP within an ensemble learning framework, aligning mutation and crossover mechanisms with boosting and stacking strategies. These insights suggest structured feature partitioning within MPGP can optimize evolutionary learning dynamics.

\section{Methodology} \label{Section:Methodology}

This section presents the MEGP framework, which partitions feature spaces into distinct views and evolves classifiers within multiple interacting populations.
The methodology incorporates ``\textit{Semantic-Preserving Feature Partitioning}'' (SPFP) \cite{khorshidi2025semantic} to ensure structured feature segmentation while remaining flexible for alternative random partitioning strategies \cite{kumar2023review}. 
The framework integrates isolated evolution within each population while leveraging an ensemble mechanism to enhance classification performance.

\subsection{Overview of MEGP}

The MEGP operates through a structured evolutionary process that consists of three primary stages: 
(\textit{i}) feature partitioning and population initialization, 
(\textit{ii}) independent evolution of populations using genetic programming, and 
(\textit{iii}) ensemble learning through linear combination of individuals across populations. 
The evolutionary cycle is illustrated in Figure \ref{fig:megp_flowchart_1}, while the population-specific and ensemble fitness evaluation process is depicted in Figure \ref{fig:megp_flowchart_2}.

The framework begins by decomposing the dataset into conditionally independent feature subsets, referred to as views, each assigned to a separate population. 
This structured partitioning preserves complementary feature information across populations while ensuring diversity in evolved solutions. 
Within each population, individuals encode genetic programs whose outputs are aggregated using a SoftMax function to yield class probabilities. 
The evolutionary process iteratively refines these programs while simultaneously optimizing an ensemble models that combines classifiers from different populations.

\subsection{Feature Partitioning and Isolated Fitness Evaluation}

The partitioning mechanism plays a fundamental role in MEGP by structuring the search space into semantically distinct representations. 
The SPFP algorithm ensures that each population receives a well-balanced feature subset that maintains feature relevance while reducing redundancy. 
Alternatively, random partitioning strategies \cite{bhowan2013reusing} can be employed, which have been shown to enhance efficiency in multi-population genetic programming by fostering diverse search trajectories.

Each population evolves independently using a standard genetic programming pipeline. 
Individuals within a population encode multiple expression trees, each representing a functional component of the final classifier. 
These outputs are aggregated using a SoftMax transformation:
\begin{equation} \label{eq:z_c}
Z_c = \sum_k \left( w^{iso}_{k} \cdot g_k \right) + b_c,
\end{equation}
where $w^{iso}_{k}$ represents the weight assigned to the $k$-th gene’s output $g_k$, and $b_c$ is the bias term for class $c$. 
The probability of class $c$ is computed as:
\begin{equation} \label{eq:pr_iso}
Pr^{iso}_{c} = \frac{\exp(Z_c)}{\sum_{C} \exp(Z_c)}.
\end{equation}

The weights and biases are optimized using a gradient-based method, providing an auxiliary heuristic that enables localized adaptation of classifiers within each population while maintaining an interpretable structure. 
Then, the isolated fitness is defined as:
\begin{equation} \label{eq:ft_iso}
FT_{iso} = -\sum_{i} y_{i,c} \log Pr^{iso}_{i,c},
\end{equation}
where $y_{(i,c)}$ represents the true class probability of instance $i$ within the dataset.

\subsection{Linear Ensembling and Ensemble Fitness Evaluation}

To integrate predictions across distinct views, MEGP employs a linear ensembling strategy that combines classifiers from different populations. 
The ensemble prediction probability is computed as:
\begin{equation} \label{eq:pr_en}
Pr^{en}_{c} = \sum_{p} \left( w^{en}_{p,c} \cdot Pr^{iso}_{p,c} \right),
\end{equation}
where $w^{en}_{p,c}$ represents the weight assigned to the isolated probability $Pr^{iso}_{p,c}$ of class $c$ predicted by an individual from population $p$. 
These ensemble weights are optimized by minimizing the Mean Squared Error (MSE):
\begin{equation} \label{eq:mse}
\text{MSE} = \frac{1}{N} \sum_{i=1}^{N} \sum_{c} \left( y_{i,c} - Pr^{en}_{i,c} \right)^2,
\end{equation}
where $N$ is the number of instances in the dataset. 
The optimization is performed using COBYLA \cite{powell1994direct} to efficiently compute the optimal ensemble weights.

The ensemble fitness, denoted as $FT_{en}$, is evaluated using cross-entropy loss:
\begin{equation} \label{eq:ft_en}
FT_{en} = -\sum_{i} y_{i,c} \log Pr^{en}_{i,c}.
\end{equation}
The computed $FT_{iso}$ from Eq. \ref{eq:ft_iso} provides an assessment of an individual's predictive capability based solely on its isolated evolutionary trajectory within a single population, ensuring population-specific adaptation. 
In contrast, $FT_{en}$ from Eq. \ref{eq:ft_en} evaluates the contribution of an individual to the collective predictive performance of the ensemble, emphasizing its role in inter-population cooperation and optimizing the global classification objective.

\subsection{Evolutionary Dynamics and Cooperative Coevolution}

Traditional cooperative coevolutionary paradigms operate by decomposing a problem into distinct subproblems, each assigned to an evolving population that specializes in optimizing a specific component. 
These paradigms typically incorporate explicit interaction mechanisms, such as inter-population crossover and migration, which facilitate knowledge exchange at predefined intervals---e.g., every five generations. 
This interaction mechanism enables populations to refine partial solutions before recombining them into a final program.

\input{flowchart.tex}

In contrast, MEGP adopts a multi-view evolutionary framework, wherein populations evolve on distinct feature subsets that remain fixed throughout the evolutionary process. 
This structural constraint ensures that each population develops specialized representations of the data without feature leakage between populations. 
Unlike conventional coevolutionary GP approaches, where solutions are incrementally refined across populations via periodic genetic exchanges, MEGP enforces feature-invariant population evolution while indirectly facilitating inter-population cooperation through an ensemble-based selection strategy. 
This approach preserves the search landscape within each population while simultaneously leveraging knowledge from complementary populations in the global search process.

To balance intra-population optimization with global ensemble adaptation, MEGP introduces a modified evolutionary cycle, as illustrated in Figure \ref{fig:megp_flowchart_1}. 
Instead of employing a single elitism mechanism, MEGP defines two elite fraction parameters--- $EF^{iso}$ and $EF^{en}$---to ensure that individuals excelling at different levels of the evolutionary hierarchy are retained. 
The isolated elitism fraction ($EF^{iso}$) preserves the top-performing individuals within each population based on their isolated fitness $FT_{iso}$, ensuring that strong standalone classifiers are propagated. 
Simultaneously, the ensemble elitism fraction ($EF^{en}$) selects individuals that contribute to the highest-performing ensembles based on their ensemble fitness $FT_{en}$. 
This distinction is critical because the individuals ranked highest in terms of isolated fitness may not necessarily form the most effective ensemble when combined across populations. 
Since each population evolves independently on a disjoint subset of features---each capturing complementary information---MEGP ensures that both locally and globally optimal individuals are preserved, thereby fostering a soft cooperative coevolutionary dynamic.

Following the elitism operation, MEGP proceeds with standard GP genetic operations---crossover, mutation, and reproduction---selected probabilistically based on predefined rates ($p_c$ for crossover, $p_m$ for mutation, and $p_r$ for reproduction). 
Mutation and reproduction are performed as in traditional GP, preserving genetic diversity and allowing populations to refine their solutions. 
However, MEGP introduces a modified crossover mechanism that incorporates an additional probabilistic parameter, $p_{en}$, which randomly selects between isolated fitness ($FT_{iso}$) and ensemble fitness ($FT_{en}$) as the selection criterion for tournament-based parent selection. 
This stochastic selection mechanism ensures that individuals contributing significantly to either isolated performance or ensemble synergy have an opportunity to propagate their genetic material, striking a balance between local adaptation and ensemble-driven generalization.

By integrating dual-level elitism and adaptive selection criteria, MEGP implicitly facilitates inter-population cooperation without requiring explicit genetic exchanges between populations. 
This enables populations to co-evolve on distinct feature views while simultaneously contributing to a globally optimized ensemble model. 
The evolutionary process iterates until a predefined termination criterion is met, such as reaching a maximum number of generations or detecting convergence in ensemble fitness.

\section{Experimental Design} \label{Section:Experimental_Design}

\subsection{Data Description} 

To rigorously evaluate the effectiveness of MEGP, eight publicly available benchmark datasets were selected, covering a diverse range of classification challenges in terms of feature dimensionality, instance count, and class distribution. 
These datasets were chosen to assess MEGP’s ability to generalize across different data complexities, including high-dimensional spaces, imbalanced class distributions, and real-world sensor-based or biomedical classification tasks.

Table \ref{tab:data} summarizes the datasets used in this study: APS Failure at Scania Trucks (APSF) \cite{apsf}, Activity Recognition Using Wearable Physiological Measurements (ARWPM) \cite{arwpm}, Gene Expression Cancer RNA-Seq (GECR) \cite{gecr}, Grammatical Facial Expressions (GFE) \cite{gfe}, Gas Sensor Array Drift at Different Concentrations (GSAD) \cite{gsad}, Smartphone-Based Recognition of Human Activities and Postural Transitions (HAPT) \cite{hapt}, ISOLET \cite{isolet}, and Parkinson’s Disease (PD) \cite{pd}. 
These datasets represent a broad spectrum of applications, ranging from physiological signal classification to speech and industrial fault detection, making them well-suited to evaluate the adaptability and robustness of MEGP.

\input{data.tex}

\subsection{Experimental Setup}

The empirical study evaluates MEGP against a baseline Genetic Programming model (BGP), which follows the same genetic framework but operates as a single-population GP, omitting multi-population coevolution and ensemble learning. 
This ensures an isolated assessment of MEGP’s contributions.

Feature partitioning for MEGP was conducted using the SPFP algorithm \cite{khorshidi2025semantic}, generating two distinct feature subsets per dataset. 
This partitioning ensures each population receives complementary yet semantically consistent data representations. 
To analyze the impact of cooperative coevolution on optimization efficiency and population diversity, we experiment with five MEGP configurations: MEGP\textsubscript{0}, MEGP\textsubscript{25}, MEGP\textsubscript{50}, MEGP\textsubscript{75}, and MEGP\textsubscript{100}, each varying in elite fraction distribution ($EF^{iso}$ vs. $EF^{en}$) and ensemble selection probability ($p_{en}$).

\subsubsection{MEGP Configurations}

Each MEGP variant is parameterized by:

\begin{itemize} \item Elite Fraction ($EF^{en}$ vs. $EF^{iso}$): The proportion of elite individuals selected based on ensemble fitness versus isolated fitness, maintaining $EF^{iso} + EF^{en} = const.$
\item Ensemble Selection Probability ($p_{en}$): Governs whether tournament selection prioritizes isolated fitness ($FT_{iso}$) or ensemble fitness ($FT_{en}$).
\end{itemize}

The configurations are defined as follows:

\begin{itemize} \item MEGP\textsubscript{0}: $EF^{en} = 0$, $p_{en} = 0$ (purely isolated fitness-driven selection).
\item MEGP\textsubscript{25}: $EF^{en} = 0.25EF$, $p_{en} = 0.25$.
\item MEGP\textsubscript{50}: $EF^{en} = 0.5EF$, $p_{en} = 0.5$.
\item MEGP\textsubscript{75}: $EF^{en} = 0.75EF$, $p_{en} = 0.75$.
\item MEGP\textsubscript{100}: $EF^{en} = EF$, $p_{en} = 1$ (purely ensemble fitness-driven selection).
\end{itemize}

This experimental design enables a systematic analysis of ensemble fitness’s influence on evolutionary dynamics and classification performance.

\subsubsection{Training and Evaluation Procedure}

Each experiment consists of 30 independent runs to ensure statistical robustness. Datasets are randomly partitioned as follows:

\begin{itemize} \item 54\% training
\item 13\% validation (prevents overfitting of individual and ensemble weights)
\item 33\% testing
\end{itemize}

Parameter settings for BGP and MEGP are detailed in Table \ref{tab:parameters}, ensuring consistency across experiments.

\input{parameters.tex}

\subsection{Evaluation Metrics}

The performance of the proposed MEGP framework is assessed through a two-tier evaluation strategy:

\begin{itemize} \item Evolutionary Progression Analysis – Investigates the trajectory of evolutionary search and convergence behavior relative to BGP.
\item Generalization Performance – Evaluates predictive efficacy and robustness on unseen test data.
\end{itemize}

\subsubsection{Evolutionary Progression Metrics}

To elucidate the evolutionary dynamics of MEGP, we compute the following quantitative measures at distinct generational intervals ($0$–$50$, $51$–$100$, $101$–$150$) and across the entire evolutionary process ($0$–$150$):

\begin{itemize} \item Fitness ($FT$): The primary optimization criterion, measured via log-loss to quantify classification performance (lower values denote superior solutions).
\item Convergence Rate ($CR$): Captures the rate of fitness improvement across successive generations, indicating search efficiency.
\item Crossover Convergence Rate ($CCR$): Evaluates the efficacy of crossover in enhancing offspring fitness relative to parental solutions.
\item Entropy of Final Population: Quantifies diversity preservation in the terminal population to mitigate premature convergence and maintain search-space exploration.
\item Total Computational Overhead (seconds): Measures the runtime efficiency of MEGP relative to BGP.
\end{itemize}

\subsubsection{Generalization Performance Metrics}

The evolved models' predictive generalization is assessed using standard classification metrics:

\begin{itemize} \item Log-Loss: A probabilistic scoring function quantifying prediction confidence (lower values signify more decisive classifications).
\item Precision: The fraction of correctly identified positive instances among predicted positives, indicating resistance to false positives.
\item Recall: The fraction of correctly identified positives among actual positives, reflecting sensitivity to minority classes.
\item $F_1$ Score: The harmonic mean of precision and recall, providing a balanced assessment of predictive reliability.
\item Area Under the Curve (AUC): Measures the model’s discriminatory capacity, with higher values denoting superior separability between classes.
\end{itemize}

\subsection{Statistical Analysis}

To ensure statistical rigor, we employ non-parametric tests suited for the stochastic nature of evolutionary algorithms, which often exhibit non-normal distributions and heteroskedasticity. 
The Friedman test, a rank-based alternative to ANOVA, assesses whether performance differences between BGP and MEGP configurations are statistically significant across multiple datasets.
Given the multiple comparisons, we apply Bonferroni correction to control Type I errors (false positives), though its conservativeness may increase Type II errors (false negatives). 
To refine significance testing, we conduct Conover post-hoc tests for pairwise comparisons and use Benjamini-Hochberg (BH) correction to balance statistical power and false discovery control.

Statistical significance alone does not quantify effect magnitude. 
We compute Cliff’s Delta ($\delta$) to measure dominance between distributions, categorizing effects as negligible ($|\delta| < 0.147$), small ($0.147 \leq |\delta| < 0.33$), medium ($0.33 \leq |\delta| < 0.474$), or large ($|\delta| \geq 0.474$).
By integrating significance testing, post-hoc analysis, and effect size estimation, we provide a robust evaluation of MEGP’s optimization dynamics, model diversity, and predictive generalization relative to BGP.

\subsection{Computational Resources}

The experiments were executed on a high-performance computing cluster comprising 30 nodes, with each independent GP run assigned to a dedicated node. Parallelization across the cluster expedited computation while ensuring isolated execution. The hardware configuration of each node was as follows:

\begin{itemize} \item Processor: AMD EPYC 9354P, 32-core @ 3.25 GHz.
\item Memory: 128 GB RAM.
\item Storage: 1.92 TB NVMe SSD.
\item Software: Python-based MEGP implementation. 
\end{itemize}

Each dataset underwent 30 independent GP runs, with dedicated node allocation preventing resource contention. 
While nodes were accessed via a virtual machine (VM), potential constraints on CPU availability, memory bandwidth, and I/O operations remain unverified. 
Nonetheless, uniform resource allocation across all experiments ensured fair comparisons between MEGP configurations and the BGP baseline. 
This parallelized execution strategy markedly reduced computational overhead, enabling large-scale GP experimentation while preserving statistical rigor in performance assessment.

\section{Results and Discussion} \label{Section:Results_Discussion}

To evaluate the efficacy of MEGP relative to BGP, we conduct two primary analyses. 
First, we assess algorithmic efficiency by examining convergence dynamics, analyzing fitness progression, convergence rate, crossover effectiveness, diversity retention, and computational efficiency across different generational intervals. 
Second, we evaluate generalization performance by comparing the classification accuracy of the best-evolved solutions using standard metrics. This section is structured as follows: Section \ref{Section:convergence} presents and discusses evolutionary progression and convergence metrics, while Section \ref{Section:generalization} analyzes generalization performance metrics.

\subsection{Evolutionary Progression Metrics}\label{Section:convergence}

Figure \ref{fig:fig_congen_paper} illustrates the mean and median fitness trajectories of the best individuals across generations for the GFE and HAPT datasets, while Figure \ref{fig:fig_congen}, provided in the Supplementary Document (SD), presents the corresponding results for all eight datasets. 
These visualizations offer insights into MEGP’s convergence behavior under varying data complexities and dimensionalities.
MEGP configurations consistently achieve faster and lower convergence compared to BGP, particularly in early generations, indicating more effective exploration of high-dimensional search spaces. 

Higher ensemble selection probabilities (MEGP\textsubscript{75} and MEGP\textsubscript{100}) exhibit superior convergence rates and lower final fitness values across datasets, emphasizing the role of cooperative ensemble-based selection in accelerating optimization.
For datasets with high feature dimensionality, such as GECR and HAPT, MEGP\textsubscript{75} and MEGP\textsubscript{100} demonstrate clear advantages over both BGP and lower ensemble-probability MEGP variants. 
Similarly, in APSF and PD, MEGP models rapidly surpass BGP and stabilize at lower fitness values, suggesting that cooperative coevolution enables efficient exploration-exploitation balance, mitigating premature convergence.

\input{fig_congen_paper.tex}

MEGP\textsubscript{50} often performs comparably to MEGP\textsubscript{75} and MEGP\textsubscript{100}, implying that moderate ensemble contributions can yield substantial improvements. 
In contrast, MEGP\textsubscript{0}, which relies exclusively on isolated fitness, exhibits slower convergence, particularly in datasets with complex class structures like ISOLET, where the absence of inter-population cooperation hampers optimization efficiency.
The comparative evaluation extends to quantitative convergence metrics, including Fitness ($FT$), Convergence Rate ($CR$), Crossover Convergence Rate ($CCR$), final population entropy, and total execution time. 
These metrics were analyzed over three generational phases (0–50, 51–100, and 101–150) and across the full evolutionary trajectory (0–150).

Table~\ref{tab:table_friedman_convergence} (provided in SD) presents the Friedman test results with Bonferroni-adjusted p-values ($\alpha = 0.05$), where bold values indicate statistically significant differences between models.
Analysis of the Crossover Convergence Rate (CCR) reveals significant improvements in MEGP compared to BGP. 
As shown in Table~\ref{tab:table_friedman_convergence} (SD), MEGP configurations with higher ensemble selection probabilities (50\%, 75\%, and 100\%) consistently achieve higher CCR values across all generational intervals (0–50, 51–100, 101–150) and overall (0–150). 
This indicates that MEGP's multi-population structure enables more effective crossover operations, expediting convergence toward optimal solutions. 
The observed CCR improvements, particularly at higher ensemble probabilities, highlight the role of cooperative interactions in promoting diverse exploration and accelerating convergence. 
The Bonferroni-adjusted p-values confirm the statistical significance of these enhancements, with values well below the $\alpha$ threshold.

For Convergence Rate (CR), MEGP demonstrates marginal improvements over BGP, though less pronounced than in CCR. Notably, MEGP\textsubscript{50} and MEGP\textsubscript{75} exhibit slightly higher CR values in certain generational intervals, suggesting that cooperative evolution can accelerate convergence, though its impact varies with ensemble probability. 
This trend indicates that while MEGP enhances crossover efficiency, overall convergence is influenced by factors such as selection pressure and genetic diversity maintenance. 
Statistical analysis reinforces these findings, with significant differences observed primarily for intermediate ensemble probabilities, suggesting that a balanced combination of isolated and cooperative selection enhances convergence efficiency.

Entropy of the final population serves as a key indicator of solution diversity. 
MEGP configurations with moderate ensemble probabilities (50\% and 75\%) maintain significantly higher entropy than BGP, as evidenced by the bolded adjusted p-values in Table~\ref{tab:table_friedman_convergence} (SD). 
This suggests that MEGP mitigates premature convergence, allowing a broader exploration of the search space. 
Cooperative interactions across subpopulations likely contribute to this diversity preservation, preventing any single subpopulation from dominating the search process. 
By sustaining a more balanced exploration-exploitation trade-off, MEGP fosters multiple promising solutions, which is crucial for addressing complex, multimodal optimization landscapes.

Regarding computational efficiency, MEGP introduces a modest increase in running time compared to BGP, particularly at higher ensemble probabilities. 
This overhead stems from the additional computations required for ensemble fitness evaluations and cooperative optimization. 
However, the increase remains manageable and is not statistically significant, as indicated by the adjusted p-values. 
Importantly, the observed performance gains in CCR and entropy justify this computational cost, demonstrating that MEGP enhances evolutionary search without imposing a prohibitive computational burden.

The SD provide comprehensive statistical visualizations for the 14 performance metrics, including raincloud plots, heatmaps of adjusted Conover p-values, and Cliff’s $\delta$ analysis with 10,000 bootstraps and a 95\% confidence interval. 
These visualizations offer deeper insights into the comparative performance of MEGP and BGP.

We conducted effect size analysis to quantify the magnitude of performance differences between models, offering insights beyond statistical significance. 
While p-values indicate whether differences are statistically detectable, Cliff’s $\delta$ assesses their practical relevance, particularly in large datasets or cases where subtle performance improvements matter. 
This analysis helps distinguish meaningful model enhancements from statistically significant but negligible variations, guiding future model refinements. 
The combined statistical and visual comparisons further clarify how different ensemble configurations influence key metrics, reinforcing the robustness of the observed results.

\input{winners_convergence}

Figure~\ref{fig:winners_convergence} provides a comprehensive comparison of the top-performing MEGP configurations against BGP across multiple convergence criteria: $FT$ (Figures \ref{fig:ft50_win}, \ref{fig:ft100_win}, \ref{fig:ft150_win}, and \ref{fig:ftall_win}) $CR$ (Figures \ref{fig:cr50_win}, \ref{fig:cr100_win}, \ref{fig:cr150_win}, and \ref{fig:crall_win}), $CCR$ (Figures \ref{fig:ccr50_win}, \ref{fig:ccr100_win}, \ref{fig:ccr150_win}, and \ref{fig:ccrall_win}), Entropy (Figure \ref{fig:entropy_win}), and Running Time (Figure \ref{fig:time_win}). 
Each circle represents a metric, with segments corresponding to performance across different datasets. Color coding indicates statistical significance and the direction of performance differences.

\begin{itemize}
\item Grey segments indicate no significant difference ($P_{fr} > 0.05$, $P_{cn} > 0.05$), suggesting that MEGP performs comparably to BGP for these dataset-metric combinations, demonstrating stability in results.
\item Blue segments highlight cases where MEGP significantly outperforms BGP ($P_{fr} < 0.05$, $P_{cn} < 0.05$, and $\delta > 0$), emphasizing the advantage of multi-population ensembles in improving specific convergence metrics.
\item Red segments denote instances where BGP surpasses MEGP ($P_{fr} < 0.05$, $P_{cn} < 0.05$, and $\delta < 0$), indicating scenarios where MEGP’s added complexity may not yield a net performance gain.
\end{itemize}

Examining fitness trends ($FT_{50}$, $FT_{100}$, and $FT_{150}$) in Figures \ref{fig:ft50_win}, \ref{fig:ft100_win}, and \ref{fig:ft150_win}, MEGP exhibits superior fitness progression across most datasets, as indicated by the predominance of blue segments. 
This suggests that the multi-population cooperative framework facilitates more effective exploration and optimization throughout the evolutionary process. 
Figure \ref{fig:ftall_win} further reinforces this trend, demonstrating that $FT_{all}$ favors MEGP across the full evolutionary trajectory.

However, the APSF and GECR datasets deviate from this pattern, where neither MEGP nor BGP exhibits a clear fitness advantage. 
Further analysis of final best-model fitness in Figure \ref{fig:fig_congen} (see SD) reveals that for these two datasets, all models (BGP and MEGP) converge to highly accurate classification solutions with log-loss values around 0.1. 
This suggests that the classification task for APSF and GECR is inherently less complex, making it more challenging for MEGP to demonstrate a distinct advantage over BGP. 
When models from both frameworks already achieve near-optimal solutions, the incremental benefits of multi-population coevolution become less pronounced.

In contrast, for datasets where the log-loss of the evolved models remains higher, MEGP consistently surpasses BGP across all fitness metrics ($FT_{50}$, $FT_{100}$, $FT_{150}$, and $FT_{all}$). 
This reinforces the premise that MEGP’s multi-view cooperative ensemble mechanism is particularly advantageous in more challenging classification scenarios, where evolutionary diversity and enhanced optimization dynamics play a crucial role in improving model performance.

For convergence rate ($CR_{50}$, $CR_{100}$, $CR_{150}$) in Figures \ref{fig:cr50_win}, \ref{fig:cr100_win}, and \ref{fig:cr150_win}, the results reveal a dataset-dependent trend. 
At $CR_{50}$ (Figure \ref{fig:cr50_win}), three datasets favor BGP (red), while five show no significant difference (grey), suggesting that MEGP’s cooperative framework may introduce additional constraints in early generations, limiting immediate convergence gains. 
As evolution progresses, $CR_{100}$ (Figure \ref{fig:cr100_win}) shows a shift, with one dataset favoring MEGP (blue) and the remaining seven displaying no statistical difference, indicating that ensemble interactions may begin stabilizing improvements over time. 
By $CR_{150}$ (Figure \ref{fig:cr150_win}), three datasets favor MEGP, while the remaining five remain neutral, implying that cooperative selection benefits tend to emerge in later evolutionary stages rather than at the outset.

However, examining the full evolutionary trajectory ($CR_{all}$) in Figure \ref{fig:crall_win}, MEGP outperforms BGP only on GFE, while results for PD and APSF show no significant difference, and BGP demonstrates superiority in the remaining datasets. 
Investigating Figure \ref{fig:fig_congen} reveals a key factor influencing these results: the initial fitness (generation 0) for MEGP models is considerably lower than that of BGP models. 
Since BGP begins with higher fitness values and a larger search space to explore—alongside a significantly larger population size (60 for BGP vs. 30 per population for MEGP)—it experiences a greater overall fitness improvement over generations, despite the genetic operations in both frameworks being fundamentally identical. 
This suggests that while MEGP facilitates cooperative evolution and structured search, the larger population and broader search landscape of BGP allow for more substantial fitness gains over time.

These findings underscore that MEGP’s effectiveness in accelerating convergence is highly dependent on dataset complexity and initial search space conditions. 
Interestingly, the datasets where BGP demonstrates superior convergence rate are precisely those where MEGP starts with significantly lower initial fitness and also achieves lower final fitness, with the log-loss of the best final classifiers ranging from 0.6 to 1.6 (Figure \ref{fig:fig_congen}). 
This suggests that problems that are more challenging for BGP tend to be more tractable for MEGP, albeit with a lower convergence rate.
While MEGP’s ensemble-driven optimization enhances search diversity, it does not always translate into faster convergence, particularly in cases where BGP benefits from a broader evolutionary landscape and a larger population size to navigate complex fitness landscapes more effectively.

Since we modified the crossover operation by introducing $p_{en}$, which determines whether tournament selection prioritizes $FT^{iso}$ (isolated fitness) or $FT^{en}$ (ensemble fitness), it is essential to examine how this adjustment influences fitness improvements driven solely by crossover. 
Figures \ref{fig:ccr50_win}, \ref{fig:ccr100_win}, \ref{fig:ccr150_win}, and \ref{fig:ccrall_win} illustrate the crossover convergence rate trends ($CCR_{50}$, $CCR_{100}$, $CCR_{150}$, and $CCR_{all}$), where MEGP consistently outperforms BGP across all datasets. 
The universal presence of blue segments highlights the robustness of MEGP’s crossover mechanism in sustaining fitness improvements across generations.

The performance gap becomes more pronounced in $CCR_{100}$ and $CCR_{150}$, suggesting that as evolution progresses, ensemble-driven selection enhances recombination efficiency, leading to increasingly superior offspring. 
Unlike BGP, where genetic material is confined within a single population, MEGP leverages inter-population interactions to maintain diversity, ensuring crossover remains an active driver of search progression. 
The same pattern persists in $CCR_{all}$, where MEGP achieves sustained cumulative fitness improvements, reinforcing the long-term advantage of multi-population cooperation in evolutionary search.

The results suggest that tournament selection guided by $FT^{en}$ enables a more structured and adaptive crossover process, balancing localized exploitation with broader exploration. 
By allowing populations to exchange and refine genetic material more effectively, MEGP ensures that crossover convergence remains efficient across diverse problem landscapes, consistently outperforming BGP throughout the evolutionary process.

The entropy metric (Figure \ref{fig:entropy_win}) shows almost exclusively blue segments, indicating that MEGP models maintain significantly higher population diversity than BGP.
 Statistical analysis of the final population entropy confirms this advantage, demonstrating that despite both models having the same total number of individuals (60 in BGP vs. 2×30 in MEGP), the multi-population structure of MEGP fosters a more diverse genetic pool. 
 This suggests that inter-population variation introduced by independent evolutionary processes contributes to broader solution exploration, mitigating the risk of premature convergence and improving adaptation to complex fitness landscapes.

Running Time (Figure \ref{fig:time_win}) exhibits a dominance of red segments, particularly for larger datasets, indicating that BGP consistently achieves lower computational overhead. 
This performance gap is primarily attributed to the additional ensemble weight learning stage in MEGP, which introduces an optimization step beyond the standard evolutionary process in BGP. 
The increased computational complexity stems from the necessity to optimize ensemble coefficients in MEGP, which enhances predictive performance but incurs additional cost. 
While MEGP demonstrates clear advantages in convergence stability and diversity preservation, these benefits come at the expense of higher computational demands, highlighting the trade-off between model performance and efficiency in large-scale evolutionary tasks.

\input{wtl_convergence}

Table \ref{tab:wtl_convergence} presents the win-tie-loss (W-T-L) analysis comparing MEGP configurations ($MEGP_0$ to $MEGP_{100}$) against BGP across 14 convergence metrics, using adjusted Friedman and Conover p-values.
For fitness tracking metrics ($FT_{50}$, $FT_{100}$, $FT_{150}$, $FT_{all}$), MEGP configurations either match or surpass BGP, with no losses reported. 
Higher ensemble selection probabilities ($MEGP_{50}$–$MEGP_{100}$) yield the most wins, suggesting that ensemble selection enhances fitness progression over generations.

In convergence rate metrics ($CR_{50}$–$CR_{all}$), results vary. MEGP shows a high number of ties, with few wins and multiple losses, indicating that BGP remains competitive in convergence speed. 
The lack of improvement in higher ensemble selection probabilities ($MEGP_{75}$, $MEGP_{100}$) suggests that increasing ensemble reliance does not necessarily accelerate convergence.
For crossover convergence metrics ($CCR_{50}$–$CCR_{all}$), $MEGP_0$ and $MEGP_{25}$ achieve the most wins, while higher ensemble probabilities lead to more ties and occasional losses. 
This suggests that moderate ensemble selection aids cumulative fitness improvements, but excessive reliance may yield diminishing returns.

Entropy results confirm that MEGP configurations maintain higher diversity than BGP, with wins increasing as ensemble probability rises. 
This supports the role of multi-population evolution in preserving diversity, reducing premature convergence. $MEGP_{50}$–$MEGP_{100}$ exhibit the strongest trends, reinforcing ensemble selection’s contribution to solution exploration.
In Running Time, MEGP consistently records ties or losses, with no wins. 
This reflects the additional computational cost of ensemble weight learning, emphasizing the trade-off between performance gains in fitness and diversity and increased execution time

\input{rank_gen_megp_100}

Figure \ref{fig:rank_gen_megp_100} presents the ranking trajectories of individuals forming the best ensemble in each generation for MEGP\textsubscript{100}, illustrating their isolated rankings within their respective populations over 30 runs. 
A key observation is that the individuals contributing to the best ensemble are not necessarily the ones with the lowest fitness within their populations. 
This phenomenon highlights the nuanced interplay between individual performance in isolated fitness ($FT^{iso}$) and the collective predictive capability of the ensemble ($FT^{en}$).

Unlike traditional single-population GP, where selection is directly tied to individual fitness, MEGP\textsubscript{100} operates under a cooperative coevolution framework, where the final ensemble is optimized to minimize classification error holistically rather than selecting individuals purely based on isolated fitness. 
This occurs because ensemble fitness ($FT^{en}$) is influenced by the complementary contributions of individuals from both populations. In some cases, an individual with suboptimal isolated fitness may provide unique feature interactions or corrective influence, improving the ensemble’s overall predictive capacity. 
Conversely, individuals with the lowest $FT^{iso}$ may be redundant or overfit to specific training patterns, reducing ensemble generalization.

This effect is particularly pronounced in MEGP\textsubscript{100}, where ensemble selection is the dominant guiding force ($p_{en} = 1$), leading to an evolutionary process that prioritizes inter-population synergy over individual dominance. 
Consequently, the ranking distributions in Figure \ref{fig:rank_gen_megp_100} show a broader spread of contributing individuals across different fitness levels, reinforcing that ensemble formation in MEGP does not follow a strict fitness-based hierarchy but rather an adaptive, cooperative selection mechanism.

A similar trend is observed in other MEGP configurations, as shown in SD Figures \ref{fig:rank_gen_megp_0}, \ref{fig:rank_gen_megp_25}, \ref{fig:rank_gen_megp_50}, and \ref{fig:rank_gen_megp_75}, with varying degrees of ensemble-driven selection influence. 
Notably, in MEGP\textsubscript{0}, where ensemble fitness is not considered ($p_{en} = 0$), the rankings align more closely with isolated fitness, reinforcing the role of ensemble-driven optimization in shaping individual selection dynamics. 
As $p_{en}$ increases, ensemble fitness gradually gains dominance, leading to the divergence between isolated rankings and ensemble-contributing individuals seen in MEGP\textsubscript{100}.

\subsection{Generalization Metrics}\label{Section:generalization}

Table \ref{tab:table_friedman_generalization} (See SD) presents a statistical comparison of BGP and MEGP models across Log-Loss, Precision, Recall, $F_{1}$ score, and AUC, with MEGP variants evaluated at 0\%, 25\%, 50\%, 75\%, and 100\% ensemble selection probabilities.
Performance is assessed using the Friedman test with Bonferroni-adjusted p-values, with mean and standard deviation computed from the best final solutions in testing data.

\input{winners_generalization}

Figure \ref{fig:winners_generalization} provides a visual comparison of MEGP vs. BGP across generalization metrics, with circular plots indicating dataset-specific performance for each metric: Log-Loss (Figure \ref{fig:loss_win}), Precision (Figure \ref{fig:precision_win}), Recall (Figure \ref{fig:recall_win}), $F_{1}$ score (Figure \ref{fig:f1_win}), and AUC (Figure \ref{fig:auc_win}). 
The results reveal a consistent advantage for MEGP models, particularly in complex datasets like ISOLET and HAPT, where ensemble learning significantly improves classification robustness and predictive reliability.

Across nearly all metrics, MEGP models achieve either superior or equivalent performance compared to BGP, with few cases showing no significant difference. 
Notably, higher ensemble selection probabilities ($MEGP_{50}$–$MEGP_{100}$) exhibit dominant improvements in Log-Loss, Precision, and Recall, highlighting the effectiveness of cooperative learning in refining decision boundaries and reducing misclassification errors. 
The absence of red segments further confirms that MEGP never underperforms relative to BGP, reinforcing the stability and adaptability of the multi-population approach.

Although datasets such as APSF, GFE, and PD show no statistically significant improvement (grey segments), the overall trend indicates that ensemble-driven optimization consistently enhances generalization. 
This suggests that MEGP’s structural advantages—feature partitioning, inter-population diversity, and ensemble selection—contribute to a more generalized decision-making process, particularly in high-dimensional and complex classification tasks.

The SD provide raincloud plots for individual runs, heatmaps of adjusted Conover p-values, and Cliff’s $\delta$ effect sizes, offering deeper statistical insights into the generalization performance of MEGP across configurations.

\input{wtl_generalization}

Table \ref{tab:wtl_generalization} summarizes the win-tie-loss (W-T-L) comparison between MEGP and BGP across Log-Loss, Precision, Recall, $F_{1}$ score, and AUC, based on adjusted Friedman and Conover p-values. The results indicate a clear generalization advantage for MEGP, with zero losses across all metrics.
For Log-Loss, MEGP achieves six wins and two ties, consistently outperforming BGP in minimizing prediction error. 
This result underscores MEGP’s superior reliability in probabilistic classification, particularly in scenarios requiring reduced prediction uncertainty. The absence of losses confirms its robustness in generalization.

In Precision, Recall, $F_{1}$ score, and AUC, MEGP exhibits four wins and four ties per metric, demonstrating consistently competitive performance. 
The results suggest that MEGP effectively preserves high Precision, ensuring accurate classification of positive instances, while maintaining strong Recall, making it well-suited for sensitivity-driven tasks. 
The balanced wins in $F_{1}$ score confirm MEGP’s ability to optimize both metrics simultaneously, avoiding trade-offs that could otherwise hinder generalization.
AUC results further validate MEGP’s enhanced class separation capability, with no cases where BGP outperforms MEGP. This highlights the stability and adaptability of MEGP’s ensemble-driven optimization across diverse datasets.

The consistent zero-loss pattern across all generalization metrics suggests that MEGP’s ensemble-driven optimization effectively balances exploitation and exploration across evolutionary cycles. 
The improvement in Log-Loss indicates that MEGP configurations reduce uncertainty in probabilistic classification, leading to more calibrated decision boundaries. 
The parity in Precision and Recall wins suggests that MEGP enhances both positive class identification and error minimization, a critical advantage in imbalanced datasets where isolated selection may lead to biased optimization. 
The AUC results confirm that MEGP maintains robust class discrimination across decision thresholds, reinforcing its adaptability across diverse problem landscapes. 
These findings demonstrate that multi-population coevolution, when combined with ensemble fitness evaluation, provides a structural advantage in optimizing generalization across classification tasks.

\section{Conclusion}\label{Section:Conclusion}

This study systematically evaluated Multi-population Ensemble Genetic Programming (MEGP) against Baseline Genetic Programming (BGP) across multiple datasets and performance metrics. 
Statistical analysis using Friedman and Conover post-hoc tests confirmed that MEGP consistently achieves either superior or equivalent performance in both convergence behavior and generalization ability. 
In terms of convergence metrics, MEGP demonstrated faster and more stable fitness improvement, enhanced crossover efficiency, and superior diversity preservation, indicating that its ensemble-driven selection facilitates a more structured and efficient evolutionary trajectory.
In generalization metrics, MEGP configurations exhibited lower Log-Loss and more stable Precision, Recall, $F_{1}$ scores, and AUC reinforcing their ability to produce better-calibrated and more adaptable classifiers.

A key factor in MEGP’s effectiveness is its multi-view learning structure, where feature partitioning across populations ensures that evolutionary search operates on conditionally independent representations of the data. 
This structure promotes diversity within individual populations while enabling the ensemble to integrate complementary perspectives, improving generalization. 
The ability to maintain distinct yet cooperative evolutionary processes mitigates feature redundancy and overfitting, allowing MEGP to explore richer hypothesis spaces than traditional GP models constrained by a single feature representation.

These findings highlight the structural advantages of MEGP’s multi-population and multi-view framework. 
By maintaining inter-population diversity and cooperative evolution, MEGP reduces the risk of premature convergence, a limitation often observed in single-population GP. 
The ensemble fitness selection mechanism further refines evolutionary search by integrating isolated and collective performance criteria, ensuring that ensemble models benefit from both local adaptations and global synergy. 
This dual-fitness evaluation enhances solution diversity, stability, and adaptability, making MEGP well-suited for high-dimensional and complex problem landscapes where traditional GP may struggle.

Several research directions emerge from these findings. 
Adaptive population sizing and dynamic ensemble selection strategies could improve efficiency by reducing computational overhead while maintaining search flexibility.
Investigating alternative fitness evaluation criteria, such as entropy-based or information-theoretic measures, could further enhance ensemble robustness in highly multi-modal spaces. 
Additionally, extending MEGP to multi-objective optimization could leverage its multi-population design to explore trade-offs across conflicting objectives. 
Future work should also consider scalability in distributed and high-performance computing environments to enable large-scale applications where evolutionary models interact with real-time data streams. 
Hybridizing MEGP with deep learning architectures or other optimization heuristics could further enhance its ability to solve complex AI-driven optimization problems.

The integration of multi-view learning with cooperative coevolution introduces a structurally adaptive and scalable approach to genetic programming. 
This framework expands the applicability of evolutionary algorithms to diverse problem domains, from feature selection in machine learning to large-scale combinatorial optimization. 
By refining ensemble selection mechanisms and extending its adaptability to multi-objective and hybrid AI-driven paradigms, MEGP provides a powerful foundation for advancing genetic programming and evolutionary computation.

\bibliography{Refs}

\clearpage
\appendix
\renewcommand{\thefigure}{A.\arabic{figure}}
\renewcommand{\thetable}{A.\arabic{table}}
\renewcommand{\theequation}{A.\arabic{equation}}
\setcounter{figure}{0}
\setcounter{table}{0}
\setcounter{equation}{0}
\section*{Supplementary Material}
\input{SuppDoc_MultiViewGP}

\end{document}

%% file: flowchart.tex
\begin{figure*}[!htbp]
  \centering
  \subfloat[Evolutionary cycle]{\includegraphics[width=0.5\textwidth]{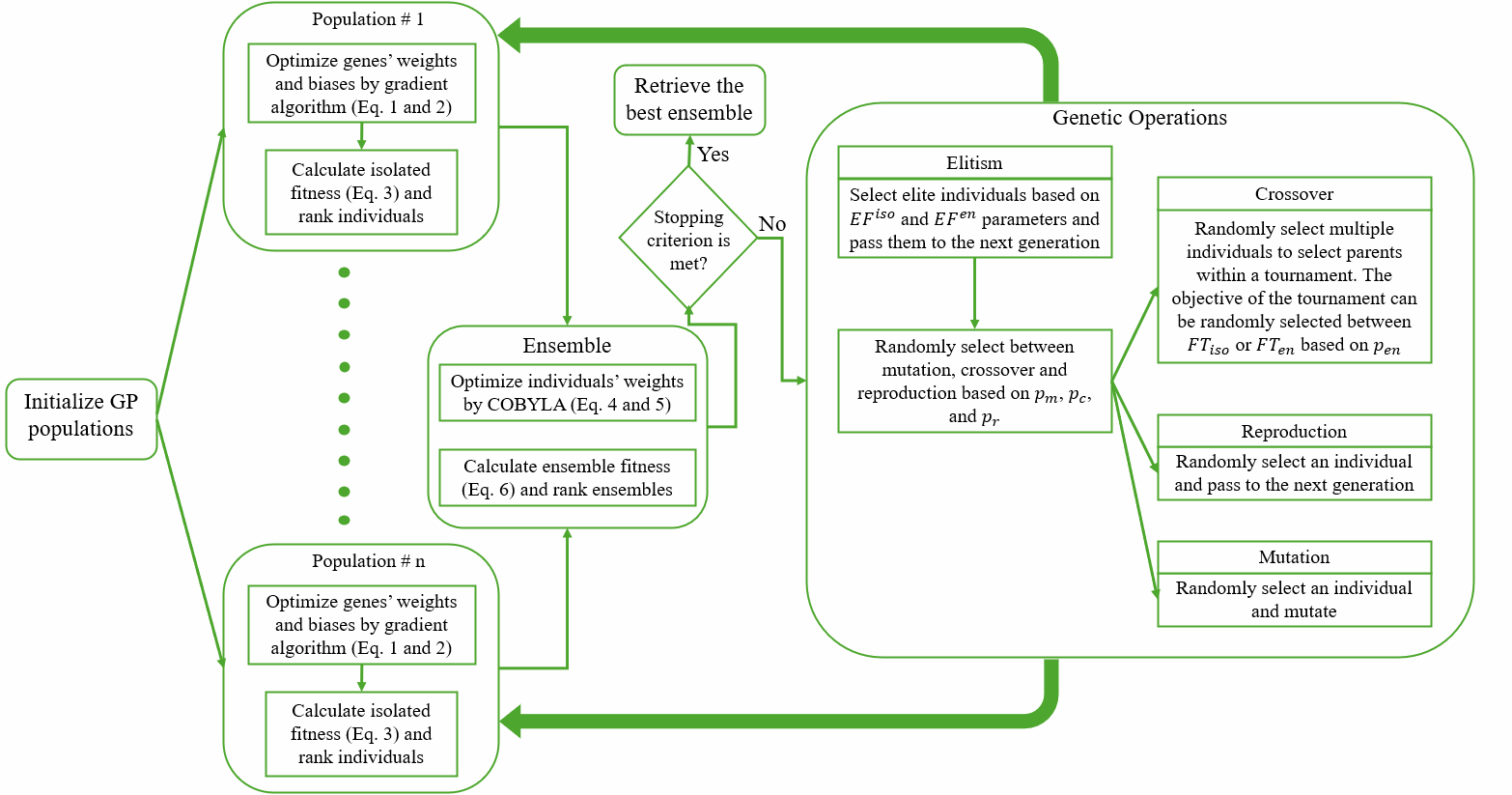}\label{fig:megp_flowchart_1}}%
\hfill
\subfloat[Fitness evaluation]{\includegraphics[width=0.5\textwidth]{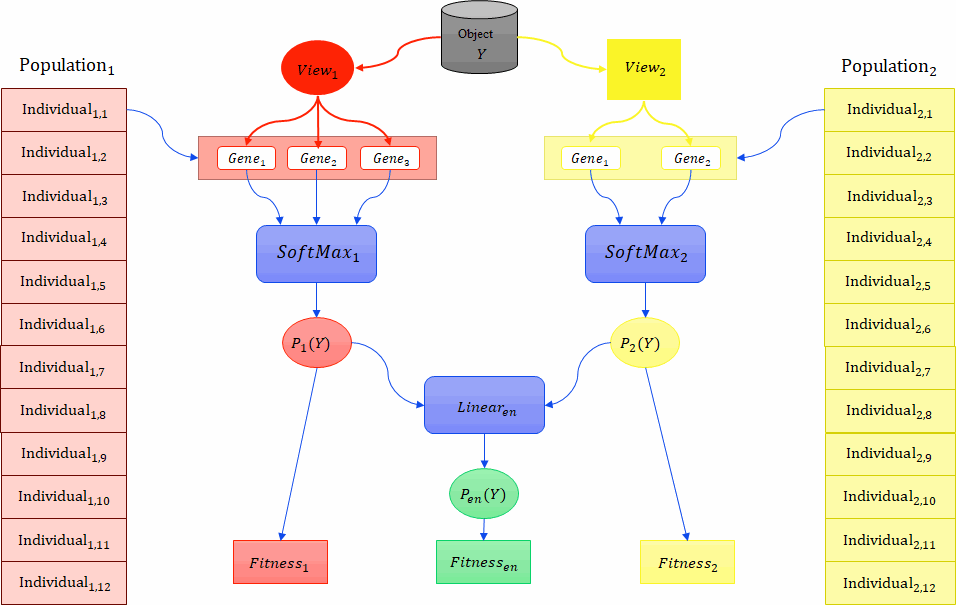}\label{fig:megp_flowchart_2}}%
  \caption{Overview of the Multi-Population Ensemble Genetic Programming (MEGP) framework. (a) Illustrates the evolutionary process, including initialization, population-specific optimization, genetic operations, and selection mechanisms. (b) Highlights the structured feature partitioning, population-specific classifier evolution, and ensemble fitness computation..}
  \label{fig:flowchart}
\end{figure*}

%% file: data.tex
\begin{table}[!ht]
\caption{Datasets' description.}
\label{tab:data}
\centering
\begin{tabular}{|c|c|c|c|}
\hline
\textbf{Dataset} & \textbf{\# Instances} & \textbf{\# Features} & \textbf{\# Classes}\\
\hline
APSF    & \(75,994\)    & \(170\)       & \(2\)  \\
ARWPM   & \(4,480\)     & \(533\)       & \(4\)  \\
GECR    & \(801\)       & \(20,531\)    & \(5\)  \\
GFE     & \(27,965\)    & \(301\)       & \(2\)  \\
GSAD    & \(13,910\)    & \(129\)       & \(6\)  \\
HAPT    & \(10,929\)    & \(561\)       & \(12\) \\
ISOLET  & \(7,797\)     & \(617\)       & \(26\) \\
PD      & \(756\)       & \(753\)       & \(2\)  \\
\hline
\end{tabular}
\end{table}

%% file: parameters.tex
\begin{table*}[!ht]
\caption{Parameter configuration for BGP and MEGP models.}
\label{tab:parameters}
\centering
\resizebox{\textwidth}{!}{%
\begin{tabular}{|l|c|c|c|c|c|c|}
\hline
\textbf{Parameter} & \textbf{BGP} & \textbf{MEGP\textsubscript{0}} & \textbf{MEGP\textsubscript{25}} & \textbf{MEGP\textsubscript{50}} & \textbf{MEGP\textsubscript{75}} & \textbf{MEGP\textsubscript{100}} \\
\hline
Number of Runs           & 30    & 30      & 30      & 30      & 30      & 30      \\
Number of Populations    & 1     & 2       & 2       & 2       & 2       & 2       \\
Population Size          & 60    & 30      & 30      & 30      & 30      & 30      \\
Max Generations          & 150   & 150     & 150     & 150     & 150     & 150     \\
Stall Generation         & 30    & 30      & 30      & 30      & 30      & 30      \\
Genes per Individual     & 10    & 10      & 10      & 10      & 10      & 10      \\
Max Tree Depth           & 10    & 10      & 10      & 10      & 10      & 10      \\
Initialization           & Half-and-Half & Half-and-Half & Half-and-Half & Half-and-Half & Half-and-Half & Half-and-Half \\
$EF^{iso}$ & 0.133 & 0.133 & 0.1 & 0.066 & 0.033 & 0 \\
$EF^{en}$  & N/A   & 0 & 0.033 & 0.066 & 0.1 & 0.133 \\
$p_{en}$             & N/A   & 0       & 0.25      & 0.5      & 0.75      & 1     \\
$p_{c}$            & 0.84    & 0.84      & 0.84      & 0.84      & 0.84      & 0.84      \\
$p_{m}$             & 0.14    & 0.14      & 0.14      & 0.14      & 0.14      & 0.14      \\
$p_{r}$         & 0.02     & 0.02       & 0.02       & 0.02       & 0.02       & 0.02       \\
Constant Range                        & [-10, 10] & [-10, 10] & [-10, 10] & [-10, 10] & [-10, 10] & [-10, 10] \\
Functional nodes		& $+$ $-$ $/$ $\times$ & $+$ $-$ $/$ $\times$ & $+$ $-$ $/$ $\times$ & $+$ $-$ $/$ $\times$ & $+$ $-$ $/$ $\times$ & $+$ $-$ $/$ $\times$ \\
Batch Size                            & Data size/50 & Data size/50 & Data size/50 & Data size/50 & Data size/50 & Data size/50 \\
Epochs                                & 1000  & 1000    & 1000    & 1000    & 1000    & 1000    \\
Learning Rate                         & 0.001 & 0.001   & 0.001   & 0.001   & 0.001   & 0.001   \\
\hline
\end{tabular}
}
\end{table*}

%% file: fig_congen_paper.tex
\begin{figure*}[ht] 
\centering
\subfloat[GFE]{\includegraphics[width=0.5\textwidth]{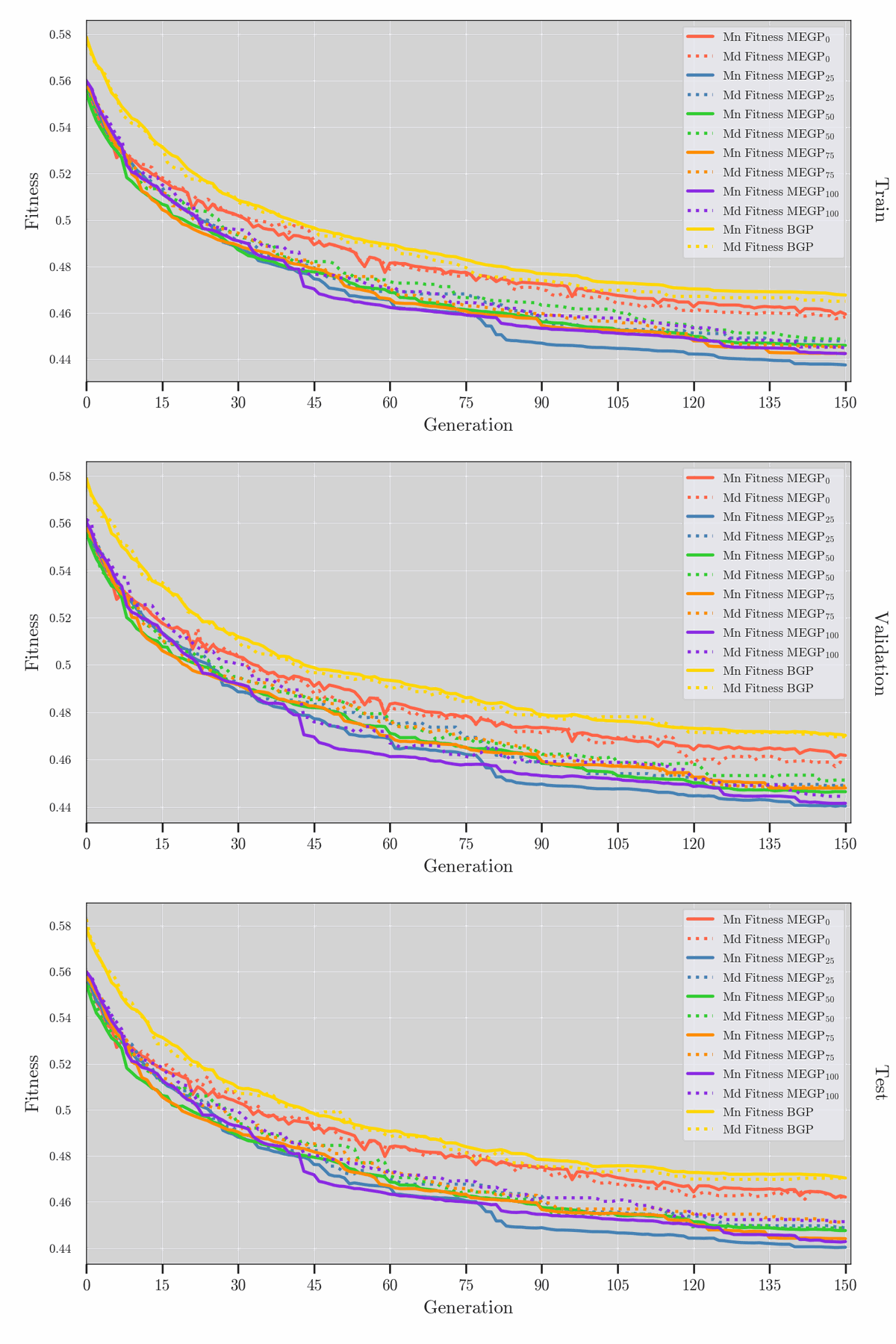}\label{fig:gfe_gen}}
\hfill
\subfloat[HAPT]{\includegraphics[width=0.5\textwidth]{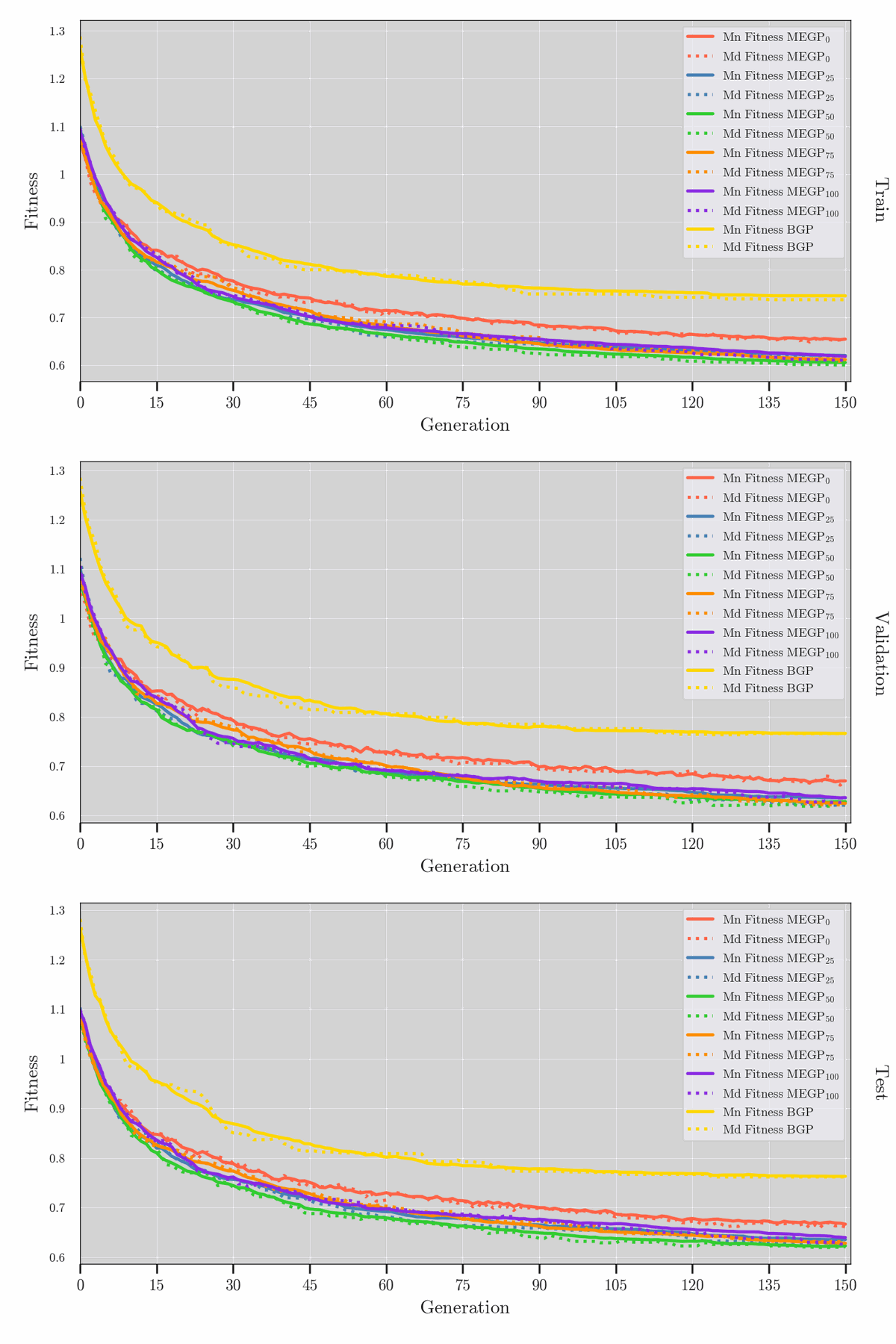}\label{fig:hapt_gen}}%
\hfill

\caption{The progression of mean and median best fitness values over 150 generations for Baseline Genetic Programming (BGP) and different Multi-Population Ensemble Genetic Programming (MEGP\textsubscript{0}, MEGP\textsubscript{25}, MEGP\textsubscript{50}, MEGP\textsubscript{75}, and MEGP\textsubscript{100}) configurations on: (a) GFE, and (b) HAPT datasets.}

\label{fig:fig_congen_paper}
\end{figure*}

%% file: winners_convergence.tex
\begin{figure*}[!htb]
\centering
\subfloat[$FT_{50}$]{\includegraphics[width=0.19\textwidth]{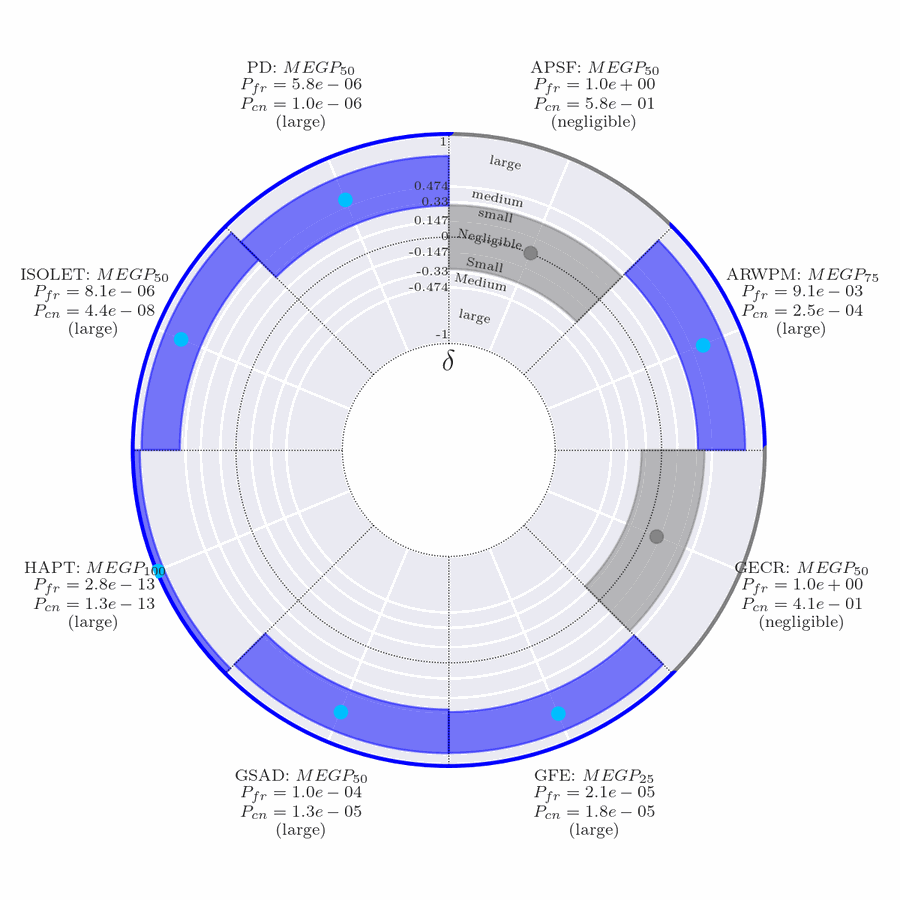}\label{fig:ft50_win}}\hfill
\subfloat[$FT_{100}$]{\includegraphics[width=0.19\textwidth]{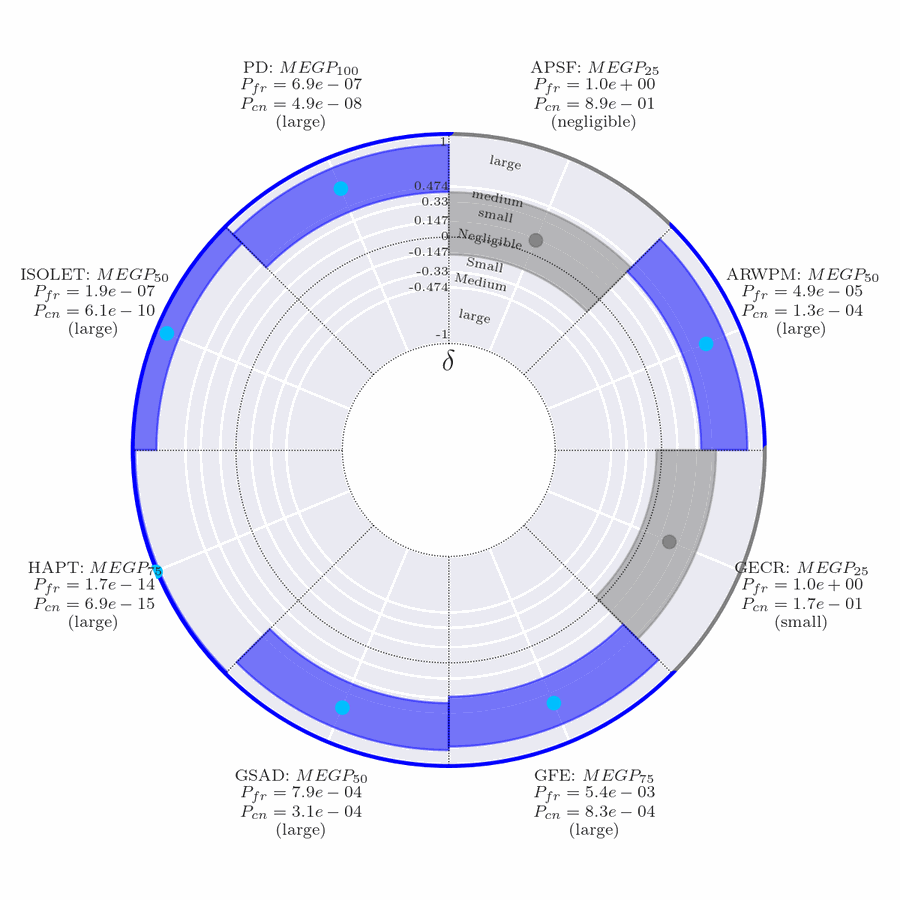}\label{fig:ft100_win}}\hfill
\subfloat[$FT_{150}$]{\includegraphics[width=0.19\textwidth]{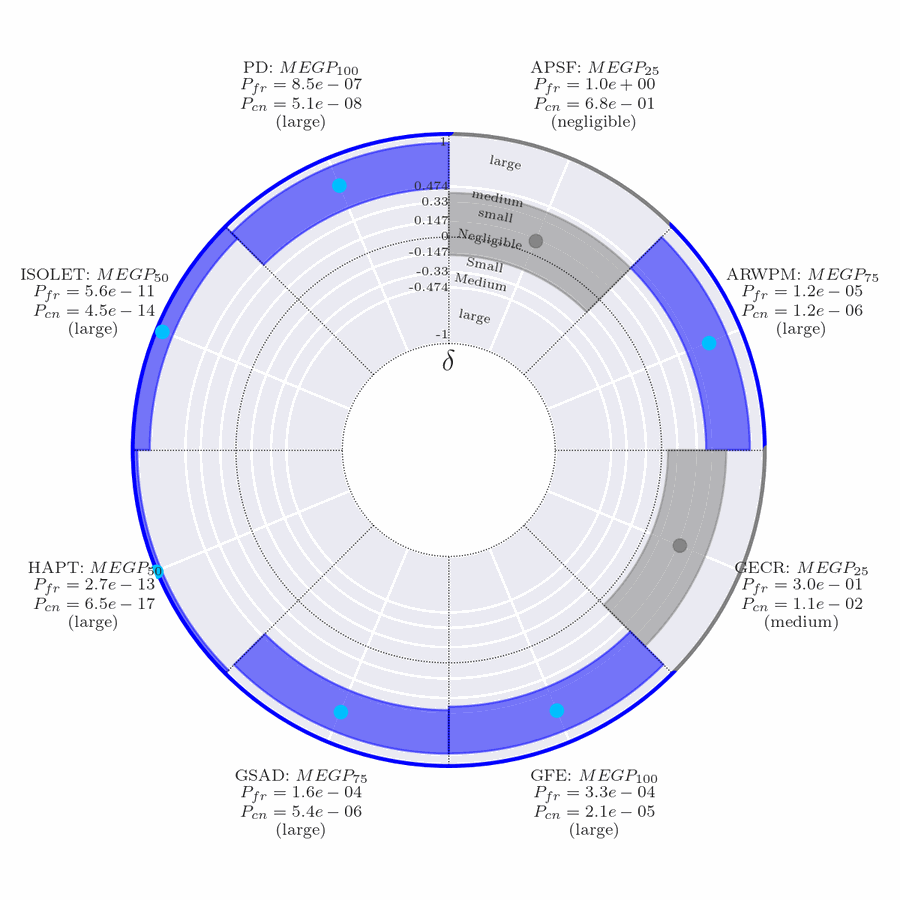}\label{fig:ft150_win}}\hfill
\subfloat[$FT_{all}$]{\includegraphics[width=0.19\textwidth]{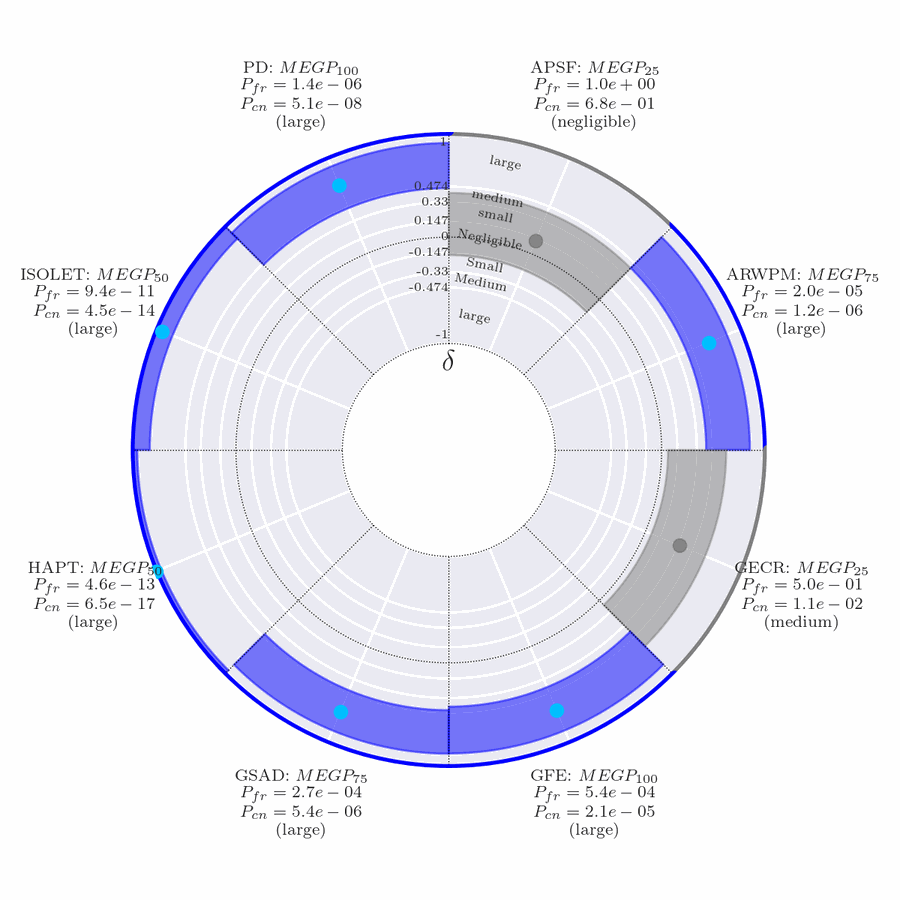}\label{fig:ftall_win}}\hfill
\subfloat[Entropy]{\includegraphics[width=0.19\textwidth]{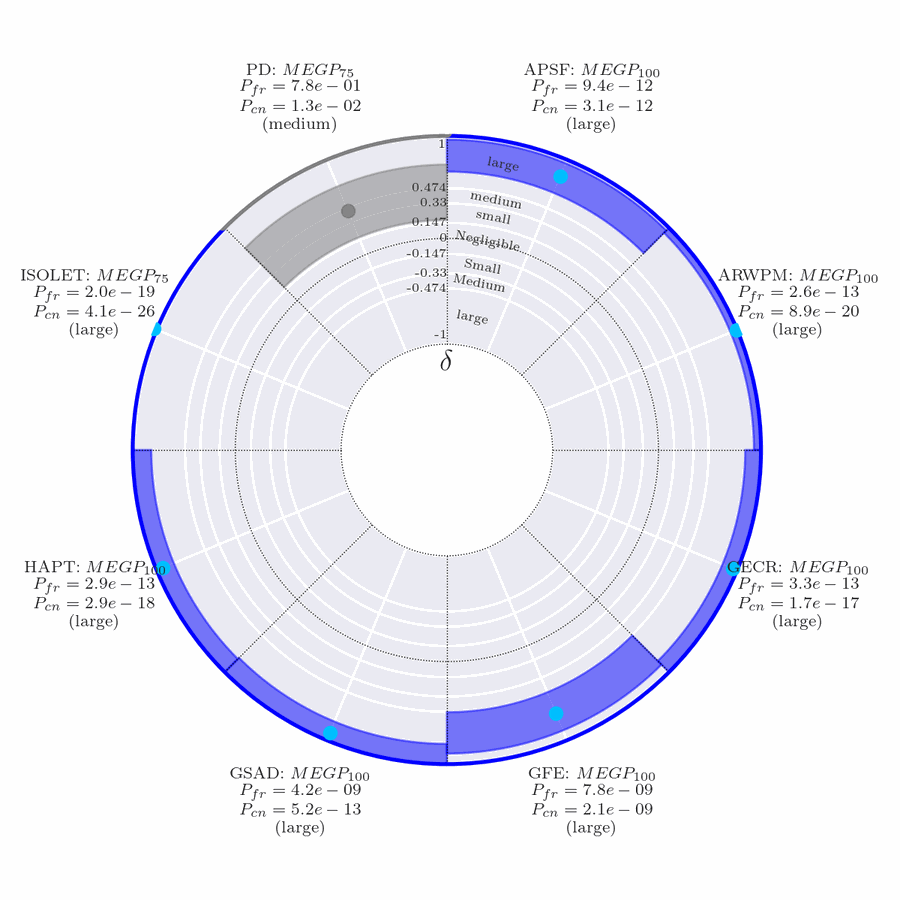}\label{fig:entropy_win}}

\vspace{1ex}
\subfloat[$CR_{50}$]{\includegraphics[width=0.19\textwidth]{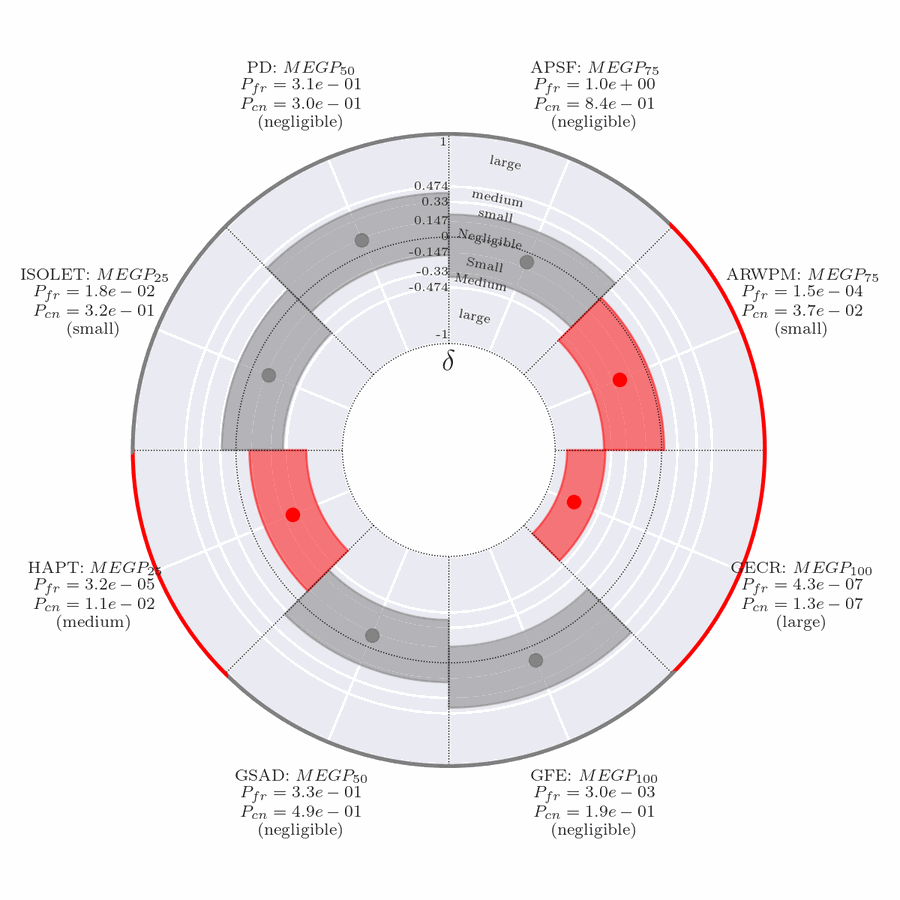}\label{fig:cr50_win}}\hfill
\subfloat[$CR_{100}$]{\includegraphics[width=0.19\textwidth]{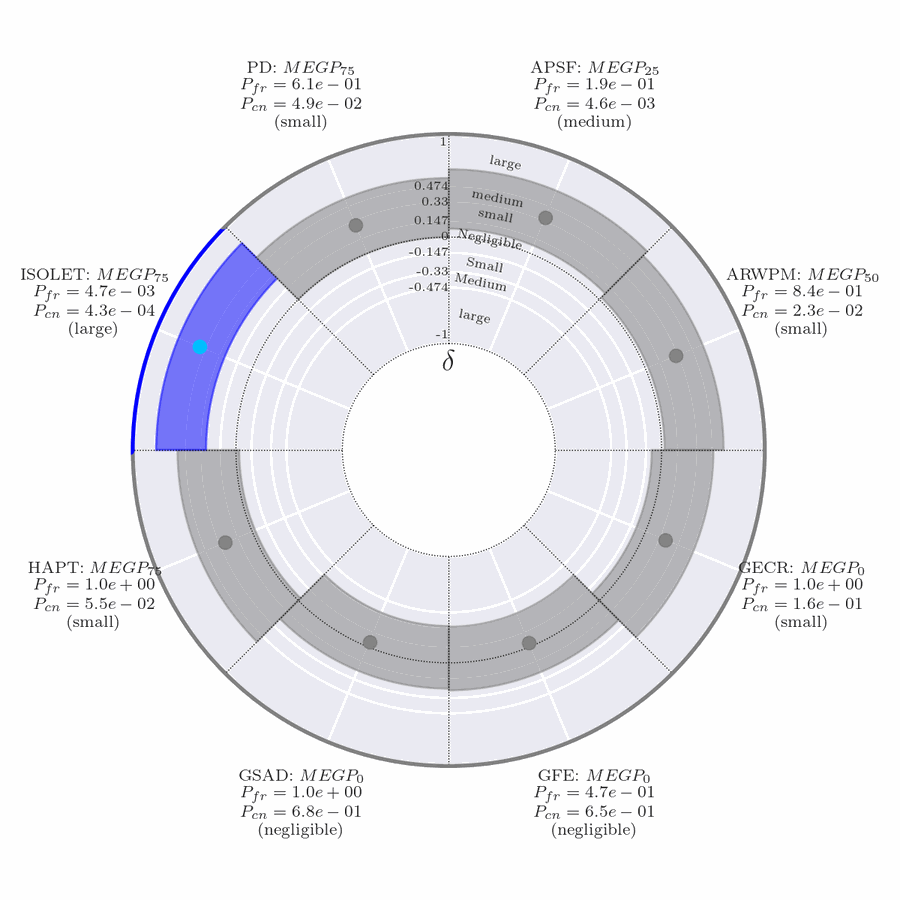}\label{fig:cr100_win}}\hfill
\subfloat[$CR_{150}$]{\includegraphics[width=0.19\textwidth]{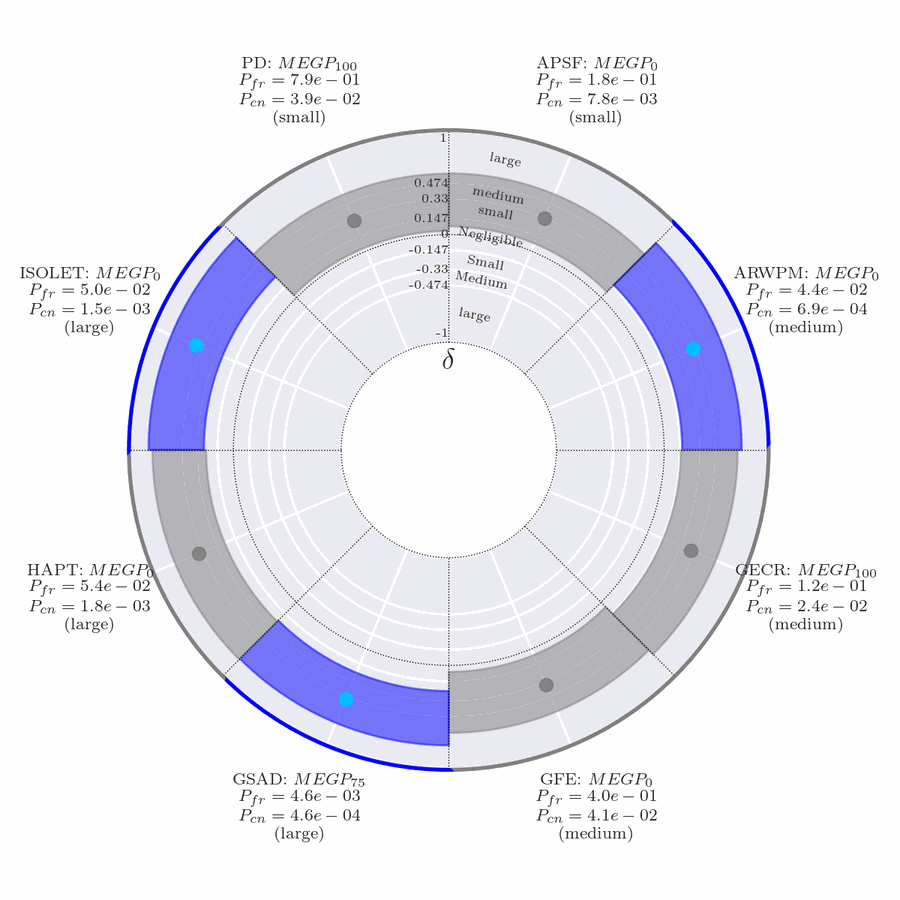}\label{fig:cr150_win}}\hfill
\subfloat[$CR_{all}$]{\includegraphics[width=0.19\textwidth]{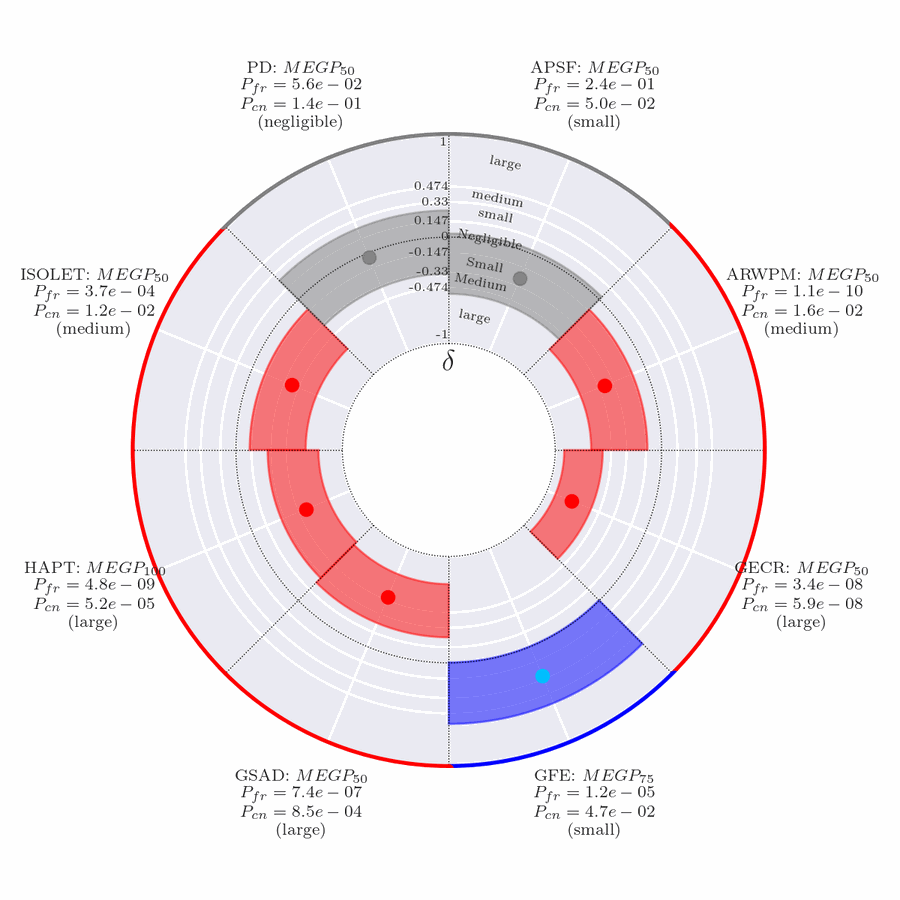}\label{fig:crall_win}}
\hspace{0.2\textwidth}

\vspace{1ex}
\subfloat[$CCR_{50}$]{\includegraphics[width=0.19\textwidth]{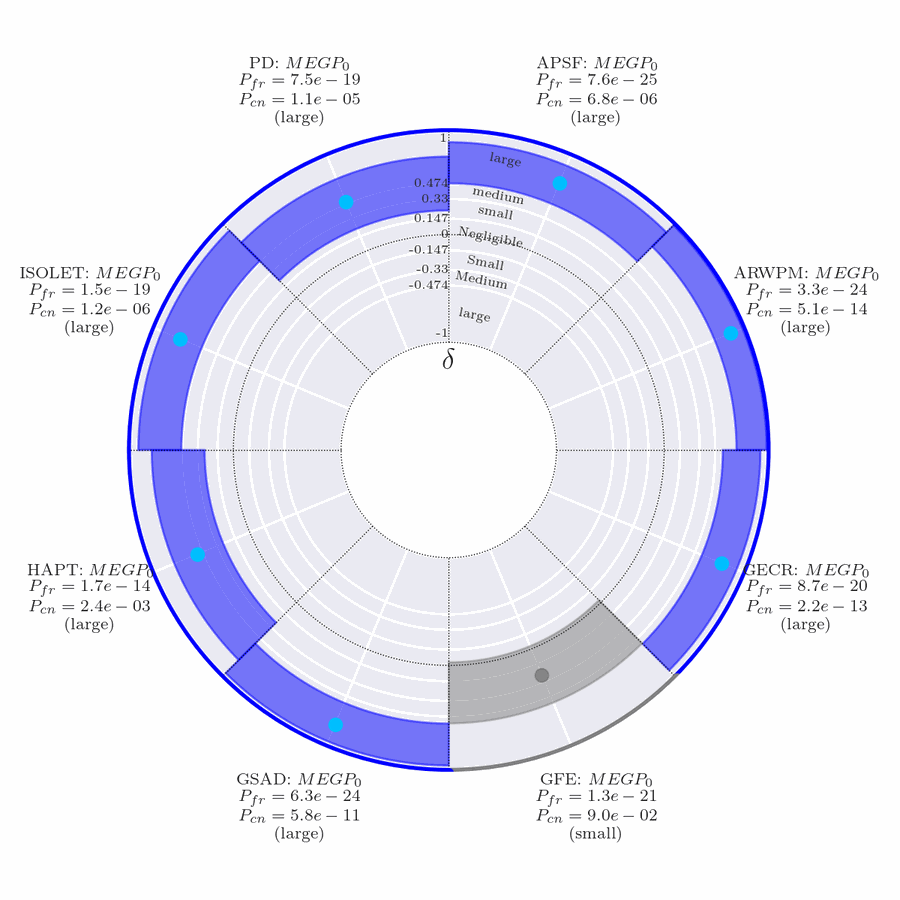}\label{fig:ccr50_win}}\hfill
\subfloat[$CCR_{100}$]{\includegraphics[width=0.19\textwidth]{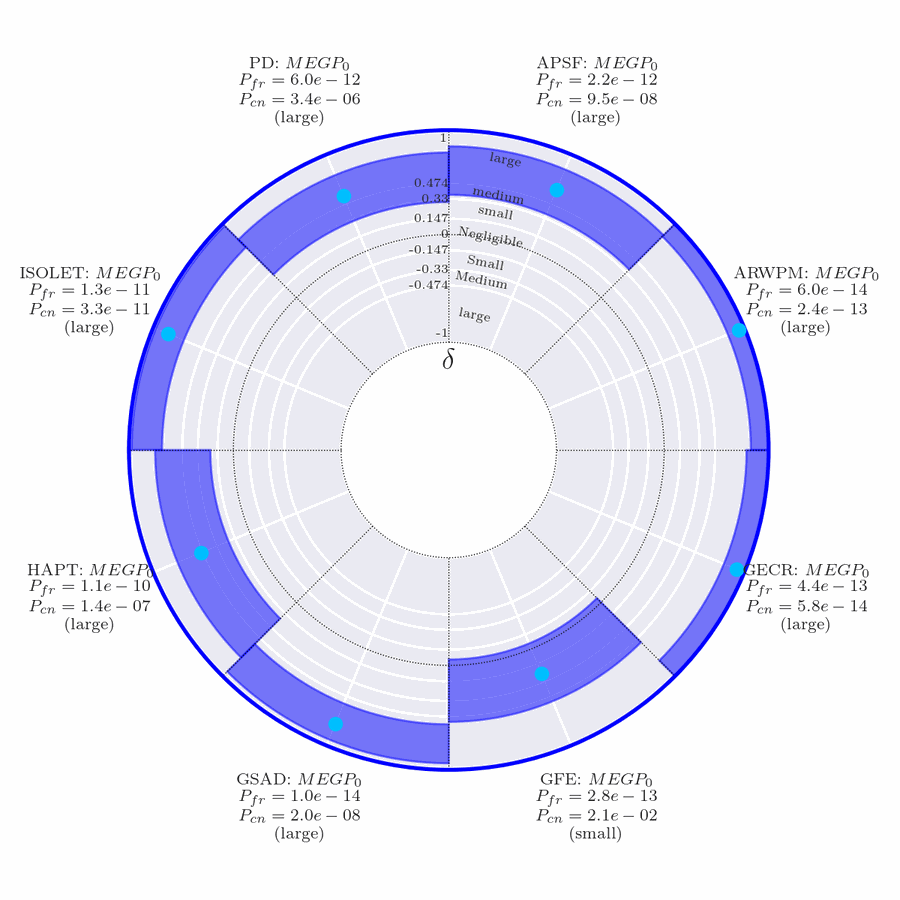}\label{fig:ccr100_win}}\hfill
\subfloat[$CCR_{150}$]{\includegraphics[width=0.19\textwidth]{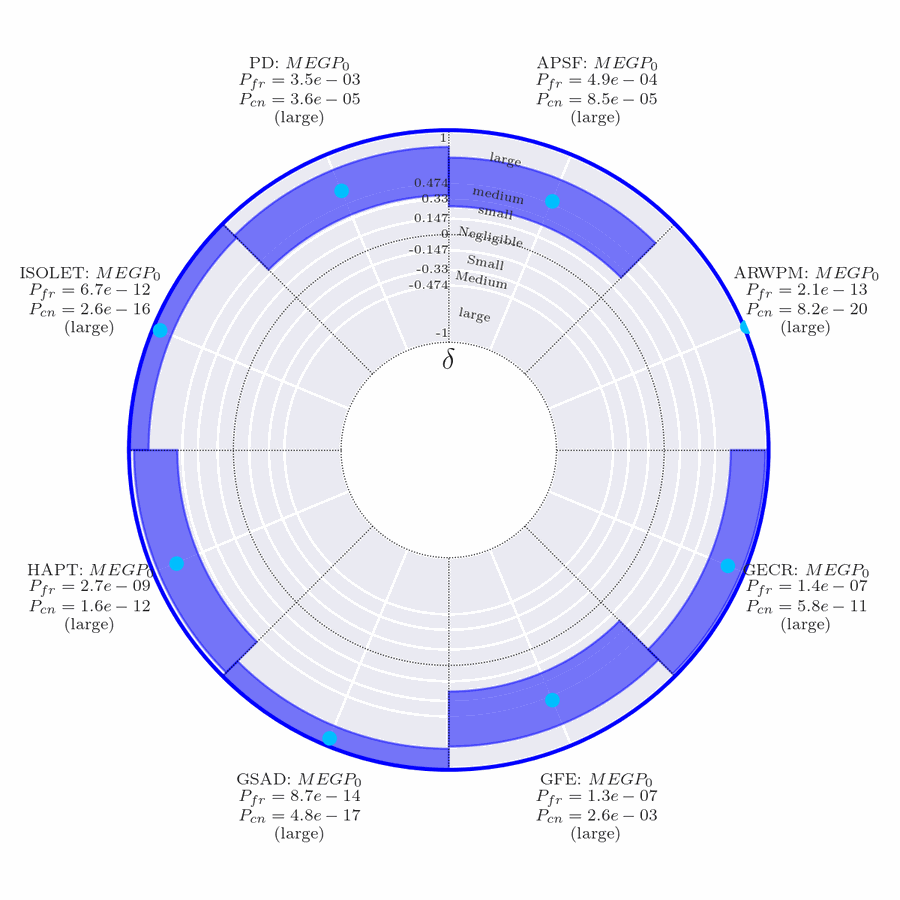}\label{fig:ccr150_win}}\hfill
\subfloat[$CCR_{all}$]{\includegraphics[width=0.19\textwidth]{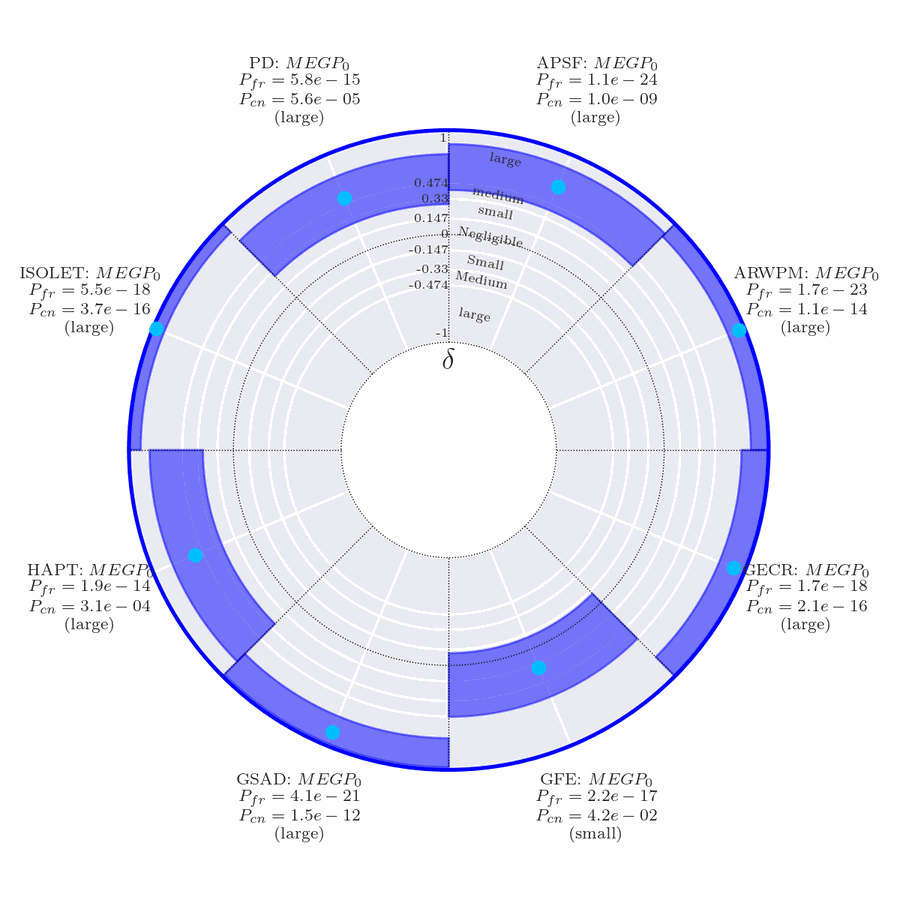}\label{fig:ccrall_win}}\hfill
\subfloat[Running Time]{\includegraphics[width=0.19\textwidth]{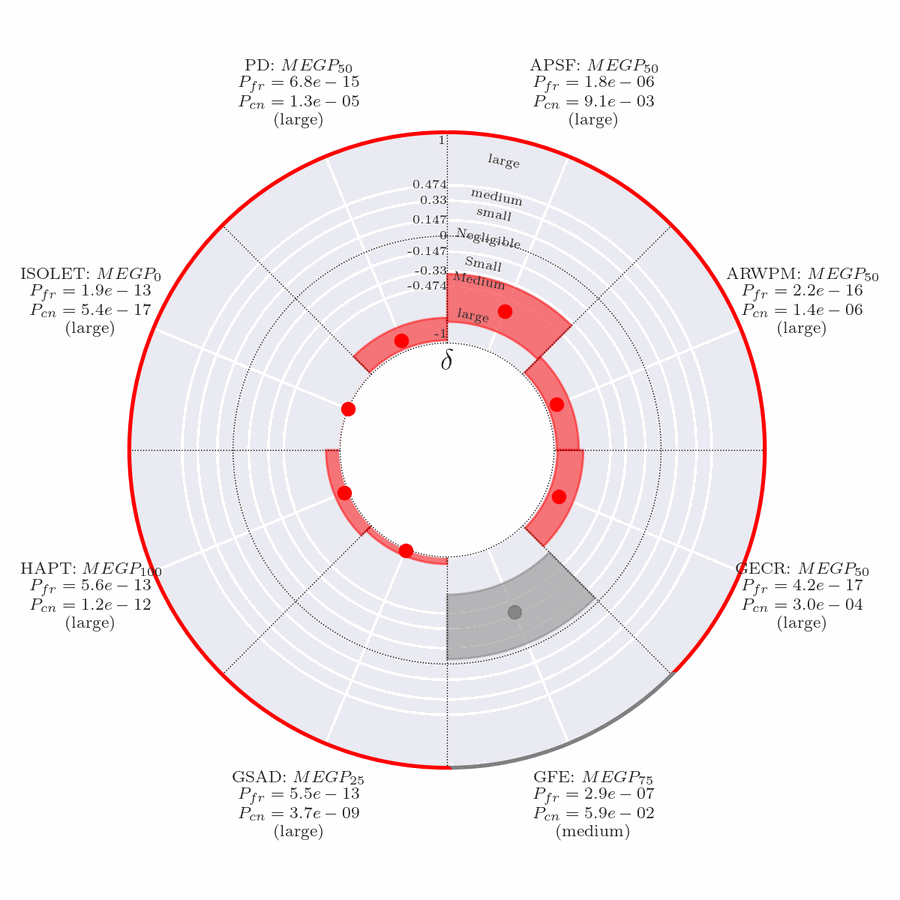}\label{fig:time_win}}

\caption{The figure illustrates the best performing MEGP models in comparison to the BGP model based on convergence criteria (FT, CR, CCR, Entropy, and Running time), with the segments representing the 95\% confidence interval of Cliff's $\delta$ (center point). Grey segments indicate cases where either $P_{fr}>0.05$ or $P_{cn}>0.05$, suggesting no significant difference from the BGP. Blue segments denote instances where the MEGP model outperforms the BGP ($P_{fr}<0.05$, $P_{cn}<0.05$, and $\delta>0$), while red segments indicate that the BGP model outperforms the corresponding MEGP model ($P_{fr}<0.05$, $P_{cn}<0.05$, and $\delta<0$).}
\label{fig:winners_convergence}
\end{figure*}

%% file: wtl_convergence.tex
\begin{table*}
\centering
\caption[Summary of statistical comparison for training data obtained from MEGP and BGP Runs.]{Summary of statistical comparison for training data obtained from MEGP and BGP Runs. W, T, and L denote win, tie, and loss based on adjusted Friedman and Conover's p-values.}
\label{tab:wtl_convergence}
\begin{tabular}{cccccc}
\hline
\multicolumn{6}{c}{ (Win - Tie - Loss)}\\
\hline
Metric & $MEGP_{0}$ & $MEGP_{25}$ & $MEGP_{50}$ & $MEGP_{75}$ & $MEGP_{100}$ \\
\hline
$FT_{50}$ & 3 - 5 - 0 & 5 - 3 - 0 & 6 - 2 - 0 & 6 - 2 - 0 & 6 - 2 - 0 \\
$CR_{50}$ & 0 - 3 - 5 & 0 - 5 - 3 & 0 - 5 - 3 & 0 - 5 - 3 & 0 - 4 - 4 \\
$CCR_{50}$ & 7 - 1 - 0 & 3 - 5 - 0 & 0 - 2 - 6 & 0 - 0 - 8 & 0 - 0 - 8 \\
$FT_{100}$ & 2 - 6 - 0 & 6 - 2 - 0 & 6 - 2 - 0 & 6 - 2 - 0 & 6 - 2 - 0 \\
$CR_{100}$ & 1 - 7 - 0 & 0 - 8 - 0 & 1 - 7 - 0 & 1 - 7 - 0 & 1 - 7 - 0 \\
$CCR_{100}$ & 8 - 0 - 0 & 7 - 1 - 0 & 2 - 5 - 1 & 0 - 5 - 3 & 0 - 1 - 7 \\
$FT_{150}$ & 4 - 4 - 0 & 6 - 2 - 0 & 6 - 2 - 0 & 6 - 2 - 0 & 6 - 2 - 0 \\
$CR_{150}$ & 3 - 5 - 0 & 2 - 6 - 0 & 2 - 6 - 0 & 2 - 6 - 0 & 2 - 6 - 0 \\
$CCR_{150}$ & 8 - 0 - 0 & 7 - 1 - 0 & 6 - 1 - 1 & 5 - 2 - 1 & 2 - 5 - 1 \\
$FT_{all}$ & 4 - 4 - 0 & 6 - 2 - 0 & 6 - 2 - 0 & 6 - 2 - 0 & 6 - 2 - 0 \\
$CR_{all}$ & 0 - 2 - 6 & 0 - 3 - 5 & 0 - 3 - 5 & 1 - 2 - 5 & 0 - 3 - 5 \\
$CCR_{all}$ & 8 - 0 - 0 & 4 - 4 - 0 & 1 - 3 - 4 & 0 - 2 - 6 & 0 - 0 - 8 \\
Entropy & 4 - 4 - 0 & 5 - 3 - 0 & 7 - 1 - 0 & 7 - 1 - 0 & 7 - 1 - 0 \\
Running Time (sec) & 0 - 0 - 8 & 0 - 0 - 8 & 0 - 1 - 7 & 0 - 1 - 7 & 0 - 0 - 8 \\
\hline
\end{tabular}
\end{table*}

%% file: rank_gen_megp_100.tex
\begin{figure*}[ht] 
\centering
\includegraphics[width=\textwidth]{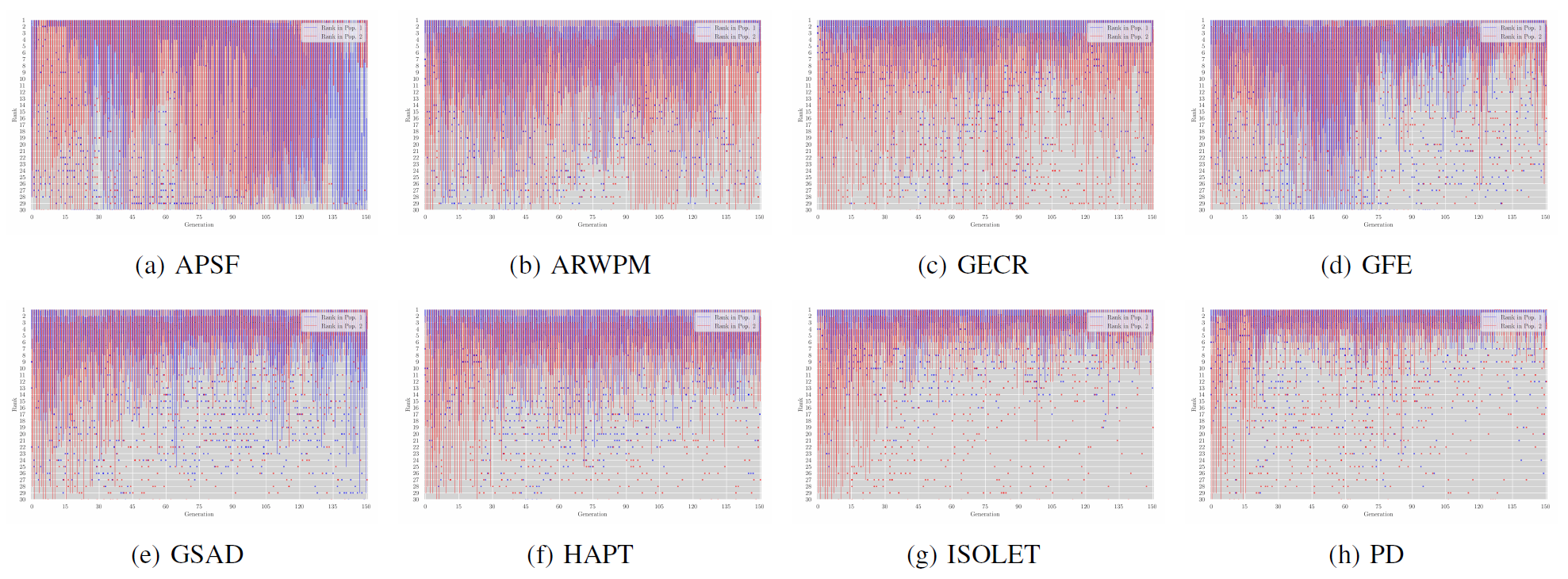}

\caption[Boxplots of isolated rankings for best ensemble individuals in MEGP\textsubscript{100}]{Boxplots of isolated rankings for individuals forming the best ensemble within their respective populations across generations (0–150) over 30 runs of MEGP\textsubscript{100} for the benchmark datasets: (a) APSF, (b) ARWPM, (c) GECR, (d) GFE, (e) GSAD, (f) HAPT, (g) ISOLET, and (h) PD. The y-axis represents the ranking position, where lower values indicate better placement, while the x-axis denotes the generation number.}

\label{fig:rank_gen_megp_100}
\end{figure*}

%% file: winners_generalization.tex
\begin{figure*}[ht] 
\centering
\subfloat[Log-Loss]{\includegraphics[width=0.33\textwidth]{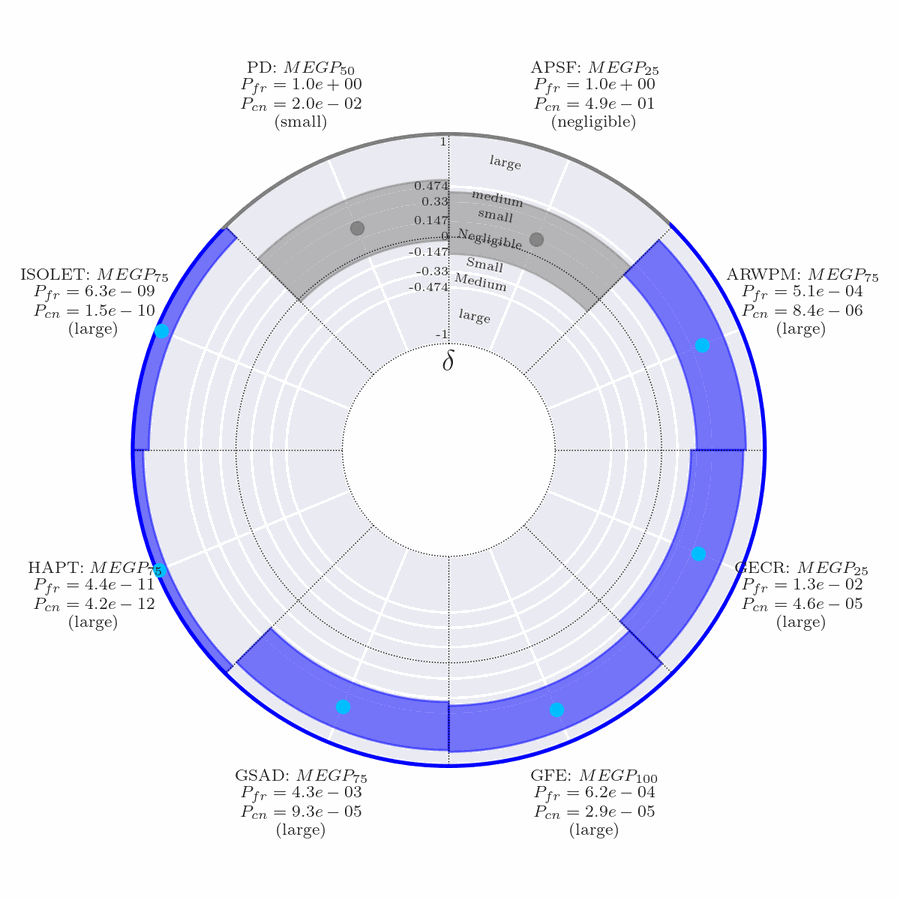}\label{fig:loss_win}}%
\hfill
\subfloat[Precision]{\includegraphics[width=0.33\textwidth]{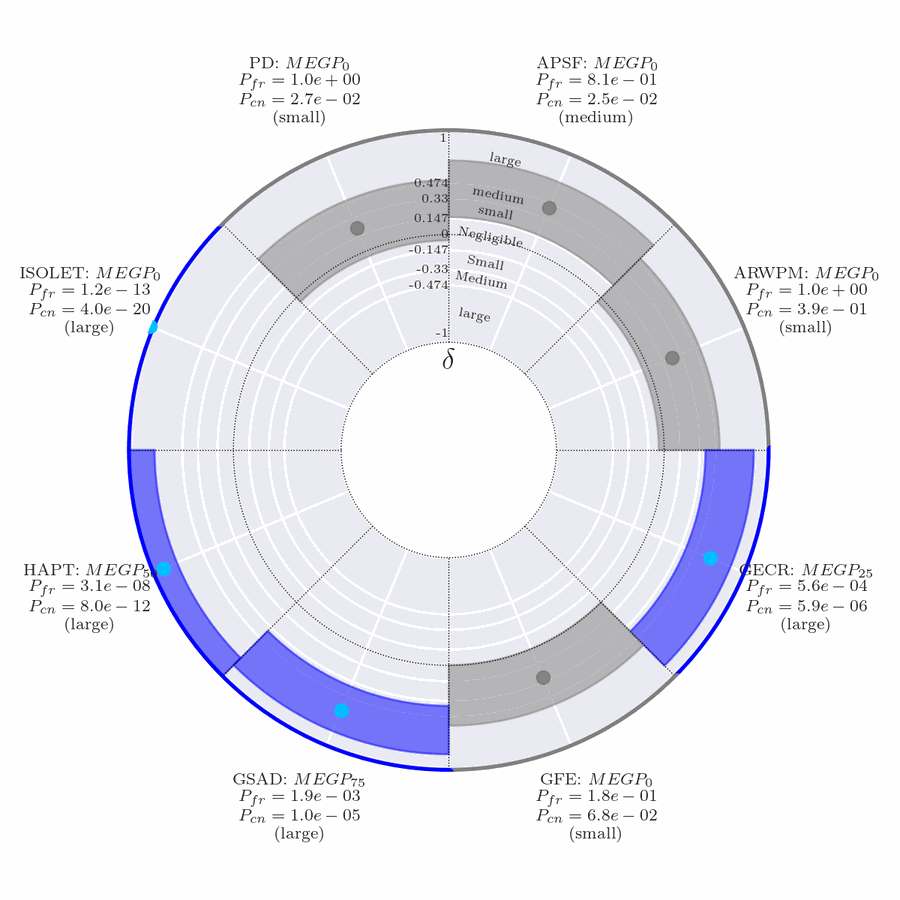}\label{fig:precision_win}}%
\hfill
\subfloat[Recall]{\includegraphics[width=0.33\textwidth]{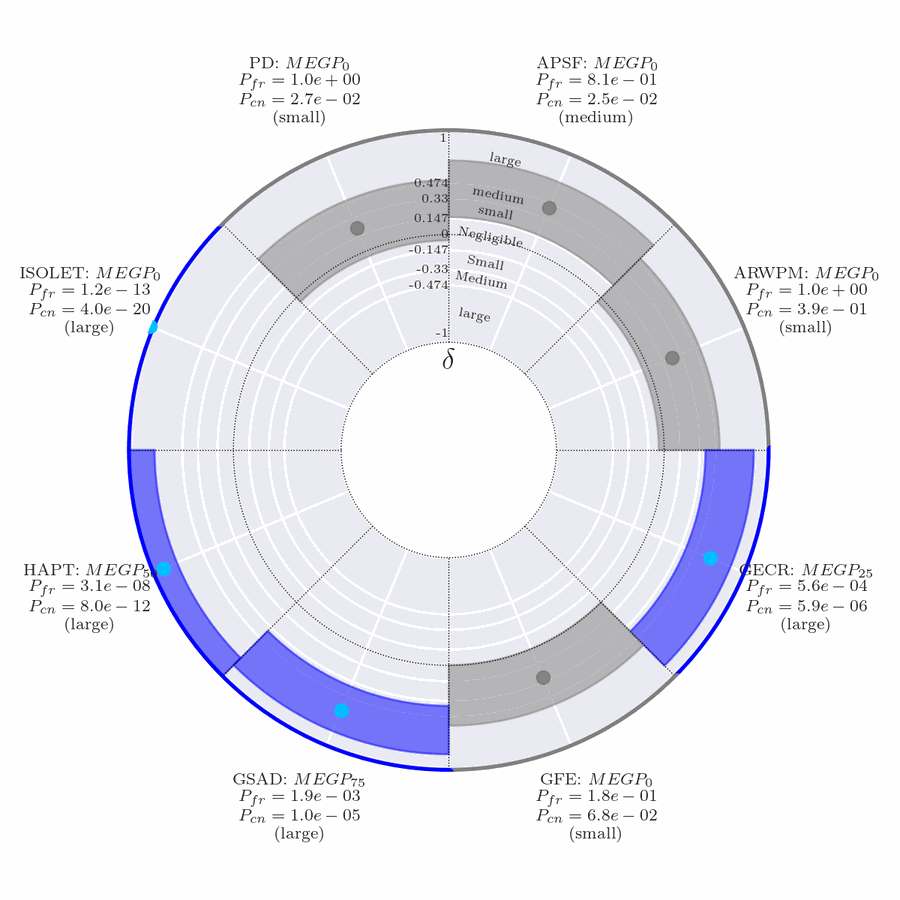}\label{fig:recall_win}}

\subfloat[$F_{1}$]{\includegraphics[width=0.33\textwidth]{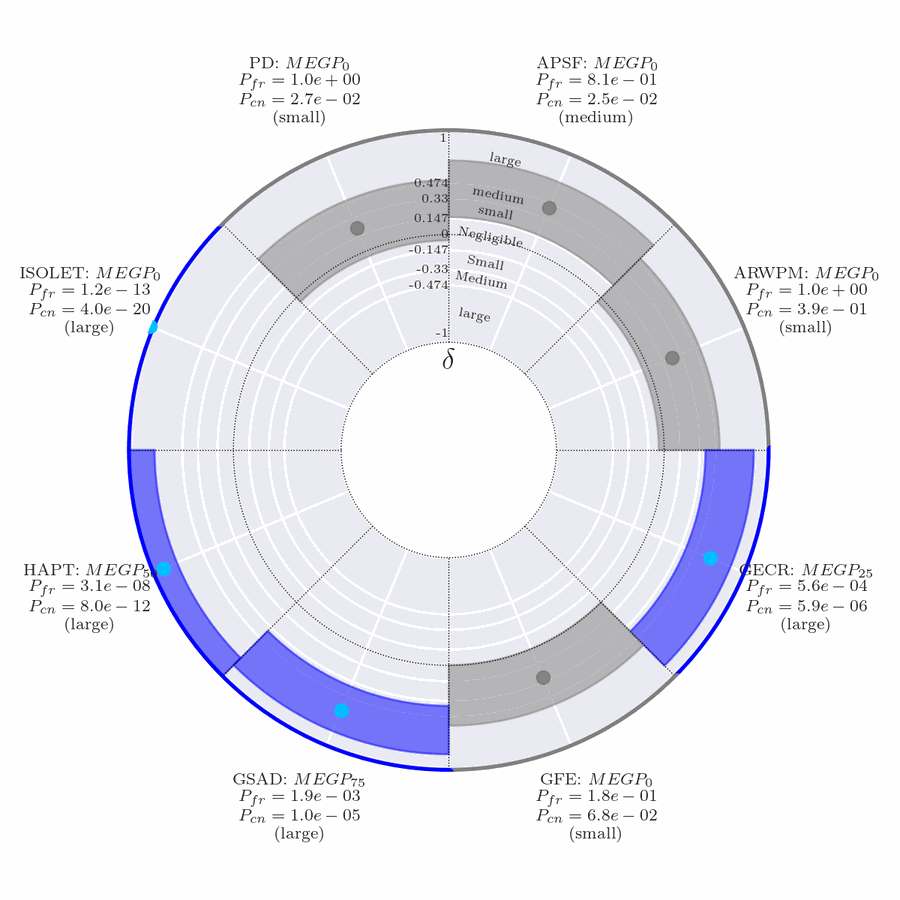}\label{fig:f1_win}}%
\hfill
\subfloat[AUC]{\includegraphics[width=0.33\textwidth]{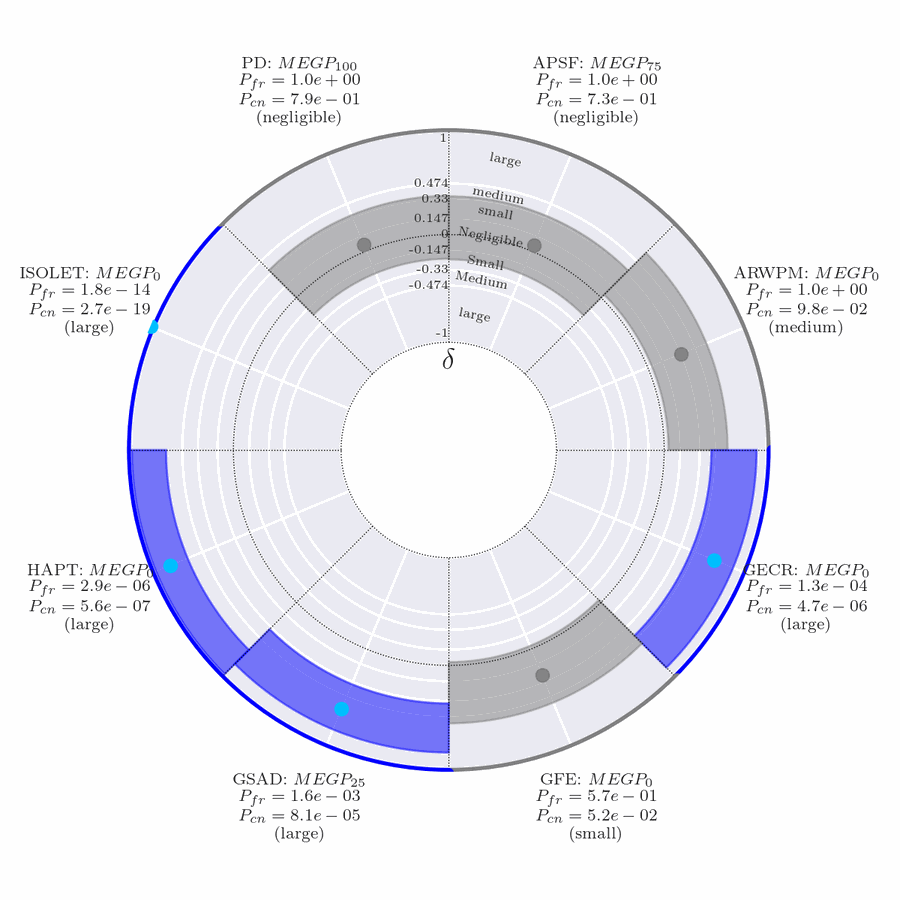}\label{fig:auc_win}}%

\caption{The figure illustrates the best performing MEGP models in comparison to the BGP model based on generalization criteria (Log-Loss, Precision, Recall, $F_{1}$ score, and AUC), with the segments representing the 95\% confidence interval of Cliff's $\delta$ (center point). Grey segments indicate cases where either $P_{fr}>0.05$ or $P_{cn}>0.05$, suggesting no significant difference from the BGP. Blue segments denote instances where the MEGP model outperforms the BGP ($P_{fr}<0.05$, $P_{cn}<0.05$, and $\delta>0$), while red segments indicate that the BGP model outperforms the corresponding MEGP model ($P_{fr}<0.05$, $P_{cn}<0.05$, and $\delta<0$).}

\label{fig:winners_generalization}
\end{figure*}

%% file: wtl_generalization.tex
\begin{table*}
\centering
\caption[Summary of statistical comparison for testing data obtained from MEGP and BGP Runs.]{Summary of statistical comparison for testing data obtained from MEGP and BGP Runs. W, T, and L denote win, tie, and loss based on adjusted Friedman and Conover's p-values.}
\label{tab:wtl_generalization}
\begin{tabular}{cccccc}
\hline
\multicolumn{6}{c}{ (Win - Tie - Loss)}\\
\hline
Metric & $MEGP_{0}$ & $MEGP_{25}$ & $MEGP_{50}$ & $MEGP_{75}$ & $MEGP_{100}$ \\
\hline
Log-Loss & 4 - 4 - 0 & 6 - 2 - 0 & 6 - 2 - 0 & 6 - 2 - 0 & 6 - 2 - 0 \\
Precision & 4 - 4 - 0 & 4 - 4 - 0 & 3 - 5 - 0 & 4 - 4 - 0 & 4 - 4 - 0 \\
Recall & 4 - 4 - 0 & 4 - 4 - 0 & 3 - 5 - 0 & 4 - 4 - 0 & 4 - 4 - 0 \\
$F_{1}$ & 4 - 4 - 0 & 4 - 4 - 0 & 3 - 5 - 0 & 4 - 4 - 0 & 4 - 4 - 0 \\
AUC & 4 - 4 - 0 & 4 - 4 - 0 & 4 - 4 - 0 & 4 - 4 - 0 & 4 - 4 - 0 \\
\hline
\end{tabular}
\end{table*}

%% file: SuppDoc_MultiViewGP.tex
\FloatBarrier
\input{fig_congen}
\input{table_friedman_convergence}

\FloatBarrier
\input{box_ft50}
\input{con_ft50}
\FloatBarrier
\input{cliff_ft50}
\input{wtl_ft50}

\FloatBarrier
\input{box_cr50}
\input{con_cr50}
\FloatBarrier
\input{cliff_cr50}
\input{wtl_cr50}

\FloatBarrier
\input{box_ccr50}

\input{con_ccr50}
\FloatBarrier
\input{cliff_ccr50}
\input{wtl_ccr50}
\FloatBarrier
\input{box_ft100}
\input{con_ft100}
\FloatBarrier
\input{cliff_ft100}
\input{wtl_ft100}

\FloatBarrier
\input{box_cr100}
\input{con_cr100}
\FloatBarrier
\input{cliff_cr100}
\input{wtl_cr100}

\FloatBarrier
\input{box_ccr100}
\input{con_ccr100}
\FloatBarrier
\input{cliff_ccr100}
\input{wtl_ccr100}

\FloatBarrier
\input{box_ft150}
\input{con_ft150}
\FloatBarrier
\input{cliff_ft150}
\input{wtl_ft150}

\FloatBarrier
\input{box_cr150}
\input{con_cr150}
\FloatBarrier
\input{cliff_cr150}
\input{wtl_cr150}

\FloatBarrier
\input{box_ccr150}
\input{con_ccr150}
\FloatBarrier
\input{cliff_ccr150}
\input{wtl_ccr150}

\FloatBarrier
\input{box_ftall}
\input{con_ftall}
\FloatBarrier
\input{cliff_ftall}
\input{wtl_ftall}

\FloatBarrier
\input{box_crall}
\input{con_crall}
\FloatBarrier
\input{cliff_crall}
\input{wtl_crall}

\FloatBarrier
\input{box_ccrall}

\input{con_ccrall}
\FloatBarrier
\input{cliff_ccrall}
\input{wtl_ccrall}

\FloatBarrier
\input{box_entropy}
\input{con_entropy}
\FloatBarrier
\input{cliff_entropy}
\input{wtl_entropy}

\FloatBarrier
\input{box_time}
\input{con_time}
\FloatBarrier
\input{cliff_time}
\input{wtl_time}

\FloatBarrier
\input{rank_gen_megp_0}

\input{rank_gen_megp_25}

\input{rank_gen_megp_50}

\input{rank_gen_megp_75}

\input{rank_gen_megp_100}

\input{table_friedman_generalization}
\FloatBarrier

\input{box_loss}
\input{con_loss}
\FloatBarrier
\input{cliff_loss}
\input{wtl_loss}

\FloatBarrier

\input{box_precision}
\input{con_precision}
\FloatBarrier
\input{cliff_precision}
\input{wtl_precision}

\FloatBarrier

\input{box_recall}
\input{con_recall}
\FloatBarrier
\input{cliff_recall}
\input{wtl_recall}

\FloatBarrier

\input{box_f1}

\input{con_f1}
\FloatBarrier
\input{cliff_f1}
\input{wtl_f1}

\FloatBarrier

\input{box_auc}

\input{con_auc}
\FloatBarrier
\input{cliff_auc}
\input{wtl_auc}

\FloatBarrier

%% file: fig_congen.tex
\begin{figure*}[hb] 
\centering
\includegraphics[width=\textwidth]{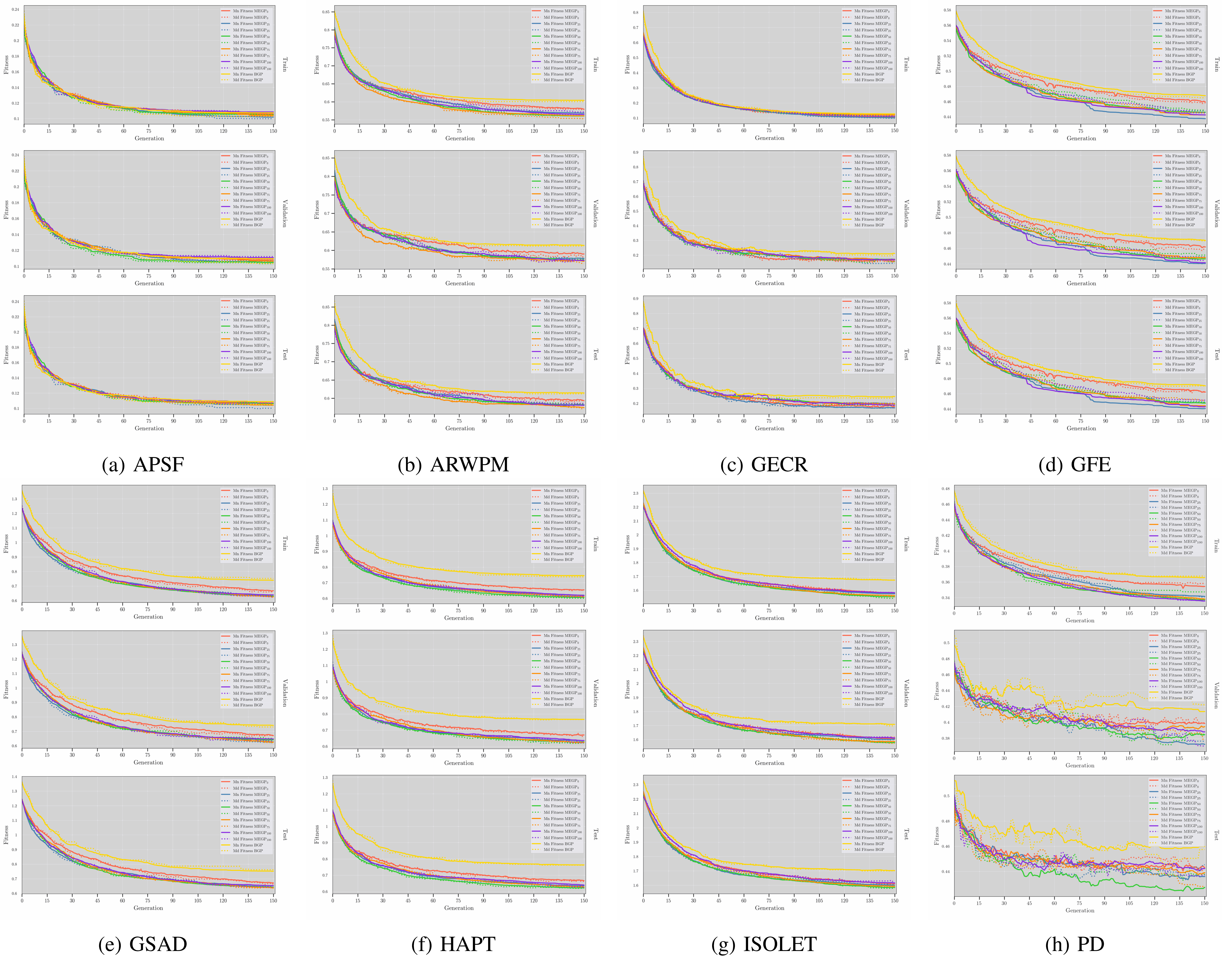}\label{fig:apsf_gen}%
\caption[The progression of mean and median best fitness values over 150 generations obtained for BGP and MEGP runs.]{The progression of mean and median best fitness values over 150 generations for Baseline Genetic Programming (BGP) and different Multi-Population Ensemble Genetic Programming (MEGP\textsubscript{0}, MEGP\textsubscript{25}, MEGP\textsubscript{50}, MEGP\textsubscript{75}, and MEGP\textsubscript{100}) configurations on: (a) APSF, (b) ARWPM, (c) GECR, (d) GFE, (e) GSAD, (f) HAPT, (g) ISOLET, and (h) PD datasets.}

\label{fig:fig_congen}
\end{figure*}

%% file: table_friedman_convergence.tex
\begin{sidewaystable*}[]
\centering
\caption{Comparison of Baseline GP ($\mathrm{MEGP}_{0}$) and Multi-population Ensemble GP (MEGP) models across various criteria. The table presents the mean and standard deviation values for each criterion, including Fitness (FT), Convergence Rate (CR), and Crossover Convergence Rate (CCR), calculated over generations 0-50, 51-100, 101-150, and across all generations (0-150). Additional metrics include the entropy of the final population and the total running time (in seconds). MEGP models are compared for different ensemble selection probabilities: 0\%, 25\%, 50\%, 75\%, and 100\%. Friedman test was performed to assess statistical significance, with p-values adjusted using the Bonferroni method.}
\label{tab:table_friedman_convergence}
\resizebox{0.8\textwidth}{!}{%
\begin{tabular}{cccccccccccccccc}
\hline
 &  & \multicolumn{14}{c}{Criteria} \\ \hline
\rowcolor[HTML]{C0C0C0}
\multicolumn{1}{c|}{\cellcolor[HTML]{C0C0C0}Dataset} &  \multicolumn{1}{c|}{\cellcolor[HTML]{C0C0C0}Model} &  $FT_{50}$ &  $CR_{50}$ &  \multicolumn{1}{c|}{\cellcolor[HTML]{C0C0C0}$CCR_{50}$} &  $FT_{100}$ &  $CR_{100}$ &  \multicolumn{1}{c|}{\cellcolor[HTML]{C0C0C0}$CCR_{100}$} &  $FT_{150}$ &  $CR_{150}$ &  \multicolumn{1}{c|}{\cellcolor[HTML]{C0C0C0}$CCR_{150}$} &  $FT_{all}$ &  $CR_{all}$ &  $CCR_{all}$ &  Entropy &  Running Time (sec) \\ \hline

\multicolumn{1}{c|}{} &  \multicolumn{1}{c|}{$\mathrm{MEGP}_{0}$} &  \multicolumn{1}{c|}{$0.118 \pm 0.012$} &  $0.002 \pm 0.001$ &  $0.129 \pm 0.016$ &  \multicolumn{1}{c|}{$0.109 \pm 0.012$} &  $0.0 \pm 0.0$ &  $0.099 \pm 0.034$ &  \multicolumn{1}{c|}{$0.106 \pm 0.013$} &  $0.0 \pm 0.0$ &  $0.062 \pm 0.052$ &  \multicolumn{1}{c|}{$0.106 \pm 0.013$} &  $0.001 \pm 0.0$ &  $0.124 \pm 0.014$ &  $0.246 \pm 0.07$ &  $3016.3 \pm 965.9$ \\ 

\multicolumn{1}{c|}{} &  \multicolumn{1}{c|}{\cellcolor[HTML]{C0C0C0}$\mathrm{MEGP}_{25}$} &  \multicolumn{1}{c|}{\cellcolor[HTML]{C0C0C0}$0.117 \pm 0.015$} &  \cellcolor[HTML]{C0C0C0}$0.002 \pm 0.001$ &  \cellcolor[HTML]{C0C0C0}$0.108 \pm 0.011$ &  \multicolumn{1}{c|}{\cellcolor[HTML]{C0C0C0}$0.105 \pm 0.015$} &  \cellcolor[HTML]{C0C0C0}$0.0 \pm 0.0$ &  \cellcolor[HTML]{C0C0C0}$0.087 \pm 0.03$ &  \multicolumn{1}{c|}{\cellcolor[HTML]{C0C0C0}$0.102 \pm 0.015$} &  \cellcolor[HTML]{C0C0C0}$0.0 \pm 0.0$ &  \cellcolor[HTML]{C0C0C0}$0.047 \pm 0.044$ &  \multicolumn{1}{c|}{\cellcolor[HTML]{C0C0C0}$0.102 \pm 0.015$} &  \cellcolor[HTML]{C0C0C0}$0.001 \pm 0.0$ &  \cellcolor[HTML]{C0C0C0}$0.106 \pm 0.008$ &  \cellcolor[HTML]{C0C0C0}$0.279 \pm 0.067$ &  \cellcolor[HTML]{C0C0C0}$2901.8 \pm 897.5$ \\ 

\multicolumn{1}{c|}{} &  \multicolumn{1}{c|}{$\mathrm{MEGP}_{50}$} &  \multicolumn{1}{c|}{$0.116 \pm 0.013$} &  $0.002 \pm 0.001$ &  $0.083 \pm 0.014$ &  \multicolumn{1}{c|}{$0.106 \pm 0.014$} &  $0.0 \pm 0.0$ &  $0.056 \pm 0.028$ &  \multicolumn{1}{c|}{$0.105 \pm 0.015$} &  $0.0 \pm 0.0$ &  $0.019 \pm 0.032$ &  \multicolumn{1}{c|}{$0.105 \pm 0.015$} &  $0.001 \pm 0.001$ &  $0.084 \pm 0.013$ &  $0.332 \pm 0.093$ &  $2254.7 \pm 833.2$ \\ 

\multicolumn{1}{c|}{} &  \multicolumn{1}{c|}{\cellcolor[HTML]{C0C0C0}$\mathrm{MEGP}_{75}$} &  \multicolumn{1}{c|}{\cellcolor[HTML]{C0C0C0}$0.119 \pm 0.012$} &  \cellcolor[HTML]{C0C0C0}$0.002 \pm 0.001$ &  \cellcolor[HTML]{C0C0C0}$0.06 \pm 0.013$ &  \multicolumn{1}{c|}{\cellcolor[HTML]{C0C0C0}$0.108 \pm 0.014$} &  \cellcolor[HTML]{C0C0C0}$0.0 \pm 0.0$ &  \cellcolor[HTML]{C0C0C0}$0.045 \pm 0.024$ &  \multicolumn{1}{c|}{\cellcolor[HTML]{C0C0C0}$0.104 \pm 0.016$} &  \cellcolor[HTML]{C0C0C0}$0.0 \pm 0.0$ &  \cellcolor[HTML]{C0C0C0}$0.018 \pm 0.025$ &  \multicolumn{1}{c|}{\cellcolor[HTML]{C0C0C0}$0.104 \pm 0.016$} &  \cellcolor[HTML]{C0C0C0}$0.001 \pm 0.001$ &  \cellcolor[HTML]{C0C0C0}$0.059 \pm 0.012$ &  \cellcolor[HTML]{C0C0C0}$0.423 \pm 0.123$ &  \cellcolor[HTML]{C0C0C0}$2456.7 \pm 890.5$ \\ 

\multicolumn{1}{c|}{} &  \multicolumn{1}{c|}{$\mathrm{MEGP}_{100}$} &  \multicolumn{1}{c|}{$0.118 \pm 0.011$} &  $0.002 \pm 0.001$ &  $0.041 \pm 0.017$ &  \multicolumn{1}{c|}{$0.11 \pm 0.012$} &  $0.0 \pm 0.0$ &  $0.03 \pm 0.021$ &  \multicolumn{1}{c|}{$0.109 \pm 0.012$} &  $0.0 \pm 0.0$ &  $0.007 \pm 0.011$ &  \multicolumn{1}{c|}{$0.109 \pm 0.012$} &  $0.001 \pm 0.001$ &  $0.041 \pm 0.018$ &  $0.454 \pm 0.115$ &  $2336.2 \pm 781.0$ \\ 

\multicolumn{1}{c|}{} &  \multicolumn{1}{c|}{\cellcolor[HTML]{C0C0C0}BGP} &  \multicolumn{1}{c|}{\cellcolor[HTML]{C0C0C0}$0.116 \pm 0.015$} &  \cellcolor[HTML]{C0C0C0}$0.002 \pm 0.001$ &  \cellcolor[HTML]{C0C0C0}$0.107 \pm 0.017$ &  \multicolumn{1}{c|}{\cellcolor[HTML]{C0C0C0}$0.109 \pm 0.017$} &  \cellcolor[HTML]{C0C0C0}$0.0 \pm 0.0$ &  \cellcolor[HTML]{C0C0C0}$0.054 \pm 0.037$ &  \multicolumn{1}{c|}{\cellcolor[HTML]{C0C0C0}$0.108 \pm 0.018$} &  \cellcolor[HTML]{C0C0C0}$0.0 \pm 0.0$ &  \cellcolor[HTML]{C0C0C0}$0.013 \pm 0.027$ &  \multicolumn{1}{c|}{\cellcolor[HTML]{C0C0C0}$0.108 \pm 0.018$} &  \cellcolor[HTML]{C0C0C0}$0.002 \pm 0.001$ &  \cellcolor[HTML]{C0C0C0}$0.104 \pm 0.017$ &  \cellcolor[HTML]{C0C0C0}$0.254 \pm 0.093$ &  \cellcolor[HTML]{C0C0C0}$1402.3 \pm 525.3$ \\ 

\multicolumn{1}{c|}{} &  \multicolumn{1}{c|}{Friedman's P-value} &  \multicolumn{1}{c|}{$0.59$} &  $0.89$ &  $\mathbf{2.53e-25}$ &  \multicolumn{1}{c|}{$0.58$} &  $0.06$ &  $\mathbf{7.40e-13}$ &  \multicolumn{1}{c|}{$0.49$} &  $0.06$ &  $\mathbf{1.64e-04}$ &  \multicolumn{1}{c|}{$0.49$} &  $\mathbf{0.05}$ &  $\mathbf{2.26e-25}$ &  $\mathbf{1.87e-12}$ &  $\mathbf{3.61e-07}$ \\ 

\multicolumn{1}{c|}{\multirow{-8}{*}{APSF}} &  \multicolumn{1}{c|}{\cellcolor[HTML]{C0C0C0}Adjusted P-value} &  \multicolumn{1}{c|}{\cellcolor[HTML]{C0C0C0}$1.00$} &  \cellcolor[HTML]{C0C0C0}$1.00$ &  \cellcolor[HTML]{C0C0C0}$\mathbf{7.58e-25}$ &  \multicolumn{1}{c|}{\cellcolor[HTML]{C0C0C0}$1.00$} &  \cellcolor[HTML]{C0C0C0}$0.19$ &  \cellcolor[HTML]{C0C0C0}$\mathbf{2.22e-12}$ &  \multicolumn{1}{c|}{\cellcolor[HTML]{C0C0C0}$1.00$} &  \cellcolor[HTML]{C0C0C0}$0.18$ &  \cellcolor[HTML]{C0C0C0}$\mathbf{4.91e-04}$ &  \multicolumn{1}{c|}{\cellcolor[HTML]{C0C0C0}$1.00$} &  \cellcolor[HTML]{C0C0C0}$0.24$ &  \cellcolor[HTML]{C0C0C0}$\mathbf{1.13e-24}$ &  \cellcolor[HTML]{C0C0C0}$\mathbf{9.37e-12}$ &  \cellcolor[HTML]{C0C0C0}$\mathbf{1.81e-06}$ \\ \hline

\multicolumn{1}{c|}{} &  \multicolumn{1}{c|}{$\mathrm{MEGP}_{0}$} &  \multicolumn{1}{c|}{$0.62 \pm 0.029$} &  $0.003 \pm 0.001$ &  $0.074 \pm 0.006$ &  \multicolumn{1}{c|}{$0.592 \pm 0.029$} &  $0.001 \pm 0.0$ &  $0.08 \pm 0.008$ &  \multicolumn{1}{c|}{$0.58 \pm 0.031$} &  $0.0 \pm 0.0$ &  $0.079 \pm 0.009$ &  \multicolumn{1}{c|}{$0.58 \pm 0.031$} &  $0.001 \pm 0.0$ &  $0.078 \pm 0.007$ &  $0.556 \pm 0.059$ &  $1480.5 \pm 485.8$ \\ 
\multicolumn{1}{c|}{} &  \multicolumn{1}{c|}{\cellcolor[HTML]{C0C0C0}$\mathrm{MEGP}_{25}$} &  \multicolumn{1}{c|}{\cellcolor[HTML]{C0C0C0}$0.611 \pm 0.021$} &  \cellcolor[HTML]{C0C0C0}$0.004 \pm 0.001$ &  \cellcolor[HTML]{C0C0C0}$0.064 \pm 0.005$ &  \multicolumn{1}{c|}{\cellcolor[HTML]{C0C0C0}$0.577 \pm 0.023$} &  \cellcolor[HTML]{C0C0C0}$0.001 \pm 0.001$ &  \cellcolor[HTML]{C0C0C0}$0.061 \pm 0.021$ &  \multicolumn{1}{c|}{\cellcolor[HTML]{C0C0C0}$0.569 \pm 0.027$} &  \cellcolor[HTML]{C0C0C0}$0.0 \pm 0.0$ &  \cellcolor[HTML]{C0C0C0}$0.04 \pm 0.031$ &  \multicolumn{1}{c|}{\cellcolor[HTML]{C0C0C0}$0.569 \pm 0.027$} &  \cellcolor[HTML]{C0C0C0}$0.002 \pm 0.001$ &  \cellcolor[HTML]{C0C0C0}$0.069 \pm 0.004$ &  \cellcolor[HTML]{C0C0C0}$0.582 \pm 0.049$ &  \cellcolor[HTML]{C0C0C0}$1195.0 \pm 460.6$ \\ 
\multicolumn{1}{c|}{} &  \multicolumn{1}{c|}{$\mathrm{MEGP}_{50}$} &  \multicolumn{1}{c|}{$0.602 \pm 0.027$} &  $0.004 \pm 0.001$ &  $0.059 \pm 0.006$ &  \multicolumn{1}{c|}{$0.569 \pm 0.033$} &  $0.001 \pm 0.0$ &  $0.054 \pm 0.019$ &  \multicolumn{1}{c|}{$0.561 \pm 0.033$} &  $0.0 \pm 0.0$ &  $0.032 \pm 0.028$ &  \multicolumn{1}{c|}{$0.561 \pm 0.033$} &  $0.002 \pm 0.0$ &  $0.062 \pm 0.006$ &  $0.586 \pm 0.06$ &  $774.7 \pm 234.6$ \\ 
\multicolumn{1}{c|}{} &  \multicolumn{1}{c|}{\cellcolor[HTML]{C0C0C0}$\mathrm{MEGP}_{75}$} &  \multicolumn{1}{c|}{\cellcolor[HTML]{C0C0C0}$0.597 \pm 0.028$} &  \cellcolor[HTML]{C0C0C0}$0.004 \pm 0.001$ &  \cellcolor[HTML]{C0C0C0}$0.049 \pm 0.007$ &  \multicolumn{1}{c|}{\cellcolor[HTML]{C0C0C0}$0.569 \pm 0.03$} &  \cellcolor[HTML]{C0C0C0}$0.001 \pm 0.0$ &  \cellcolor[HTML]{C0C0C0}$0.048 \pm 0.019$ &  \multicolumn{1}{c|}{\cellcolor[HTML]{C0C0C0}$0.561 \pm 0.031$} &  \cellcolor[HTML]{C0C0C0}$0.0 \pm 0.0$ &  \cellcolor[HTML]{C0C0C0}$0.03 \pm 0.025$ &  \multicolumn{1}{c|}{\cellcolor[HTML]{C0C0C0}$0.561 \pm 0.031$} &  \cellcolor[HTML]{C0C0C0}$0.002 \pm 0.001$ &  \cellcolor[HTML]{C0C0C0}$0.053 \pm 0.008$ &  \cellcolor[HTML]{C0C0C0}$0.612 \pm 0.052$ &  \cellcolor[HTML]{C0C0C0}$1205.8 \pm 528.7$ \\ 
\multicolumn{1}{c|}{} &  \multicolumn{1}{c|}{$\mathrm{MEGP}_{100}$} &  \multicolumn{1}{c|}{$0.605 \pm 0.032$} &  $0.003 \pm 0.001$ &  $0.042 \pm 0.008$ &  \multicolumn{1}{c|}{$0.576 \pm 0.03$} &  $0.001 \pm 0.0$ &  $0.042 \pm 0.017$ &  \multicolumn{1}{c|}{$0.565 \pm 0.032$} &  $0.0 \pm 0.0$ &  $0.033 \pm 0.021$ &  \multicolumn{1}{c|}{$0.565 \pm 0.032$} &  $0.002 \pm 0.001$ &  $0.045 \pm 0.008$ &  $0.638 \pm 0.056$ &  $1356.0 \pm 624.9$ \\ 
\multicolumn{1}{c|}{} &  \multicolumn{1}{c|}{\cellcolor[HTML]{C0C0C0}BGP} &  \multicolumn{1}{c|}{\cellcolor[HTML]{C0C0C0}$0.629 \pm 0.027$} &  \cellcolor[HTML]{C0C0C0}$0.004 \pm 0.001$ &  \cellcolor[HTML]{C0C0C0}$0.063 \pm 0.005$ &  \multicolumn{1}{c|}{\cellcolor[HTML]{C0C0C0}$0.607 \pm 0.031$} &  \cellcolor[HTML]{C0C0C0}$0.0 \pm 0.0$ &  \cellcolor[HTML]{C0C0C0}$0.052 \pm 0.019$ &  \multicolumn{1}{c|}{\cellcolor[HTML]{C0C0C0}$0.604 \pm 0.035$} &  \cellcolor[HTML]{C0C0C0}$0.0 \pm 0.0$ &  \cellcolor[HTML]{C0C0C0}$0.014 \pm 0.022$ &  \multicolumn{1}{c|}{\cellcolor[HTML]{C0C0C0}$0.604 \pm 0.035$} &  \cellcolor[HTML]{C0C0C0}$0.003 \pm 0.001$ &  \cellcolor[HTML]{C0C0C0}$0.065 \pm 0.005$ &  \cellcolor[HTML]{C0C0C0}$0.483 \pm 0.046$ &  \cellcolor[HTML]{C0C0C0}$325.8 \pm 107.4$ \\ 
\multicolumn{1}{c|}{} &  \multicolumn{1}{c|}{Friedman's P-value} &  \multicolumn{1}{c|}{$\mathbf{3.04e-03}$} &  $\mathbf{5.13e-05}$ &  $\mathbf{1.10e-24}$ &  \multicolumn{1}{c|}{$\mathbf{1.65e-05}$} &  $0.28$ &  $\mathbf{2.00e-14}$ &  \multicolumn{1}{c|}{$\mathbf{4.09e-06}$} &  $\mathbf{0.01}$ &  $\mathbf{7.10e-14}$ &  \multicolumn{1}{c|}{$\mathbf{4.09e-06}$} &  $\mathbf{2.15e-11}$ &  $\mathbf{3.48e-24}$ &  $\mathbf{5.17e-14}$ &  $\mathbf{4.44e-17}$ \\ 
\multicolumn{1}{c|}{\multirow{-8}{*}{ARWPM}} &  \multicolumn{1}{c|}{\cellcolor[HTML]{C0C0C0}Adjusted P-value} &  \multicolumn{1}{c|}{\cellcolor[HTML]{C0C0C0}$\mathbf{9.13e-03}$} &  \cellcolor[HTML]{C0C0C0}$\mathbf{1.54e-04}$ &  \cellcolor[HTML]{C0C0C0}$\mathbf{3.30e-24}$ &  \multicolumn{1}{c|}{\cellcolor[HTML]{C0C0C0}$\mathbf{4.95e-05}$} &  \cellcolor[HTML]{C0C0C0}$0.84$ &  \cellcolor[HTML]{C0C0C0}$\mathbf{5.99e-14}$ &  \multicolumn{1}{c|}{\cellcolor[HTML]{C0C0C0}$\mathbf{1.23e-05}$} &  \cellcolor[HTML]{C0C0C0}$\mathbf{0.04}$ &  \cellcolor[HTML]{C0C0C0}$\mathbf{2.13e-13}$ &  \multicolumn{1}{c|}{\cellcolor[HTML]{C0C0C0}$\mathbf{2.04e-05}$} &  \cellcolor[HTML]{C0C0C0}$\mathbf{1.08e-10}$ &  \cellcolor[HTML]{C0C0C0}$\mathbf{1.74e-23}$ &  \cellcolor[HTML]{C0C0C0}$\mathbf{2.58e-13}$ &  \cellcolor[HTML]{C0C0C0}$\mathbf{2.22e-16}$ \\ \hline

\multicolumn{1}{c|}{} &  \multicolumn{1}{c|}{$\mathrm{MEGP}_{0}$} &  \multicolumn{1}{c|}{$0.175 \pm 0.038$} &  $0.009 \pm 0.001$ &  $0.108 \pm 0.012$ &  \multicolumn{1}{c|}{$0.122 \pm 0.032$} &  $0.001 \pm 0.0$ &  $0.125 \pm 0.016$ &  \multicolumn{1}{c|}{$0.105 \pm 0.034$} &  $0.0 \pm 0.0$ &  $0.107 \pm 0.041$ &  \multicolumn{1}{c|}{$0.105 \pm 0.034$} &  $0.004 \pm 0.001$ &  $0.121 \pm 0.013$ &  $0.396 \pm 0.061$ &  $1267.4 \pm 271.0$ \\ 
\multicolumn{1}{c|}{} &  \multicolumn{1}{c|}{\cellcolor[HTML]{C0C0C0}$\mathrm{MEGP}_{25}$} &  \multicolumn{1}{c|}{\cellcolor[HTML]{C0C0C0}$0.175 \pm 0.044$} &  \cellcolor[HTML]{C0C0C0}$0.009 \pm 0.001$ &  \cellcolor[HTML]{C0C0C0}$0.101 \pm 0.009$ &  \multicolumn{1}{c|}{\cellcolor[HTML]{C0C0C0}$0.118 \pm 0.034$} &  \cellcolor[HTML]{C0C0C0}$0.001 \pm 0.001$ &  \cellcolor[HTML]{C0C0C0}$0.111 \pm 0.019$ &  \multicolumn{1}{c|}{\cellcolor[HTML]{C0C0C0}$0.101 \pm 0.033$} &  \cellcolor[HTML]{C0C0C0}$0.0 \pm 0.0$ &  \cellcolor[HTML]{C0C0C0}$0.083 \pm 0.041$ &  \multicolumn{1}{c|}{\cellcolor[HTML]{C0C0C0}$0.101 \pm 0.033$} &  \cellcolor[HTML]{C0C0C0}$0.004 \pm 0.001$ &  \cellcolor[HTML]{C0C0C0}$0.11 \pm 0.01$ &  \cellcolor[HTML]{C0C0C0}$0.418 \pm 0.08$ &  \cellcolor[HTML]{C0C0C0}$1190.9 \pm 300.1$ \\ 
\multicolumn{1}{c|}{} &  \multicolumn{1}{c|}{$\mathrm{MEGP}_{50}$} &  \multicolumn{1}{c|}{$0.172 \pm 0.044$} &  $0.009 \pm 0.002$ &  $0.093 \pm 0.008$ &  \multicolumn{1}{c|}{$0.128 \pm 0.035$} &  $0.001 \pm 0.001$ &  $0.1 \pm 0.022$ &  \multicolumn{1}{c|}{$0.111 \pm 0.03$} &  $0.0 \pm 0.0$ &  $0.076 \pm 0.044$ &  \multicolumn{1}{c|}{$0.111 \pm 0.03$} &  $0.004 \pm 0.001$ &  $0.103 \pm 0.008$ &  $0.447 \pm 0.057$ &  $834.3 \pm 194.8$ \\ 
\multicolumn{1}{c|}{} &  \multicolumn{1}{c|}{\cellcolor[HTML]{C0C0C0}$\mathrm{MEGP}_{75}$} &  \multicolumn{1}{c|}{\cellcolor[HTML]{C0C0C0}$0.185 \pm 0.029$} &  \cellcolor[HTML]{C0C0C0}$0.009 \pm 0.001$ &  \cellcolor[HTML]{C0C0C0}$0.084 \pm 0.008$ &  \multicolumn{1}{c|}{\cellcolor[HTML]{C0C0C0}$0.134 \pm 0.029$} &  \cellcolor[HTML]{C0C0C0}$0.001 \pm 0.0$ &  \cellcolor[HTML]{C0C0C0}$0.094 \pm 0.019$ &  \multicolumn{1}{c|}{\cellcolor[HTML]{C0C0C0}$0.12 \pm 0.032$} &  \cellcolor[HTML]{C0C0C0}$0.0 \pm 0.0$ &  \cellcolor[HTML]{C0C0C0}$0.076 \pm 0.039$ &  \multicolumn{1}{c|}{\cellcolor[HTML]{C0C0C0}$0.12 \pm 0.032$} &  \cellcolor[HTML]{C0C0C0}$0.004 \pm 0.001$ &  \cellcolor[HTML]{C0C0C0}$0.095 \pm 0.01$ &  \cellcolor[HTML]{C0C0C0}$0.471 \pm 0.055$ &  \cellcolor[HTML]{C0C0C0}$1198.8 \pm 322.8$ \\ 
\multicolumn{1}{c|}{} &  \multicolumn{1}{c|}{$\mathrm{MEGP}_{100}$} &  \multicolumn{1}{c|}{$0.175 \pm 0.041$} &  $0.009 \pm 0.002$ &  $0.069 \pm 0.01$ &  \multicolumn{1}{c|}{$0.126 \pm 0.031$} &  $0.001 \pm 0.0$ &  $0.083 \pm 0.019$ &  \multicolumn{1}{c|}{$0.111 \pm 0.031$} &  $0.0 \pm 0.0$ &  $0.068 \pm 0.028$ &  \multicolumn{1}{c|}{$0.111 \pm 0.031$} &  $0.004 \pm 0.001$ &  $0.079 \pm 0.014$ &  $0.547 \pm 0.134$ &  $1262.1 \pm 246.5$ \\ 
\multicolumn{1}{c|}{} &  \multicolumn{1}{c|}{\cellcolor[HTML]{C0C0C0}BGP} &  \multicolumn{1}{c|}{\cellcolor[HTML]{C0C0C0}$0.182 \pm 0.042$} &  \cellcolor[HTML]{C0C0C0}$0.012 \pm 0.002$ &  \cellcolor[HTML]{C0C0C0}$0.09 \pm 0.01$ &  \multicolumn{1}{c|}{\cellcolor[HTML]{C0C0C0}$0.135 \pm 0.039$} &  \cellcolor[HTML]{C0C0C0}$0.001 \pm 0.001$ &  \cellcolor[HTML]{C0C0C0}$0.085 \pm 0.024$ &  \multicolumn{1}{c|}{\cellcolor[HTML]{C0C0C0}$0.127 \pm 0.042$} &  \cellcolor[HTML]{C0C0C0}$0.0 \pm 0.0$ &  \cellcolor[HTML]{C0C0C0}$0.033 \pm 0.041$ &  \multicolumn{1}{c|}{\cellcolor[HTML]{C0C0C0}$0.127 \pm 0.042$} &  \cellcolor[HTML]{C0C0C0}$0.007 \pm 0.002$ &  \cellcolor[HTML]{C0C0C0}$0.097 \pm 0.01$ &  \cellcolor[HTML]{C0C0C0}$0.342 \pm 0.062$ &  \cellcolor[HTML]{C0C0C0}$439.6 \pm 119.0$ \\ 
\multicolumn{1}{c|}{} &  \multicolumn{1}{c|}{Friedman's P-value} &  \multicolumn{1}{c|}{$0.78$} &  $\mathbf{1.44e-07}$ &  $\mathbf{2.90e-20}$ &  \multicolumn{1}{c|}{$0.75$} &  $0.38$ &  $\mathbf{1.48e-13}$ &  \multicolumn{1}{c|}{$0.10$} &  $\mathbf{0.04}$ &  $\mathbf{4.54e-08}$ &  \multicolumn{1}{c|}{$0.10$} &  $\mathbf{6.76e-09}$ &  $\mathbf{3.32e-19}$ &  $\mathbf{6.67e-14}$ &  $\mathbf{8.31e-18}$ \\ 
\multicolumn{1}{c|}{\multirow{-8}{*}{GECR}} &  \multicolumn{1}{c|}{\cellcolor[HTML]{C0C0C0}Adjusted P-value} &  \multicolumn{1}{c|}{\cellcolor[HTML]{C0C0C0}$1.00$} &  \cellcolor[HTML]{C0C0C0}$\mathbf{4.32e-07}$ &  \cellcolor[HTML]{C0C0C0}$\mathbf{8.69e-20}$ &  \multicolumn{1}{c|}{\cellcolor[HTML]{C0C0C0}$1.00$} &  \cellcolor[HTML]{C0C0C0}$1.00$ &  \cellcolor[HTML]{C0C0C0}$\mathbf{4.43e-13}$ &  \multicolumn{1}{c|}{\cellcolor[HTML]{C0C0C0}$0.30$} &  \cellcolor[HTML]{C0C0C0}$0.12$ &  \cellcolor[HTML]{C0C0C0}$\mathbf{1.36e-07}$ &  \multicolumn{1}{c|}{\cellcolor[HTML]{C0C0C0}$0.50$} &  \cellcolor[HTML]{C0C0C0}$\mathbf{3.38e-08}$ &  \cellcolor[HTML]{C0C0C0}$\mathbf{1.66e-18}$ &  \cellcolor[HTML]{C0C0C0}$\mathbf{3.34e-13}$ &  \cellcolor[HTML]{C0C0C0}$\mathbf{4.15e-17}$ \\ \hline

\multicolumn{1}{c|}{} &  \multicolumn{1}{c|}{$\mathrm{MEGP}_{0}$} &  \multicolumn{1}{c|}{$0.488 \pm 0.015$} &  $0.001 \pm 0.0$ &  $0.028 \pm 0.003$ &  \multicolumn{1}{c|}{$0.469 \pm 0.016$} &  $0.0 \pm 0.0$ &  $0.031 \pm 0.003$ &  \multicolumn{1}{c|}{$0.459 \pm 0.015$} &  $0.0 \pm 0.0$ &  $0.03 \pm 0.007$ &  \multicolumn{1}{c|}{$0.459 \pm 0.015$} &  $0.001 \pm 0.0$ &  $0.031 \pm 0.003$ &  $0.29 \pm 0.046$ &  $3035.9 \pm 335.4$ \\ 
\multicolumn{1}{c|}{} &  \multicolumn{1}{c|}{\cellcolor[HTML]{C0C0C0}$\mathrm{MEGP}_{25}$} &  \multicolumn{1}{c|}{\cellcolor[HTML]{C0C0C0}$0.47 \pm 0.022$} &  \cellcolor[HTML]{C0C0C0}$0.002 \pm 0.0$ &  \cellcolor[HTML]{C0C0C0}$0.026 \pm 0.003$ &  \multicolumn{1}{c|}{\cellcolor[HTML]{C0C0C0}$0.445 \pm 0.037$} &  \cellcolor[HTML]{C0C0C0}$0.0 \pm 0.001$ &  \cellcolor[HTML]{C0C0C0}$0.024 \pm 0.01$ &  \multicolumn{1}{c|}{\cellcolor[HTML]{C0C0C0}$0.438 \pm 0.035$} &  \cellcolor[HTML]{C0C0C0}$0.0 \pm 0.0$ &  \cellcolor[HTML]{C0C0C0}$0.018 \pm 0.014$ &  \multicolumn{1}{c|}{\cellcolor[HTML]{C0C0C0}$0.438 \pm 0.035$} &  \cellcolor[HTML]{C0C0C0}$0.001 \pm 0.001$ &  \cellcolor[HTML]{C0C0C0}$0.028 \pm 0.003$ &  \cellcolor[HTML]{C0C0C0}$0.318 \pm 0.04$ &  \cellcolor[HTML]{C0C0C0}$2443.1 \pm 795.8$ \\ 
\multicolumn{1}{c|}{} &  \multicolumn{1}{c|}{$\mathrm{MEGP}_{50}$} &  \multicolumn{1}{c|}{$0.475 \pm 0.021$} &  $0.002 \pm 0.001$ &  $0.022 \pm 0.003$ &  \multicolumn{1}{c|}{$0.454 \pm 0.023$} &  $0.0 \pm 0.0$ &  $0.021 \pm 0.01$ &  \multicolumn{1}{c|}{$0.446 \pm 0.024$} &  $0.0 \pm 0.0$ &  $0.016 \pm 0.014$ &  \multicolumn{1}{c|}{$0.446 \pm 0.024$} &  $0.001 \pm 0.001$ &  $0.025 \pm 0.003$ &  $0.334 \pm 0.041$ &  $2167.6 \pm 774.4$ \\ 
\multicolumn{1}{c|}{} &  \multicolumn{1}{c|}{\cellcolor[HTML]{C0C0C0}$\mathrm{MEGP}_{75}$} &  \multicolumn{1}{c|}{\cellcolor[HTML]{C0C0C0}$0.475 \pm 0.017$} &  \cellcolor[HTML]{C0C0C0}$0.002 \pm 0.0$ &  \cellcolor[HTML]{C0C0C0}$0.019 \pm 0.004$ &  \multicolumn{1}{c|}{\cellcolor[HTML]{C0C0C0}$0.453 \pm 0.024$} &  \cellcolor[HTML]{C0C0C0}$0.0 \pm 0.0$ &  \cellcolor[HTML]{C0C0C0}$0.017 \pm 0.01$ &  \multicolumn{1}{c|}{\cellcolor[HTML]{C0C0C0}$0.442 \pm 0.03$} &  \cellcolor[HTML]{C0C0C0}$0.0 \pm 0.0$ &  \cellcolor[HTML]{C0C0C0}$0.012 \pm 0.012$ &  \multicolumn{1}{c|}{\cellcolor[HTML]{C0C0C0}$0.442 \pm 0.03$} &  \cellcolor[HTML]{C0C0C0}$0.001 \pm 0.001$ &  \cellcolor[HTML]{C0C0C0}$0.021 \pm 0.004$ &  \cellcolor[HTML]{C0C0C0}$0.343 \pm 0.045$ &  \cellcolor[HTML]{C0C0C0}$2204.0 \pm 897.8$ \\ 
\multicolumn{1}{c|}{} &  \multicolumn{1}{c|}{$\mathrm{MEGP}_{100}$} &  \multicolumn{1}{c|}{$0.466 \pm 0.028$} &  $0.002 \pm 0.001$ &  $0.016 \pm 0.004$ &  \multicolumn{1}{c|}{$0.452 \pm 0.024$} &  $0.0 \pm 0.0$ &  $0.015 \pm 0.009$ &  \multicolumn{1}{c|}{$0.442 \pm 0.024$} &  $0.0 \pm 0.0$ &  $0.013 \pm 0.012$ &  \multicolumn{1}{c|}{$0.442 \pm 0.024$} &  $0.001 \pm 0.001$ &  $0.018 \pm 0.005$ &  $0.371 \pm 0.049$ &  $2297.9 \pm 875.3$ \\ 
\multicolumn{1}{c|}{} &  \multicolumn{1}{c|}{\cellcolor[HTML]{C0C0C0}BGP} &  \multicolumn{1}{c|}{\cellcolor[HTML]{C0C0C0}$0.494 \pm 0.014$} &  \cellcolor[HTML]{C0C0C0}$0.002 \pm 0.0$ &  \cellcolor[HTML]{C0C0C0}$0.027 \pm 0.004$ &  \multicolumn{1}{c|}{\cellcolor[HTML]{C0C0C0}$0.474 \pm 0.015$} &  \cellcolor[HTML]{C0C0C0}$0.0 \pm 0.0$ &  \cellcolor[HTML]{C0C0C0}$0.029 \pm 0.007$ &  \multicolumn{1}{c|}{\cellcolor[HTML]{C0C0C0}$0.468 \pm 0.016$} &  \cellcolor[HTML]{C0C0C0}$0.0 \pm 0.0$ &  \cellcolor[HTML]{C0C0C0}$0.02 \pm 0.012$ &  \multicolumn{1}{c|}{\cellcolor[HTML]{C0C0C0}$0.468 \pm 0.016$} &  \cellcolor[HTML]{C0C0C0}$0.001 \pm 0.0$ &  \cellcolor[HTML]{C0C0C0}$0.03 \pm 0.004$ &  \cellcolor[HTML]{C0C0C0}$0.294 \pm 0.058$ &  \cellcolor[HTML]{C0C0C0}$1713.2 \pm 355.7$ \\ 
\multicolumn{1}{c|}{} &  \multicolumn{1}{c|}{Friedman's P-value} &  \multicolumn{1}{c|}{$\mathbf{7.01e-06}$} &  $\mathbf{9.92e-04}$ &  $\mathbf{4.37e-22}$ &  \multicolumn{1}{c|}{$\mathbf{1.81e-03}$} &  $0.16$ &  $\mathbf{9.20e-14}$ &  \multicolumn{1}{c|}{$\mathbf{1.08e-04}$} &  $0.13$ &  $\mathbf{4.21e-08}$ &  \multicolumn{1}{c|}{$\mathbf{1.08e-04}$} &  $\mathbf{2.32e-06}$ &  $\mathbf{4.40e-18}$ &  $\mathbf{1.56e-09}$ &  $\mathbf{5.74e-08}$ \\ 
\multicolumn{1}{c|}{\multirow{-8}{*}{GFE}} &  \multicolumn{1}{c|}{\cellcolor[HTML]{C0C0C0}Adjusted P-value} &  \multicolumn{1}{c|}{\cellcolor[HTML]{C0C0C0}$\mathbf{2.10e-05}$} &  \cellcolor[HTML]{C0C0C0}$\mathbf{2.98e-03}$ &  \cellcolor[HTML]{C0C0C0}$\mathbf{1.31e-21}$ &  \multicolumn{1}{c|}{\cellcolor[HTML]{C0C0C0}$\mathbf{5.42e-03}$} &  \cellcolor[HTML]{C0C0C0}$0.47$ &  \cellcolor[HTML]{C0C0C0}$\mathbf{2.76e-13}$ &  \multicolumn{1}{c|}{\cellcolor[HTML]{C0C0C0}$\mathbf{3.25e-04}$} &  \cellcolor[HTML]{C0C0C0}$0.40$ &  \cellcolor[HTML]{C0C0C0}$\mathbf{1.26e-07}$ &  \multicolumn{1}{c|}{\cellcolor[HTML]{C0C0C0}$\mathbf{5.42e-04}$} &  \cellcolor[HTML]{C0C0C0}$\mathbf{1.16e-05}$ &  \cellcolor[HTML]{C0C0C0}$\mathbf{2.20e-17}$ &  \cellcolor[HTML]{C0C0C0}$\mathbf{7.79e-09}$ &  \cellcolor[HTML]{C0C0C0}$\mathbf{2.87e-07}$ \\ \hline

\multicolumn{1}{c|}{} &  \multicolumn{1}{c|}{$\mathrm{MEGP}_{0}$} &  \multicolumn{1}{c|}{$0.799 \pm 0.069$} &  $0.009 \pm 0.002$ &  $0.105 \pm 0.006$ &  \multicolumn{1}{c|}{$0.711 \pm 0.07$} &  $0.002 \pm 0.001$ &  $0.133 \pm 0.013$ &  \multicolumn{1}{c|}{$0.669 \pm 0.066$} &  $0.001 \pm 0.001$ &  $0.134 \pm 0.024$ &  \multicolumn{1}{c|}{$0.669 \pm 0.066$} &  $0.004 \pm 0.001$ &  $0.127 \pm 0.01$ &  $0.847 \pm 0.076$ &  $2524.6 \pm 245.6$ \\ 
\multicolumn{1}{c|}{} &  \multicolumn{1}{c|}{\cellcolor[HTML]{C0C0C0}$\mathrm{MEGP}_{25}$} &  \multicolumn{1}{c|}{\cellcolor[HTML]{C0C0C0}$0.752 \pm 0.051$} &  \cellcolor[HTML]{C0C0C0}$0.009 \pm 0.002$ &  \cellcolor[HTML]{C0C0C0}$0.095 \pm 0.008$ &  \multicolumn{1}{c|}{\cellcolor[HTML]{C0C0C0}$0.672 \pm 0.06$} &  \cellcolor[HTML]{C0C0C0}$0.002 \pm 0.001$ &  \cellcolor[HTML]{C0C0C0}$0.124 \pm 0.016$ &  \multicolumn{1}{c|}{\cellcolor[HTML]{C0C0C0}$0.63 \pm 0.069$} &  \cellcolor[HTML]{C0C0C0}$0.001 \pm 0.001$ &  \cellcolor[HTML]{C0C0C0}$0.115 \pm 0.039$ &  \multicolumn{1}{c|}{\cellcolor[HTML]{C0C0C0}$0.63 \pm 0.069$} &  \cellcolor[HTML]{C0C0C0}$0.004 \pm 0.001$ &  \cellcolor[HTML]{C0C0C0}$0.118 \pm 0.01$ &  \cellcolor[HTML]{C0C0C0}$0.874 \pm 0.059$ &  \cellcolor[HTML]{C0C0C0}$2445.3 \pm 387.5$ \\ 
\multicolumn{1}{c|}{} &  \multicolumn{1}{c|}{$\mathrm{MEGP}_{50}$} &  \multicolumn{1}{c|}{$0.738 \pm 0.065$} &  $0.01 \pm 0.002$ &  $0.085 \pm 0.01$ &  \multicolumn{1}{c|}{$0.662 \pm 0.064$} &  $0.001 \pm 0.001$ &  $0.115 \pm 0.014$ &  \multicolumn{1}{c|}{$0.63 \pm 0.072$} &  $0.001 \pm 0.001$ &  $0.094 \pm 0.05$ &  \multicolumn{1}{c|}{$0.63 \pm 0.072$} &  $0.004 \pm 0.001$ &  $0.108 \pm 0.011$ &  $0.888 \pm 0.069$ &  $2798.2 \pm 712.3$ \\ 
\multicolumn{1}{c|}{} &  \multicolumn{1}{c|}{\cellcolor[HTML]{C0C0C0}$\mathrm{MEGP}_{75}$} &  \multicolumn{1}{c|}{\cellcolor[HTML]{C0C0C0}$0.75 \pm 0.053$} &  \cellcolor[HTML]{C0C0C0}$0.009 \pm 0.001$ &  \cellcolor[HTML]{C0C0C0}$0.071 \pm 0.008$ &  \multicolumn{1}{c|}{\cellcolor[HTML]{C0C0C0}$0.67 \pm 0.052$} &  \cellcolor[HTML]{C0C0C0}$0.002 \pm 0.001$ &  \cellcolor[HTML]{C0C0C0}$0.1 \pm 0.017$ &  \multicolumn{1}{c|}{\cellcolor[HTML]{C0C0C0}$0.626 \pm 0.057$} &  \cellcolor[HTML]{C0C0C0}$0.001 \pm 0.001$ &  \cellcolor[HTML]{C0C0C0}$0.099 \pm 0.029$ &  \multicolumn{1}{c|}{\cellcolor[HTML]{C0C0C0}$0.626 \pm 0.057$} &  \cellcolor[HTML]{C0C0C0}$0.004 \pm 0.0$ &  \cellcolor[HTML]{C0C0C0}$0.093 \pm 0.011$ &  \cellcolor[HTML]{C0C0C0}$0.887 \pm 0.052$ &  \cellcolor[HTML]{C0C0C0}$2522.0 \pm 350.6$ \\ 
\multicolumn{1}{c|}{} &  \multicolumn{1}{c|}{$\mathrm{MEGP}_{100}$} &  \multicolumn{1}{c|}{$0.759 \pm 0.06$} &  $0.009 \pm 0.001$ &  $0.06 \pm 0.01$ &  \multicolumn{1}{c|}{$0.668 \pm 0.055$} &  $0.002 \pm 0.001$ &  $0.086 \pm 0.014$ &  \multicolumn{1}{c|}{$0.64 \pm 0.06$} &  $0.001 \pm 0.001$ &  $0.072 \pm 0.033$ &  \multicolumn{1}{c|}{$0.64 \pm 0.06$} &  $0.004 \pm 0.001$ &  $0.079 \pm 0.012$ &  $0.931 \pm 0.069$ &  $2451.3 \pm 425.6$ \\ 
\multicolumn{1}{c|}{} &  \multicolumn{1}{c|}{\cellcolor[HTML]{C0C0C0}BGP} &  \multicolumn{1}{c|}{\cellcolor[HTML]{C0C0C0}$0.85 \pm 0.094$} &  \cellcolor[HTML]{C0C0C0}$0.01 \pm 0.002$ &  \cellcolor[HTML]{C0C0C0}$0.093 \pm 0.009$ &  \multicolumn{1}{c|}{\cellcolor[HTML]{C0C0C0}$0.759 \pm 0.087$} &  \cellcolor[HTML]{C0C0C0}$0.002 \pm 0.001$ &  \cellcolor[HTML]{C0C0C0}$0.101 \pm 0.035$ &  \multicolumn{1}{c|}{\cellcolor[HTML]{C0C0C0}$0.741 \pm 0.096$} &  \cellcolor[HTML]{C0C0C0}$0.0 \pm 0.001$ &  \cellcolor[HTML]{C0C0C0}$0.054 \pm 0.048$ &  \multicolumn{1}{c|}{\cellcolor[HTML]{C0C0C0}$0.741 \pm 0.096$} &  \cellcolor[HTML]{C0C0C0}$0.006 \pm 0.002$ &  \cellcolor[HTML]{C0C0C0}$0.108 \pm 0.009$ &  \cellcolor[HTML]{C0C0C0}$0.75 \pm 0.091$ &  \cellcolor[HTML]{C0C0C0}$982.2 \pm 275.2$ \\ 
\multicolumn{1}{c|}{} &  \multicolumn{1}{c|}{Friedman's P-value} &  \multicolumn{1}{c|}{$\mathbf{3.49e-05}$} &  $0.11$ &  $\mathbf{2.09e-24}$ &  \multicolumn{1}{c|}{$\mathbf{2.63e-04}$} &  $0.72$ &  $\mathbf{3.35e-15}$ &  \multicolumn{1}{c|}{$\mathbf{5.30e-05}$} &  $\mathbf{1.53e-03}$ &  $\mathbf{2.91e-14}$ &  \multicolumn{1}{c|}{$\mathbf{5.30e-05}$} &  $\mathbf{1.48e-07}$ &  $\mathbf{8.13e-22}$ &  $\mathbf{8.31e-10}$ &  $\mathbf{1.10e-13}$ \\ 
\multicolumn{1}{c|}{\multirow{-8}{*}{GSAD}} &  \multicolumn{1}{c|}{\cellcolor[HTML]{C0C0C0}Adjusted P-value} &  \multicolumn{1}{c|}{\cellcolor[HTML]{C0C0C0}$\mathbf{1.05e-04}$} &  \cellcolor[HTML]{C0C0C0}$0.33$ &  \cellcolor[HTML]{C0C0C0}$\mathbf{6.26e-24}$ &  \multicolumn{1}{c|}{\cellcolor[HTML]{C0C0C0}$\mathbf{7.90e-04}$} &  \cellcolor[HTML]{C0C0C0}$1.00$ &  \cellcolor[HTML]{C0C0C0}$\mathbf{1.01e-14}$ &  \multicolumn{1}{c|}{\cellcolor[HTML]{C0C0C0}$\mathbf{1.59e-04}$} &  \cellcolor[HTML]{C0C0C0}$\mathbf{4.60e-03}$ &  \cellcolor[HTML]{C0C0C0}$\mathbf{8.72e-14}$ &  \multicolumn{1}{c|}{\cellcolor[HTML]{C0C0C0}$\mathbf{2.65e-04}$} &  \cellcolor[HTML]{C0C0C0}$\mathbf{7.40e-07}$ &  \cellcolor[HTML]{C0C0C0}$\mathbf{4.07e-21}$ &  \cellcolor[HTML]{C0C0C0}$\mathbf{4.15e-09}$ &  \cellcolor[HTML]{C0C0C0}$\mathbf{5.51e-13}$ \\ \hline

\multicolumn{1}{c|}{} &  \multicolumn{1}{c|}{$\mathrm{MEGP}_{0}$} &  \multicolumn{1}{c|}{$0.729 \pm 0.043$} &  $0.007 \pm 0.001$ &  $0.098 \pm 0.01$ &  \multicolumn{1}{c|}{$0.679 \pm 0.04$} &  $0.001 \pm 0.001$ &  $0.114 \pm 0.012$ &  \multicolumn{1}{c|}{$0.655 \pm 0.034$} &  $0.0 \pm 0.0$ &  $0.116 \pm 0.012$ &  \multicolumn{1}{c|}{$0.655 \pm 0.034$} &  $0.003 \pm 0.0$ &  $0.11 \pm 0.009$ &  $0.678 \pm 0.048$ &  $5007.4 \pm 375.2$ \\ 
\multicolumn{1}{c|}{} &  \multicolumn{1}{c|}{\cellcolor[HTML]{C0C0C0}$\mathrm{MEGP}_{25}$} &  \multicolumn{1}{c|}{\cellcolor[HTML]{C0C0C0}$0.694 \pm 0.033$} &  \cellcolor[HTML]{C0C0C0}$0.008 \pm 0.001$ &  \cellcolor[HTML]{C0C0C0}$0.091 \pm 0.01$ &  \multicolumn{1}{c|}{\cellcolor[HTML]{C0C0C0}$0.642 \pm 0.03$} &  \cellcolor[HTML]{C0C0C0}$0.001 \pm 0.001$ &  \cellcolor[HTML]{C0C0C0}$0.101 \pm 0.022$ &  \multicolumn{1}{c|}{\cellcolor[HTML]{C0C0C0}$0.621 \pm 0.037$} &  \cellcolor[HTML]{C0C0C0}$0.0 \pm 0.0$ &  \cellcolor[HTML]{C0C0C0}$0.086 \pm 0.043$ &  \multicolumn{1}{c|}{\cellcolor[HTML]{C0C0C0}$0.621 \pm 0.037$} &  \cellcolor[HTML]{C0C0C0}$0.004 \pm 0.001$ &  \cellcolor[HTML]{C0C0C0}$0.102 \pm 0.012$ &  \cellcolor[HTML]{C0C0C0}$0.692 \pm 0.053$ &  \cellcolor[HTML]{C0C0C0}$4541.5 \pm 831.8$ \\ 
\multicolumn{1}{c|}{} &  \multicolumn{1}{c|}{$\mathrm{MEGP}_{50}$} &  \multicolumn{1}{c|}{$0.679 \pm 0.044$} &  $0.008 \pm 0.001$ &  $0.08 \pm 0.008$ &  \multicolumn{1}{c|}{$0.627 \pm 0.036$} &  $0.001 \pm 0.001$ &  $0.097 \pm 0.01$ &  \multicolumn{1}{c|}{$0.606 \pm 0.036$} &  $0.0 \pm 0.0$ &  $0.082 \pm 0.035$ &  \multicolumn{1}{c|}{$0.606 \pm 0.036$} &  $0.003 \pm 0.001$ &  $0.092 \pm 0.009$ &  $0.725 \pm 0.06$ &  $4331.4 \pm 669.8$ \\ 
\multicolumn{1}{c|}{} &  \multicolumn{1}{c|}{\cellcolor[HTML]{C0C0C0}$\mathrm{MEGP}_{75}$} &  \multicolumn{1}{c|}{\cellcolor[HTML]{C0C0C0}$0.699 \pm 0.04$} &  \cellcolor[HTML]{C0C0C0}$0.007 \pm 0.001$ &  \cellcolor[HTML]{C0C0C0}$0.079 \pm 0.012$ &  \multicolumn{1}{c|}{\cellcolor[HTML]{C0C0C0}$0.637 \pm 0.034$} &  \cellcolor[HTML]{C0C0C0}$0.001 \pm 0.001$ &  \cellcolor[HTML]{C0C0C0}$0.092 \pm 0.015$ &  \multicolumn{1}{c|}{\cellcolor[HTML]{C0C0C0}$0.612 \pm 0.033$} &  \cellcolor[HTML]{C0C0C0}$0.0 \pm 0.0$ &  \cellcolor[HTML]{C0C0C0}$0.081 \pm 0.029$ &  \multicolumn{1}{c|}{\cellcolor[HTML]{C0C0C0}$0.612 \pm 0.033$} &  \cellcolor[HTML]{C0C0C0}$0.003 \pm 0.001$ &  \cellcolor[HTML]{C0C0C0}$0.088 \pm 0.013$ &  \cellcolor[HTML]{C0C0C0}$0.757 \pm 0.057$ &  \cellcolor[HTML]{C0C0C0}$4763.4 \pm 570.8$ \\ 
\multicolumn{1}{c|}{} &  \multicolumn{1}{c|}{$\mathrm{MEGP}_{100}$} &  \multicolumn{1}{c|}{$0.692 \pm 0.034$} &  $0.008 \pm 0.001$ &  $0.069 \pm 0.008$ &  \multicolumn{1}{c|}{$0.648 \pm 0.037$} &  $0.001 \pm 0.001$ &  $0.074 \pm 0.021$ &  \multicolumn{1}{c|}{$0.619 \pm 0.045$} &  $0.001 \pm 0.001$ &  $0.067 \pm 0.034$ &  \multicolumn{1}{c|}{$0.619 \pm 0.045$} &  $0.004 \pm 0.001$ &  $0.077 \pm 0.01$ &  $0.835 \pm 0.09$ &  $4510.0 \pm 966.1$ \\ 
\multicolumn{1}{c|}{} &  \multicolumn{1}{c|}{\cellcolor[HTML]{C0C0C0}BGP} &  \multicolumn{1}{c|}{\cellcolor[HTML]{C0C0C0}$0.801 \pm 0.041$} &  \cellcolor[HTML]{C0C0C0}$0.009 \pm 0.002$ &  \cellcolor[HTML]{C0C0C0}$0.088 \pm 0.012$ &  \multicolumn{1}{c|}{\cellcolor[HTML]{C0C0C0}$0.757 \pm 0.048$} &  \cellcolor[HTML]{C0C0C0}$0.001 \pm 0.001$ &  \cellcolor[HTML]{C0C0C0}$0.083 \pm 0.036$ &  \multicolumn{1}{c|}{\cellcolor[HTML]{C0C0C0}$0.745 \pm 0.058$} &  \cellcolor[HTML]{C0C0C0}$0.0 \pm 0.0$ &  \cellcolor[HTML]{C0C0C0}$0.052 \pm 0.048$ &  \multicolumn{1}{c|}{\cellcolor[HTML]{C0C0C0}$0.745 \pm 0.058$} &  \cellcolor[HTML]{C0C0C0}$0.005 \pm 0.002$ &  \cellcolor[HTML]{C0C0C0}$0.097 \pm 0.014$ &  \cellcolor[HTML]{C0C0C0}$0.642 \pm 0.077$ &  \cellcolor[HTML]{C0C0C0}$1672.7 \pm 588.6$ \\ 
\multicolumn{1}{c|}{} &  \multicolumn{1}{c|}{Friedman's P-value} &  \multicolumn{1}{c|}{$\mathbf{9.44e-14}$} &  $\mathbf{1.07e-05}$ &  $\mathbf{5.60e-15}$ &  \multicolumn{1}{c|}{$\mathbf{5.75e-15}$} &  $0.39$ &  $\mathbf{3.77e-11}$ &  \multicolumn{1}{c|}{$\mathbf{9.10e-14}$} &  $\mathbf{0.02}$ &  $\mathbf{9.13e-10}$ &  \multicolumn{1}{c|}{$\mathbf{9.10e-14}$} &  $\mathbf{9.59e-10}$ &  $\mathbf{3.77e-15}$ &  $\mathbf{5.87e-14}$ &  $\mathbf{1.12e-13}$ \\ 
\multicolumn{1}{c|}{\multirow{-8}{*}{HAPT}} &  \multicolumn{1}{c|}{\cellcolor[HTML]{C0C0C0}Adjusted P-value} &  \multicolumn{1}{c|}{\cellcolor[HTML]{C0C0C0}$\mathbf{2.83e-13}$} &  \cellcolor[HTML]{C0C0C0}$\mathbf{3.21e-05}$ &  \cellcolor[HTML]{C0C0C0}$\mathbf{1.68e-14}$ &  \multicolumn{1}{c|}{\cellcolor[HTML]{C0C0C0}$\mathbf{1.73e-14}$} &  \cellcolor[HTML]{C0C0C0}$1.00$ &  \cellcolor[HTML]{C0C0C0}$\mathbf{1.13e-10}$ &  \multicolumn{1}{c|}{\cellcolor[HTML]{C0C0C0}$\mathbf{2.73e-13}$} &  \cellcolor[HTML]{C0C0C0}$0.05$ &  \cellcolor[HTML]{C0C0C0}$\mathbf{2.74e-09}$ &  \multicolumn{1}{c|}{\cellcolor[HTML]{C0C0C0}$\mathbf{4.55e-13}$} &  \cellcolor[HTML]{C0C0C0}$\mathbf{4.80e-09}$ &  \cellcolor[HTML]{C0C0C0}$\mathbf{1.89e-14}$ &  \cellcolor[HTML]{C0C0C0}$\mathbf{2.94e-13}$ &  \cellcolor[HTML]{C0C0C0}$\mathbf{5.61e-13}$ \\ \hline

\multicolumn{1}{c|}{} &  \multicolumn{1}{c|}{$\mathrm{MEGP}_{0}$} &  \multicolumn{1}{c|}{$1.692 \pm 0.053$} &  $0.01 \pm 0.001$ &  $0.061 \pm 0.004$ &  \multicolumn{1}{c|}{$1.617 \pm 0.045$} &  $0.001 \pm 0.001$ &  $0.082 \pm 0.005$ &  \multicolumn{1}{c|}{$1.579 \pm 0.043$} &  $0.001 \pm 0.001$ &  $0.084 \pm 0.015$ &  \multicolumn{1}{c|}{$1.579 \pm 0.043$} &  $0.004 \pm 0.001$ &  $0.077 \pm 0.004$ &  $1.421 \pm 0.049$ &  $5939.7 \pm 660.8$ \\ 
\multicolumn{1}{c|}{} &  \multicolumn{1}{c|}{\cellcolor[HTML]{C0C0C0}$\mathrm{MEGP}_{25}$} &  \multicolumn{1}{c|}{\cellcolor[HTML]{C0C0C0}$1.665 \pm 0.049$} &  \cellcolor[HTML]{C0C0C0}$0.011 \pm 0.001$ &  \cellcolor[HTML]{C0C0C0}$0.058 \pm 0.004$ &  \multicolumn{1}{c|}{\cellcolor[HTML]{C0C0C0}$1.606 \pm 0.048$} &  \cellcolor[HTML]{C0C0C0}$0.001 \pm 0.001$ &  \cellcolor[HTML]{C0C0C0}$0.074 \pm 0.009$ &  \multicolumn{1}{c|}{\cellcolor[HTML]{C0C0C0}$1.575 \pm 0.05$} &  \cellcolor[HTML]{C0C0C0}$0.001 \pm 0.001$ &  \cellcolor[HTML]{C0C0C0}$0.057 \pm 0.034$ &  \multicolumn{1}{c|}{\cellcolor[HTML]{C0C0C0}$1.575 \pm 0.05$} &  \cellcolor[HTML]{C0C0C0}$0.005 \pm 0.001$ &  \cellcolor[HTML]{C0C0C0}$0.071 \pm 0.005$ &  \cellcolor[HTML]{C0C0C0}$1.471 \pm 0.06$ &  \cellcolor[HTML]{C0C0C0}$5393.4 \pm 1268.1$ \\ 
\multicolumn{1}{c|}{} &  \multicolumn{1}{c|}{$\mathrm{MEGP}_{50}$} &  \multicolumn{1}{c|}{$1.668 \pm 0.046$} &  $0.011 \pm 0.001$ &  $0.054 \pm 0.003$ &  \multicolumn{1}{c|}{$1.591 \pm 0.049$} &  $0.001 \pm 0.001$ &  $0.07 \pm 0.008$ &  \multicolumn{1}{c|}{$1.554 \pm 0.05$} &  $0.001 \pm 0.001$ &  $0.057 \pm 0.029$ &  \multicolumn{1}{c|}{$1.554 \pm 0.05$} &  $0.005 \pm 0.001$ &  $0.066 \pm 0.004$ &  $1.462 \pm 0.06$ &  $5335.4 \pm 914.7$ \\ 
\multicolumn{1}{c|}{} &  \multicolumn{1}{c|}{\cellcolor[HTML]{C0C0C0}$\mathrm{MEGP}_{75}$} &  \multicolumn{1}{c|}{\cellcolor[HTML]{C0C0C0}$1.684 \pm 0.049$} &  \cellcolor[HTML]{C0C0C0}$0.011 \pm 0.001$ &  \cellcolor[HTML]{C0C0C0}$0.05 \pm 0.004$ &  \multicolumn{1}{c|}{\cellcolor[HTML]{C0C0C0}$1.592 \pm 0.051$} &  \cellcolor[HTML]{C0C0C0}$0.002 \pm 0.001$ &  \cellcolor[HTML]{C0C0C0}$0.07 \pm 0.006$ &  \multicolumn{1}{c|}{\cellcolor[HTML]{C0C0C0}$1.562 \pm 0.044$} &  \cellcolor[HTML]{C0C0C0}$0.001 \pm 0.0$ &  \cellcolor[HTML]{C0C0C0}$0.063 \pm 0.021$ &  \multicolumn{1}{c|}{\cellcolor[HTML]{C0C0C0}$1.562 \pm 0.044$} &  \cellcolor[HTML]{C0C0C0}$0.005 \pm 0.001$ &  \cellcolor[HTML]{C0C0C0}$0.064 \pm 0.004$ &  \cellcolor[HTML]{C0C0C0}$1.519 \pm 0.062$ &  \cellcolor[HTML]{C0C0C0}$5703.4 \pm 634.0$ \\ 
\multicolumn{1}{c|}{} &  \multicolumn{1}{c|}{$\mathrm{MEGP}_{100}$} &  \multicolumn{1}{c|}{$1.687 \pm 0.055$} &  $0.01 \pm 0.001$ &  $0.047 \pm 0.003$ &  \multicolumn{1}{c|}{$1.609 \pm 0.057$} &  $0.002 \pm 0.001$ &  $0.062 \pm 0.009$ &  \multicolumn{1}{c|}{$1.583 \pm 0.058$} &  $0.001 \pm 0.0$ &  $0.052 \pm 0.028$ &  \multicolumn{1}{c|}{$1.583 \pm 0.058$} &  $0.005 \pm 0.001$ &  $0.06 \pm 0.003$ &  $1.565 \pm 0.068$ &  $5351.4 \pm 1003.2$ \\ 
\multicolumn{1}{c|}{} &  \multicolumn{1}{c|}{\cellcolor[HTML]{C0C0C0}BGP} &  \multicolumn{1}{c|}{\cellcolor[HTML]{C0C0C0}$1.743 \pm 0.052$} &  \cellcolor[HTML]{C0C0C0}$0.011 \pm 0.002$ &  \cellcolor[HTML]{C0C0C0}$0.055 \pm 0.004$ &  \multicolumn{1}{c|}{\cellcolor[HTML]{C0C0C0}$1.688 \pm 0.045$} &  \cellcolor[HTML]{C0C0C0}$0.001 \pm 0.001$ &  \cellcolor[HTML]{C0C0C0}$0.064 \pm 0.016$ &  \multicolumn{1}{c|}{\cellcolor[HTML]{C0C0C0}$1.673 \pm 0.05$} &  \cellcolor[HTML]{C0C0C0}$0.0 \pm 0.001$ &  \cellcolor[HTML]{C0C0C0}$0.033 \pm 0.033$ &  \multicolumn{1}{c|}{\cellcolor[HTML]{C0C0C0}$1.673 \pm 0.05$} &  \cellcolor[HTML]{C0C0C0}$0.006 \pm 0.002$ &  \cellcolor[HTML]{C0C0C0}$0.066 \pm 0.005$ &  \cellcolor[HTML]{C0C0C0}$1.253 \pm 0.072$ &  \cellcolor[HTML]{C0C0C0}$1052.3 \pm 254.4$ \\ 
\multicolumn{1}{c|}{} &  \multicolumn{1}{c|}{Friedman's P-value} &  \multicolumn{1}{c|}{$\mathbf{2.69e-06}$} &  $\mathbf{5.88e-03}$ &  $\mathbf{4.91e-20}$ &  \multicolumn{1}{c|}{$\mathbf{6.27e-08}$} &  $\mathbf{1.56e-03}$ &  $\mathbf{4.21e-12}$ &  \multicolumn{1}{c|}{$\mathbf{1.88e-11}$} &  $\mathbf{0.02}$ &  $\mathbf{2.24e-12}$ &  \multicolumn{1}{c|}{$\mathbf{1.88e-11}$} &  $\mathbf{7.34e-05}$ &  $\mathbf{1.09e-18}$ &  $\mathbf{3.93e-20}$ &  $\mathbf{3.79e-14}$ \\ 
\multicolumn{1}{c|}{\multirow{-8}{*}{ISOLET}} &  \multicolumn{1}{c|}{\cellcolor[HTML]{C0C0C0}Adjusted P-value} &  \multicolumn{1}{c|}{\cellcolor[HTML]{C0C0C0}$\mathbf{8.07e-06}$} &  \cellcolor[HTML]{C0C0C0}$\mathbf{0.02}$ &  \cellcolor[HTML]{C0C0C0}$\mathbf{1.47e-19}$ &  \multicolumn{1}{c|}{\cellcolor[HTML]{C0C0C0}$\mathbf{1.88e-07}$} &  \cellcolor[HTML]{C0C0C0}$\mathbf{4.69e-03}$ &  \cellcolor[HTML]{C0C0C0}$\mathbf{1.26e-11}$ &  \multicolumn{1}{c|}{\cellcolor[HTML]{C0C0C0}$\mathbf{5.63e-11}$} &  \cellcolor[HTML]{C0C0C0}$\mathbf{0.05}$ &  \cellcolor[HTML]{C0C0C0}$\mathbf{6.72e-12}$ &  \multicolumn{1}{c|}{\cellcolor[HTML]{C0C0C0}$\mathbf{9.39e-11}$} &  \cellcolor[HTML]{C0C0C0}$\mathbf{3.67e-04}$ &  \cellcolor[HTML]{C0C0C0}$\mathbf{5.46e-18}$ &  \cellcolor[HTML]{C0C0C0}$\mathbf{1.97e-19}$ &  \cellcolor[HTML]{C0C0C0}$\mathbf{1.89e-13}$ \\ \hline

\multicolumn{1}{c|}{} &  \multicolumn{1}{c|}{$\mathrm{MEGP}_{0}$} &  \multicolumn{1}{c|}{$0.379 \pm 0.017$} &  $0.002 \pm 0.0$ &  $0.083 \pm 0.009$ &  \multicolumn{1}{c|}{$0.361 \pm 0.017$} &  $0.0 \pm 0.0$ &  $0.085 \pm 0.011$ &  \multicolumn{1}{c|}{$0.354 \pm 0.019$} &  $0.0 \pm 0.0$ &  $0.067 \pm 0.032$ &  \multicolumn{1}{c|}{$0.354 \pm 0.019$} &  $0.001 \pm 0.0$ &  $0.086 \pm 0.01$ &  $0.407 \pm 0.079$ &  $1105.2 \pm 258.2$ \\ 
\multicolumn{1}{c|}{} &  \multicolumn{1}{c|}{\cellcolor[HTML]{C0C0C0}$\mathrm{MEGP}_{25}$} &  \multicolumn{1}{c|}{\cellcolor[HTML]{C0C0C0}$0.37 \pm 0.016$} &  \cellcolor[HTML]{C0C0C0}$0.002 \pm 0.0$ &  \cellcolor[HTML]{C0C0C0}$0.072 \pm 0.008$ &  \multicolumn{1}{c|}{\cellcolor[HTML]{C0C0C0}$0.351 \pm 0.019$} &  \cellcolor[HTML]{C0C0C0}$0.0 \pm 0.0$ &  \cellcolor[HTML]{C0C0C0}$0.074 \pm 0.015$ &  \multicolumn{1}{c|}{\cellcolor[HTML]{C0C0C0}$0.342 \pm 0.022$} &  \cellcolor[HTML]{C0C0C0}$0.0 \pm 0.0$ &  \cellcolor[HTML]{C0C0C0}$0.054 \pm 0.029$ &  \multicolumn{1}{c|}{\cellcolor[HTML]{C0C0C0}$0.342 \pm 0.022$} &  \cellcolor[HTML]{C0C0C0}$0.001 \pm 0.0$ &  \cellcolor[HTML]{C0C0C0}$0.074 \pm 0.01$ &  \cellcolor[HTML]{C0C0C0}$0.4 \pm 0.089$ &  \cellcolor[HTML]{C0C0C0}$1051.5 \pm 206.7$ \\ 
\multicolumn{1}{c|}{} &  \multicolumn{1}{c|}{$\mathrm{MEGP}_{50}$} &  \multicolumn{1}{c|}{$0.363 \pm 0.025$} &  $0.002 \pm 0.001$ &  $0.065 \pm 0.008$ &  \multicolumn{1}{c|}{$0.345 \pm 0.025$} &  $0.0 \pm 0.0$ &  $0.064 \pm 0.018$ &  \multicolumn{1}{c|}{$0.337 \pm 0.026$} &  $0.0 \pm 0.0$ &  $0.047 \pm 0.033$ &  \multicolumn{1}{c|}{$0.337 \pm 0.026$} &  $0.001 \pm 0.0$ &  $0.068 \pm 0.009$ &  $0.413 \pm 0.06$ &  $833.2 \pm 217.3$ \\ 
\multicolumn{1}{c|}{} &  \multicolumn{1}{c|}{\cellcolor[HTML]{C0C0C0}$\mathrm{MEGP}_{75}$} &  \multicolumn{1}{c|}{\cellcolor[HTML]{C0C0C0}$0.368 \pm 0.019$} &  \cellcolor[HTML]{C0C0C0}$0.002 \pm 0.0$ &  \cellcolor[HTML]{C0C0C0}$0.06 \pm 0.01$ &  \multicolumn{1}{c|}{\cellcolor[HTML]{C0C0C0}$0.346 \pm 0.022$} &  \cellcolor[HTML]{C0C0C0}$0.0 \pm 0.0$ &  \cellcolor[HTML]{C0C0C0}$0.061 \pm 0.012$ &  \multicolumn{1}{c|}{\cellcolor[HTML]{C0C0C0}$0.338 \pm 0.024$} &  \cellcolor[HTML]{C0C0C0}$0.0 \pm 0.0$ &  \cellcolor[HTML]{C0C0C0}$0.044 \pm 0.026$ &  \multicolumn{1}{c|}{\cellcolor[HTML]{C0C0C0}$0.338 \pm 0.024$} &  \cellcolor[HTML]{C0C0C0}$0.001 \pm 0.0$ &  \cellcolor[HTML]{C0C0C0}$0.062 \pm 0.009$ &  \cellcolor[HTML]{C0C0C0}$0.442 \pm 0.067$ &  \cellcolor[HTML]{C0C0C0}$1034.8 \pm 257.7$ \\ 
\multicolumn{1}{c|}{} &  \multicolumn{1}{c|}{$\mathrm{MEGP}_{100}$} &  \multicolumn{1}{c|}{$0.365 \pm 0.018$} &  $0.002 \pm 0.0$ &  $0.052 \pm 0.008$ &  \multicolumn{1}{c|}{$0.344 \pm 0.019$} &  $0.0 \pm 0.0$ &  $0.055 \pm 0.012$ &  \multicolumn{1}{c|}{$0.336 \pm 0.021$} &  $0.0 \pm 0.0$ &  $0.042 \pm 0.024$ &  \multicolumn{1}{c|}{$0.336 \pm 0.021$} &  $0.001 \pm 0.0$ &  $0.055 \pm 0.009$ &  $0.419 \pm 0.067$ &  $1082.2 \pm 282.4$ \\ 
\multicolumn{1}{c|}{} &  \multicolumn{1}{c|}{\cellcolor[HTML]{C0C0C0}BGP} &  \multicolumn{1}{c|}{\cellcolor[HTML]{C0C0C0}$0.386 \pm 0.02$} &  \cellcolor[HTML]{C0C0C0}$0.002 \pm 0.0$ &  \cellcolor[HTML]{C0C0C0}$0.074 \pm 0.01$ &  \multicolumn{1}{c|}{\cellcolor[HTML]{C0C0C0}$0.37 \pm 0.021$} &  \cellcolor[HTML]{C0C0C0}$0.0 \pm 0.0$ &  \cellcolor[HTML]{C0C0C0}$0.064 \pm 0.023$ &  \multicolumn{1}{c|}{\cellcolor[HTML]{C0C0C0}$0.366 \pm 0.022$} &  \cellcolor[HTML]{C0C0C0}$0.0 \pm 0.0$ &  \cellcolor[HTML]{C0C0C0}$0.033 \pm 0.029$ &  \multicolumn{1}{c|}{\cellcolor[HTML]{C0C0C0}$0.366 \pm 0.022$} &  \cellcolor[HTML]{C0C0C0}$0.001 \pm 0.0$ &  \cellcolor[HTML]{C0C0C0}$0.074 \pm 0.012$ &  \cellcolor[HTML]{C0C0C0}$0.374 \pm 0.095$ &  \cellcolor[HTML]{C0C0C0}$407.1 \pm 117.0$ \\ 
\multicolumn{1}{c|}{} &  \multicolumn{1}{c|}{Friedman's P-value} &  \multicolumn{1}{c|}{$\mathbf{1.95e-06}$} &  $0.10$ &  $\mathbf{2.50e-19}$ &  \multicolumn{1}{c|}{$\mathbf{2.30e-07}$} &  $0.20$ &  $\mathbf{2.00e-12}$ &  \multicolumn{1}{c|}{$\mathbf{2.82e-07}$} &  $0.26$ &  $\mathbf{1.18e-03}$ &  \multicolumn{1}{c|}{$\mathbf{2.82e-07}$} &  $\mathbf{0.01}$ &  $\mathbf{1.16e-15}$ &  $0.16$ &  $\mathbf{1.36e-15}$ \\ 
\multicolumn{1}{c|}{\multirow{-8}{*}{PD}} &  \multicolumn{1}{c|}{\cellcolor[HTML]{C0C0C0}Adjusted P-value} &  \multicolumn{1}{c|}{\cellcolor[HTML]{C0C0C0}$\mathbf{5.84e-06}$} &  \cellcolor[HTML]{C0C0C0}$0.31$ &  \cellcolor[HTML]{C0C0C0}$\mathbf{7.49e-19}$ &  \multicolumn{1}{c|}{\cellcolor[HTML]{C0C0C0}$\mathbf{6.91e-07}$} &  \cellcolor[HTML]{C0C0C0}$0.61$ &  \cellcolor[HTML]{C0C0C0}$\mathbf{5.99e-12}$ &  \multicolumn{1}{c|}{\cellcolor[HTML]{C0C0C0}$\mathbf{8.46e-07}$} &  \cellcolor[HTML]{C0C0C0}$0.79$ &  \cellcolor[HTML]{C0C0C0}$\mathbf{3.54e-03}$ &  \multicolumn{1}{c|}{\cellcolor[HTML]{C0C0C0}$\mathbf{1.41e-06}$} &  \cellcolor[HTML]{C0C0C0}$0.06$ &  \cellcolor[HTML]{C0C0C0}$\mathbf{5.78e-15}$ &  \cellcolor[HTML]{C0C0C0}$0.78$ &  \cellcolor[HTML]{C0C0C0}$\mathbf{6.82e-15}$ \\ \hline
\end{tabular}%
}
\end{sidewaystable*}

%% file: box_ft50.tex
\begin{figure*}[ht] 
\centering
\includegraphics[width=\textwidth]{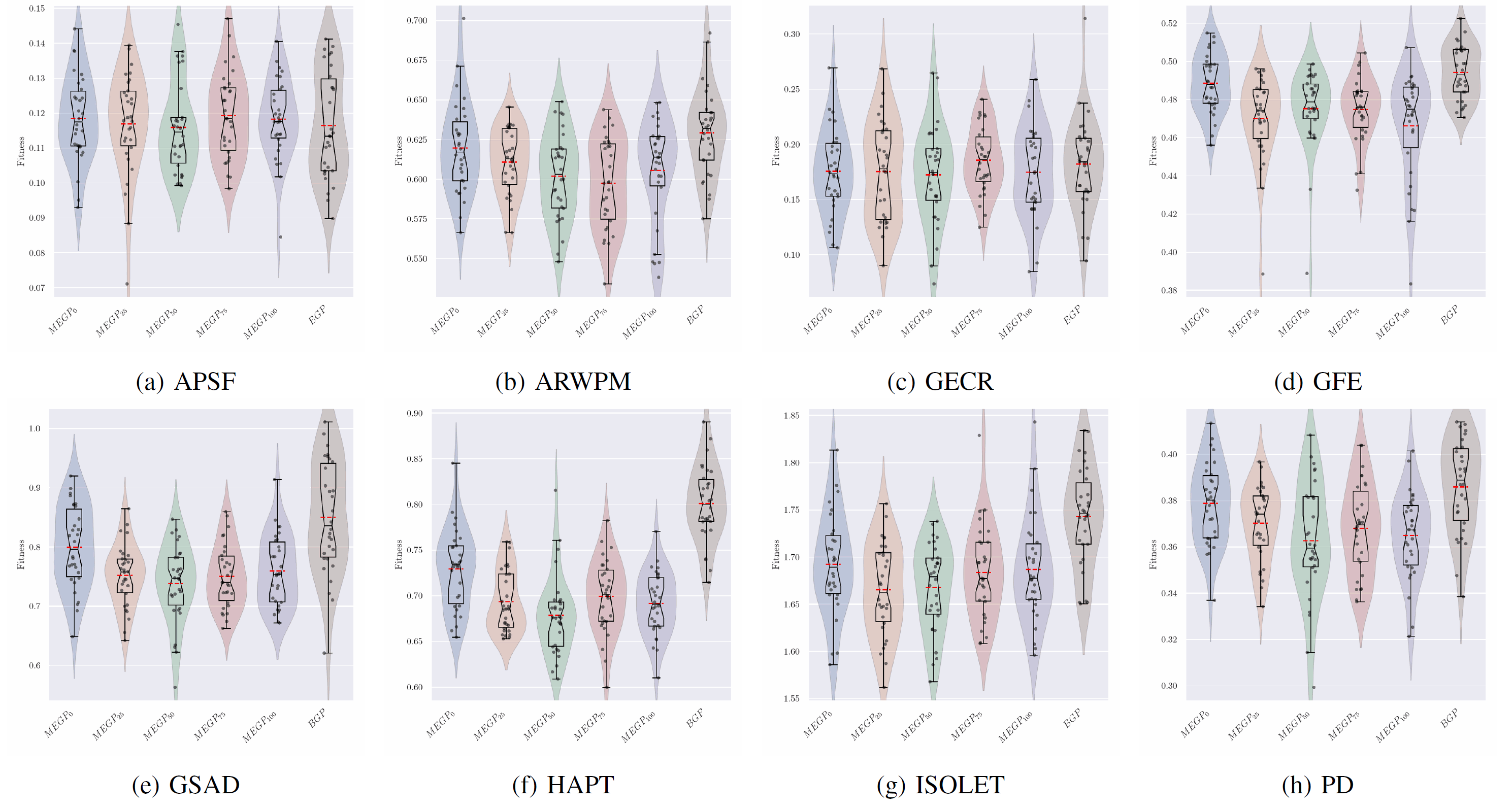}\label{fig:apsf_box_ft50}%

\caption[The distribution of FT\(_{50}\) across BGP and MEGP models over 30 runs.]{Raincloud plots showing the distribution of Fitness (FT\(_{50}\)) across 30 runs for BGP and MEGP models with different ensemble selection probabilities (0\%, 25\%, 50\%, 75\%, 100\%). Each plot illustrates the variability and central tendency of model performance over generations 1 to 50.}

\label{fig:box_ft50}
\end{figure*}

%% file: con_ft50.tex
\begin{figure*}[ht] 
\centering
\includegraphics[width=\textwidth]{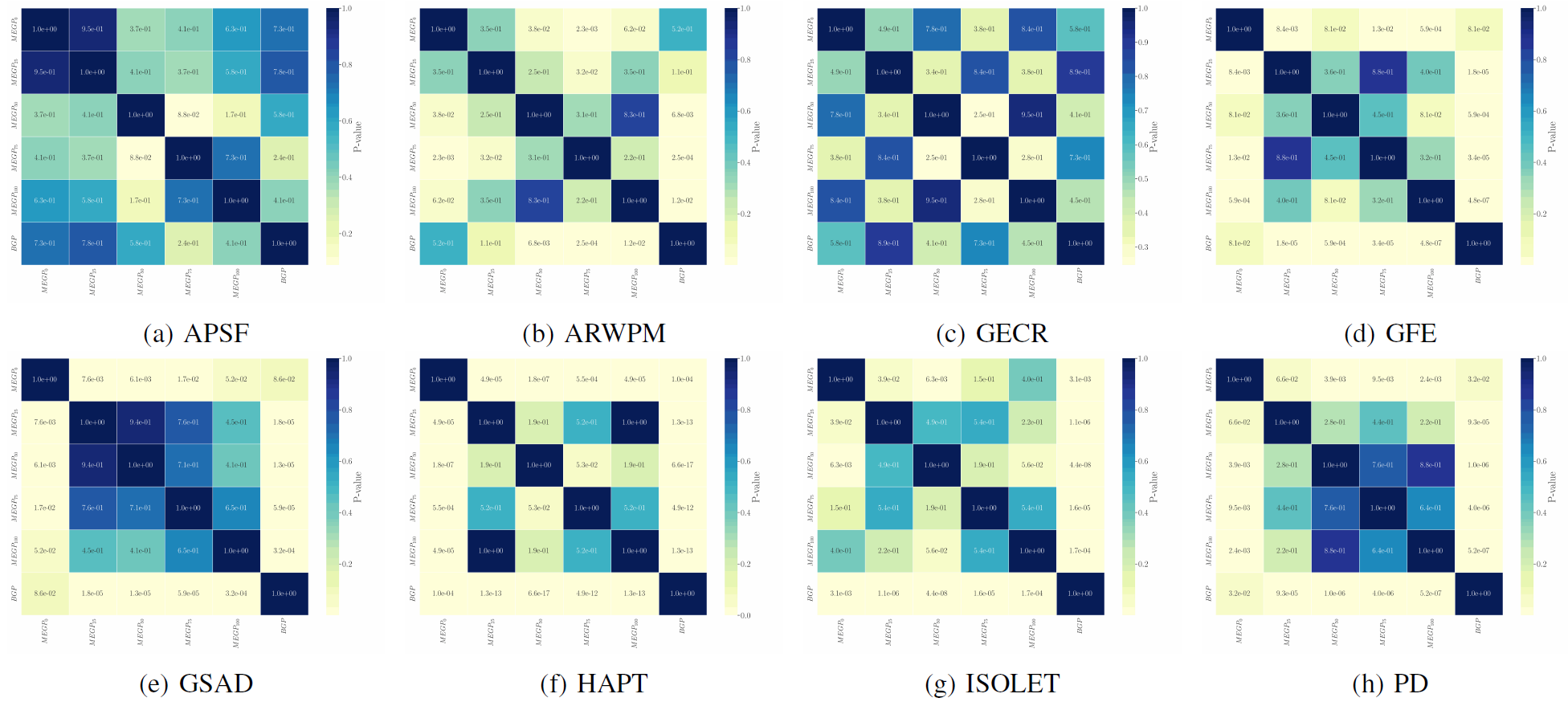}\label{fig:apsf_con_ft50}%

\caption[Heatmap of adjusted pairwise significance from Conover post-hoc test for FT\(_{50}\).]{Heatmap of adjusted p-values from the pairwise Conover post-hoc test for FT\(_{50}\), corrected using the Benjamini-Hochberg method. The heatmap highlights the statistical significance of pairwise comparisons between BGP and MEGP models with varying ensemble selection probabilities (0\%, 25\%, 50\%, 75\%, 100\%) over generations 1 to 50.}

\label{fig:con_ft50}
\end{figure*}

%% file: cliff_ft50.tex
\begin{figure*}[htbp] 
\centering
\includegraphics[width=\textwidth]{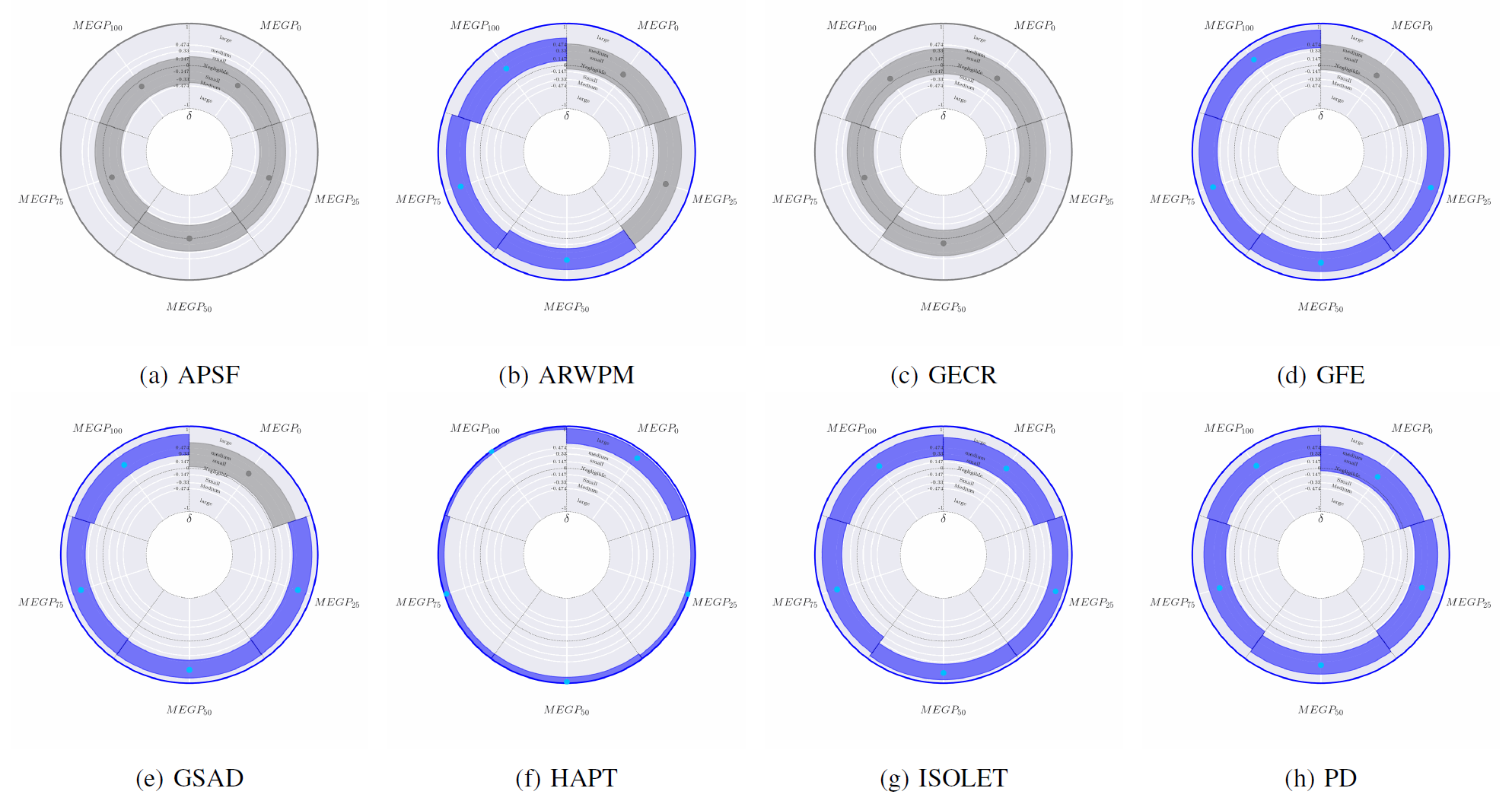}

\caption[The Cliff's $\delta$ effect size measure and its 95\% confidence intervals for FT\(_{50}\) obtained from 30 BGP and MEGP runs.]{Effect size analysis of FT\(_{50}\) across 30 runs for BGP and MEGP models using Cliff's $\delta$. Each point represents the actual FT\(_{50}\) value obtained, with segments denoting 95\% confidence intervals based on 10,000 bootstrap resamplings. The outer ring color visualizes statistical significance: grey illustrates no significant difference (adjusted Friedman's P-value $>0.05$), while color indicates significant differences; blue indicates that at least one MEGP configuration outperforms BGP (adjusted Conover's p-value $< 0.05$, Cliff's $\delta > 0$), and red signifies that all MEGP configurations underperform relative to BGP (adjusted Conover's p-value $< 0.05$, Cliff's $\delta < 0$). Segment colors show performance differences against BGP: grey for no significant difference (adjusted Conover's p-value $> 0.05$), blue for better performance (Cliff's $\delta > 0$), and red for worse performance (Cliff's $\delta < 0$).}

\label{fig:cliff_ft50}
\end{figure*}

%% file: wtl_ft50.tex
\begin{table*}
\centering
\caption[The results of Friedman and Conover tests and Cliff's $\delta$ analysis for the $FT_{50}$ obtained from MEGP and BGP runs.]{Statistical comparison of $FT_{50}$ results for training data obtained from MEGP and BGP Runs. W, T, and L denote win, tie, and loss based on adjusted Friedman and Conover's p-values. Effect sizes are calculated using Cliff's Delta method and are categorized as negligible, small, medium, or large.}
\label{tab:wtl_ft50}
\begin{tabular}{cccccc}
\hline
\multicolumn{6}{c}{$FT_{50}$} \\
\hline
Dataset & $MEGP_{0}$ & $MEGP_{25}$ & $MEGP_{50}$ & $MEGP_{75}$ & $MEGP_{100}$ \\
\hline
APSF & T (negligible) & T (negligible) & T (negligible) & T (negligible) & T (negligible) \\
ARWPM & T (small) & T (medium) & W (large) & W (large) & W (medium) \\
GECR & T (negligible) & T (negligible) & T (negligible) & T (negligible) & T (negligible) \\
GFE & T (small) & W (large) & W (large) & W (large) & W (large) \\
GSAD & T (small) & W (large) & W (large) & W (large) & W (large) \\
HAPT & W (large) & W (large) & W (large) & W (large) & W (large) \\
ISOLET & W (medium) & W (large) & W (large) & W (large) & W (large) \\
PD & W (small) & W (medium) & W (large) & W (medium) & W (large) \\
\hline
W - T - L & 3 - 5 - 0 & 5 - 3 - 0 & 6 - 2 - 0 & 6 - 2 - 0 & 6 - 2 - 0 \\
\hline
\end{tabular}
\end{table*}

%% file: box_cr50.tex
\begin{figure*}[ht] 
\centering
\includegraphics[width=\textwidth]{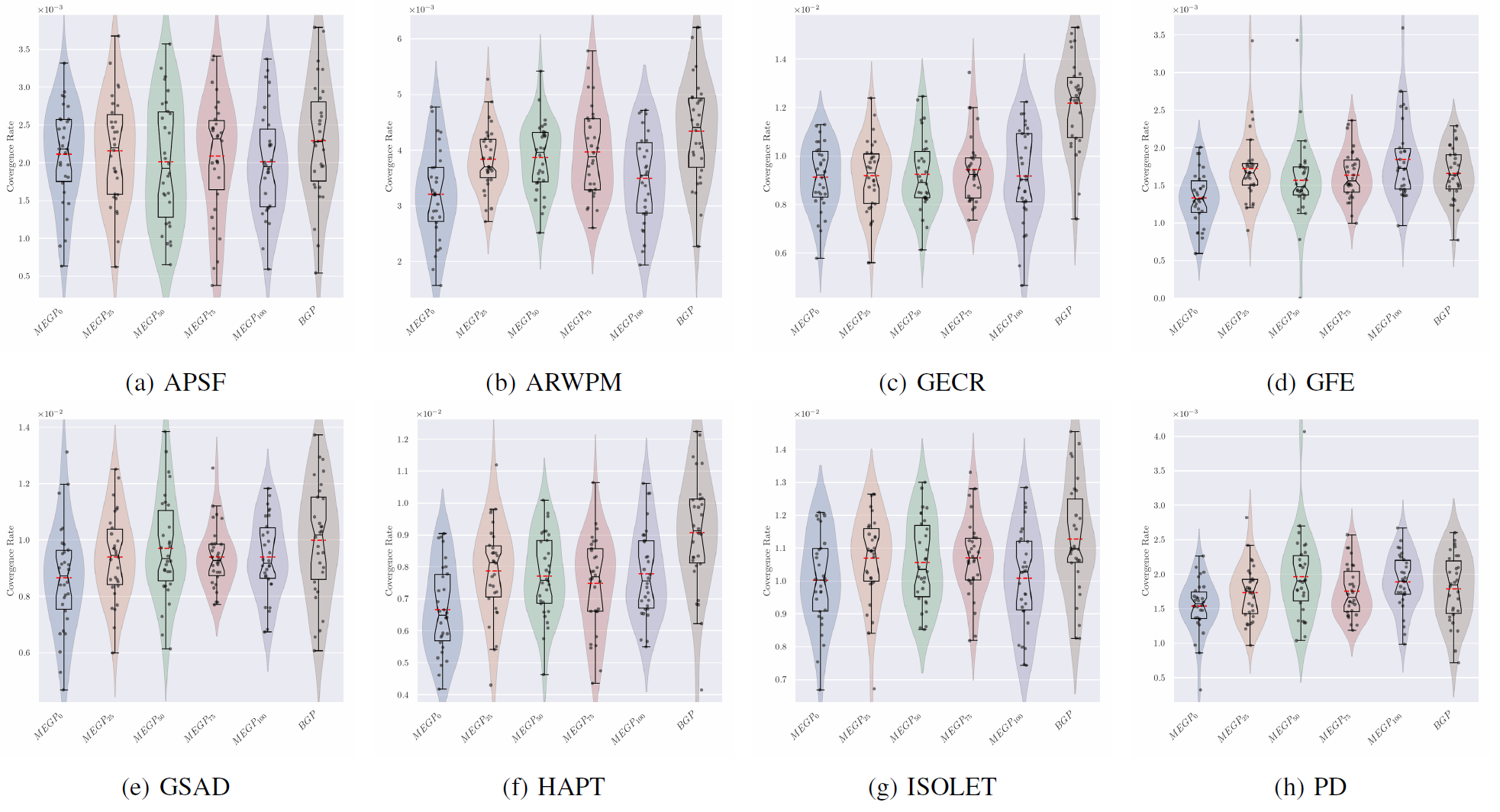}
\caption[The distribution of CR\(_{50}\) across BGP and MEGP models over 30 runs.]{Raincloud plots showing the distribution of Convergence Rate (CR\(_{50}\)) across 30 runs for BGP and MEGP models with different ensemble selection probabilities (0\%, 25\%, 50\%, 75\%, 100\%). Each plot illustrates the variability and central tendency of model performance over generations 1 to 50.}

\label{fig:box_cr50}
\end{figure*}

%% file: con_cr50.tex
\begin{figure*}[ht] 
\centering
\includegraphics[width=\textwidth]{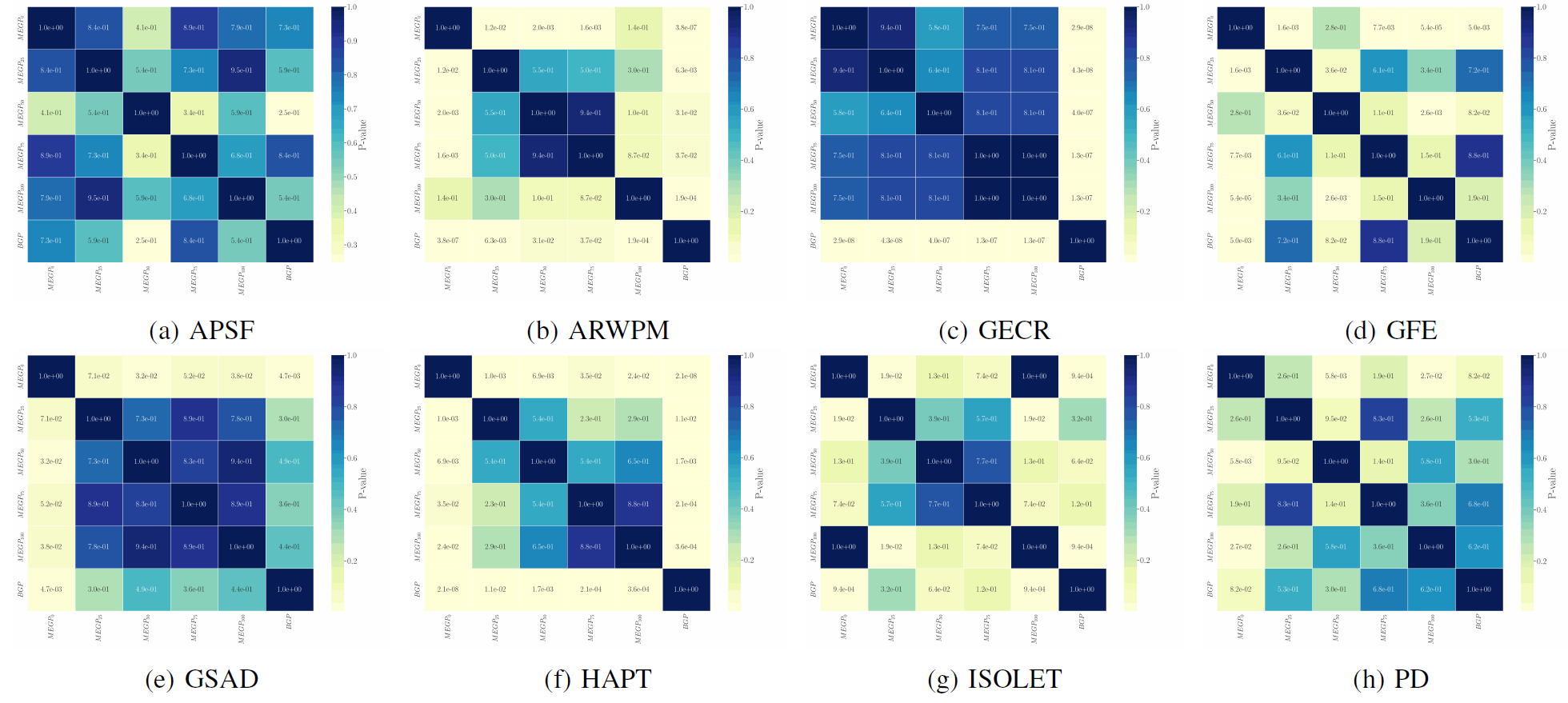}\label{fig:apsf_con_cr50}%

\caption[Heatmap of adjusted pairwise significance from Conover post-hoc test for CR\(_{50}\).]{Heatmap of adjusted p-values from the pairwise Conover post-hoc test for CR\(_{50}\), corrected using the Benjamini-Hochberg method. The heatmap highlights the statistical significance of pairwise comparisons between BGP and MEGP models with varying ensemble selection probabilities (0\%, 25\%, 50\%, 75\%, 100\%) over generations 1 to 50.}

\label{fig:con_cr50}
\end{figure*}

%% file: cliff_cr50.tex
\begin{figure*}[htbp] 
\centering
\includegraphics[width=\textwidth]{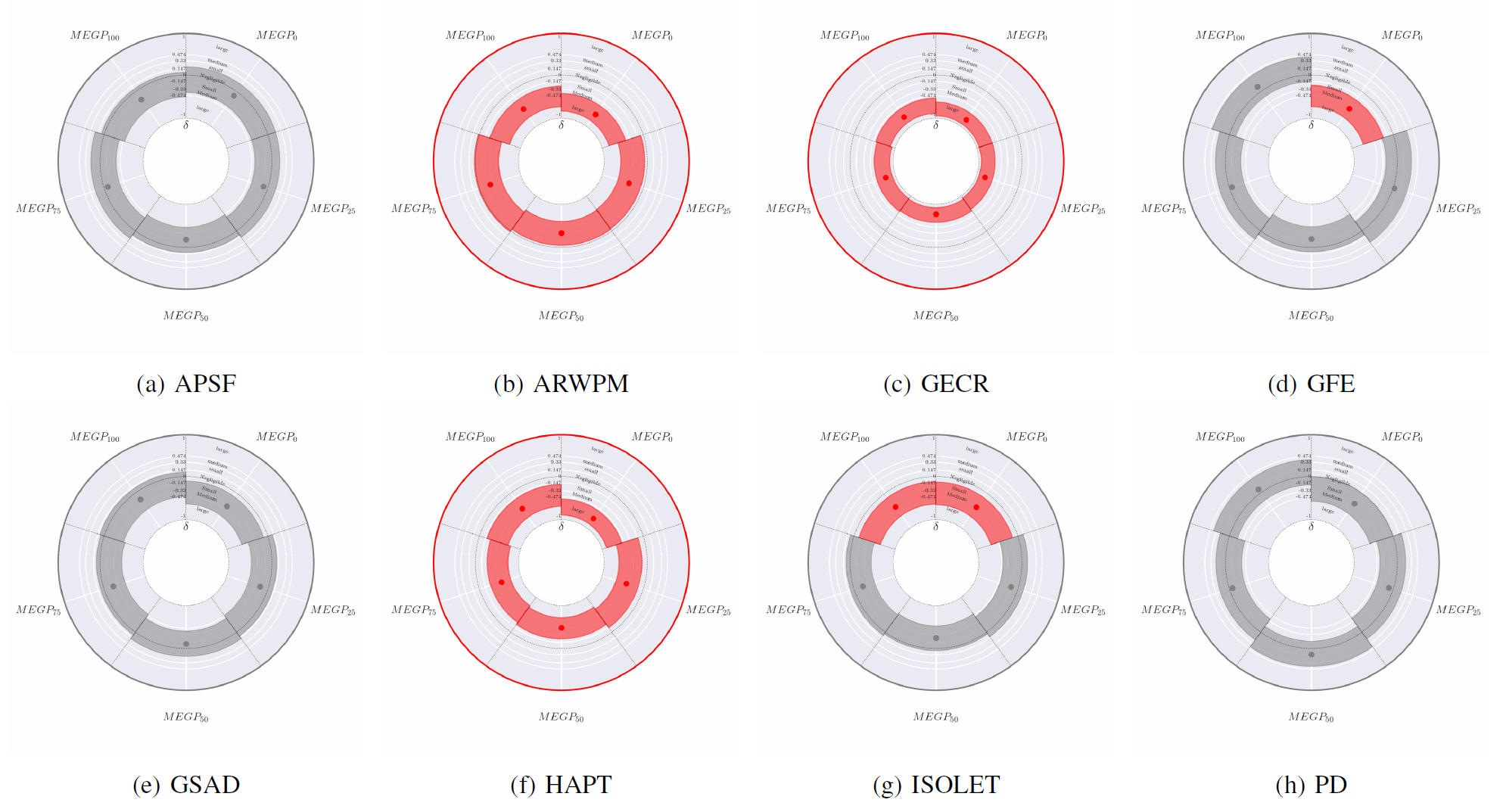}\label{fig:apsf_cliff_cr50}%

\caption[The Cliff's $\delta$ effect size measure and its 95\% confidence intervals for CR\(_{50}\) obtained from 30 BGP and MEGP runs.]{Effect size analysis of CR\(_{50}\) across 30 runs for BGP and MEGP models using Cliff's $\delta$. Each point represents the actual CR\(_{50}\) value obtained, with segments denoting 95\% confidence intervals based on 10,000 bootstrap resamplings. The outer ring color visualizes statistical significance: grey illustrates no significant difference (adjusted Friedman's P-value $>0.05$), while color indicates significant differences; blue indicates that at least one MEGP configuration outperforms BGP (adjusted Conover's p-value $< 0.05$, Cliff's $\delta > 0$), and red signifies that all MEGP configurations underperform relative to BGP (adjusted Conover's p-value $< 0.05$, Cliff's $\delta < 0$). Segment colors show performance differences against BGP: grey for no significant difference (adjusted Conover's p-value $> 0.05$), blue for better performance (Cliff's $\delta > 0$), and red for worse performance (Cliff's $\delta < 0$).}

\label{fig:cliff_cr50}
\end{figure*}

%% file: wtl_cr50.tex
\begin{table*}
\centering
\caption[The results of Friedman and Conover tests and Cliff's $\delta$ analysis for the $CR_{50}$ obtained from MEGP and BGP runs.]{Statistical comparison of $CR_{50}$ results for training data obtained from MEGP and BGP Runs. W, T, and L denote win, tie, and loss based on adjusted Friedman and Conover's p-values. Effect sizes are calculated using Cliff's Delta method and are categorized as negligible, small, medium, or large.}
\label{tab:wtl_cr50}
\begin{tabular}{cccccc}
\hline
\multicolumn{6}{c}{$CR_{50}$} \\
\hline
Dataset & $MEGP_{0}$ & $MEGP_{25}$ & $MEGP_{50}$ & $MEGP_{75}$ & $MEGP_{100}$ \\
\hline
APSF & T (negligible) & T (negligible) & T (small) & T (negligible) & T (small) \\
ARWPM & L (large) & L (medium) & L (medium) & L (small) & L (large) \\
GECR & L (large) & L (large) & L (large) & L (large) & L (large) \\
GFE & L (large) & T (negligible) & T (small) & T (negligible) & T (negligible) \\
GSAD & T (medium) & T (small) & T (negligible) & T (small) & T (small) \\
HAPT & L (large) & L (medium) & L (large) & L (large) & L (medium) \\
ISOLET & L (medium) & T (small) & T (small) & T (small) & L (medium) \\
PD & T (small) & T (negligible) & T (negligible) & T (negligible) & T (negligible) \\
\hline
W - T - L & 0 - 3 - 5 & 0 - 5 - 3 & 0 - 5 - 3 & 0 - 5 - 3 & 0 - 4 - 4 \\
\hline
\end{tabular}
\end{table*}

%% file: box_ccr50.tex
\begin{figure*}[ht] 
\centering
\includegraphics[width=\textwidth]{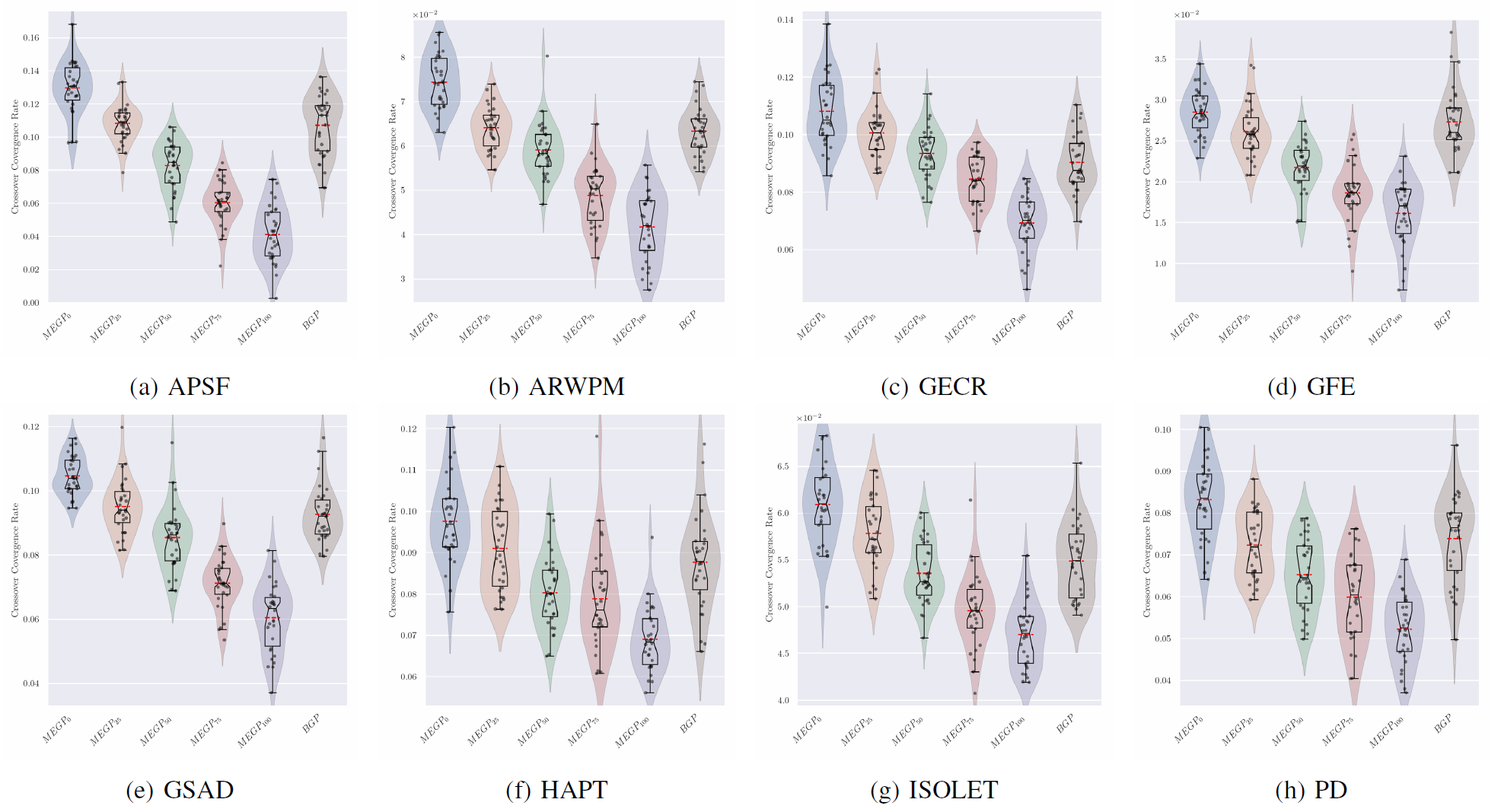}\label{fig:apsf_box_ccr50}%

\caption[The distribution of CCR\(_{50}\) across BGP and MEGP models over 30 runs.]{Raincloud plots showing the distribution of Crossover Convergence Rate (CCR\(_{50}\)) across 30 runs for BGP and MEGP models with different ensemble selection probabilities (0\%, 25\%, 50\%, 75\%, 100\%). Each plot illustrates the variability and central tendency of model performance over generations 1 to 50.}

\label{fig:box_ccr50}
\end{figure*}

%% file: con_ccr50.tex
\begin{figure*}[ht] 
\centering
\includegraphics[width=\textwidth]{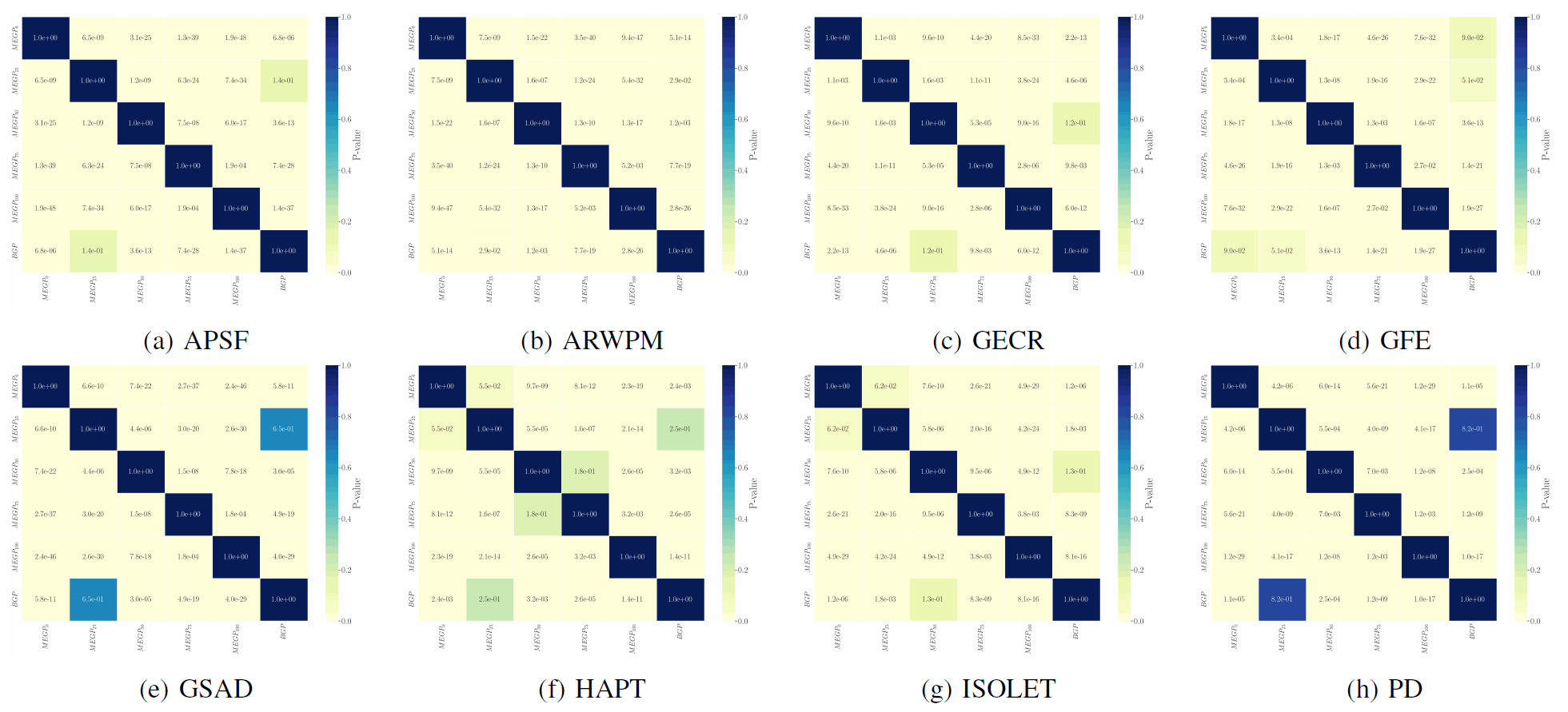}\label{fig:apsf_con_ccr50}%

\caption[Heatmap of adjusted pairwise significance from Conover post-hoc test for CCR\(_{50}\).]{Heatmap of adjusted p-values from the pairwise Conover post-hoc test for CCR\(_{50}\), corrected using the Benjamini-Hochberg method. The heatmap highlights the statistical significance of pairwise comparisons between BGP and MEGP models with varying ensemble selection probabilities (0\%, 25\%, 50\%, 75\%, 100\%) over generations 1 to 50.}

\label{fig:con_ccr50}
\end{figure*}

%% file: cliff_ccr50.tex
\begin{figure*}[htbp] 
\centering
\includegraphics[width=\textwidth]{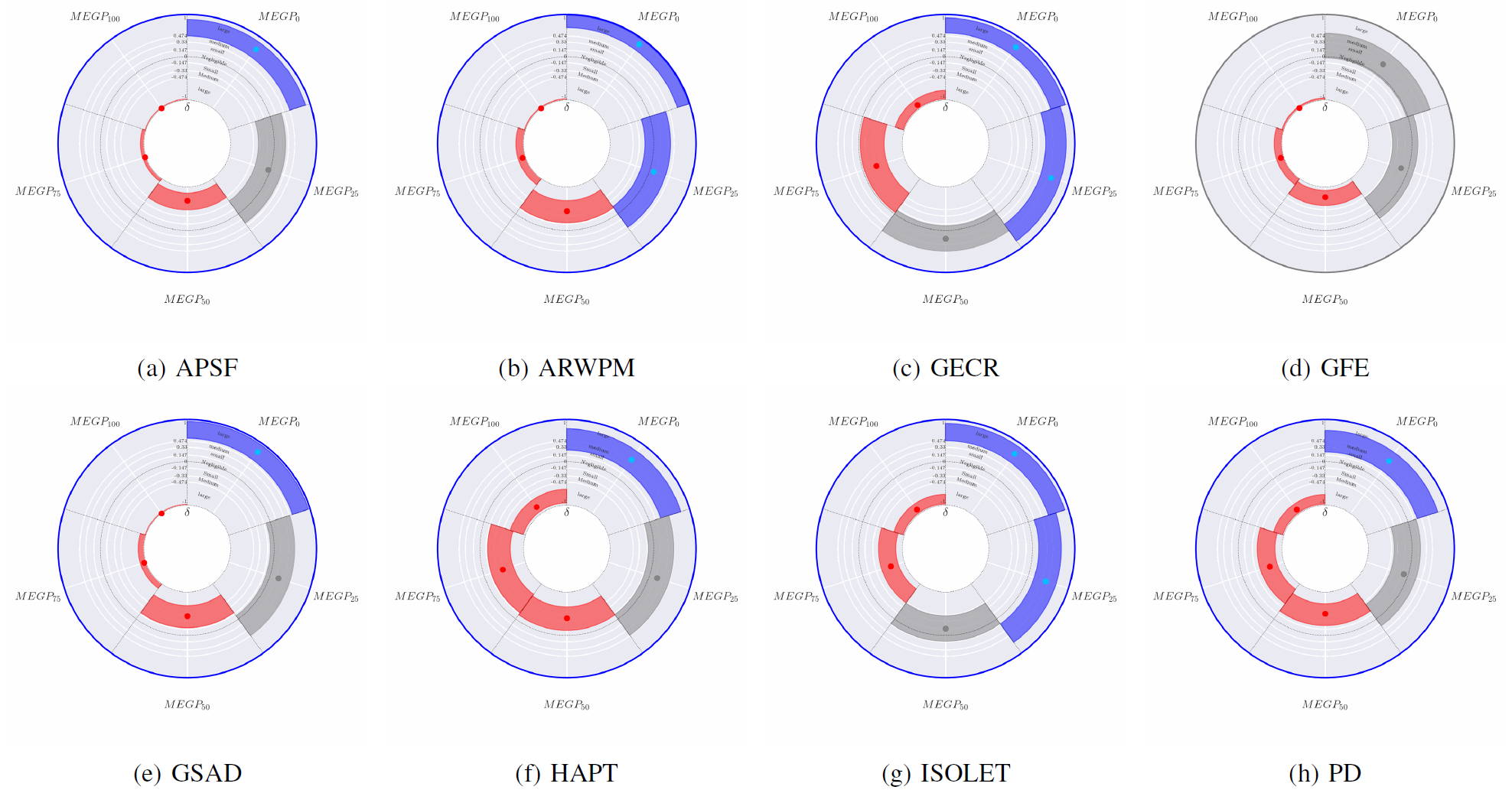}\label{fig:apsf_cliff_ccr50}%

\caption[The Cliff's $\delta$ effect size measure and its 95\% confidence intervals for CCR\(_{50}\) obtained from 30 BGP and MEGP runs.]{Effect size analysis of CCR\(_{50}\) across 30 runs for BGP and MEGP models using Cliff's $\delta$. Each point represents the actual CCR\(_{50}\) value obtained, with segments denoting 95\% confidence intervals based on 10,000 bootstrap resamplings. The outer ring color visualizes statistical significance: grey illustrates no significant difference (adjusted Friedman's P-value $>0.05$), while color indicates significant differences; blue indicates that at least one MEGP configuration outperforms BGP (adjusted Conover's p-value $< 0.05$, Cliff's $\delta > 0$), and red signifies that all MEGP configurations underperform relative to BGP (adjusted Conover's p-value $< 0.05$, Cliff's $\delta < 0$). Segment colors show performance differences against BGP: grey for no significant difference (adjusted Conover's p-value $> 0.05$), blue for better performance (Cliff's $\delta > 0$), and red for worse performance (Cliff's $\delta < 0$).}

\label{fig:cliff_ccr50}
\end{figure*}

%% file: wtl_ccr50.tex
\begin{table*}
\centering
\caption[The results of Friedman and Conover tests and Cliff's $\delta$ analysis for the $CCR_{50}$ obtained from MEGP and BGP runs.]{Statistical comparison of $CCR_{50}$ results for training data obtained from MEGP and BGP Runs. W, T, and L denote win, tie, and loss based on adjusted Friedman and Conover's p-values. Effect sizes are calculated using Cliff's Delta method and are categorized as negligible, small, medium, or large.}
\label{tab:wtl_ccr50}
\begin{tabular}{cccccc}
\hline
\multicolumn{6}{c}{$CCR_{50}$} \\
\hline
Dataset & $MEGP_{0}$ & $MEGP_{25}$ & $MEGP_{50}$ & $MEGP_{75}$ & $MEGP_{100}$ \\
\hline
APSF & W (large) & T (negligible) & L (large) & L (large) & L (large) \\
ARWPM & W (large) & W (negligible) & L (medium) & L (large) & L (large) \\
GECR & W (large) & W (large) & T (small) & L (small) & L (large) \\
GFE & T (small) & T (small) & L (large) & L (large) & L (large) \\
GSAD & W (large) & T (small) & L (medium) & L (large) & L (large) \\
HAPT & W (large) & T (small) & L (medium) & L (medium) & L (large) \\
ISOLET & W (large) & W (medium) & T (small) & L (large) & L (large) \\
PD & W (large) & T (negligible) & L (large) & L (large) & L (large) \\
\hline
W - T - L & 7 - 1 - 0 & 3 - 5 - 0 & 0 - 2 - 6 & 0 - 0 - 8 & 0 - 0 - 8 \\
\hline
\end{tabular}
\end{table*}

%% file: box_ft100.tex
\begin{figure*}[ht] 
\centering
\includegraphics[width=\textwidth]{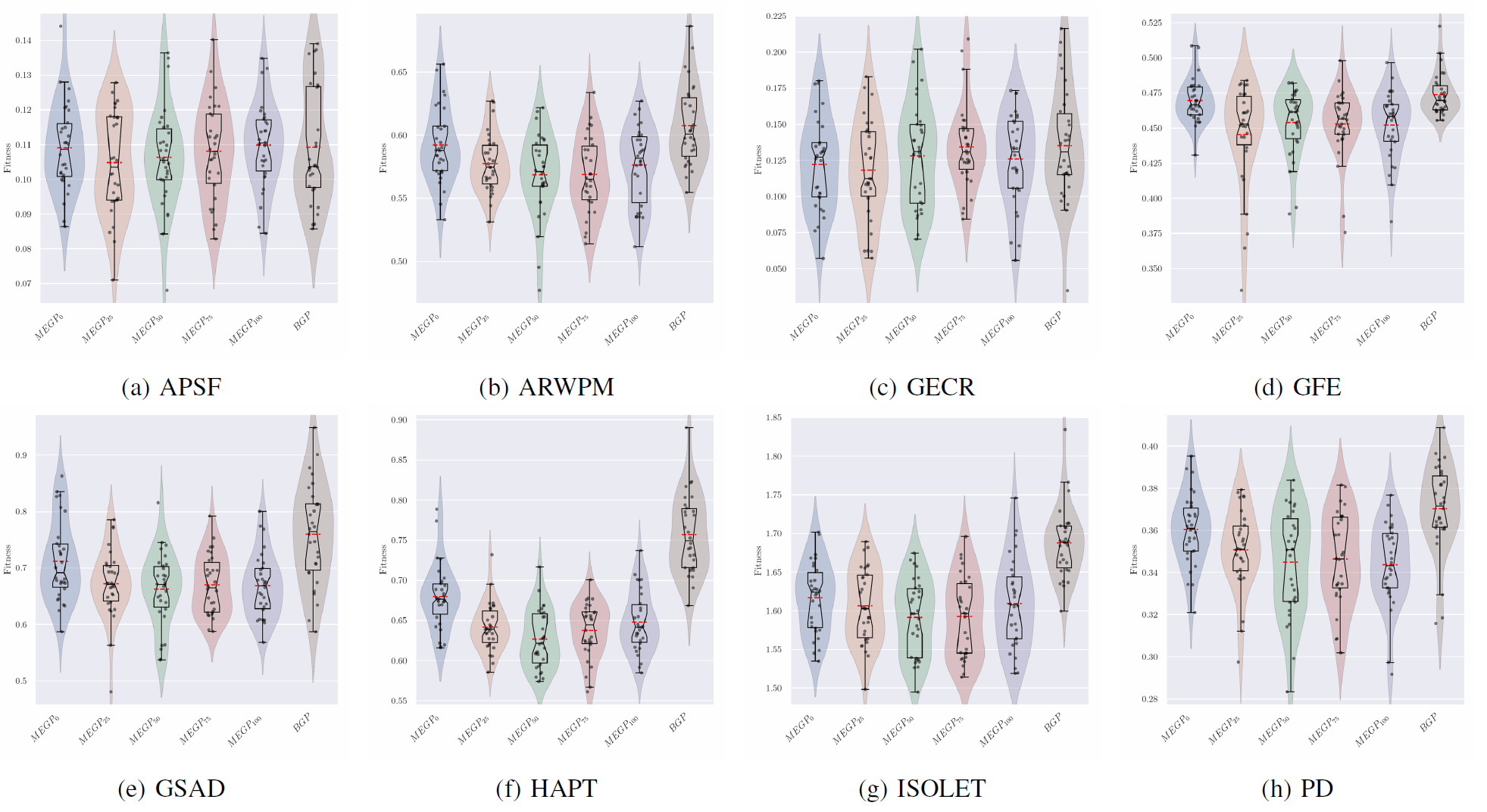}\label{fig:apsf_box_ft100}%

\caption[The distribution of FT\(_{100}\) across BGP and MEGP models over 30 runs.]{Raincloud plots showing the distribution of Fitness (FT\(_{100}\)) across 30 runs for BGP and MEGP models with different ensemble selection probabilities (0\%, 25\%, 50\%, 75\%, 100\%). Each plot illustrates the variability and central tendency of model performance over generations 51 to 100.}

\label{fig:box_ft100}
\end{figure*}

%% file: con_ft100.tex
\begin{figure*}[ht] 
\centering
\includegraphics[width=\textwidth]{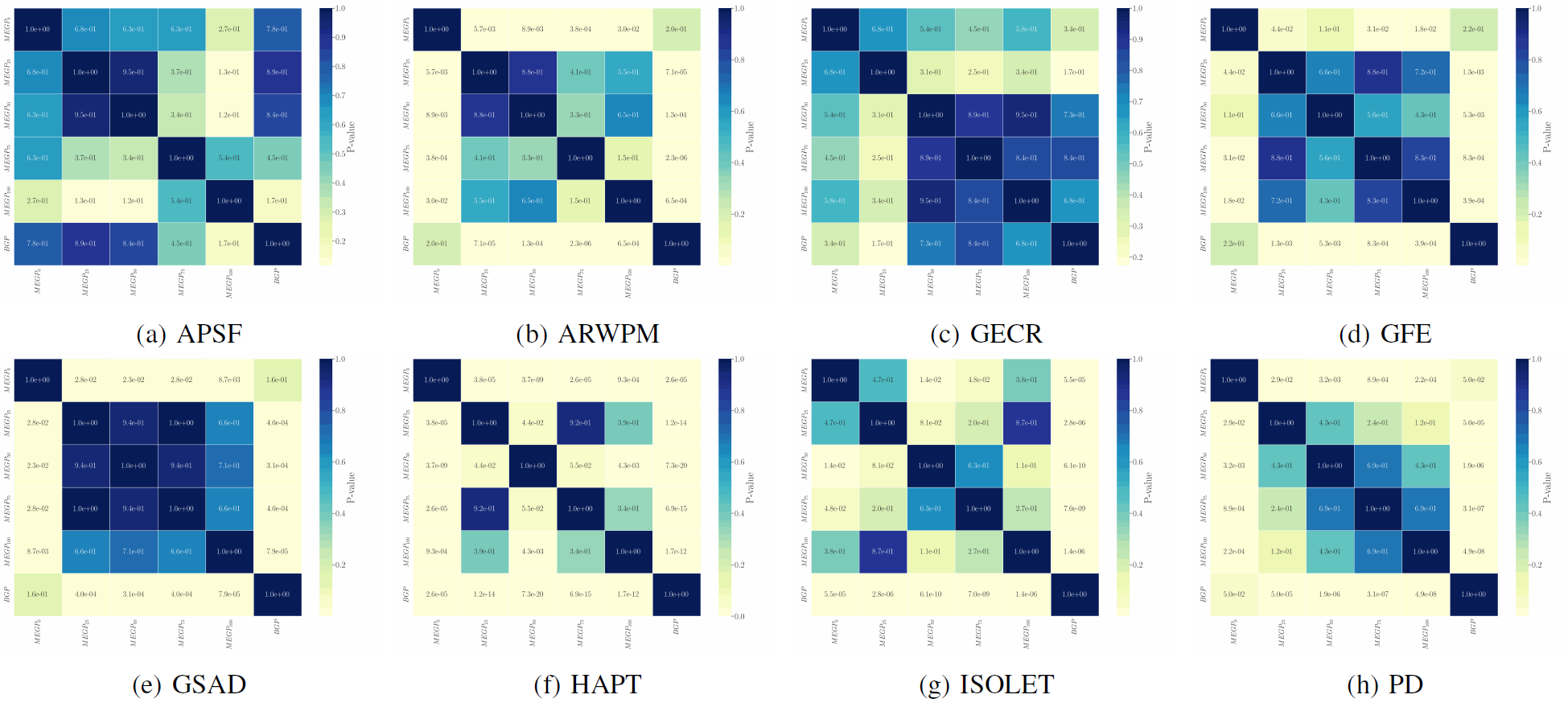}\label{fig:apsf_con_ft100}%

\caption[Heatmap of adjusted pairwise significance from Conover post-hoc test for FT\(_{100}\).]{Heatmap of adjusted p-values from the pairwise Conover post-hoc test for FT\(_{100}\), corrected using the Benjamini-Hochberg method. The heatmap highlights the statistical significance of pairwise comparisons between BGP and MEGP models with varying ensemble selection probabilities (0\%, 25\%, 50\%, 75\%, 100\%) over generations 51 to 100.}

\label{fig:con_ft100}
\end{figure*}

%% file: cliff_ft100.tex
\begin{figure*}[htbp] 
\centering
\includegraphics[width=\textwidth]{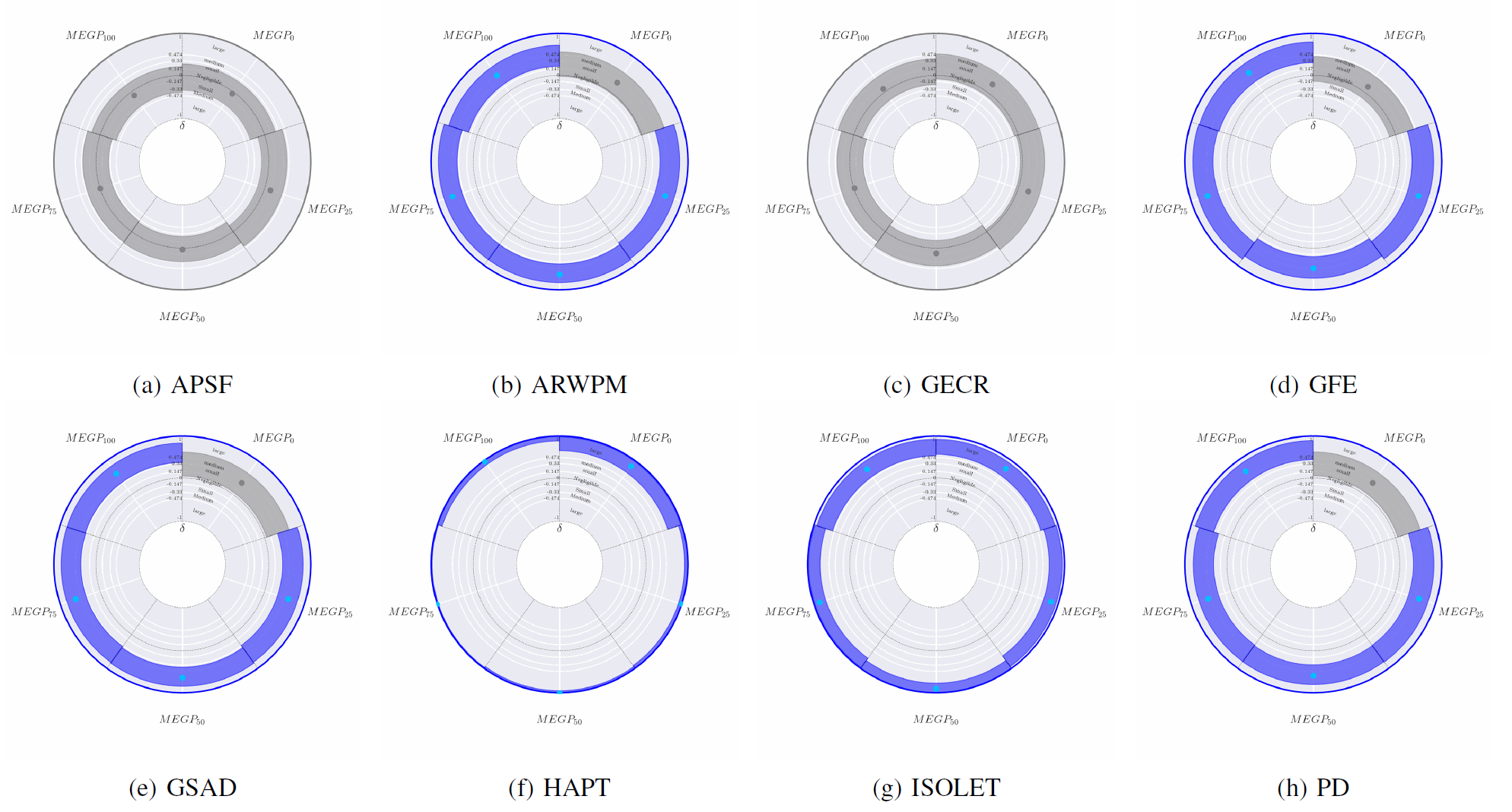}\label{fig:apsf_cliff_ft100}%

\caption[The Cliff's $\delta$ effect size measure and its 95\% confidence intervals for FT\(_{100}\) obtained from 30 BGP and MEGP runs.]{Effect size analysis of FT\(_{100}\) across 30 runs for BGP and MEGP models using Cliff's $\delta$. Each point represents the actual FT\(_{100}\) value obtained, with segments denoting 95\% confidence intervals based on 10,000 bootstrap resamplings. The outer ring color visualizes statistical significance: grey illustrates no significant difference (adjusted Friedman's P-value $>0.05$), while color indicates significant differences; blue indicates that at least one MEGP configuration outperforms BGP (adjusted Conover's p-value $< 0.05$, Cliff's $\delta > 0$), and red signifies that all MEGP configurations underperform relative to BGP (adjusted Conover's p-value $< 0.05$, Cliff's $\delta < 0$). Segment colors show performance differences against BGP: grey for no significant difference (adjusted Conover's p-value $> 0.05$), blue for better performance (Cliff's $\delta > 0$), and red for worse performance (Cliff's $\delta < 0$).}

\label{fig:cliff_ft100}
\end{figure*}

%% file: wtl_ft100.tex
\begin{table*}
\centering
\caption[The results of Friedman and Conover tests and Cliff's $\delta$ analysis for the $FT_{100}$ obtained from MEGP and BGP runs.]{Statistical comparison of $FT_{100}$ results for training data obtained from MEGP and BGP Runs. W, T, and L denote win, tie, and loss based on adjusted Friedman and Conover's p-values. Effect sizes are calculated using Cliff's Delta method and are categorized as negligible, small, medium, or large.}
\label{tab:wtl_ft100}
\begin{tabular}{cccccc}
\hline
\multicolumn{6}{c}{$FT_{100}$} \\
\hline
Dataset & $MEGP_{0}$ & $MEGP_{25}$ & $MEGP_{50}$ & $MEGP_{75}$ & $MEGP_{100}$ \\
\hline
APSF & T (negligible) & T (negligible) & T (negligible) & T (negligible) & T (negligible) \\
ARWPM & T (small) & W (large) & W (large) & W (large) & W (medium) \\
GECR & T (small) & T (small) & T (negligible) & T (negligible) & T (negligible) \\
GFE & T (small) & W (large) & W (medium) & W (large) & W (large) \\
GSAD & T (medium) & W (large) & W (large) & W (large) & W (large) \\
HAPT & W (large) & W (large) & W (large) & W (large) & W (large) \\
ISOLET & W (large) & W (large) & W (large) & W (large) & W (large) \\
PD & T (medium) & W (large) & W (large) & W (large) & W (large) \\
\hline
W - T - L & 2 - 6 - 0 & 6 - 2 - 0 & 6 - 2 - 0 & 6 - 2 - 0 & 6 - 2 - 0 \\
\hline
\end{tabular}
\end{table*}

%% file: box_cr100.tex
\begin{figure*}[ht] 
\centering
\subfloat[APSF]{\includegraphics[width=\textwidth]{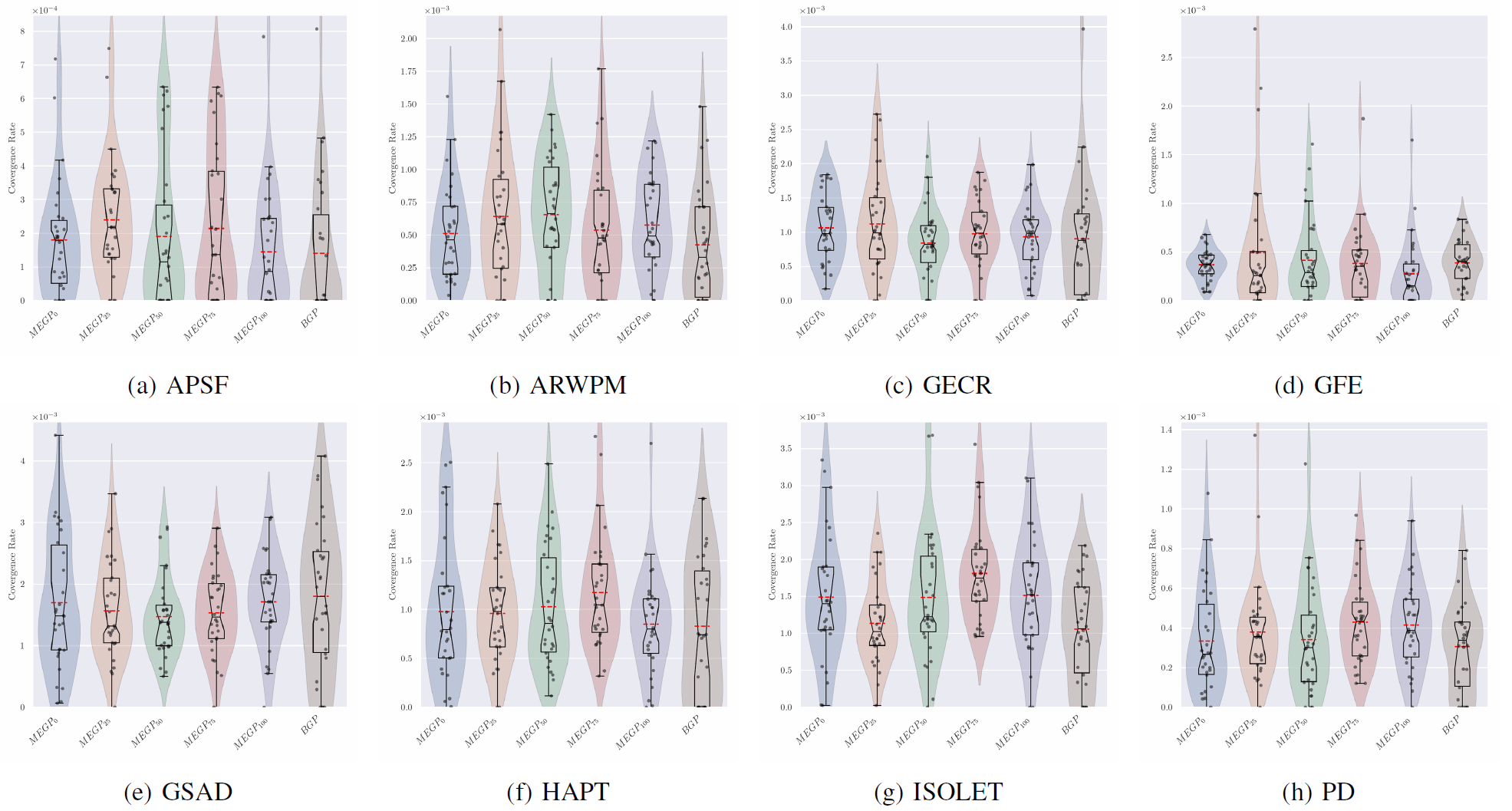}\label{fig:apsf_box_cr100}}%

\caption[The distribution of CR\(_{100}\) across BGP and MEGP models over 30 runs.]{Raincloud plots showing the distribution of Convergence Rate (CR\(_{100}\)) across 30 runs for BGP and MEGP models with different ensemble selection probabilities (0\%, 25\%, 50\%, 75\%, 100\%). Each plot illustrates the variability and central tendency of model performance over generations 51 to 100.}

\label{fig:box_cr100}
\end{figure*}

%% file: con_cr100.tex
\begin{figure*}[ht] 
\centering
\includegraphics[width=\textwidth]{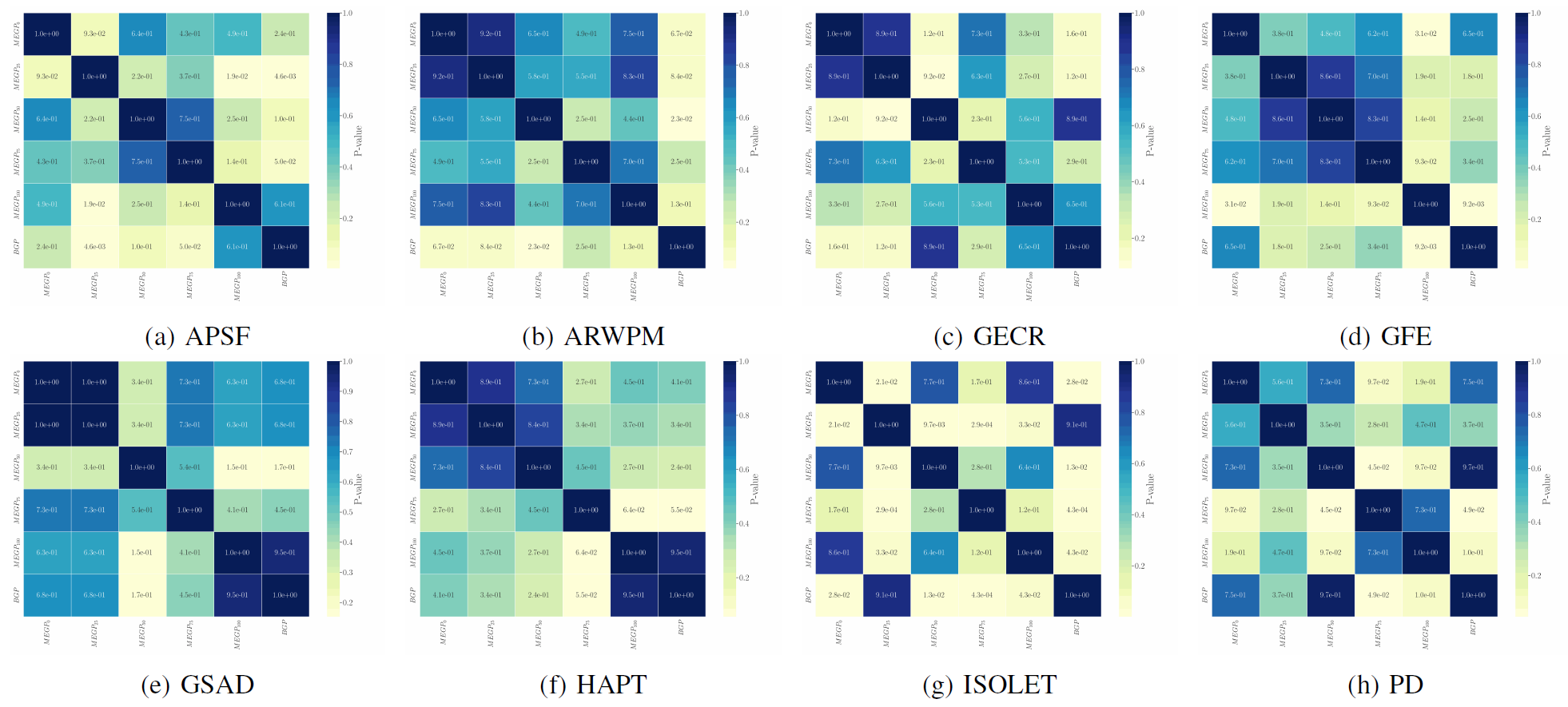}\label{fig:apsf_con_cr100}%

\caption[Heatmap of adjusted pairwise significance from Conover post-hoc test for CR\(_{100}\).]{Heatmap of adjusted p-values from the pairwise Conover post-hoc test for CR\(_{100}\), corrected using the Benjamini-Hochberg method. The heatmap highlights the statistical significance of pairwise comparisons between BGP and MEGP models with varying ensemble selection probabilities (0\%, 25\%, 50\%, 75\%, 100\%) over generations 51 to 100.}

\label{fig:con_cr100}
\end{figure*}

%% file: cliff_cr100.tex
\begin{figure*}[htbp] 
\centering
\includegraphics[width=\textwidth]{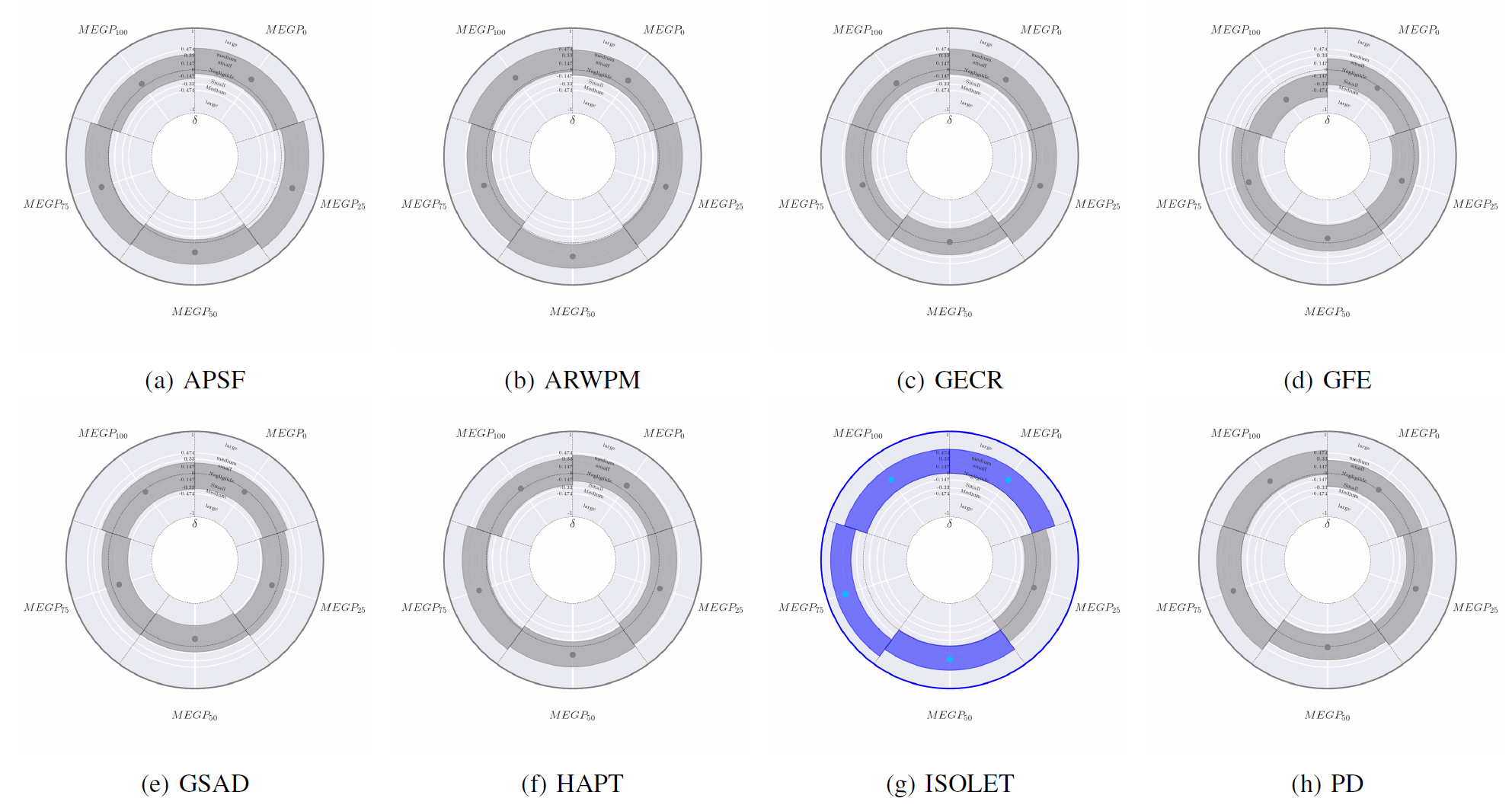}\label{fig:apsf_cliff_cr100}%

\caption[The Cliff's $\delta$ effect size measure and its 95\% confidence intervals for CR\(_{100}\) obtained from 30 BGP and MEGP runs.]{Effect size analysis of CR\(_{100}\) across 30 runs for BGP and MEGP models using Cliff's $\delta$. Each point represents the actual CR\(_{100}\) value obtained, with segments denoting 95\% confidence intervals based on 10,000 bootstrap resamplings. The outer ring color visualizes statistical significance: grey illustrates no significant difference (adjusted Friedman's P-value $>0.05$), while color indicates significant differences; blue indicates that at least one MEGP configuration outperforms BGP (adjusted Conover's p-value $< 0.05$, Cliff's $\delta > 0$), and red signifies that all MEGP configurations underperform relative to BGP (adjusted Conover's p-value $< 0.05$, Cliff's $\delta < 0$). Segment colors show performance differences against BGP: grey for no significant difference (adjusted Conover's p-value $> 0.05$), blue for better performance (Cliff's $\delta > 0$), and red for worse performance (Cliff's $\delta < 0$).}

\label{fig:cliff_cr100}
\end{figure*}

%% file: wtl_cr100.tex
\begin{table*}
\centering
\caption[The results of Friedman and Conover tests and Cliff's $\delta$ analysis for the $CR_{100}$ obtained from MEGP and BGP runs.]{Statistical comparison of $CR_{100}$ results for training data obtained from MEGP and BGP Runs. W, T, and L denote win, tie, and loss based on adjusted Friedman and Conover's p-values. Effect sizes are calculated using Cliff's Delta method and are categorized as negligible, small, medium, or large.}
\label{tab:wtl_cr100}
\begin{tabular}{cccccc}
\hline
\multicolumn{6}{c}{$CR_{100}$} \\
\hline
Dataset & $MEGP_{0}$ & $MEGP_{25}$ & $MEGP_{50}$ & $MEGP_{75}$ & $MEGP_{100}$ \\
\hline
APSF & T (small) & T (medium) & T (small) & T (small) & T (negligible) \\
ARWPM & T (small) & T (small) & T (small) & T (small) & T (small) \\
GECR & T (small) & T (small) & T (negligible) & T (negligible) & T (negligible) \\
GFE & T (negligible) & T (small) & T (negligible) & T (negligible) & T (medium) \\
GSAD & T (negligible) & T (negligible) & T (small) & T (small) & T (negligible) \\
HAPT & T (negligible) & T (negligible) & T (small) & T (small) & T (negligible) \\
ISOLET & W (small) & T (negligible) & W (small) & W (large) & W (small) \\
PD & T (negligible) & T (negligible) & T (negligible) & T (small) & T (small) \\
\hline
W - T - L & 1 - 7 - 0 & 0 - 8 - 0 & 1 - 7 - 0 & 1 - 7 - 0 & 1 - 7 - 0 \\
\hline
\end{tabular}
\end{table*}

%% file: box_ccr100.tex
\begin{figure*}[ht] 
\centering
\includegraphics[width=\textwidth]{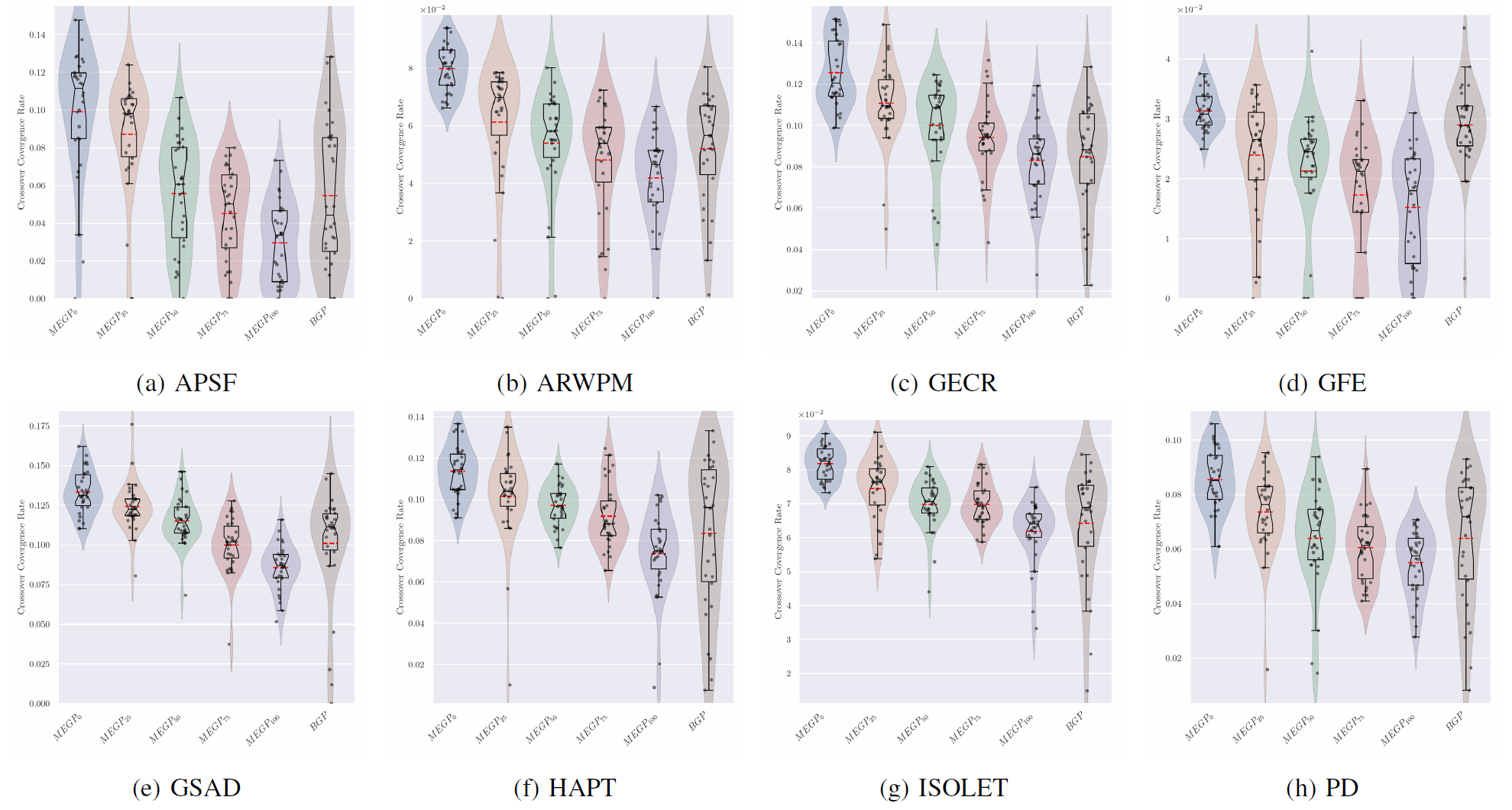}\label{fig:apsf_box_ccr100}%

\caption[The distribution of CCR\(_{100}\) across BGP and MEGP models over 30 runs.]{Raincloud plots showing the distribution of Crossover Convergence Rate (CCR\(_{100}\)) across 30 runs for BGP and MEGP models with different ensemble selection probabilities (0\%, 25\%, 50\%, 75\%, 100\%). Each plot illustrates the variability and central tendency of model performance over generations 51 to 100.}

\label{fig:box_ccr100}
\end{figure*}

%% file: con_ccr100.tex
\begin{figure*}[ht] 
\centering
\includegraphics[width=\textwidth]{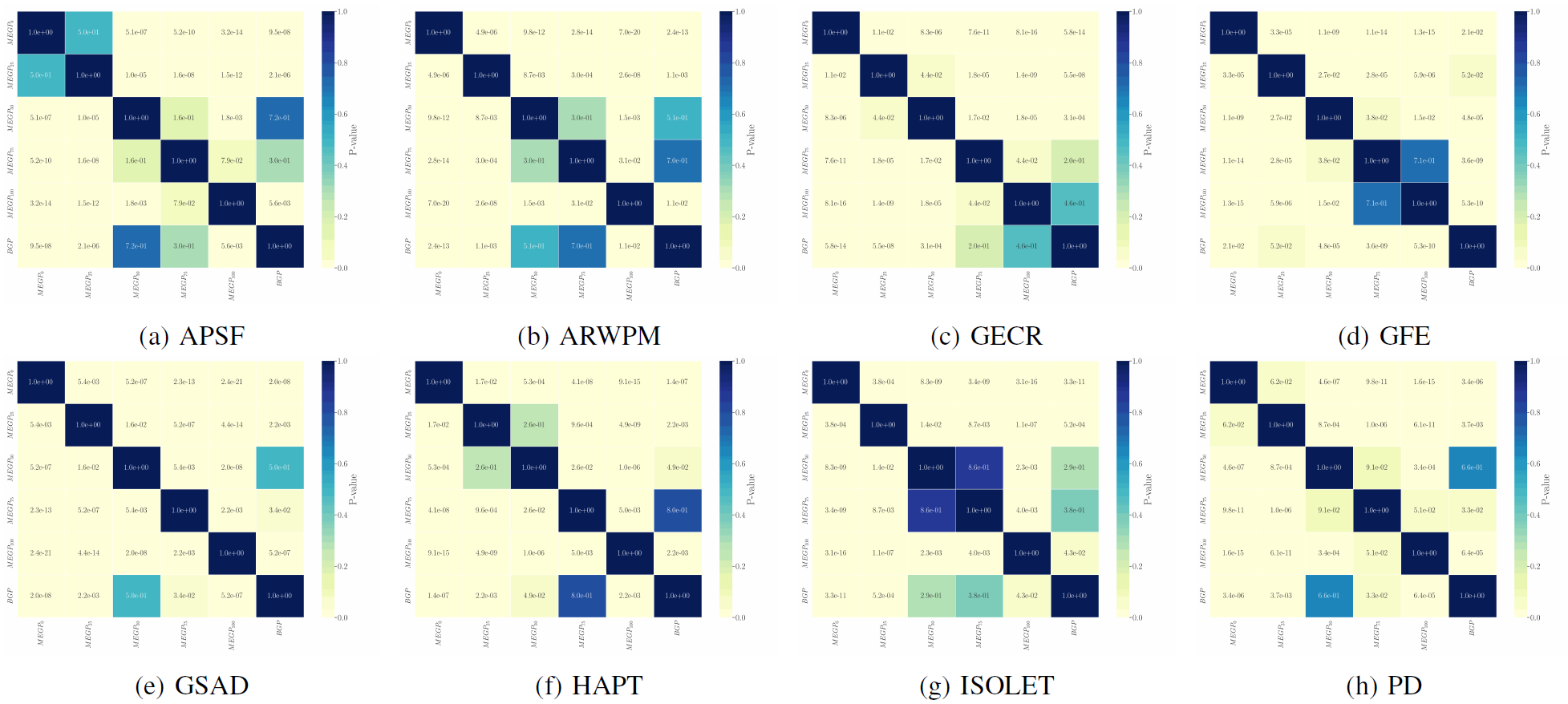}\label{fig:apsf_con_ccr100}%

\caption[Heatmap of adjusted pairwise significance from Conover post-hoc test for CCR\(_{100}\).]{Heatmap of adjusted p-values from the pairwise Conover post-hoc test for CCR\(_{100}\), corrected using the Benjamini-Hochberg method. The heatmap highlights the statistical significance of pairwise comparisons between BGP and MEGP models with varying ensemble selection probabilities (0\%, 25\%, 50\%, 75\%, 100\%) over generations 51 to 100.}

\label{fig:con_ccr100}
\end{figure*}

%% file: cliff_ccr100.tex
\begin{figure*}[htbp] 
\centering
\includegraphics[width=\textwidth]{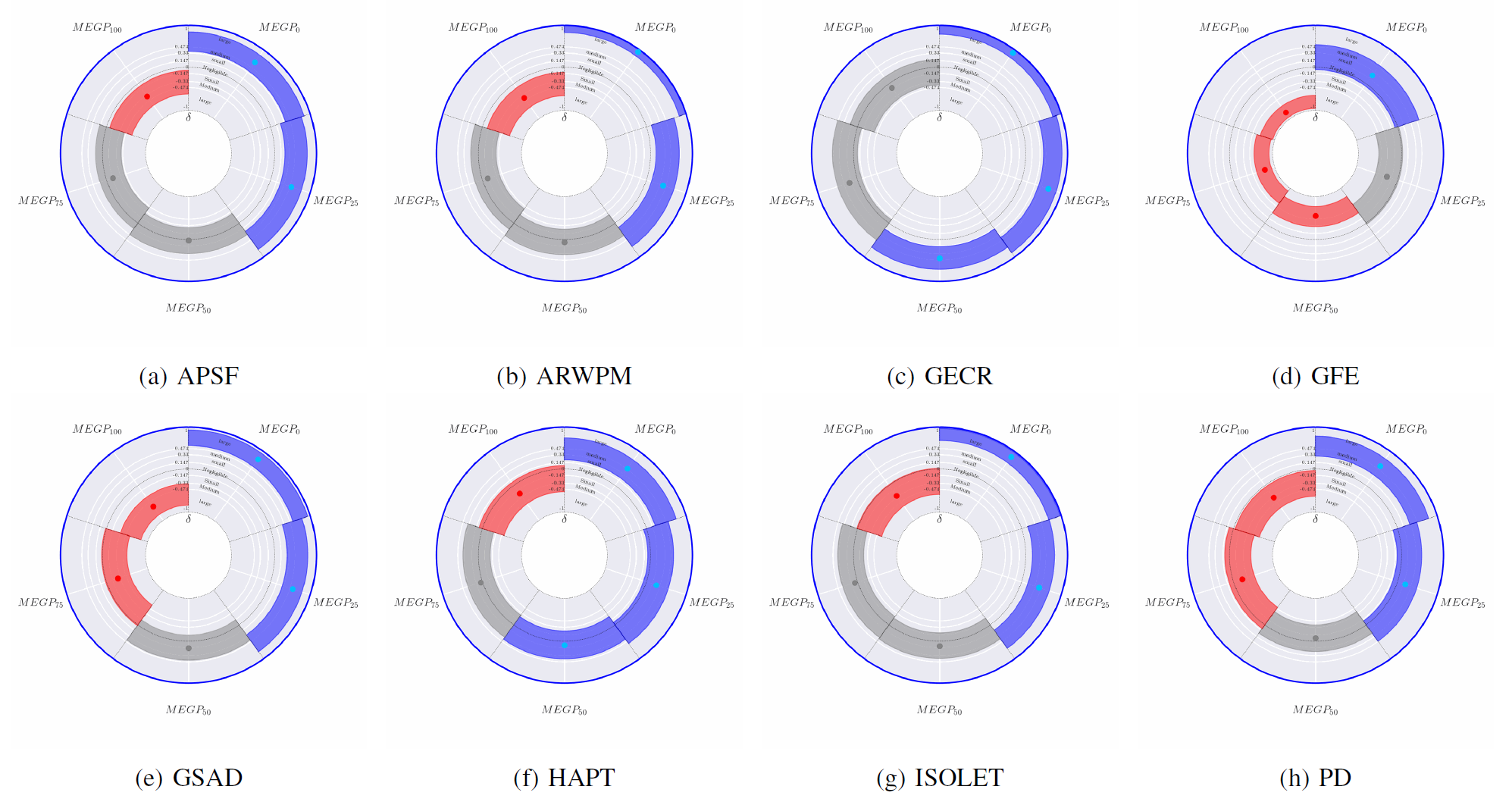}\label{fig:apsf_cliff_ccr100}%

\caption[The Cliff's $\delta$ effect size measure and its 95\% confidence intervals for CCR\(_{100}\) obtained from 30 BGP and MEGP runs.]{Effect size analysis of CCR\(_{100}\) across 30 runs for BGP and MEGP models using Cliff's $\delta$. Each point represents the actual CCR\(_{100}\) value obtained, with segments denoting 95\% confidence intervals based on 10,000 bootstrap resamplings. The outer ring color visualizes statistical significance: grey illustrates no significant difference (adjusted Friedman's P-value $>0.05$), while color indicates significant differences; blue indicates that at least one MEGP configuration outperforms BGP (adjusted Conover's p-value $< 0.05$, Cliff's $\delta > 0$), and red signifies that all MEGP configurations underperform relative to BGP (adjusted Conover's p-value $< 0.05$, Cliff's $\delta < 0$). Segment colors show performance differences against BGP: grey for no significant difference (adjusted Conover's p-value $> 0.05$), blue for better performance (Cliff's $\delta > 0$), and red for worse performance (Cliff's $\delta < 0$).}

\label{fig:cliff_ccr100}
\end{figure*}

%% file: wtl_ccr100.tex
\begin{table*}
\centering
\caption[The results of Friedman and Conover tests and Cliff's $\delta$ analysis for the $CCR_{100}$ obtained from MEGP and BGP runs.]{Statistical comparison of $CCR_{100}$ results for training data obtained from MEGP and BGP Runs. W, T, and L denote win, tie, and loss based on adjusted Friedman and Conover's p-values. Effect sizes are calculated using Cliff's Delta method and are categorized as negligible, small, medium, or large.}
\label{tab:wtl_ccr100}
\begin{tabular}{cccccc}
\hline
\multicolumn{6}{c}{$CCR_{100}$} \\
\hline
Dataset & $MEGP_{0}$ & $MEGP_{25}$ & $MEGP_{50}$ & $MEGP_{75}$ & $MEGP_{100}$ \\
\hline
APSF & W (large) & W (large) & T (negligible) & T (small) & L (medium) \\
ARWPM & W (large) & W (medium) & T (negligible) & T (negligible) & L (medium) \\
GECR & W (large) & W (large) & W (medium) & T (small) & T (negligible) \\
GFE & W (small) & T (small) & L (large) & L (large) & L (large) \\
GSAD & W (large) & W (large) & T (small) & L (small) & L (large) \\
HAPT & W (large) & W (small) & W (negligible) & T (negligible) & L (small) \\
ISOLET & W (large) & W (medium) & T (negligible) & T (negligible) & L (small) \\
PD & W (large) & W (small) & T (negligible) & L (small) & L (medium) \\
\hline
W - T - L & 8 - 0 - 0 & 7 - 1 - 0 & 2 - 5 - 1 & 0 - 5 - 3 & 0 - 1 - 7 \\
\hline
\end{tabular}
\end{table*}

%% file: box_ft150.tex
\begin{figure*}[ht] 
\centering
\includegraphics[width=\textwidth]{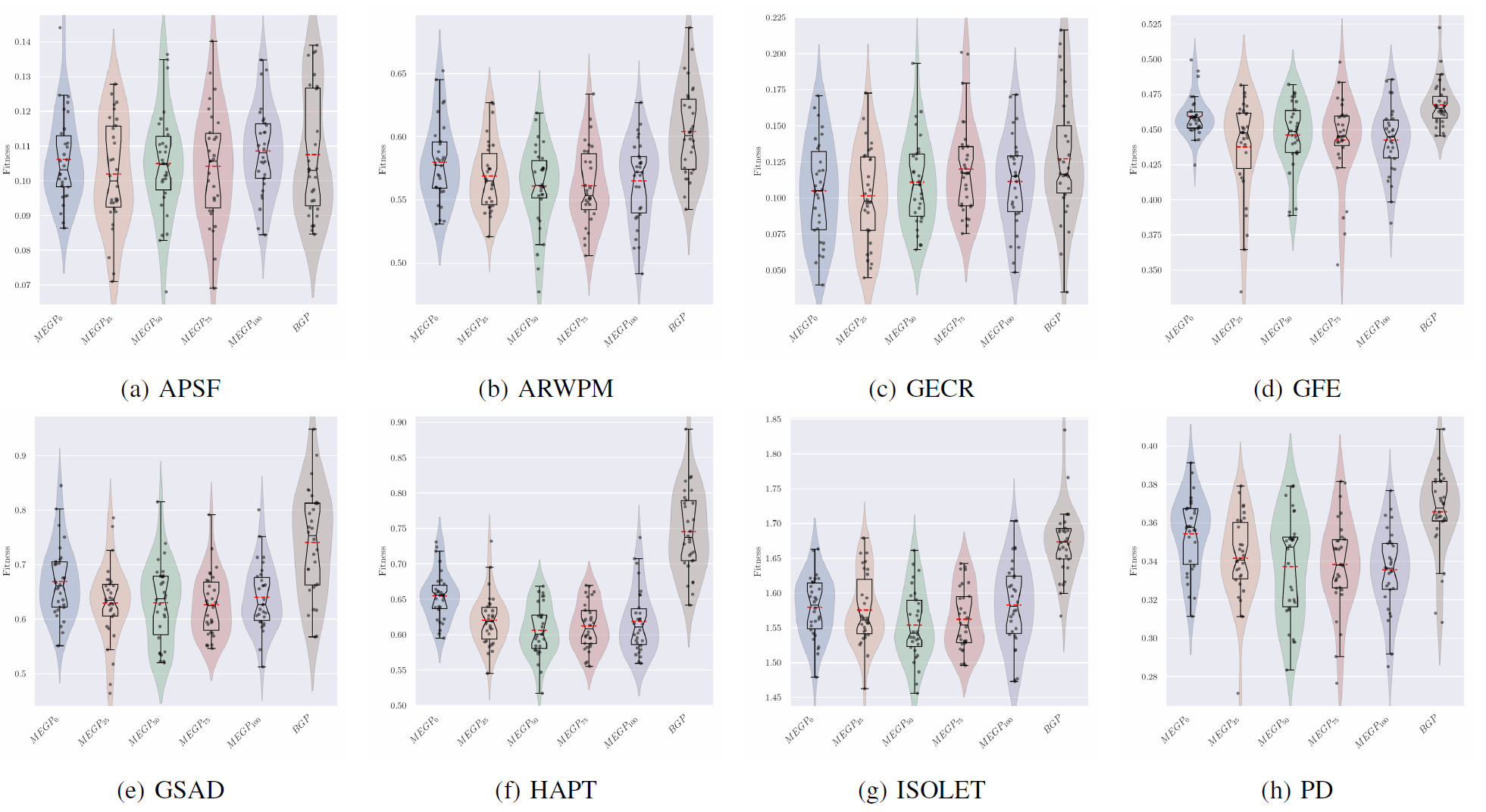}\label{fig:apsf_box_ft150}%

\caption[The distribution of FT\(_{150}\) across BGP and MEGP models over 30 runs.]{Raincloud plots showing the distribution of Fitness (FT\(_{150}\)) across 30 runs for BGP and MEGP models with different ensemble selection probabilities (0\%, 25\%, 50\%, 75\%, 100\%). Each plot illustrates the variability and central tendency of model performance over generations 101 to 150.}

\label{fig:box_ft150}
\end{figure*}

%% file: con_ft150.tex
\begin{figure*}[ht] 
\centering
\includegraphics[width=\textwidth]{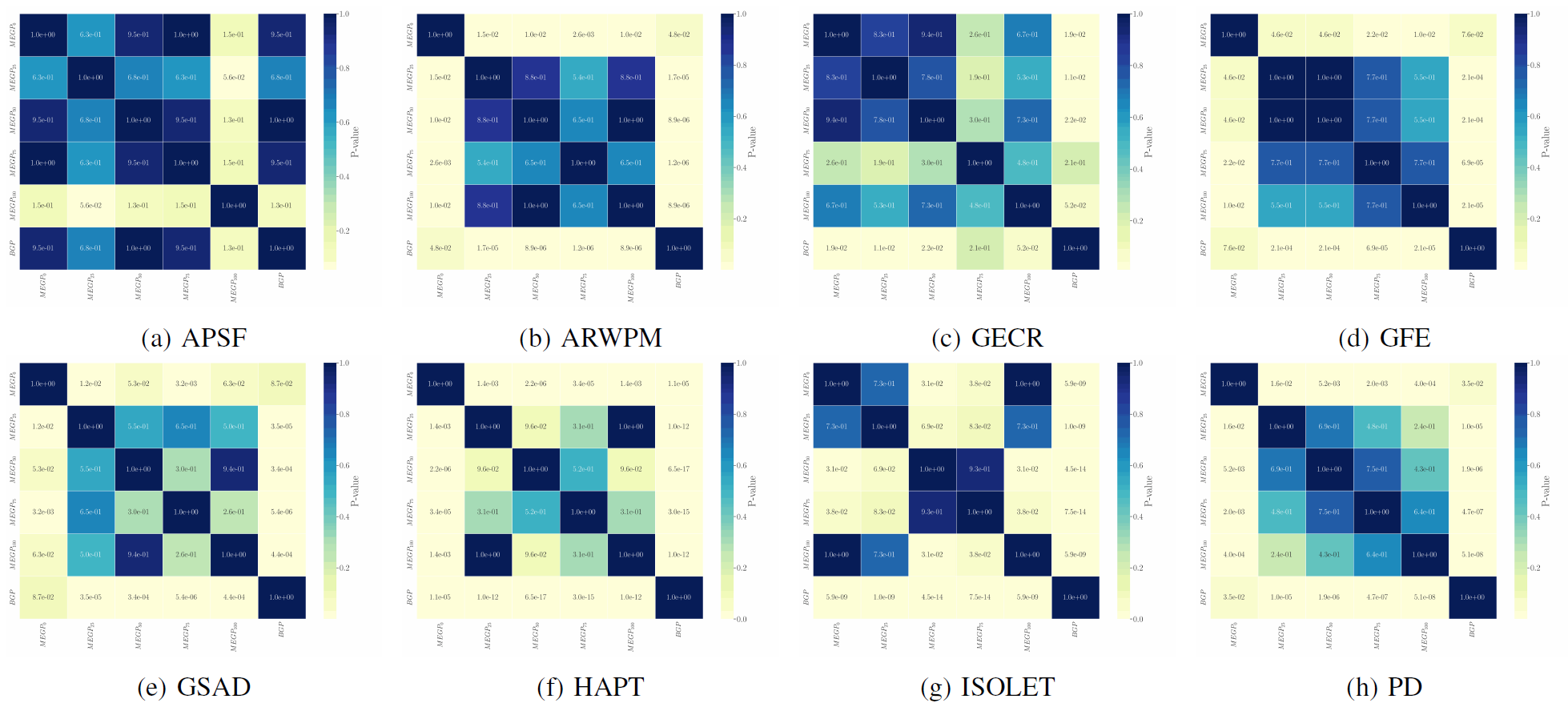}\label{fig:apsf_con_ft150}%

\caption[Heatmap of adjusted pairwise significance from Conover post-hoc test for FT\(_{150}\).]{Heatmap of adjusted p-values from the pairwise Conover post-hoc test for FT\(_{150}\), corrected using the Benjamini-Hochberg method. The heatmap highlights the statistical significance of pairwise comparisons between BGP and MEGP models with varying ensemble selection probabilities (0\%, 25\%, 50\%, 75\%, 100\%) over generations 101 to 150.}

\label{fig:con_ft150}
\end{figure*}

%% file: cliff_ft150.tex
\begin{figure*}[htbp] 
\centering
\includegraphics[width=\textwidth]{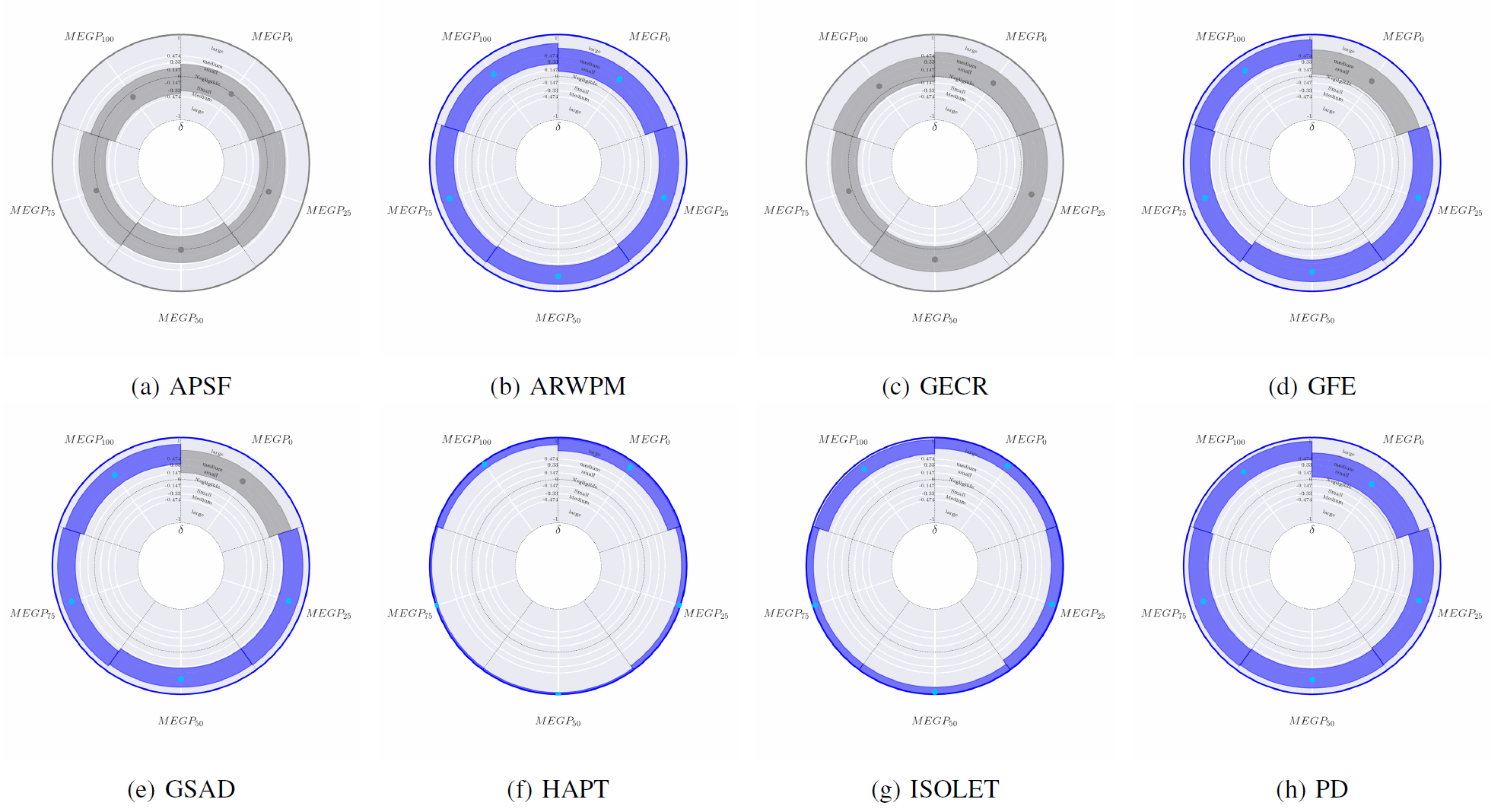}\label{fig:apsf_cliff_ft150}%

\caption[The Cliff's $\delta$ effect size measure and its 95\% confidence intervals for FT\(_{150}\) obtained from 30 BGP and MEGP runs.]{Effect size analysis of FT\(_{150}\) across 30 runs for BGP and MEGP models using Cliff's $\delta$. Each point represents the actual FT\(_{150}\) value obtained, with segments denoting 95\% confidence intervals based on 10,000 bootstrap resamplings. The outer ring color visualizes statistical significance: grey illustrates no significant difference (adjusted Friedman's P-value $>0.05$), while color indicates significant differences; blue indicates that at least one MEGP configuration outperforms BGP (adjusted Conover's p-value $< 0.05$, Cliff's $\delta > 0$), and red signifies that all MEGP configurations underperform relative to BGP (adjusted Conover's p-value $< 0.05$, Cliff's $\delta < 0$). Segment colors show performance differences against BGP: grey for no significant difference (adjusted Conover's p-value $> 0.05$), blue for better performance (Cliff's $\delta > 0$), and red for worse performance (Cliff's $\delta < 0$).}

\label{fig:cliff_ft150}
\end{figure*}

%% file: wtl_ft150.tex
\begin{table*}
\centering
\caption[The results of Friedman and Conover tests and Cliff's $\delta$ analysis for the $FT_{150}$ obtained from MEGP and BGP runs.]{Statistical comparison of $FT_{150}$ results for training data obtained from MEGP and BGP Runs. W, T, and L denote win, tie, and loss based on adjusted Friedman and Conover's p-values. Effect sizes are calculated using Cliff's Delta method and are categorized as negligible, small, medium, or large.}
\label{tab:wtl_ft150}
\begin{tabular}{cccccc}
\hline
\multicolumn{6}{c}{$FT_{150}$} \\
\hline
Dataset & $MEGP_{0}$ & $MEGP_{25}$ & $MEGP_{50}$ & $MEGP_{75}$ & $MEGP_{100}$ \\
\hline
APSF & T (negligible) & T (negligible) & T (negligible) & T (negligible) & T (negligible) \\
ARWPM & W (medium) & W (large) & W (large) & W (large) & W (large) \\
GECR & T (small) & T (medium) & T (small) & T (negligible) & T (small) \\
GFE & T (medium) & W (large) & W (large) & W (large) & W (large) \\
GSAD & T (medium) & W (large) & W (large) & W (large) & W (large) \\
HAPT & W (large) & W (large) & W (large) & W (large) & W (large) \\
ISOLET & W (large) & W (large) & W (large) & W (large) & W (large) \\
PD & W (medium) & W (large) & W (large) & W (large) & W (large) \\
\hline
W - T - L & 4 - 4 - 0 & 6 - 2 - 0 & 6 - 2 - 0 & 6 - 2 - 0 & 6 - 2 - 0 \\
\hline
\end{tabular}
\end{table*}

%% file: box_cr150.tex
\begin{figure*}[ht] 
\centering
\includegraphics[width=\textwidth]{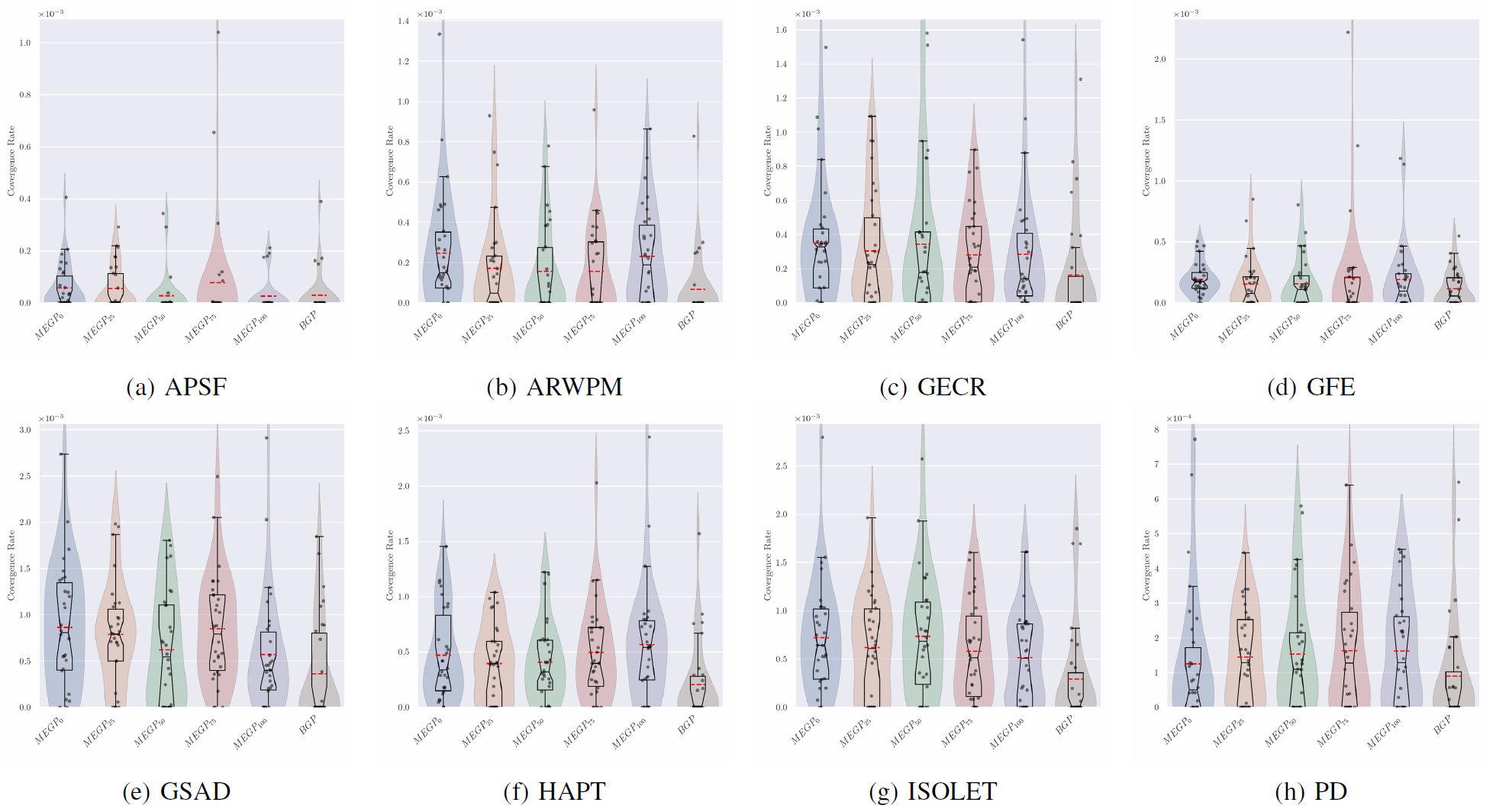}\label{fig:apsf_box_cr150}%

\caption[The distribution of CR\(_{150}\) across BGP and MEGP models over 30 runs.]{Raincloud plots showing the distribution of Convergence Rate (CR\(_{150}\)) across 30 runs for BGP and MEGP models with different ensemble selection probabilities (0\%, 25\%, 50\%, 75\%, 100\%). Each plot illustrates the variability and central tendency of model performance over generations 101 to 150.}

\label{fig:box_cr150}
\end{figure*}

%% file: con_cr150.tex
\begin{figure*}[ht] 
\centering
\includegraphics[width=\textwidth]{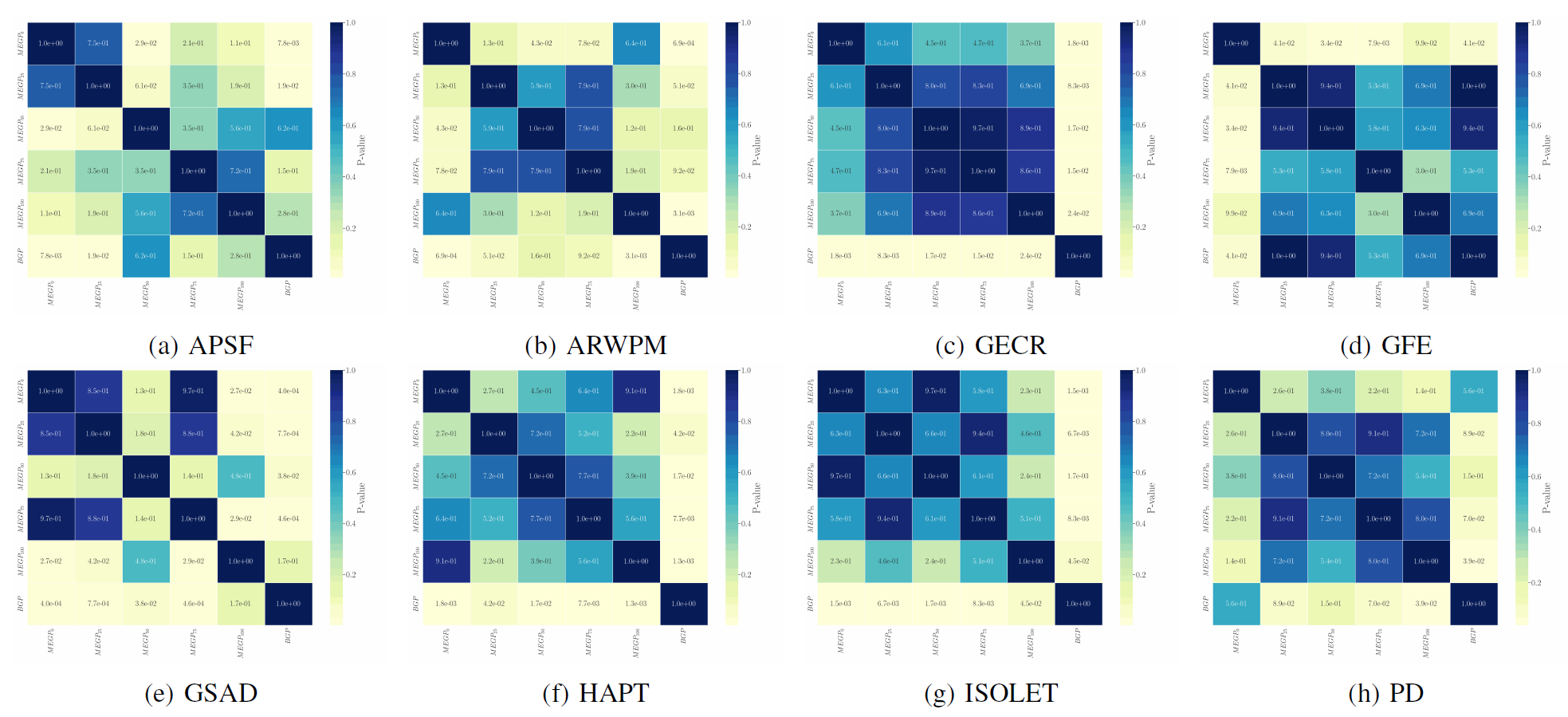}\label{fig:apsf_con_cr150}%
\caption[Heatmap of adjusted pairwise significance from Conover post-hoc test for CR\(_{150}\).]{Heatmap of adjusted p-values from the pairwise Conover post-hoc test for CR\(_{150}\), corrected using the Benjamini-Hochberg method. The heatmap highlights the statistical significance of pairwise comparisons between BGP and MEGP models with varying ensemble selection probabilities (0\%, 25\%, 50\%, 75\%, 100\%) over generations 101 to 150.}

\label{fig:con_cr150}
\end{figure*}

%% file: cliff_cr150.tex
\begin{figure*}[htbp] 
\centering
\includegraphics[width=\textwidth]{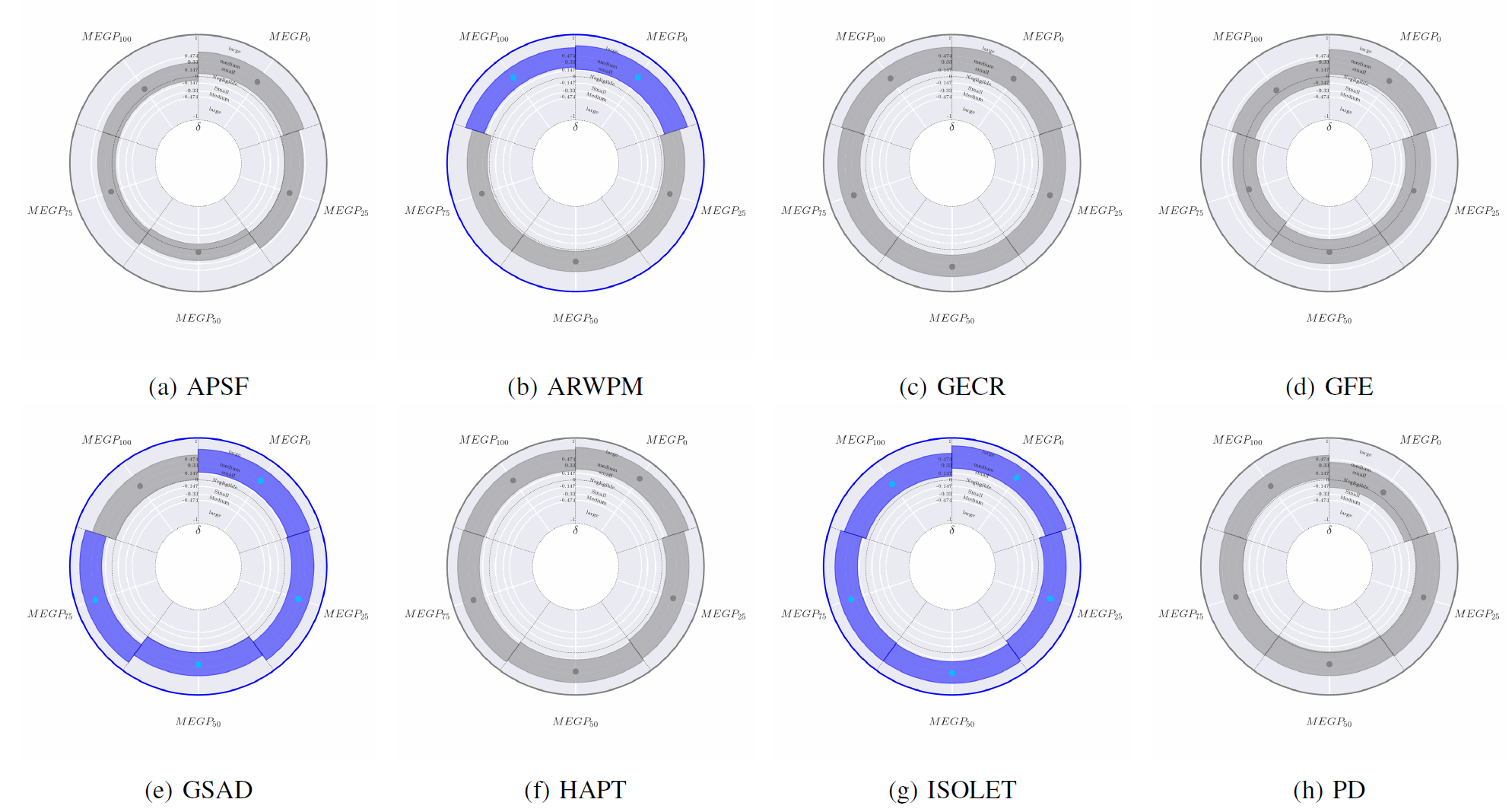}\label{fig:apsf_cliff_cr150}%

\caption[The Cliff's $\delta$ effect size measure and its 95\% confidence intervals for CR\(_{150}\) obtained from 30 BGP and MEGP runs.]{Effect size analysis of CR\(_{150}\) across 30 runs for BGP and MEGP models using Cliff's $\delta$. Each point represents the actual CR\(_{150}\) value obtained, with segments denoting 95\% confidence intervals based on 10,000 bootstrap resamplings. The outer ring color visualizes statistical significance: grey illustrates no significant difference (adjusted Friedman's P-value $>0.05$), while color indicates significant differences; blue indicates that at least one MEGP configuration outperforms BGP (adjusted Conover's p-value $< 0.05$, Cliff's $\delta > 0$), and red signifies that all MEGP configurations underperform relative to BGP (adjusted Conover's p-value $< 0.05$, Cliff's $\delta < 0$). Segment colors show performance differences against BGP: grey for no significant difference (adjusted Conover's p-value $> 0.05$), blue for better performance (Cliff's $\delta > 0$), and red for worse performance (Cliff's $\delta < 0$).}

\label{fig:cliff_cr150}
\end{figure*}

%% file: wtl_cr150.tex
\begin{table*}
\centering
\caption[The results of Friedman and Conover tests and Cliff's $\delta$ analysis for the $CR_{150}$ obtained from MEGP and BGP runs.]{Statistical comparison of $CR_{150}$ results for training data obtained from MEGP and BGP Runs. W, T, and L denote win, tie, and loss based on adjusted Friedman and Conover's p-values. Effect sizes are calculated using Cliff's Delta method and are categorized as negligible, small, medium, or large.}
\label{tab:wtl_cr150}
\begin{tabular}{cccccc}
\hline
\multicolumn{6}{c}{$CR_{150}$} \\
\hline
Dataset & $MEGP_{0}$ & $MEGP_{25}$ & $MEGP_{50}$ & $MEGP_{75}$ & $MEGP_{100}$ \\
\hline
APSF & T (small) & T (small) & T (negligible) & T (negligible) & T (negligible) \\
ARWPM & W (medium) & T (small) & T (small) & T (small) & W (medium) \\
GECR & T (medium) & T (medium) & T (medium) & T (medium) & T (medium) \\
GFE & T (medium) & T (negligible) & T (negligible) & T (negligible) & T (negligible) \\
GSAD & W (medium) & W (medium) & W (small) & W (large) & T (small) \\
HAPT & T (large) & T (medium) & T (medium) & T (large) & T (medium) \\
ISOLET & W (large) & W (medium) & W (medium) & W (medium) & W (medium) \\
PD & T (negligible) & T (small) & T (small) & T (small) & T (small) \\
\hline
W - T - L & 3 - 5 - 0 & 2 - 6 - 0 & 2 - 6 - 0 & 2 - 6 - 0 & 2 - 6 - 0 \\
\hline
\end{tabular}
\end{table*}

%% file: box_ccr150.tex
\begin{figure*}[ht] 
\centering
\includegraphics[width=\textwidth]{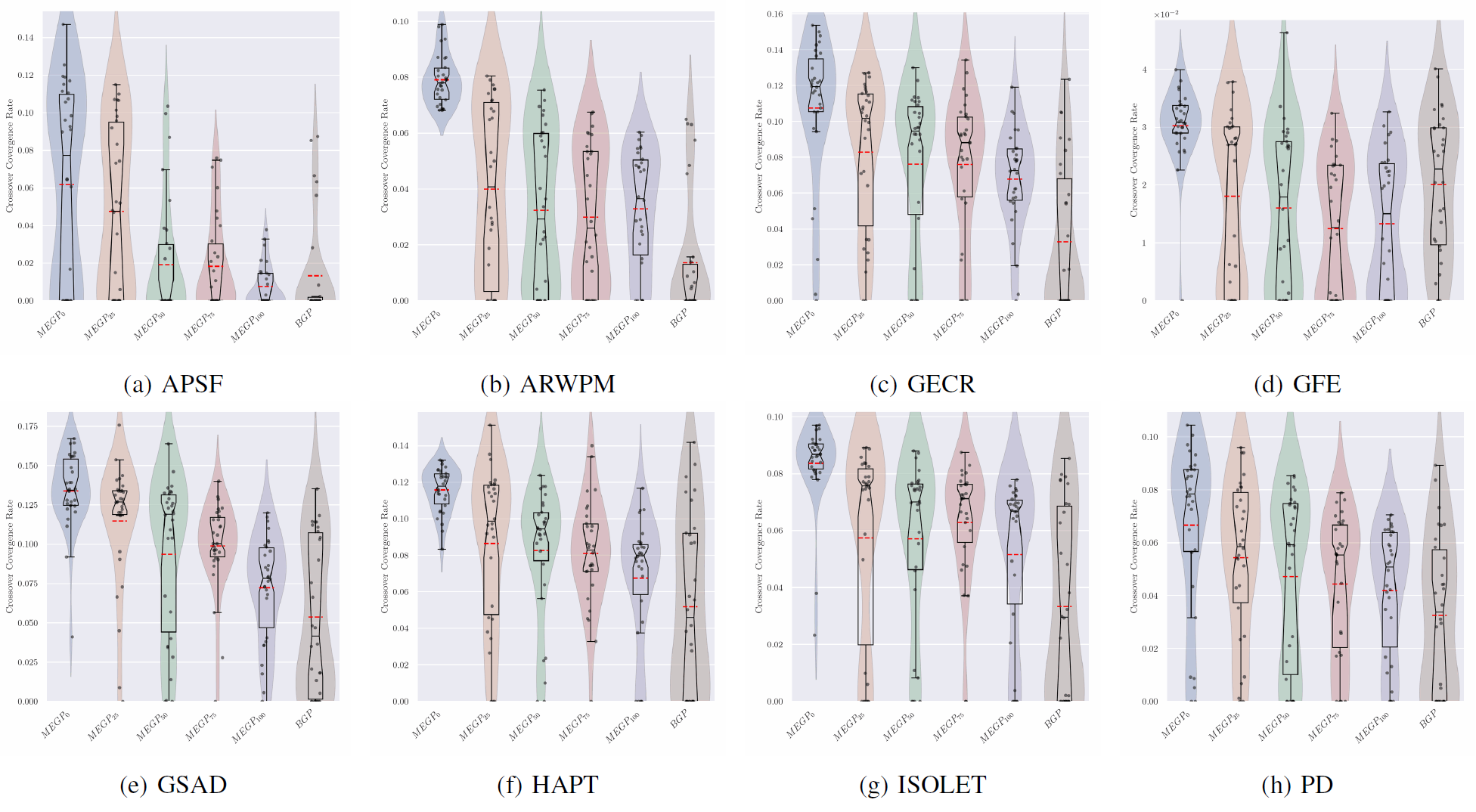}\label{fig:apsf_box_ccr150}%

\caption[The distribution of CCR\(_{150}\) across BGP and MEGP models over 30 runs.]{Raincloud plots showing the distribution of Crossover Convergence Rate (CCR\(_{150}\)) across 30 runs for BGP and MEGP models with different ensemble selection probabilities (0\%, 25\%, 50\%, 75\%, 100\%). Each plot illustrates the variability and central tendency of model performance over generations 101 to 150.}

\label{fig:box_ccr150}
\end{figure*}

%% file: con_ccr150.tex
\begin{figure*}[ht] 
\centering
\includegraphics[width=\textwidth]{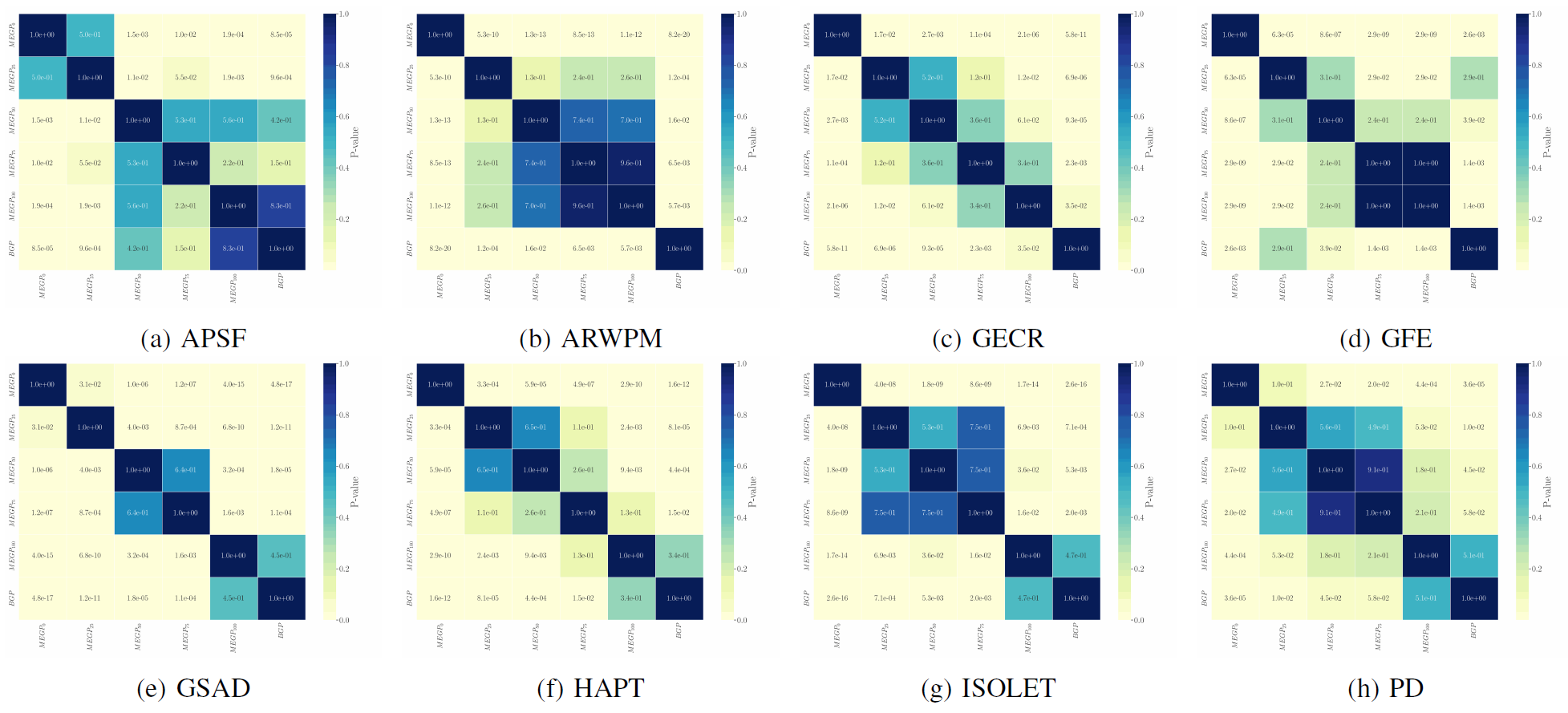}\label{fig:apsf_con_ccr150}%

\caption[Heatmap of adjusted pairwise significance from Conover post-hoc test for CCR\(_{150}\).]{Heatmap of adjusted p-values from the pairwise Conover post-hoc test for CCR\(_{150}\), corrected using the Benjamini-Hochberg method. The heatmap highlights the statistical significance of pairwise comparisons between BGP and MEGP models with varying ensemble selection probabilities (0\%, 25\%, 50\%, 75\%, 100\%) over generations 101 to 150.}

\label{fig:con_ccr150}
\end{figure*}

%% file: cliff_ccr150.tex
\begin{figure*}[htbp] 
\centering
\includegraphics[width=\textwidth]{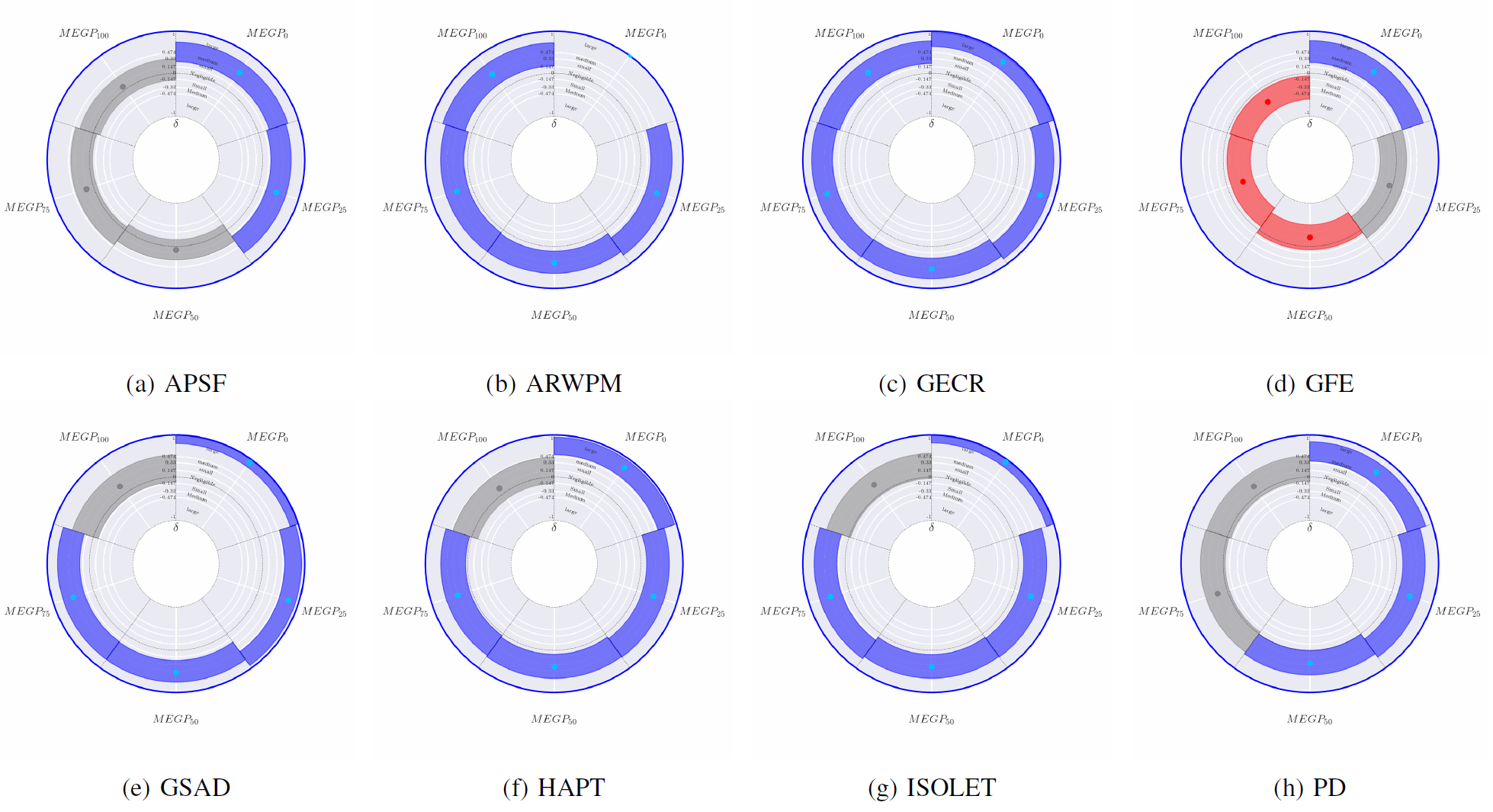}

\caption[The Cliff's $\delta$ effect size measure and its 95\% confidence intervals for CCR\(_{150}\) obtained from 30 BGP and MEGP runs.]{Effect size analysis of CCR\(_{150}\) across 30 runs for BGP and MEGP models using Cliff's $\delta$. Each point represents the actual CCR\(_{150}\) value obtained, with segments denoting 95\% confidence intervals based on 10,000 bootstrap resamplings. The outer ring color visualizes statistical significance: grey illustrates no significant difference (adjusted Friedman's P-value $>0.05$), while color indicates significant differences; blue indicates that at least one MEGP configuration outperforms BGP (adjusted Conover's p-value $< 0.05$, Cliff's $\delta > 0$), and red signifies that all MEGP configurations underperform relative to BGP (adjusted Conover's p-value $< 0.05$, Cliff's $\delta < 0$). Segment colors show performance differences against BGP: grey for no significant difference (adjusted Conover's p-value $> 0.05$), blue for better performance (Cliff's $\delta > 0$), and red for worse performance (Cliff's $\delta < 0$).}

\label{fig:cliff_ccr150}
\end{figure*}

%% file: wtl_ccr150.tex
\begin{table*}
\centering
\caption[The results of Friedman and Conover tests and Cliff's $\delta$ analysis for the $CCR_{150}$ obtained from MEGP and BGP runs.]{Statistical comparison of $CCR_{150}$ results for training data obtained from MEGP and BGP Runs. W, T, and L denote win, tie, and loss based on adjusted Friedman and Conover's p-values. Effect sizes are calculated using Cliff's Delta method and are categorized as negligible, small, medium, or large.}
\label{tab:wtl_ccr150}
\begin{tabular}{cccccc}
\hline
\multicolumn{6}{c}{$CCR_{150}$} \\
\hline
Dataset & $MEGP_{0}$ & $MEGP_{25}$ & $MEGP_{50}$ & $MEGP_{75}$ & $MEGP_{100}$ \\
\hline
APSF & W (large) & W (medium) & T (negligible) & T (small) & T (negligible) \\
ARWPM & W (large) & W (large) & W (medium) & W (medium) & W (medium) \\
GECR & W (large) & W (large) & W (large) & W (large) & W (large) \\
GFE & W (large) & T (negligible) & L (small) & L (medium) & L (medium) \\
GSAD & W (large) & W (large) & W (large) & W (large) & T (small) \\
HAPT & W (large) & W (medium) & W (medium) & W (medium) & T (small) \\
ISOLET & W (large) & W (medium) & W (medium) & W (medium) & T (small) \\
PD & W (large) & W (medium) & W (small) & T (small) & T (small) \\
\hline
W - T - L & 8 - 0 - 0 & 7 - 1 - 0 & 6 - 1 - 1 & 5 - 2 - 1 & 2 - 5 - 1 \\
\hline
\end{tabular}
\end{table*}

%% file: box_ftall.tex
\begin{figure*}[ht] 
\centering
\includegraphics[width=\textwidth]{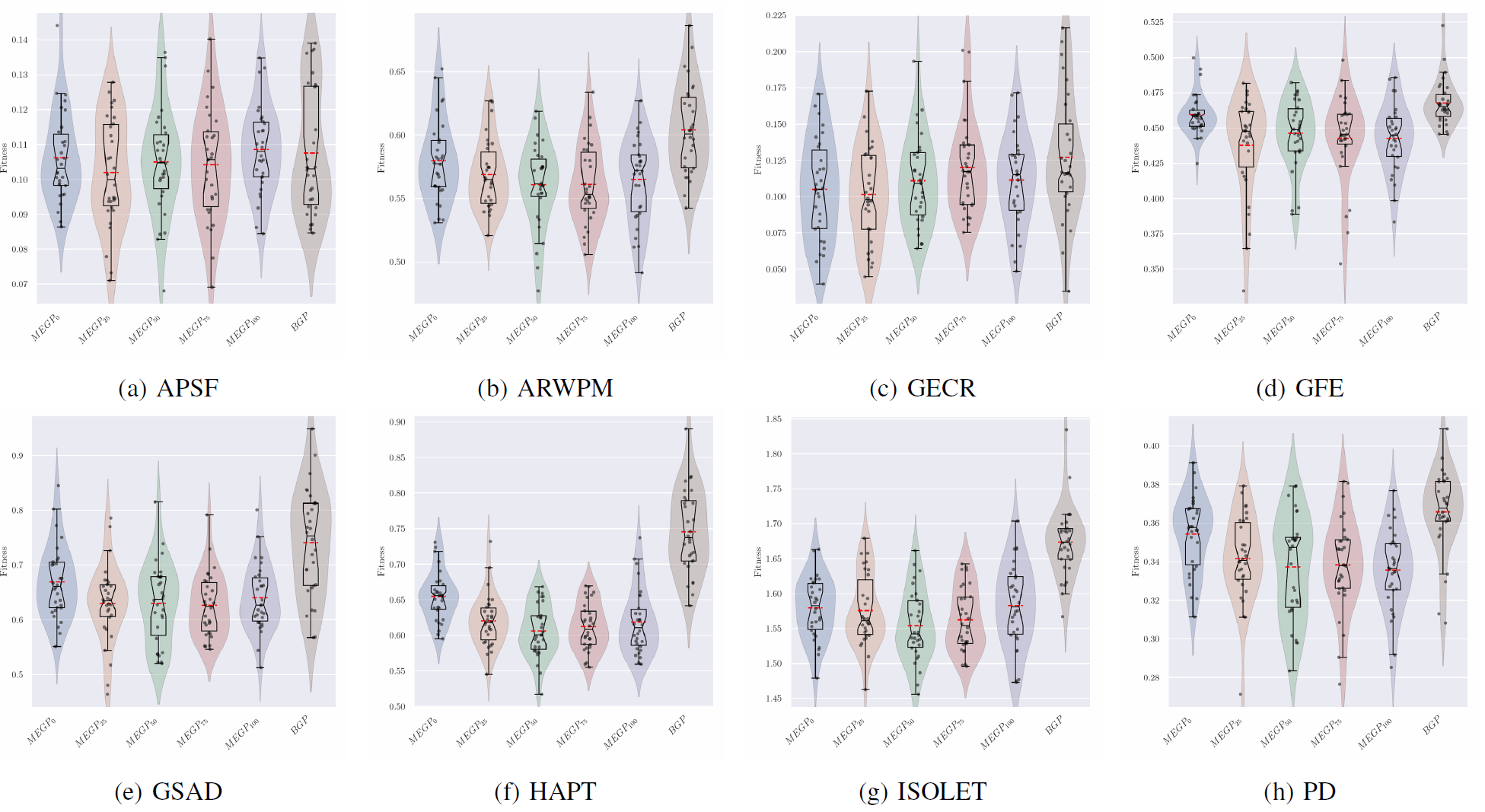}\label{fig:apsf_box_ftall}%

\caption[The distribution of FT\(_{all}\) across BGP and MEGP models over 30 runs.]{Raincloud plots showing the distribution of Fitness (FT\(_{all}\)) across 30 runs for BGP and MEGP models with different ensemble selection probabilities (0\%, 25\%, 50\%, 75\%, 100\%). Each plot illustrates the variability and central tendency of model performance over all generations.}

\label{fig:box_ftall}
\end{figure*}

%% file: con_ftall.tex
\begin{figure*}[ht] 
\centering
\includegraphics[width=\textwidth]{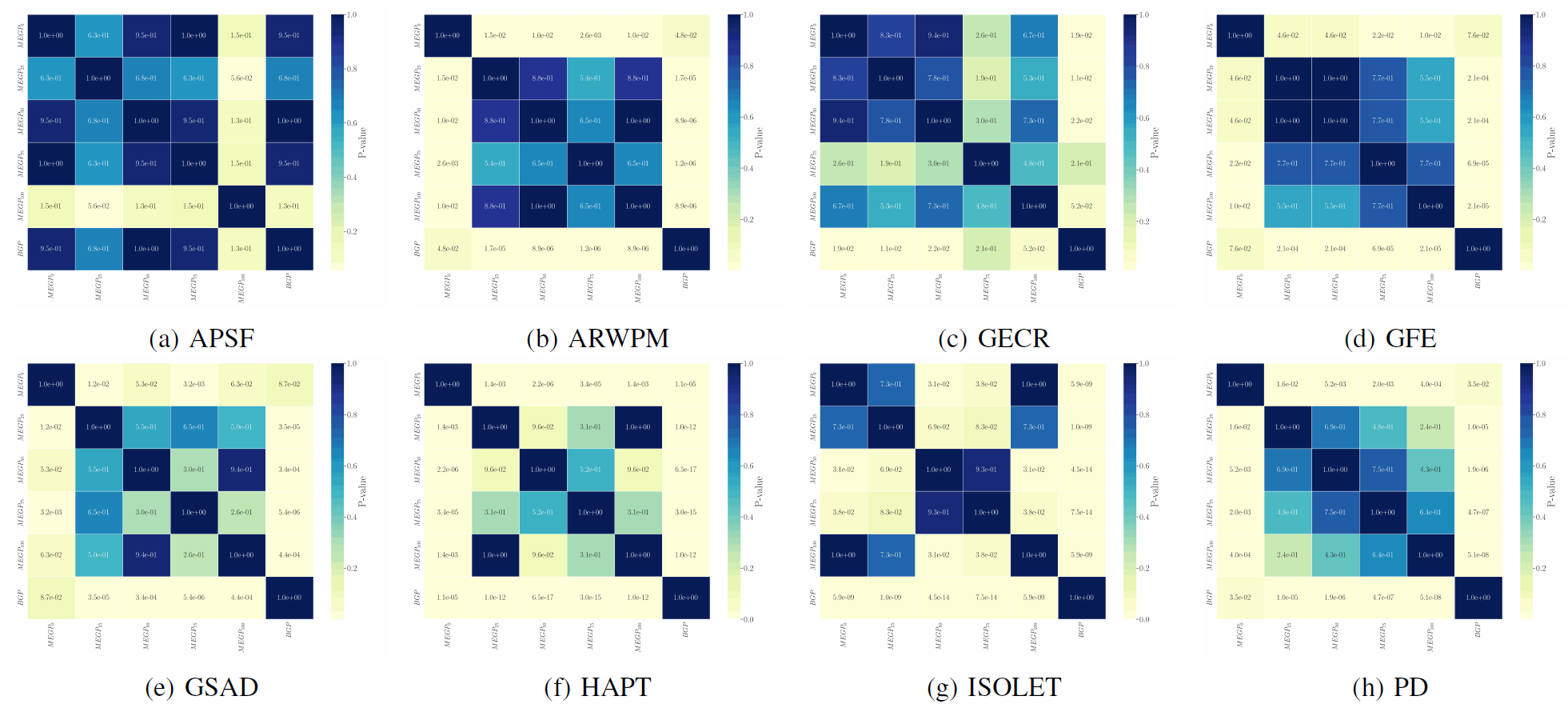}\label{fig:apsf_con_ftall}%

\caption[Heatmap of adjusted pairwise significance from Conover post-hoc test for FT\(_{all}\).]{Heatmap of adjusted p-values from the pairwise Conover post-hoc test for FT\(_{all}\), corrected using the Benjamini-Hochberg method. The heatmap highlights the statistical significance of pairwise comparisons between BGP and MEGP models with varying ensemble selection probabilities (0\%, 25\%, 50\%, 75\%, 100\%) over all generations.}

\label{fig:con_ftall}
\end{figure*}

%% file: cliff_ftall.tex
\begin{figure*}[htbp] 
\centering
\includegraphics[width=\textwidth]{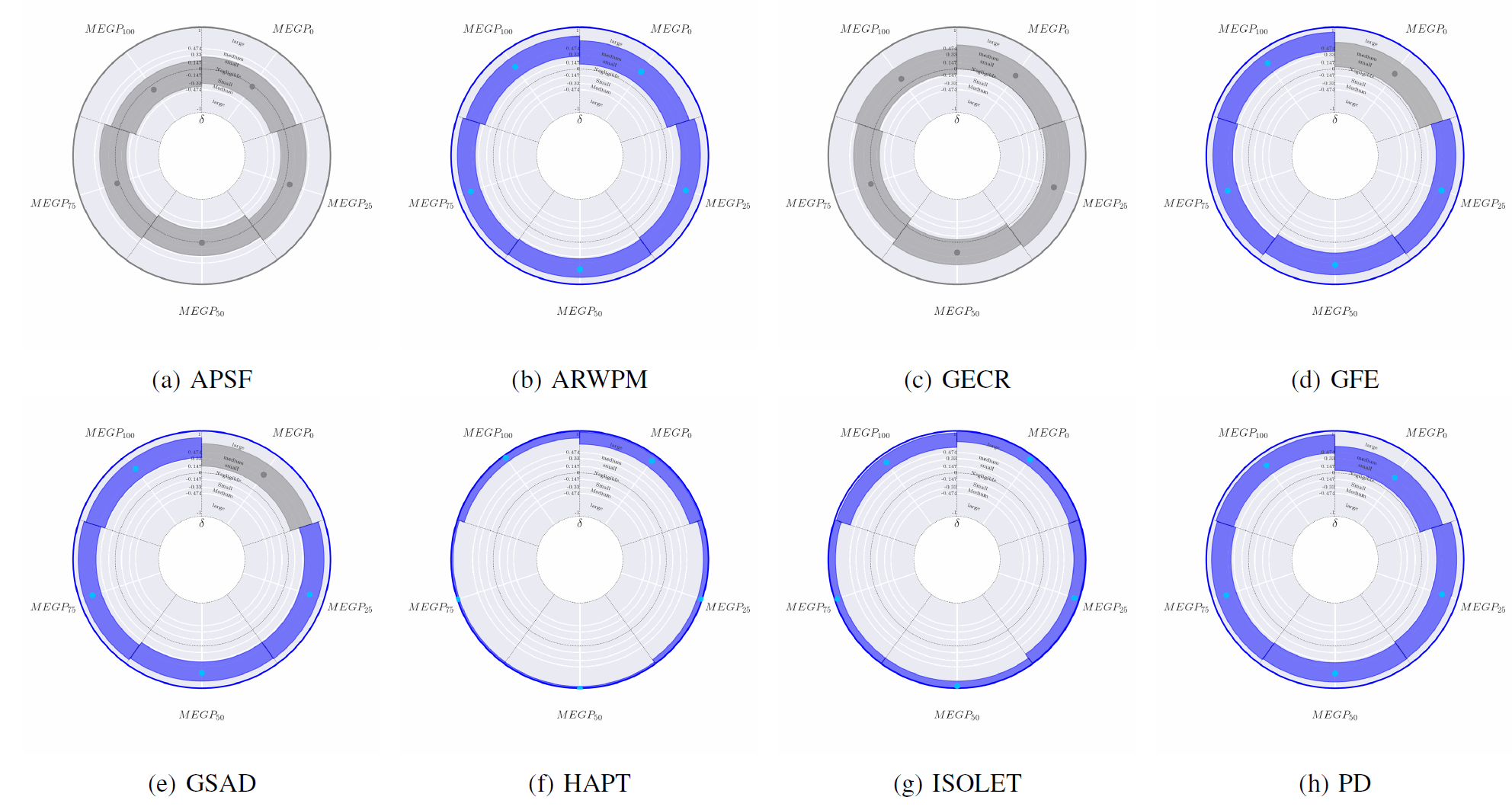}\label{fig:apsf_cliff_ftall}%

\caption[The Cliff's $\delta$ effect size measure and its 95\% confidence intervals for FT\(_{all}\) obtained from 30 BGP and MEGP runs.]{Effect size analysis of FT\(_{all}\) across 30 runs for BGP and MEGP models using Cliff's $\delta$. Each point represents the actual FT\(_{all}\) value obtained, with segments denoting 95\% confidence intervals based on 10,000 bootstrap resamplings. The outer ring color visualizes statistical significance: grey illustrates no significant difference (adjusted Friedman's P-value $>0.05$), while color indicates significant differences; blue indicates that at least one MEGP configuration outperforms BGP (adjusted Conover's p-value $< 0.05$, Cliff's $\delta > 0$), and red signifies that all MEGP configurations underperform relative to BGP (adjusted Conover's p-value $< 0.05$, Cliff's $\delta < 0$). Segment colors show performance differences against BGP: grey for no significant difference (adjusted Conover's p-value $> 0.05$), blue for better performance (Cliff's $\delta > 0$), and red for worse performance (Cliff's $\delta < 0$).}

\label{fig:cliff_ftall}
\end{figure*}

%% file: wtl_ftall.tex
\begin{table*}
\centering
\caption[The results of Friedman and Conover tests and Cliff's $\delta$ analysis for the $FT_{all}$ obtained from MEGP and BGP runs.]{Statistical comparison of $FT_{all}$ results for training data obtained from MEGP and BGP Runs. W, T, and L denote win, tie, and loss based on adjusted Friedman and Conover's p-values. Effect sizes are calculated using Cliff's Delta method and are categorized as negligible, small, medium, or large.}
\label{tab:wtl_ftall}
\begin{tabular}{cccccc}
\hline
\multicolumn{6}{c}{$FT_{all}$} \\
\hline
Dataset & $MEGP_{0}$ & $MEGP_{25}$ & $MEGP_{50}$ & $MEGP_{75}$ & $MEGP_{100}$ \\
\hline
APSF & T (negligible) & T (negligible) & T (negligible) & T (negligible) & T (negligible) \\
ARWPM & W (medium) & W (large) & W (large) & W (large) & W (large) \\
GECR & T (small) & T (medium) & T (small) & T (negligible) & T (small) \\
GFE & T (medium) & W (large) & W (large) & W (large) & W (large) \\
GSAD & T (medium) & W (large) & W (large) & W (large) & W (large) \\
HAPT & W (large) & W (large) & W (large) & W (large) & W (large) \\
ISOLET & W (large) & W (large) & W (large) & W (large) & W (large) \\
PD & W (medium) & W (large) & W (large) & W (large) & W (large) \\
\hline
W - T - L & 4 - 4 - 0 & 6 - 2 - 0 & 6 - 2 - 0 & 6 - 2 - 0 & 6 - 2 - 0 \\
\hline
\end{tabular}
\end{table*}

%% file: box_crall.tex
\begin{figure*}[ht] 
\centering
\includegraphics[width=\textwidth]{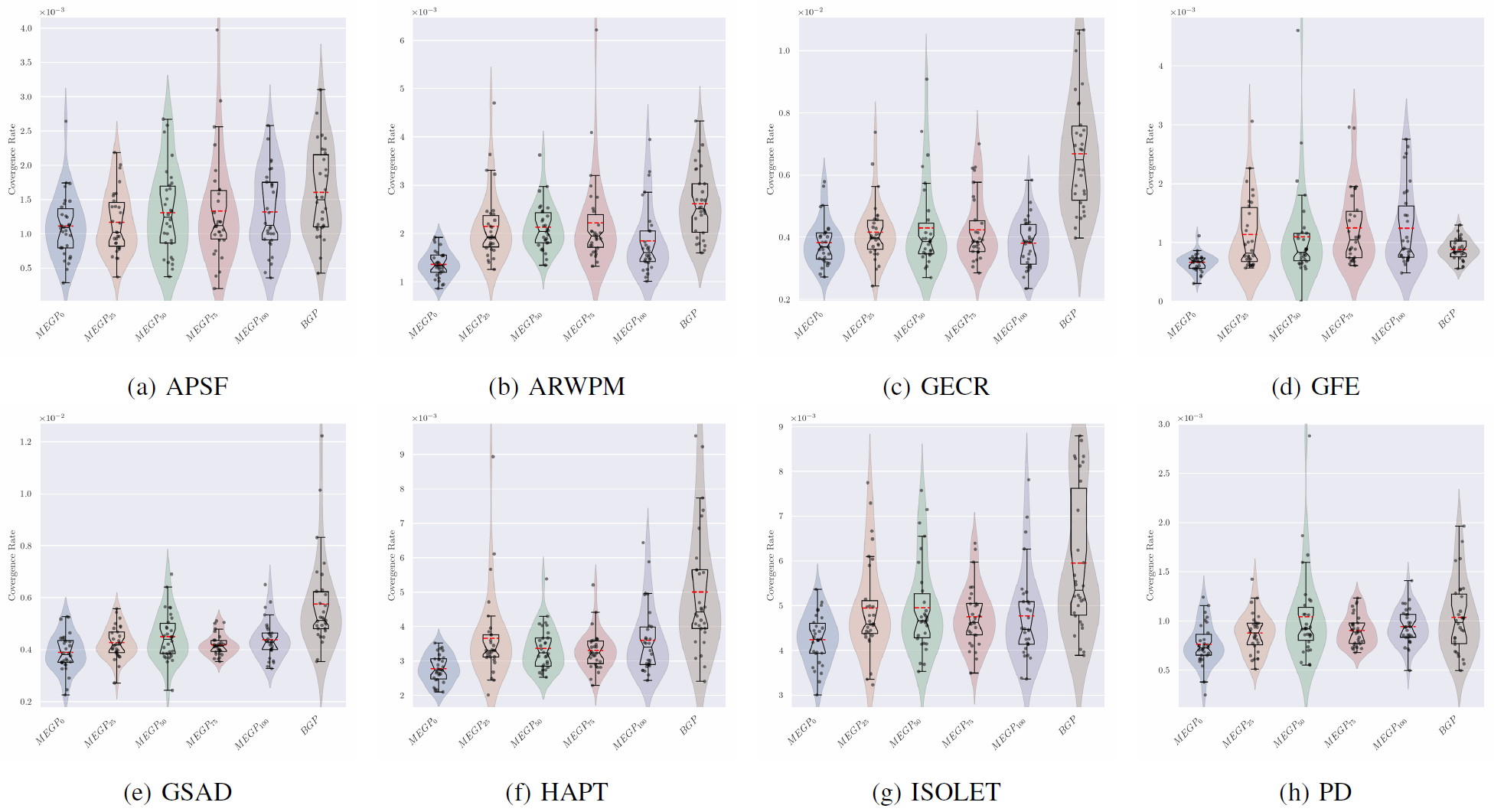}

\caption[The distribution of CR\(_{all}\) across BGP and MEGP models over 30 runs.]{Raincloud plots showing the distribution of Convergence Rate (CR\(_{all}\)) across 30 runs for BGP and MEGP models with different ensemble selection probabilities (0\%, 25\%, 50\%, 75\%, 100\%). Each plot illustrates the variability and central tendency of model performance over generations 101 to 150.}

\label{fig:box_crall}
\end{figure*}

%% file: con_crall.tex
\begin{figure*}[ht] 
\centering
\includegraphics[width=\textwidth]{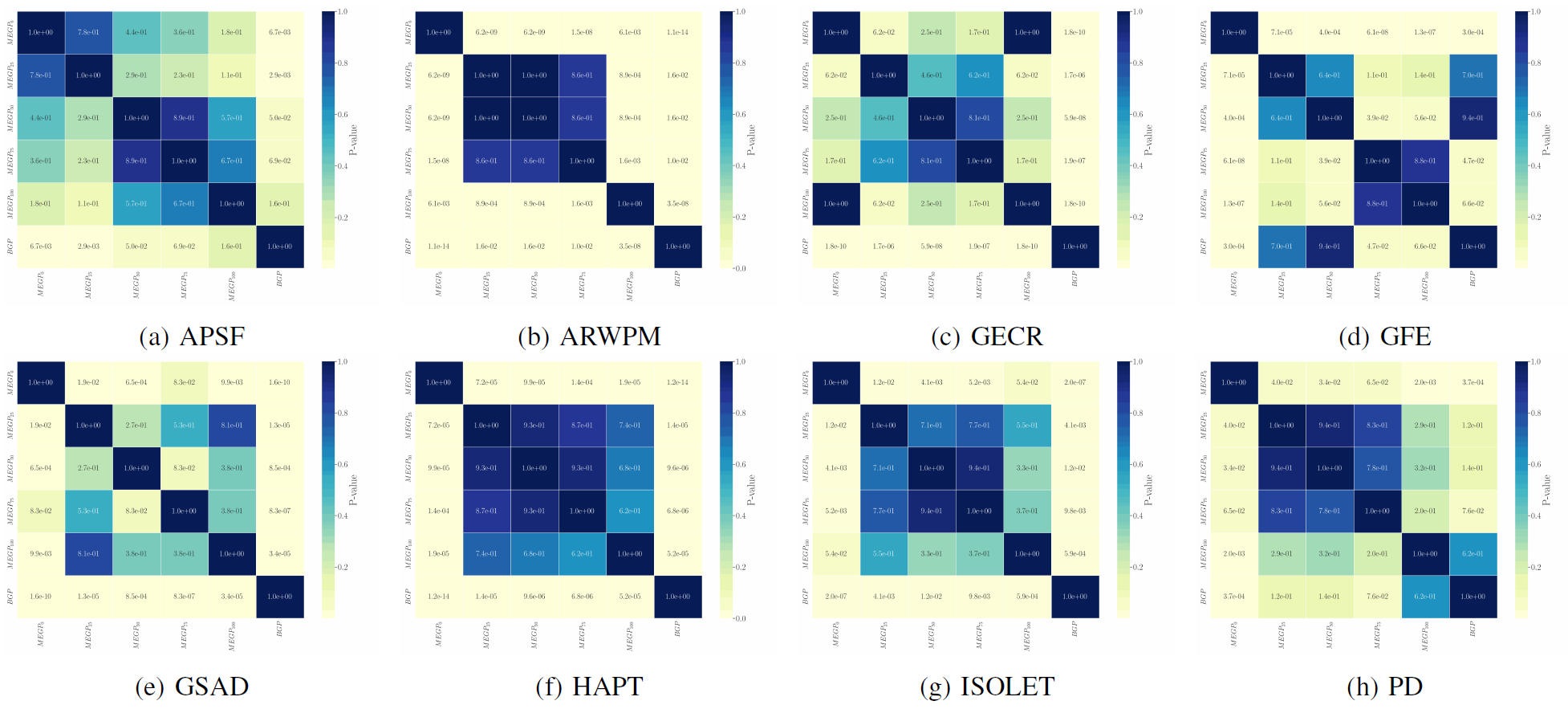}\label{fig:apsf_con_crall}%

\caption[Heatmap of CR\(_{all}\) adjusted pairwise significance from Conover post-hoc test.]{Heatmap of adjusted p-values from the pairwise Conover post-hoc test for CR\(_{all}\), corrected using the Benjamini-Hochberg method. The heatmap highlights the statistical significance of pairwise comparisons between BGP and MEGP models with varying ensemble selection probabilities (0\%, 25\%, 50\%, 75\%, 100\%) over all generations.}

\label{fig:con_crall}
\end{figure*}

%% file: cliff_crall.tex
\begin{figure*}[htbp] 
\centering
\includegraphics[width=\textwidth]{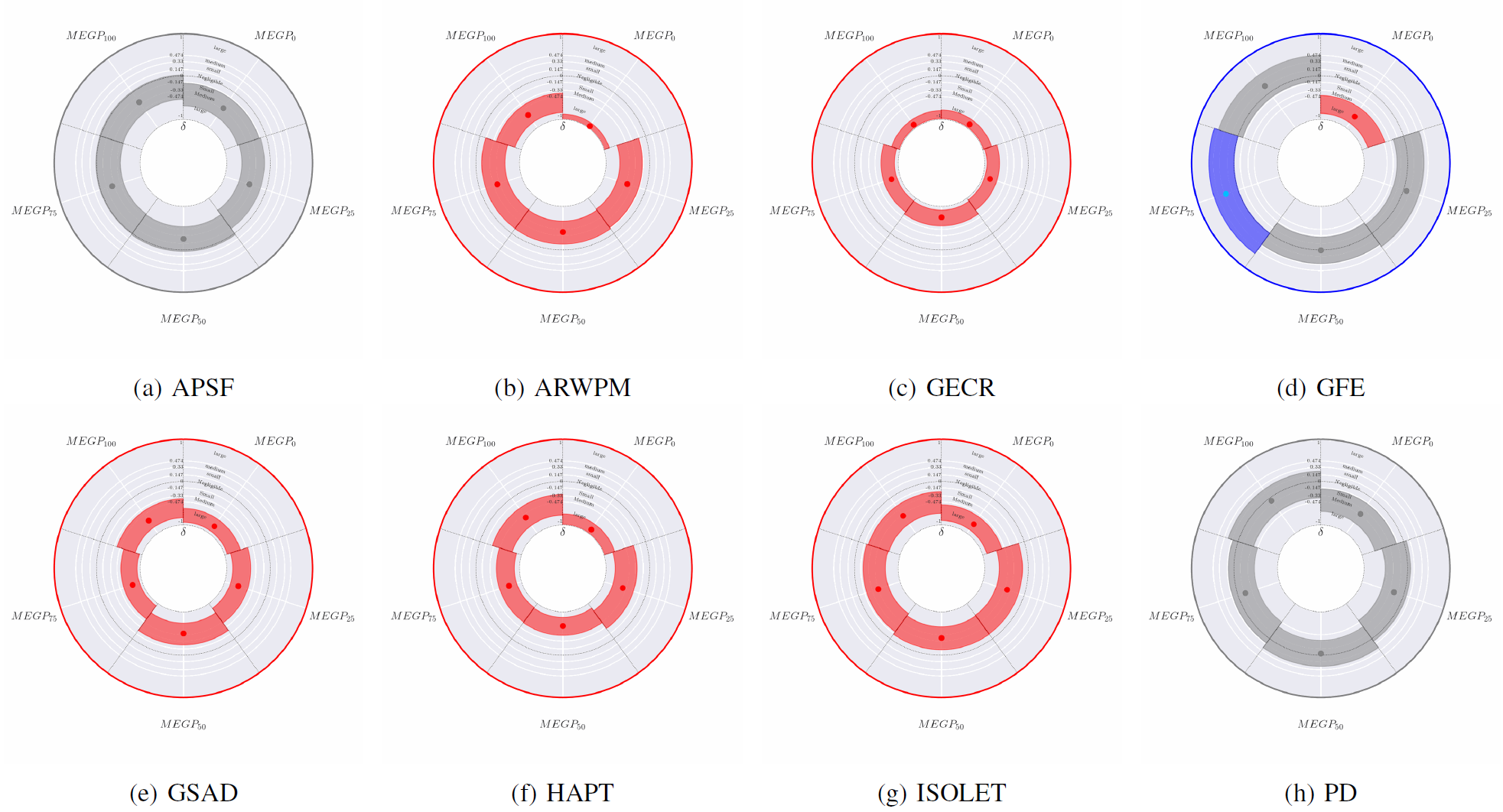}\label{fig:apsf_cliff_crall}%

\caption[The Cliff's $\delta$ effect size measure and its 95\% confidence intervals for CR\(_{all}\) obtained from 30 BGP and MEGP runs.]{Effect size analysis of CR\(_{all}\) across 30 runs for BGP and MEGP models using Cliff's $\delta$. Each point represents the actual CR\(_{all}\) value obtained, with segments denoting 95\% confidence intervals based on 10,000 bootstrap resamplings. The outer ring color visualizes statistical significance: grey illustrates no significant difference (adjusted Friedman's P-value $>0.05$), while color indicates significant differences; blue indicates that at least one MEGP configuration outperforms BGP (adjusted Conover's p-value $< 0.05$, Cliff's $\delta > 0$), and red signifies that all MEGP configurations underperform relative to BGP (adjusted Conover's p-value $< 0.05$, Cliff's $\delta < 0$). Segment colors show performance differences against BGP: grey for no significant difference (adjusted Conover's p-value $> 0.05$), blue for better performance (Cliff's $\delta > 0$), and red for worse performance (Cliff's $\delta < 0$).}

\label{fig:cliff_crall}
\end{figure*}

%% file: wtl_crall.tex
\begin{table*}
\centering
\caption[The results of Friedman and Conover tests and Cliff's $\delta$ analysis for the $CR_{all}$ obtained from MEGP and BGP runs.]{Statistical comparison of $CR_{all}$ results for training data obtained from MEGP and BGP Runs. W, T, and L denote win, tie, and loss based on adjusted Friedman and Conover's p-values. Effect sizes are calculated using Cliff's Delta method and are categorized as negligible, small, medium, or large.}
\label{tab:wtl_crall}
\begin{tabular}{cccccc}
\hline
\multicolumn{6}{c}{$CR_{all}$} \\
\hline
Dataset & $MEGP_{0}$ & $MEGP_{25}$ & $MEGP_{50}$ & $MEGP_{75}$ & $MEGP_{100}$ \\
\hline
APSF & T (medium) & T (medium) & T (small) & T (small) & T (small) \\
ARWPM & L (large) & L (medium) & L (medium) & L (medium) & L (large) \\
GECR & L (large) & L (large) & L (large) & L (large) & L (large) \\
GFE & L (large) & T (negligible) & T (negligible) & W (small) & T (small) \\
GSAD & L (large) & L (large) & L (large) & L (large) & L (large) \\
HAPT & L (large) & L (large) & L (large) & L (large) & L (large) \\
ISOLET & L (large) & L (medium) & L (medium) & L (medium) & L (large) \\
PD & T (medium) & T (small) & T (negligible) & T (small) & T (negligible) \\
\hline
W - T - L & 0 - 2 - 6 & 0 - 3 - 5 & 0 - 3 - 5 & 1 - 2 - 5 & 0 - 3 - 5 \\
\hline
\end{tabular}
\end{table*}

%% file: box_ccrall.tex
\begin{figure*}[ht] 
\centering
\includegraphics[width=\textwidth]{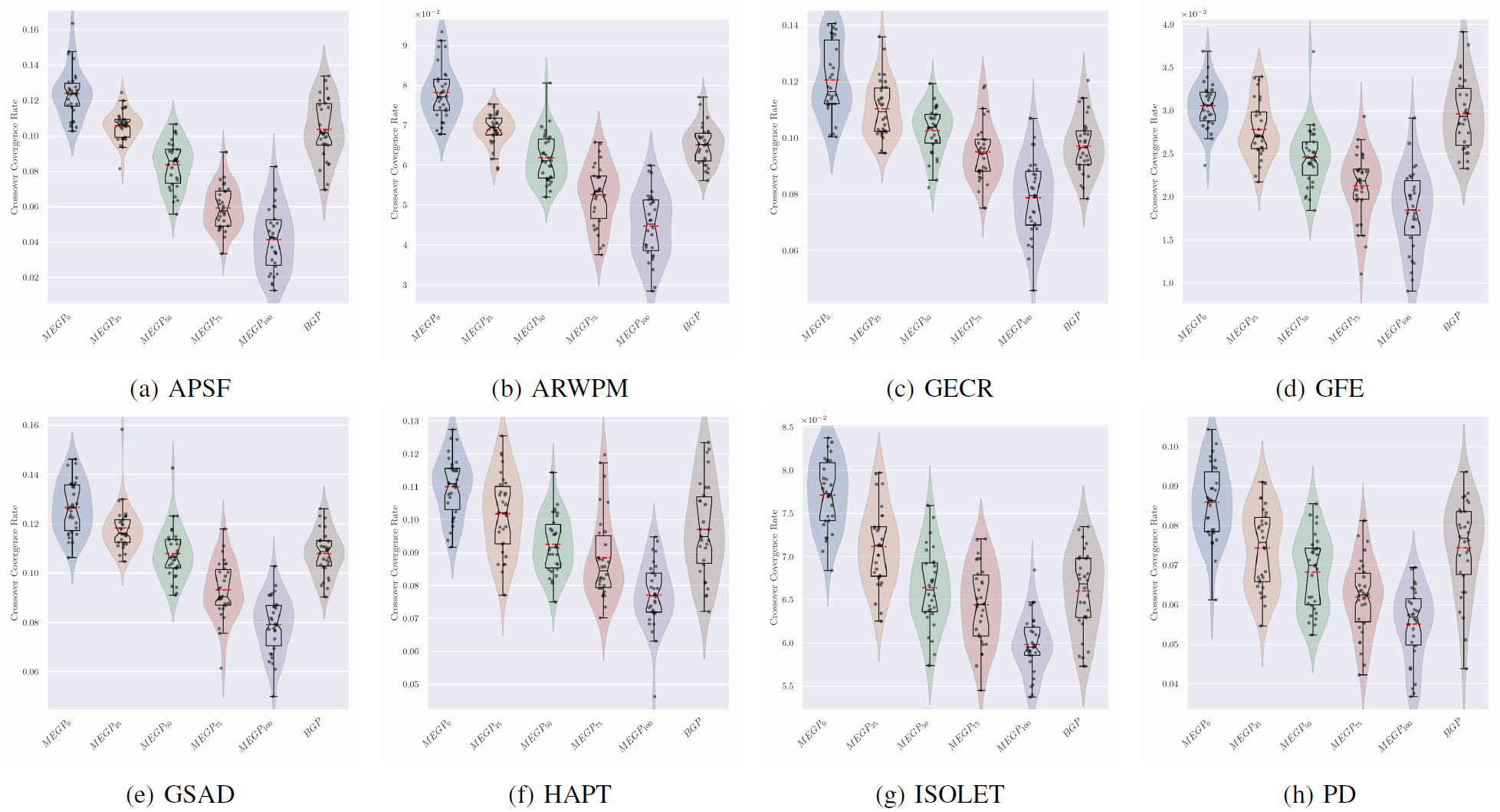}\label{fig:apsf_box_ccrall}%

\caption[The distribution of CCR\(_{all}\) across BGP and MEGP models over 30 runs.]{Raincloud plots showing the distribution of Crossover Convergence Rate (CCR\(_{all}\)) across 30 runs for BGP and MEGP models with different ensemble selection probabilities (0\%, 25\%, 50\%, 75\%, 100\%). Each plot illustrates the variability and central tendency of model performance over all generations.}

\label{fig:box_ccrall}
\end{figure*}

%% file: con_ccrall.tex
\begin{figure*}[ht] 
\centering
\includegraphics[width=\textwidth]{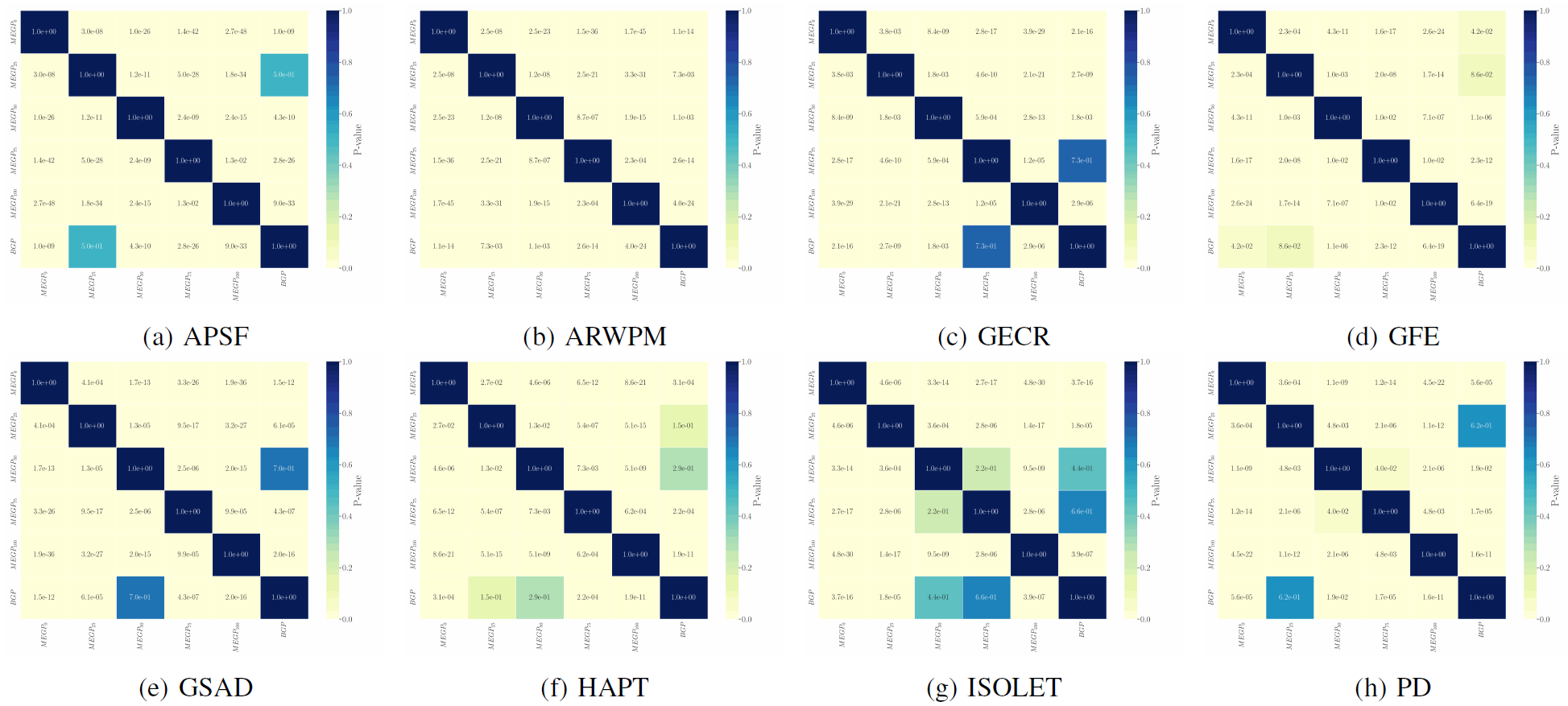}\label{fig:apsf_con_ccrall}%

\caption[Heatmap of adjusted pairwise significance from Conover post-hoc test for CCR\(_{all}\).]{Heatmap of adjusted p-values from the pairwise Conover post-hoc test for CCR\(_{all}\), corrected using the Benjamini-Hochberg method. The heatmap highlights the statistical significance of pairwise comparisons between BGP and MEGP models with varying ensemble selection probabilities (0\%, 25\%, 50\%, 75\%, 100\%) over all generations.}

\label{fig:con_ccrall}
\end{figure*}

%% file: cliff_ccrall.tex
\begin{figure*}[htbp] 
\centering
\includegraphics[width=\textwidth]{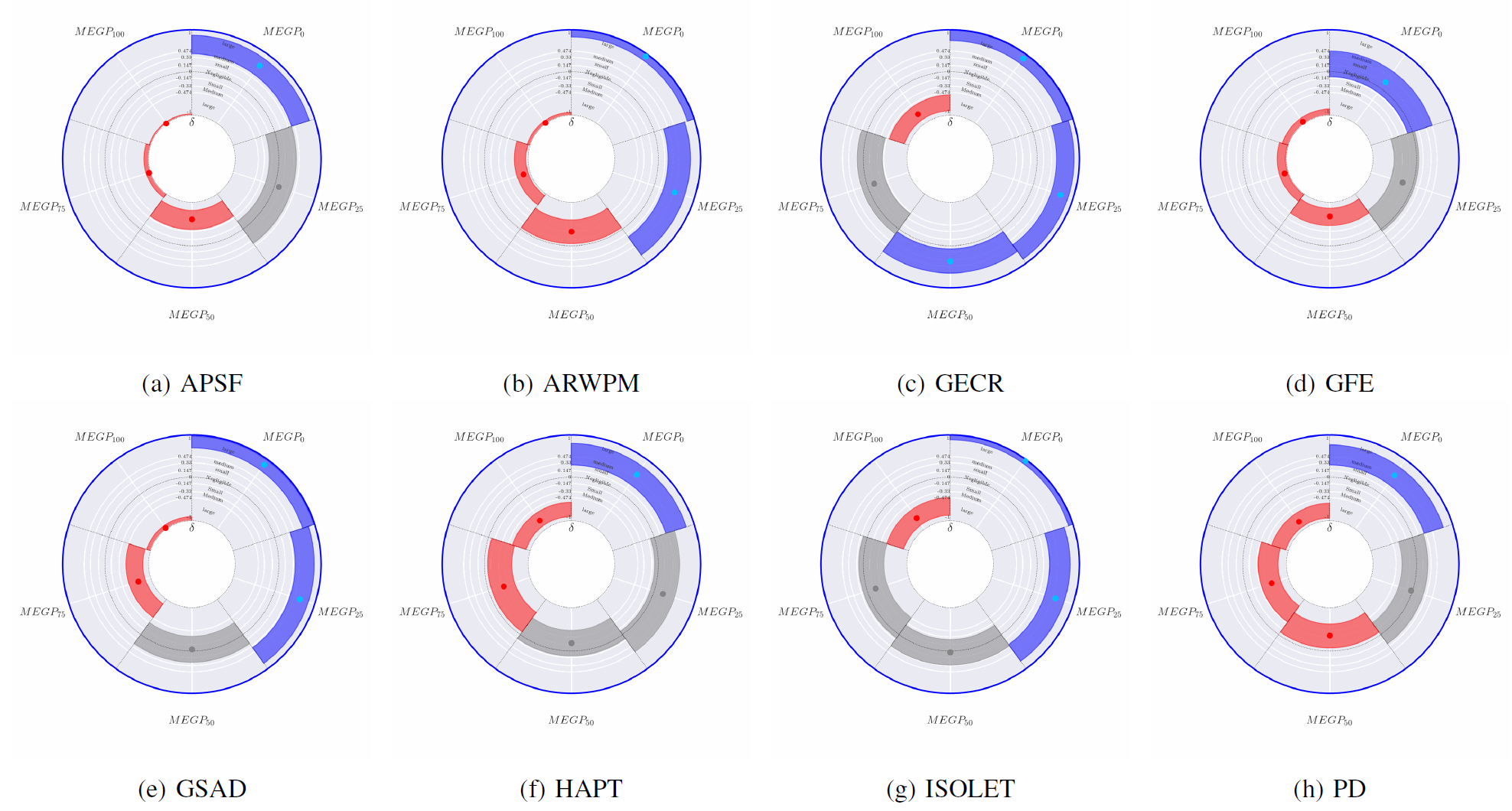}\label{fig:apsf_cliff_ccrall}%

\caption[The Cliff's $\delta$ effect size measure and its 95\% confidence intervals for CCR\(_{all}\) obtained from 30 BGP and MEGP runs.]{Effect size analysis of CCR\(_{all}\) across 30 runs for BGP and MEGP models using Cliff's $\delta$. Each point represents the actual CCR\(_{all}\) value obtained, with segments denoting 95\% confidence intervals based on 10,000 bootstrap resamplings. The outer ring color visualizes statistical significance: grey illustrates no significant difference (adjusted Friedman's P-value $>0.05$), while color indicates significant differences; blue indicates that at least one MEGP configuration outperforms BGP (adjusted Conover's p-value $< 0.05$, Cliff's $\delta > 0$), and red signifies that all MEGP configurations underperform relative to BGP (adjusted Conover's p-value $< 0.05$, Cliff's $\delta < 0$). Segment colors show performance differences against BGP: grey for no significant difference (adjusted Conover's p-value $> 0.05$), blue for better performance (Cliff's $\delta > 0$), and red for worse performance (Cliff's $\delta < 0$).}

\label{fig:cliff_ccrall}
\end{figure*}

%% file: wtl_ccrall.tex
\begin{table*}
\centering
\caption[The results of Friedman and Conover tests and Cliff's $\delta$ analysis for the $CCR_{all}$ obtained from MEGP and BGP runs.]{Statistical comparison of $CCR_{all}$ results for training data obtained from MEGP and BGP Runs. W, T, and L denote win, tie, and loss based on adjusted Friedman and Conover's p-values. Effect sizes are calculated using Cliff's Delta method and are categorized as negligible, small, medium, or large.}
\label{tab:wtl_ccrall}
\begin{tabular}{cccccc}
\hline
\multicolumn{6}{c}{$CCR_{all}$} \\
\hline
Dataset & $MEGP_{0}$ & $MEGP_{25}$ & $MEGP_{50}$ & $MEGP_{75}$ & $MEGP_{100}$ \\
\hline
APSF & W (large) & T (negligible) & L (large) & L (large) & L (large) \\
ARWPM & W (large) & W (large) & L (small) & L (large) & L (large) \\
GECR & W (large) & W (large) & W (medium) & T (small) & L (large) \\
GFE & W (small) & T (small) & L (large) & L (large) & L (large) \\
GSAD & W (large) & W (large) & T (negligible) & L (large) & L (large) \\
HAPT & W (large) & T (small) & T (small) & L (medium) & L (large) \\
ISOLET & W (large) & W (large) & T (negligible) & T (small) & L (large) \\
PD & W (large) & T (negligible) & L (medium) & L (large) & L (large) \\
\hline
W - T - L & 8 - 0 - 0 & 4 - 4 - 0 & 1 - 3 - 4 & 0 - 2 - 6 & 0 - 0 - 8 \\
\hline
\end{tabular}
\end{table*}

%% file: box_entropy.tex
\begin{figure*}[ht] 
\centering
\includegraphics[width=\textwidth]{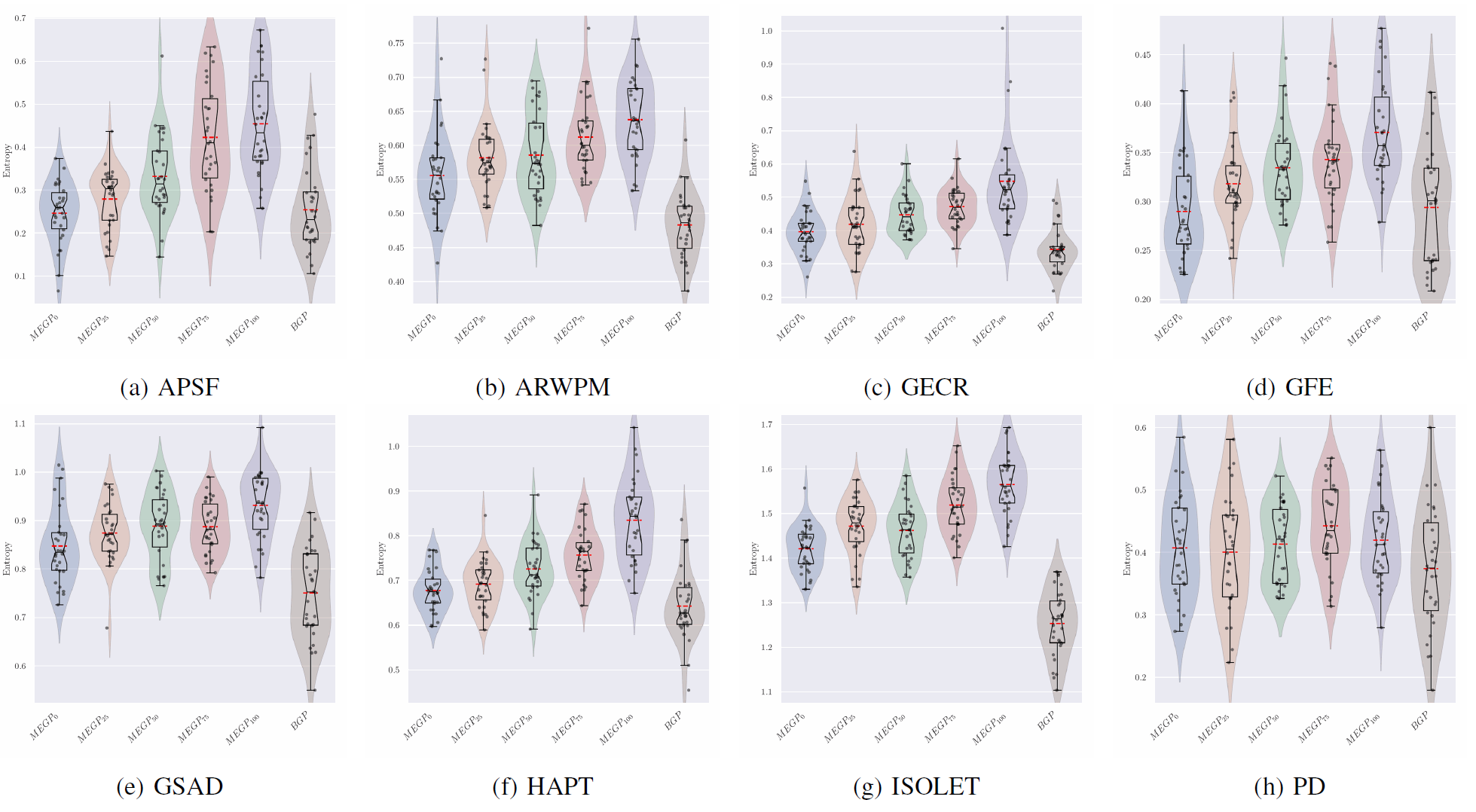}\label{fig:apsf_box_entropy}%

\caption[Distribution of final population entropy across BGP and MEGP models over 30 runs.]{Raincloud plots showing the distribution of final population entropy across 30 runs for BGP and MEGP models with varying ensemble selection probabilities (0\%, 25\%, 50\%, 75\%, 100\%). Each plot illustrates the diversity variability and central tendency within the final population.}

\label{fig:box_entropy}
\end{figure*}

%% file: con_entropy.tex
\begin{figure*}[ht] 
\centering
\includegraphics[width=\textwidth]{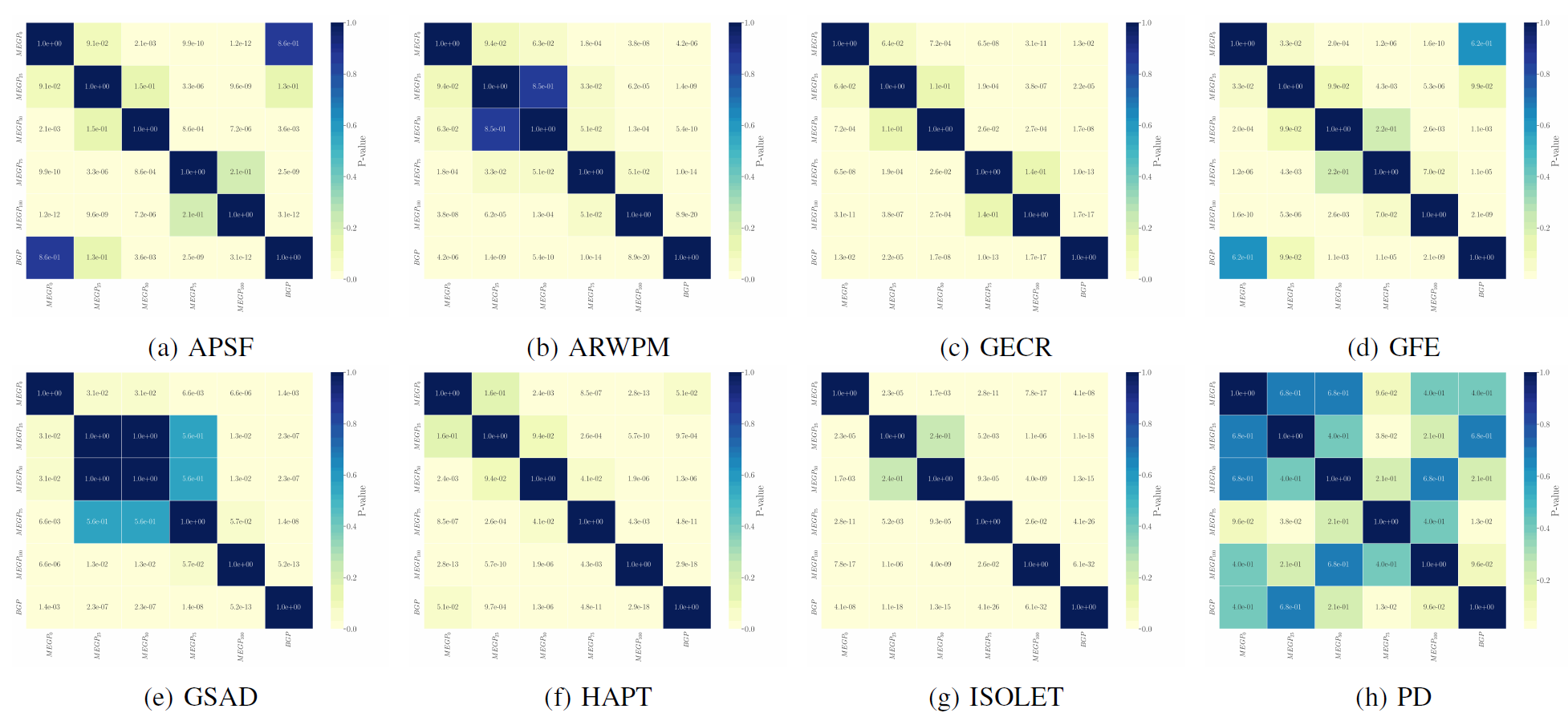}\label{fig:apsf_con_entropy}%

\caption[Heatmap of adjusted pairwise significance from Conover post-hoc test for final population entropy.]{Heatmap of adjusted p-values from the pairwise Conover post-hoc test for final population entropy, corrected using the Benjamini-Hochberg method. The heatmap highlights the statistical significance of pairwise comparisons between BGP and MEGP models with varying ensemble selection probabilities (0\%, 25\%, 50\%, 75\%, 100\%) over 30 runs.}

\label{fig:con_entropy}
\end{figure*}

%% file: cliff_entropy.tex
\begin{figure*}[htbp] 
\centering
\includegraphics[width=\textwidth]{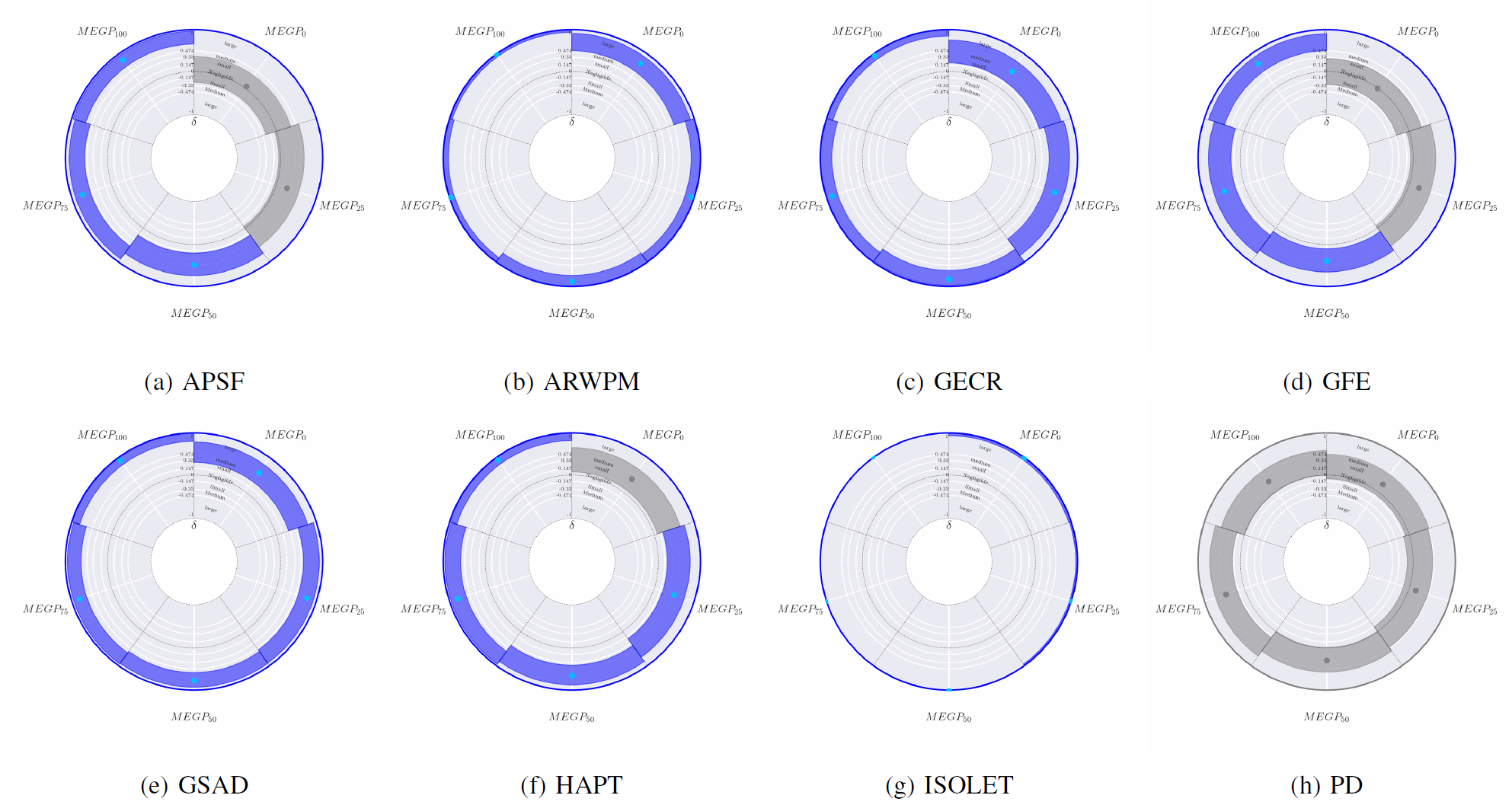}\label{fig:apsf_cliff_entropy}%

\caption[The Cliff's $\delta$ effect size measure and its 95\% confidence intervals for the final population entropy across 30 BGP and MEGP runs.]{Effect size analysis of final population entropy across 30 runs for BGP and MEGP models using Cliff's $\delta$. Each point represents the actual entropy value obtained, with segments denoting 95\% confidence intervals based on 10,000 bootstrap resamplings. The outer ring color visualizes statistical significance: grey illustrates no significant difference (adjusted Friedman's P-value $>0.05$), while color indicates significant differences; blue indicates that at least one MEGP configuration shows greater diversity than BGP (adjusted Conover's p-value $< 0.05$, Cliff's $\delta > 0$), and red signifies that all MEGP configurations have lower diversity relative to BGP (adjusted Conover's p-value $< 0.05$, Cliff's $\delta < 0$). Segment colors show diversity differences against BGP: grey for no significant difference (adjusted Conover's p-value $> 0.05$), blue for greater diversity (Cliff's $\delta > 0$), and red for lower diversity (Cliff's $\delta < 0$).}

\label{fig:cliff_entropy}
\end{figure*}

%% file: wtl_entropy.tex
\begin{table*}
\centering
\caption[The results of Friedman and Conover tests and Cliff's $\delta$ analysis for the Entropy obtained from MEGP and BGP runs.]{Statistical comparison of Entropy results obtained from MEGP and BGP Runs. W, T, and L denote win, tie, and loss based on adjusted Friedman and Conover's p-values. Effect sizes are calculated using Cliff's Delta method and are categorized as negligible, small, medium, or large.}
\label{tab:wtl_entropy}
\begin{tabular}{cccccc}
\hline
\multicolumn{6}{c}{Entropy} \\
\hline
Dataset & $MEGP_{0}$ & $MEGP_{25}$ & $MEGP_{50}$ & $MEGP_{75}$ & $MEGP_{100}$ \\
\hline
APSF & T (negligible) & T (small) & W (medium) & W (large) & W (large) \\
ARWPM & W (large) & W (large) & W (large) & W (large) & W (large) \\
GECR & W (large) & W (large) & W (large) & W (large) & W (large) \\
GFE & T (negligible) & T (small) & W (medium) & W (large) & W (large) \\
GSAD & W (large) & W (large) & W (large) & W (large) & W (large) \\
HAPT & T (medium) & W (large) & W (large) & W (large) & W (large) \\
ISOLET & W (large) & W (large) & W (large) & W (large) & W (large) \\
PD & T (small) & T (small) & T (small) & T (medium) & T (small) \\
\hline
W - T - L & 4 - 4 - 0 & 5 - 3 - 0 & 7 - 1 - 0 & 7 - 1 - 0 & 7 - 1 - 0 \\
\hline
\end{tabular}
\end{table*}

%% file: box_time.tex
\begin{figure*}[ht] 
\centering
\includegraphics[width=\textwidth]{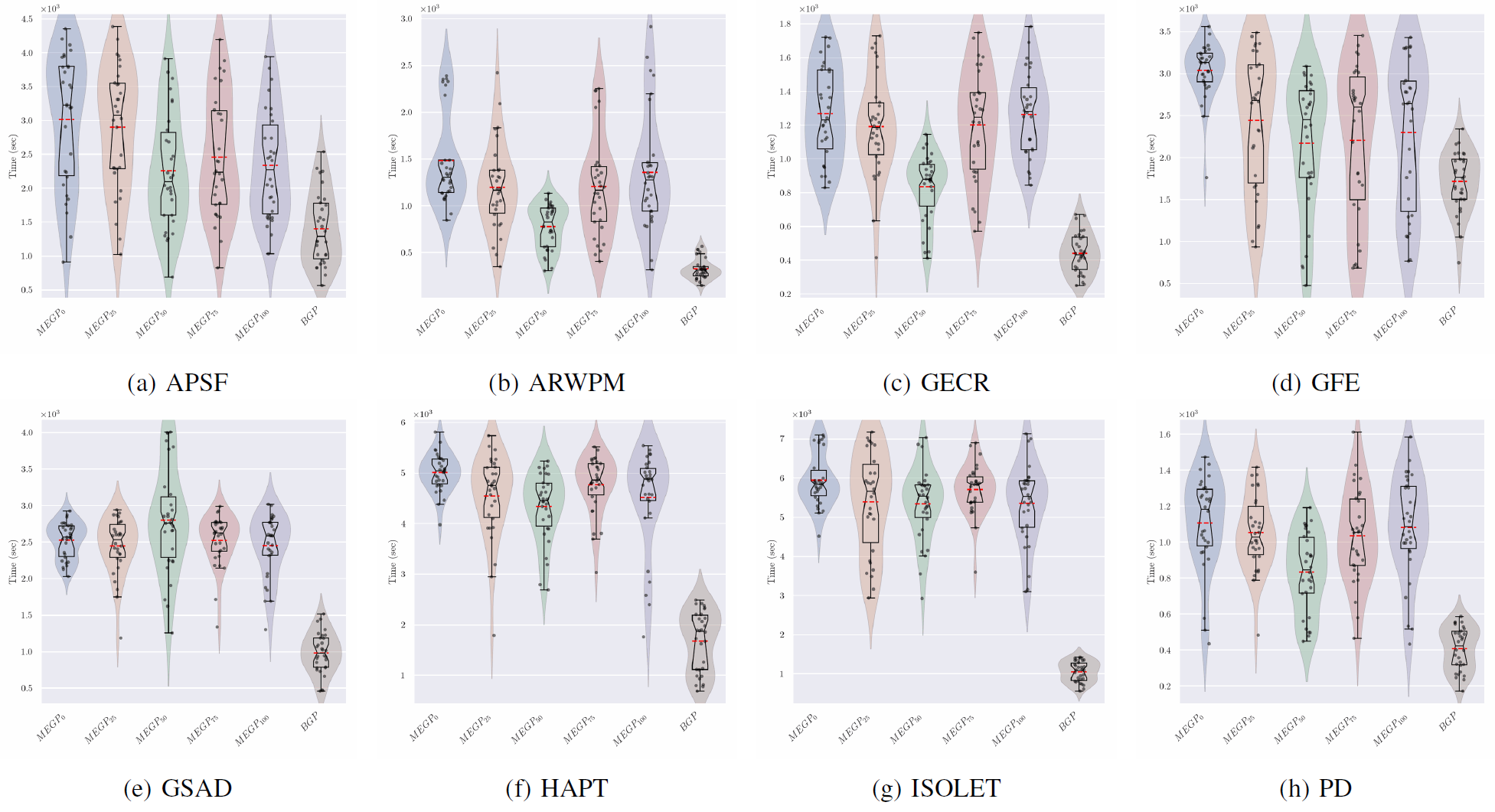}\label{fig:apsf_box_time}%

\caption[The distribution of elapsed time across BGP and MEGP models over 30 runs.]{Raincloud plots showing the distribution of elapsed time across 30 runs for BGP and MEGP models with different ensemble selection probabilities (0\%, 25\%, 50\%, 75\%, 100\%). Each plot illustrates the variability and central tendency of runtime performance.}

\label{fig:box_time}
\end{figure*}

%% file: con_time.tex
\begin{figure*}[ht] 
\centering
\includegraphics[width=\textwidth]{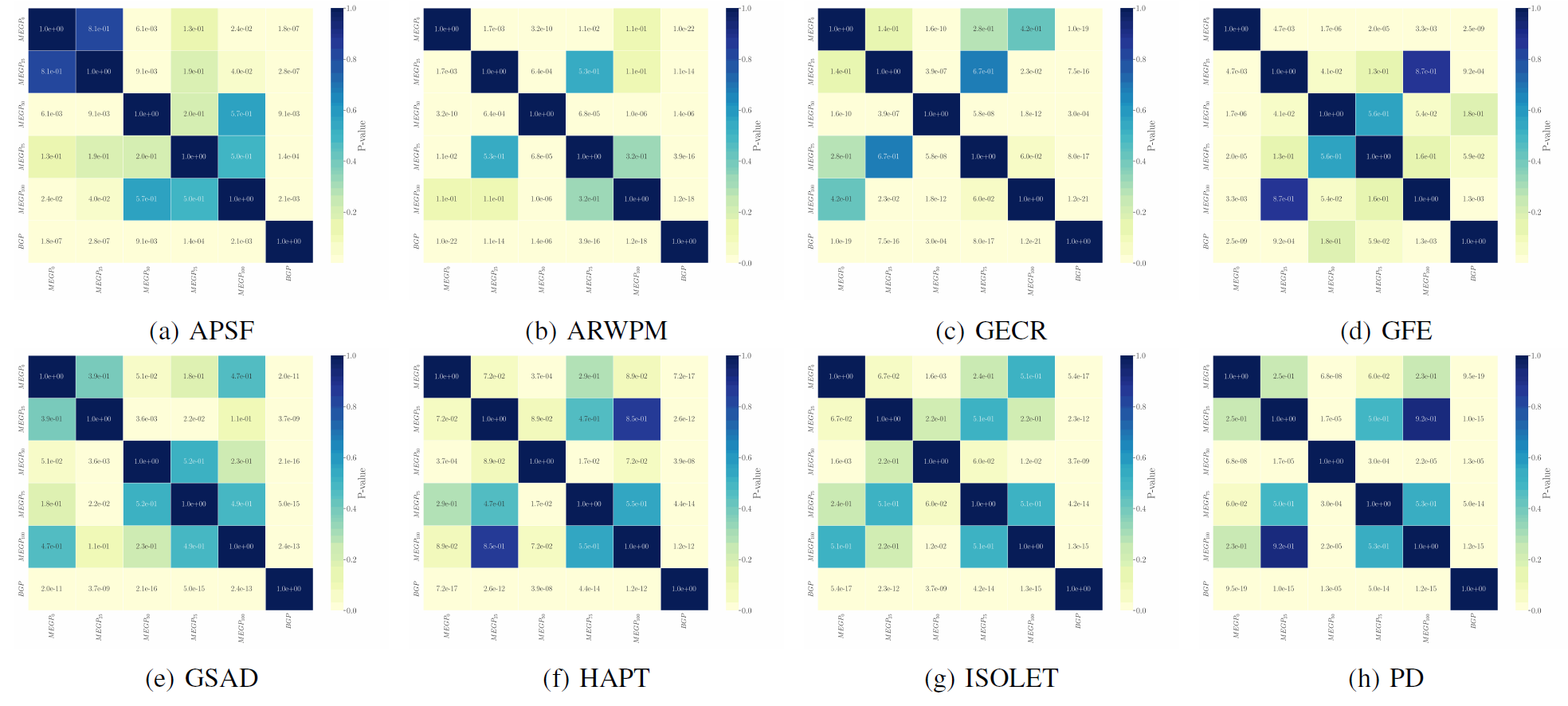}\label{fig:apsf_con_time}%

\caption[Heatmap of adjusted pairwise significance from Conover post-hoc test for Running Time (sec).]{Heatmap of adjusted p-values from the pairwise Conover post-hoc test for Running Time (sec), corrected using the Benjamini-Hochberg method. The heatmap highlights the statistical significance of pairwise comparisons between BGP and MEGP models with varying ensemble selection probabilities (0\%, 25\%, 50\%, 75\%, 100\%) over 30 runs.}

\label{fig:con_time}
\end{figure*}

%% file: cliff_time.tex
\begin{figure*}[htbp] 
\centering
\includegraphics[width=\textwidth]{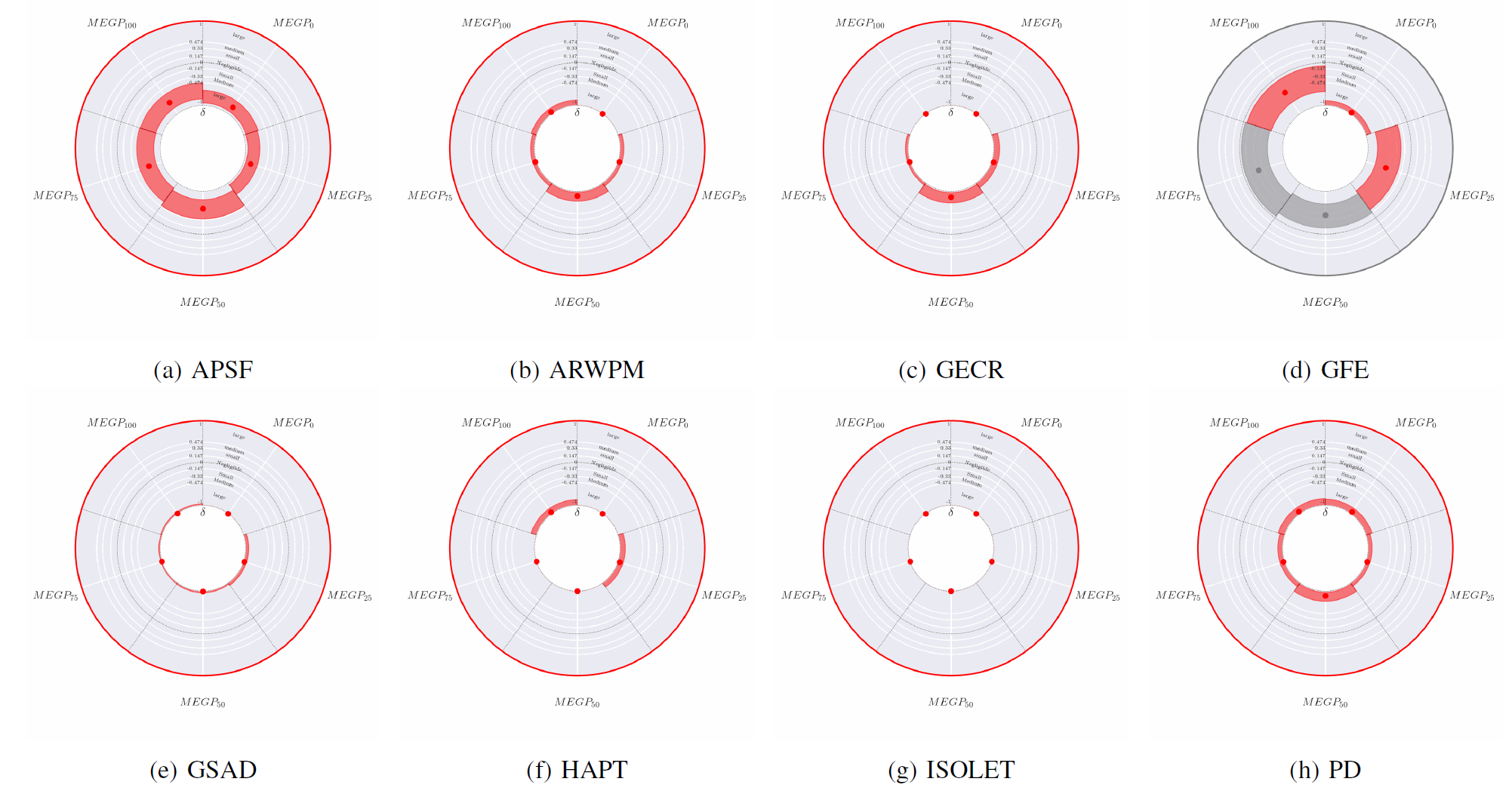}\label{fig:apsf_cliff_time}%

\caption[The Cliff's $\delta$ effect size measure and its 95\% confidence intervals for elapsed time obtained from 30 BGP and MEGP runs.]{Effect size analysis of elapsed time across 30 runs for BGP and MEGP models using Cliff's $\delta$. Each point represents the actual elapsed time value obtained, with segments denoting 95\% confidence intervals based on 10,000 bootstrap resamplings. The outer ring color visualizes statistical significance: grey illustrates no significant difference (adjusted Friedman's P-value $>0.05$), while color indicates significant differences; blue indicates that at least one MEGP configuration runs faster than BGP (adjusted Conover's p-value $< 0.05$, Cliff's $\delta > 0$), and red signifies that all MEGP configurations run slower relative to BGP (adjusted Conover's p-value $< 0.05$, Cliff's $\delta < 0$). Segment colors show elapsed time differences against BGP: grey for no significant difference (adjusted Conover's p-value $> 0.05$), blue for faster performance (Cliff's $\delta > 0$), and red for slower performance (Cliff's $\delta < 0$).}

\label{fig:cliff_time}
\end{figure*}

%% file: wtl_time.tex
\begin{table*}
\centering
\caption[The results of Friedman and Conover tests and Cliff's $\delta$ analysis for the Running Time (sec) obtained from MEGP and BGP runs.]{Statistical comparison of Running Time (seconds) for testing data obtained from MEGP and BGP Runs. W, T, and L denote win, tie, and loss based on adjusted Friedman and Conover's p-values. Effect sizes are calculated using Cliff's Delta method and are categorized as negligible, small, medium, or large.}
\label{tab:wtl_time}
\begin{tabular}{cccccc}
\hline
\multicolumn{6}{c}{Running Time (sec)} \\
\hline
Dataset & $MEGP_{0}$ & $MEGP_{25}$ & $MEGP_{50}$ & $MEGP_{75}$ & $MEGP_{100}$ \\
\hline
APSF & L (large) & L (large) & L (large) & L (large) & L (large) \\
ARWPM & L (large) & L (large) & L (large) & L (large) & L (large) \\
GECR & L (large) & L (large) & L (large) & L (large) & L (large) \\
GFE & L (large) & L (large) & T (medium) & T (medium) & L (medium) \\
GSAD & L (large) & L (large) & L (large) & L (large) & L (large) \\
HAPT & L (large) & L (large) & L (large) & L (large) & L (large) \\
ISOLET & L (large) & L (large) & L (large) & L (large) & L (large) \\
PD & L (large) & L (large) & L (large) & L (large) & L (large) \\
\hline
W - T - L & 0 - 0 - 8 & 0 - 0 - 8 & 0 - 1 - 7 & 0 - 1 - 7 & 0 - 0 - 8 \\
\hline
\end{tabular}
\end{table*}

%% file: rank_gen_megp_0.tex
\begin{figure*}[ht] 
\centering
\includegraphics[width=\textwidth]{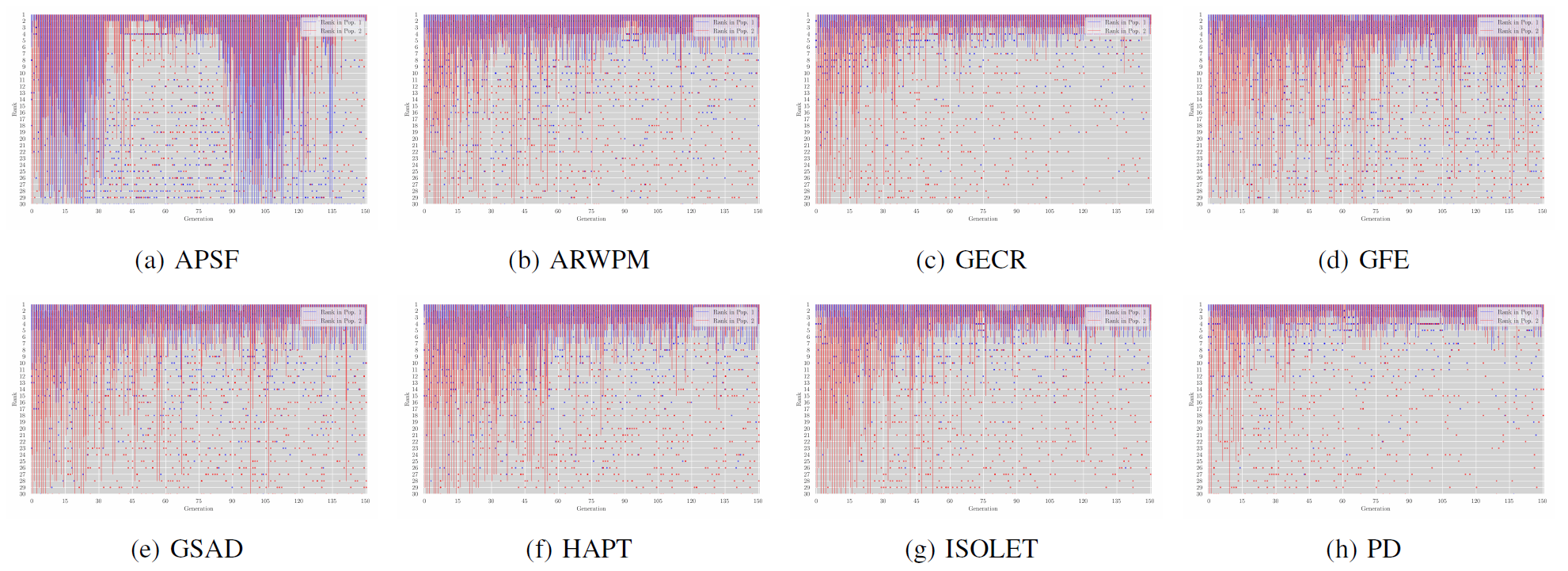}

\caption[Boxplots of isolated rankings for best ensemble individuals in MEGP\textsubscript{0}]{Boxplots of isolated rankings for individuals forming the best ensemble within their respective populations across generations (0–150) over 30 runs of MEGP\textsubscript{0} for the benchmark datasets: (a) APSF, (b) ARWPM, (c) GECR, (d) GFE, (e) GSAD, (f) HAPT, (g) ISOLET, and (h) PD. The y-axis represents the ranking position, where lower values indicate better placement, while the x-axis denotes the generation number.}

\label{fig:rank_gen_megp_0}
\end{figure*}

%% file: rank_gen_megp_25.tex
\begin{figure*}[ht] 
\centering
\includegraphics[width=\textwidth]{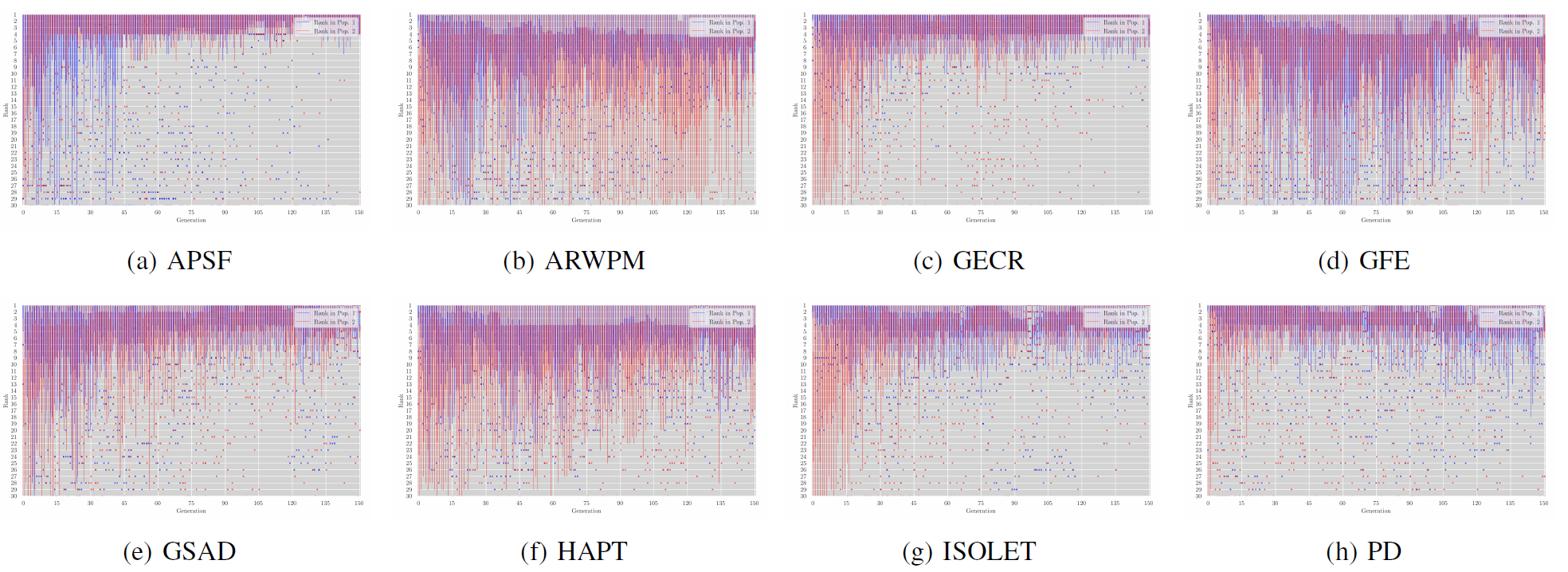}

\caption[Boxplots of isolated rankings for best ensemble individuals in MEGP\textsubscript{25}]{Boxplots of isolated rankings for individuals forming the best ensemble within their respective populations across generations (0–150) over 30 runs of MEGP\textsubscript{25} for the benchmark datasets: (a) APSF, (b) ARWPM, (c) GECR, (d) GFE, (e) GSAD, (f) HAPT, (g) ISOLET, and (h) PD. The y-axis represents the ranking position, where lower values indicate better placement, while the x-axis denotes the generation number.}

\label{fig:rank_gen_megp_25}
\end{figure*}

%% file: rank_gen_megp_50.tex
\begin{figure*}[ht] 
\centering
\includegraphics[width=\textwidth]{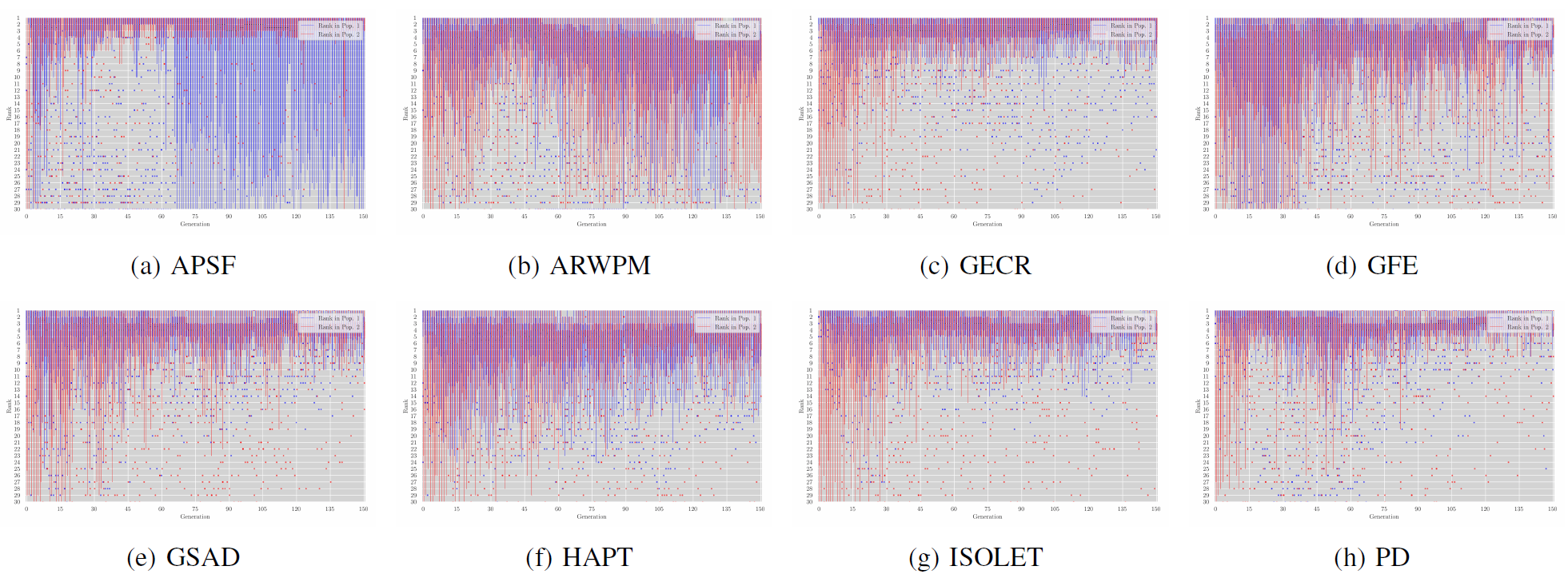}

\caption[Boxplots of isolated rankings for best ensemble individuals in MEGP\textsubscript{50}]{Boxplots of isolated rankings for individuals forming the best ensemble within their respective populations across generations (0–150) over 30 runs of MEGP\textsubscript{50} for the benchmark datasets: (a) APSF, (b) ARWPM, (c) GECR, (d) GFE, (e) GSAD, (f) HAPT, (g) ISOLET, and (h) PD. The y-axis represents the ranking position, where lower values indicate better placement, while the x-axis denotes the generation number.}

\label{fig:rank_gen_megp_50}
\end{figure*}

%% file: rank_gen_megp_75.tex
\begin{figure*}[ht] 
\centering
\includegraphics[width=\textwidth]{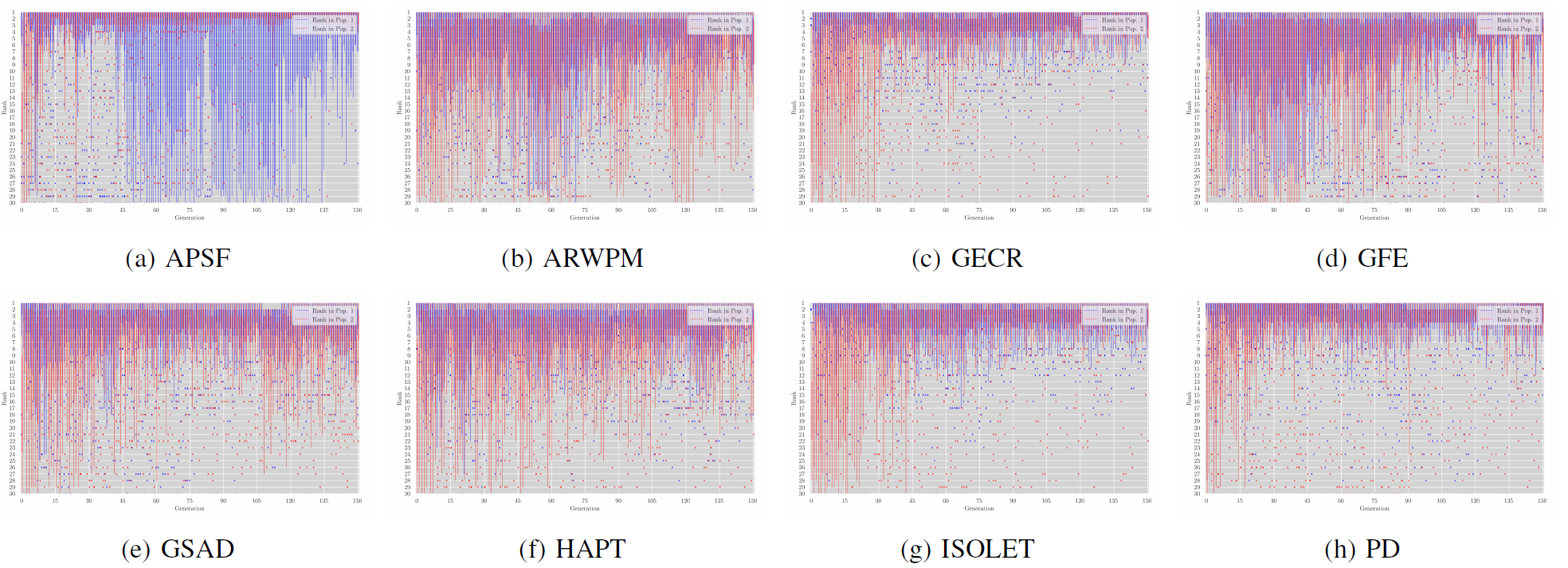}

\caption[Boxplots of isolated rankings for best ensemble individuals in MEGP\textsubscript{75}]{Boxplots of isolated rankings for individuals forming the best ensemble within their respective populations across generations (0–150) over 30 runs of MEGP\textsubscript{75} for the benchmark datasets: (a) APSF, (b) ARWPM, (c) GECR, (d) GFE, (e) GSAD, (f) HAPT, (g) ISOLET, and (h) PD. The y-axis represents the ranking position, where lower values indicate better placement, while the x-axis denotes the generation number.}

\label{fig:rank_gen_megp_75}
\end{figure*}

%% file: table_friedman_generalization.tex
\begin{table*}[]
\centering
\caption{Comparison of Baseline GP (BGP) and Multi-population Ensemble GP (MEGP) models across various criteria. The table presents the mean and standard deviation values for each criterion, including Log-Loss, Precision, Recall, $F_{1}$ score, and AUC calculated based on the testing data of the best final solution. MEGP models are compared for different ensemble selection probabilities: 0\%, 25\%, 50\%, 75\%, and 100\%. Friedman test was performed to assess statistical significance, with p-values adjusted using the Bonferroni method.}
\label{tab:table_friedman_generalization}
\resizebox{0.7\textwidth}{!}{%
\begin{tabular}{ccccccc}
\hline
 &
   &
  \multicolumn{5}{c}{Criteria} \\ \hline
\rowcolor[HTML]{C0C0C0} 
\multicolumn{1}{c|}{\cellcolor[HTML]{C0C0C0}Dataset} &
  \multicolumn{1}{c|}{\cellcolor[HTML]{C0C0C0}Model} &
  Log-Loss &
  Precision &
  Recall &
  $F_{1}$ &
  AUC \\ \hline
\multicolumn{1}{c|}{} &
  \multicolumn{1}{c|}{$\mathrm{MEGP}_{0}$} &
  $0.107 \pm 0.012$ &
  $0.982 \pm 0.011$ &
  $0.982 \pm 0.011$ &
  $0.982 \pm 0.011$ &
  $0.818 \pm 0.115$ \\
\multicolumn{1}{c|}{} &
  \multicolumn{1}{c|}{\cellcolor[HTML]{C0C0C0}$\mathrm{MEGP}_{25}$} &
  \cellcolor[HTML]{C0C0C0}$0.104 \pm 0.017$ &
  \cellcolor[HTML]{C0C0C0}$0.967 \pm 0.04$ &
  \cellcolor[HTML]{C0C0C0}$0.967 \pm 0.04$ &
  \cellcolor[HTML]{C0C0C0}$0.967 \pm 0.04$ &
  \cellcolor[HTML]{C0C0C0}$0.822 \pm 0.118$ \\
\multicolumn{1}{c|}{} &
  \multicolumn{1}{c|}{$\mathrm{MEGP}_{50}$} &
  $0.106 \pm 0.015$ &
  $0.977 \pm 0.024$ &
  $0.977 \pm 0.024$ &
  $0.977 \pm 0.024$ &
  $0.82 \pm 0.138$ \\
\multicolumn{1}{c|}{} &
  \multicolumn{1}{c|}{\cellcolor[HTML]{C0C0C0}$\mathrm{MEGP}_{75}$} &
  \cellcolor[HTML]{C0C0C0}$0.104 \pm 0.016$ &
  \cellcolor[HTML]{C0C0C0}$0.973 \pm 0.019$ &
  \cellcolor[HTML]{C0C0C0}$0.973 \pm 0.019$ &
  \cellcolor[HTML]{C0C0C0}$0.973 \pm 0.019$ &
  \cellcolor[HTML]{C0C0C0}$0.844 \pm 0.101$ \\
\multicolumn{1}{c|}{} &
  \multicolumn{1}{c|}{$\mathrm{MEGP}_{100}$} &
  $0.109 \pm 0.011$ &
  $0.972 \pm 0.026$ &
  $0.972 \pm 0.026$ &
  $0.972 \pm 0.026$ &
  $0.821 \pm 0.153$ \\
\multicolumn{1}{c|}{} &
  \multicolumn{1}{c|}{\cellcolor[HTML]{C0C0C0}BGP} &
  \cellcolor[HTML]{C0C0C0}$0.109 \pm 0.018$ &
  \cellcolor[HTML]{C0C0C0}$0.973 \pm 0.015$ &
  \cellcolor[HTML]{C0C0C0}$0.973 \pm 0.015$ &
  \cellcolor[HTML]{C0C0C0}$0.973 \pm 0.015$ &
  \cellcolor[HTML]{C0C0C0}$0.817 \pm 0.141$ \\
\multicolumn{1}{c|}{} &
  \multicolumn{1}{c|}{Friedman's P-value} &
  $0.52$ &
  $0.16$ &
  $0.16$ &
  $0.16$ &
  $0.80$ \\
\multicolumn{1}{c|}{\multirow{-8}{*}{APSF}} &
  \multicolumn{1}{c|}{\cellcolor[HTML]{C0C0C0}Adjusted P-value} &
  \cellcolor[HTML]{C0C0C0}$1.00$ &
  \cellcolor[HTML]{C0C0C0}$0.81$ &
  \cellcolor[HTML]{C0C0C0}$0.81$ &
  \cellcolor[HTML]{C0C0C0}$0.81$ &
  \cellcolor[HTML]{C0C0C0}$1.00$ \\ \hline
\multicolumn{1}{c|}{\cellcolor[HTML]{C0C0C0}} &
  \multicolumn{1}{c|}{$\mathrm{MEGP}_{0}$} &
  $0.593 \pm 0.032$ &
  $0.721 \pm 0.014$ &
  $0.721 \pm 0.014$ &
  $0.721 \pm 0.014$ &
  $0.911 \pm 0.009$ \\
\rowcolor[HTML]{C0C0C0} 
\multicolumn{1}{c|}{\cellcolor[HTML]{C0C0C0}} &
  \multicolumn{1}{c|}{\cellcolor[HTML]{C0C0C0}$\mathrm{MEGP}_{25}$} &
  $0.582 \pm 0.025$ &
  $0.71 \pm 0.022$ &
  $0.71 \pm 0.022$ &
  $0.71 \pm 0.022$ &
  $0.903 \pm 0.014$ \\
\multicolumn{1}{c|}{\cellcolor[HTML]{C0C0C0}} &
  \multicolumn{1}{c|}{$\mathrm{MEGP}_{50}$} &
  $0.581 \pm 0.03$ &
  $0.703 \pm 0.031$ &
  $0.703 \pm 0.031$ &
  $0.703 \pm 0.031$ &
  $0.902 \pm 0.015$ \\
\rowcolor[HTML]{C0C0C0} 
\multicolumn{1}{c|}{\cellcolor[HTML]{C0C0C0}} &
  \multicolumn{1}{c|}{\cellcolor[HTML]{C0C0C0}$\mathrm{MEGP}_{75}$} &
  $0.574 \pm 0.035$ &
  $0.705 \pm 0.051$ &
  $0.705 \pm 0.051$ &
  $0.705 \pm 0.051$ &
  $0.903 \pm 0.025$ \\
\multicolumn{1}{c|}{\cellcolor[HTML]{C0C0C0}} &
  \multicolumn{1}{c|}{$\mathrm{MEGP}_{100}$} &
  $0.581 \pm 0.034$ &
  $0.711 \pm 0.021$ &
  $0.711 \pm 0.021$ &
  $0.711 \pm 0.021$ &
  $0.905 \pm 0.011$ \\
\rowcolor[HTML]{C0C0C0} 
\multicolumn{1}{c|}{\cellcolor[HTML]{C0C0C0}} &
  \multicolumn{1}{c|}{\cellcolor[HTML]{C0C0C0}BGP} &
  $0.614 \pm 0.037$ &
  $0.713 \pm 0.019$ &
  $0.713 \pm 0.019$ &
  $0.713 \pm 0.019$ &
  $0.906 \pm 0.009$ \\
\multicolumn{1}{c|}{\cellcolor[HTML]{C0C0C0}} &
  \multicolumn{1}{c|}{Friedman's P-value} &
  $\mathbf{1.02e-04}$ &
  $0.43$ &
  $0.43$ &
  $0.43$ &
  $0.22$ \\
\rowcolor[HTML]{C0C0C0} 
\multicolumn{1}{c|}{\multirow{-8}{*}{\cellcolor[HTML]{C0C0C0}ARWPM}} &
  \multicolumn{1}{c|}{\cellcolor[HTML]{C0C0C0}Adjusted P-value} &
  $\mathbf{5.11e-04}$ &
  $1.00$ &
  $1.00$ &
  $1.00$ &
  $1.00$ \\ \hline
\multicolumn{1}{c|}{} &
  \multicolumn{1}{c|}{$\mathrm{MEGP}_{0}$} &
  $0.179 \pm 0.054$ &
  $0.95 \pm 0.02$ &
  $0.95 \pm 0.02$ &
  $0.95 \pm 0.02$ &
  $0.996 \pm 0.003$ \\
\multicolumn{1}{c|}{} &
  \multicolumn{1}{c|}{\cellcolor[HTML]{C0C0C0}$\mathrm{MEGP}_{25}$} &
  \cellcolor[HTML]{C0C0C0}$0.17 \pm 0.053$ &
  \cellcolor[HTML]{C0C0C0}$0.956 \pm 0.019$ &
  \cellcolor[HTML]{C0C0C0}$0.956 \pm 0.019$ &
  \cellcolor[HTML]{C0C0C0}$0.956 \pm 0.019$ &
  \cellcolor[HTML]{C0C0C0}$0.996 \pm 0.003$ \\
\multicolumn{1}{c|}{} &
  \multicolumn{1}{c|}{$\mathrm{MEGP}_{50}$} &
  $0.199 \pm 0.062$ &
  $0.943 \pm 0.019$ &
  $0.943 \pm 0.019$ &
  $0.943 \pm 0.019$ &
  $0.995 \pm 0.003$ \\
\multicolumn{1}{c|}{} &
  \multicolumn{1}{c|}{\cellcolor[HTML]{C0C0C0}$\mathrm{MEGP}_{75}$} &
  \cellcolor[HTML]{C0C0C0}$0.195 \pm 0.064$ &
  \cellcolor[HTML]{C0C0C0}$0.948 \pm 0.019$ &
  \cellcolor[HTML]{C0C0C0}$0.948 \pm 0.019$ &
  \cellcolor[HTML]{C0C0C0}$0.948 \pm 0.019$ &
  \cellcolor[HTML]{C0C0C0}$0.995 \pm 0.004$ \\
\multicolumn{1}{c|}{} &
  \multicolumn{1}{c|}{$\mathrm{MEGP}_{100}$} &
  $0.188 \pm 0.05$ &
  $0.949 \pm 0.02$ &
  $0.949 \pm 0.02$ &
  $0.949 \pm 0.02$ &
  $0.995 \pm 0.004$ \\
\multicolumn{1}{c|}{} &
  \multicolumn{1}{c|}{\cellcolor[HTML]{C0C0C0}BGP} &
  \cellcolor[HTML]{C0C0C0}$0.246 \pm 0.088$ &
  \cellcolor[HTML]{C0C0C0}$0.924 \pm 0.029$ &
  \cellcolor[HTML]{C0C0C0}$0.924 \pm 0.029$ &
  \cellcolor[HTML]{C0C0C0}$0.924 \pm 0.029$ &
  \cellcolor[HTML]{C0C0C0}$0.99 \pm 0.006$ \\
\multicolumn{1}{c|}{} &
  \multicolumn{1}{c|}{Friedman's P-value} &
  $\mathbf{2.53e-03}$ &
  $\mathbf{1.12e-04}$ &
  $\mathbf{1.12e-04}$ &
  $\mathbf{1.12e-04}$ &
  $\mathbf{2.52e-05}$ \\
\multicolumn{1}{c|}{\multirow{-8}{*}{GECR}} &
  \multicolumn{1}{c|}{\cellcolor[HTML]{C0C0C0}Adjusted P-value} &
  \cellcolor[HTML]{C0C0C0}$\mathbf{0.01}$ &
  \cellcolor[HTML]{C0C0C0}$\mathbf{5.61e-04}$ &
  \cellcolor[HTML]{C0C0C0}$\mathbf{5.61e-04}$ &
  \cellcolor[HTML]{C0C0C0}$\mathbf{5.61e-04}$ &
  \cellcolor[HTML]{C0C0C0}$\mathbf{1.26e-04}$ \\ \hline
\multicolumn{1}{c|}{\cellcolor[HTML]{C0C0C0}} &
  \multicolumn{1}{c|}{$\mathrm{MEGP}_{0}$} &
  $0.462 \pm 0.016$ &
  $0.795 \pm 0.01$ &
  $0.795 \pm 0.01$ &
  $0.795 \pm 0.01$ &
  $0.838 \pm 0.016$ \\
\rowcolor[HTML]{C0C0C0} 
\multicolumn{1}{c|}{\cellcolor[HTML]{C0C0C0}} &
  \multicolumn{1}{c|}{\cellcolor[HTML]{C0C0C0}$\mathrm{MEGP}_{25}$} &
  $0.44 \pm 0.034$ &
  $0.778 \pm 0.03$ &
  $0.778 \pm 0.03$ &
  $0.778 \pm 0.03$ &
  $0.819 \pm 0.034$ \\
\multicolumn{1}{c|}{\cellcolor[HTML]{C0C0C0}} &
  \multicolumn{1}{c|}{$\mathrm{MEGP}_{50}$} &
  $0.448 \pm 0.025$ &
  $0.777 \pm 0.036$ &
  $0.777 \pm 0.036$ &
  $0.777 \pm 0.036$ &
  $0.813 \pm 0.048$ \\
\rowcolor[HTML]{C0C0C0} 
\multicolumn{1}{c|}{\cellcolor[HTML]{C0C0C0}} &
  \multicolumn{1}{c|}{\cellcolor[HTML]{C0C0C0}$\mathrm{MEGP}_{75}$} &
  $0.444 \pm 0.034$ &
  $0.776 \pm 0.034$ &
  $0.776 \pm 0.034$ &
  $0.776 \pm 0.034$ &
  $0.808 \pm 0.054$ \\
\multicolumn{1}{c|}{\cellcolor[HTML]{C0C0C0}} &
  \multicolumn{1}{c|}{$\mathrm{MEGP}_{100}$} &
  $0.443 \pm 0.026$ &
  $0.775 \pm 0.032$ &
  $0.775 \pm 0.032$ &
  $0.775 \pm 0.032$ &
  $0.814 \pm 0.04$ \\
\rowcolor[HTML]{C0C0C0} 
\multicolumn{1}{c|}{\cellcolor[HTML]{C0C0C0}} &
  \multicolumn{1}{c|}{\cellcolor[HTML]{C0C0C0}BGP} &
  $0.47 \pm 0.015$ &
  $0.789 \pm 0.011$ &
  $0.789 \pm 0.011$ &
  $0.789 \pm 0.011$ &
  $0.831 \pm 0.016$ \\
\multicolumn{1}{c|}{\cellcolor[HTML]{C0C0C0}} &
  \multicolumn{1}{c|}{Friedman's P-value} &
  $\mathbf{1.24e-04}$ &
  $\mathbf{0.04}$ &
  $\mathbf{0.04}$ &
  $\mathbf{0.04}$ &
  $0.11$ \\
\rowcolor[HTML]{C0C0C0} 
\multicolumn{1}{c|}{\multirow{-8}{*}{\cellcolor[HTML]{C0C0C0}GFE}} &
  \multicolumn{1}{c|}{\cellcolor[HTML]{C0C0C0}Adjusted P-value} &
  $\mathbf{6.21e-04}$ &
  $0.18$ &
  $0.18$ &
  $0.18$ &
  $0.57$ \\ \hline
\multicolumn{1}{c|}{} &
  \multicolumn{1}{c|}{$\mathrm{MEGP}_{0}$} &
  $0.672 \pm 0.068$ &
  $0.782 \pm 0.027$ &
  $0.782 \pm 0.027$ &
  $0.782 \pm 0.027$ &
  $0.953 \pm 0.009$ \\
\multicolumn{1}{c|}{} &
  \multicolumn{1}{c|}{\cellcolor[HTML]{C0C0C0}$\mathrm{MEGP}_{25}$} &
  \cellcolor[HTML]{C0C0C0}$0.641 \pm 0.069$ &
  \cellcolor[HTML]{C0C0C0}$0.788 \pm 0.029$ &
  \cellcolor[HTML]{C0C0C0}$0.788 \pm 0.029$ &
  \cellcolor[HTML]{C0C0C0}$0.788 \pm 0.029$ &
  \cellcolor[HTML]{C0C0C0}$0.955 \pm 0.01$ \\
\multicolumn{1}{c|}{} &
  \multicolumn{1}{c|}{$\mathrm{MEGP}_{50}$} &
  $0.642 \pm 0.068$ &
  $0.786 \pm 0.029$ &
  $0.786 \pm 0.029$ &
  $0.786 \pm 0.029$ &
  $0.955 \pm 0.01$ \\
\multicolumn{1}{c|}{} &
  \multicolumn{1}{c|}{\cellcolor[HTML]{C0C0C0}$\mathrm{MEGP}_{75}$} &
  \cellcolor[HTML]{C0C0C0}$0.638 \pm 0.058$ &
  \cellcolor[HTML]{C0C0C0}$0.789 \pm 0.025$ &
  \cellcolor[HTML]{C0C0C0}$0.789 \pm 0.025$ &
  \cellcolor[HTML]{C0C0C0}$0.789 \pm 0.025$ &
  \cellcolor[HTML]{C0C0C0}$0.955 \pm 0.009$ \\
\multicolumn{1}{c|}{} &
  \multicolumn{1}{c|}{$\mathrm{MEGP}_{100}$} &
  $0.654 \pm 0.052$ &
  $0.784 \pm 0.027$ &
  $0.784 \pm 0.027$ &
  $0.784 \pm 0.027$ &
  $0.952 \pm 0.008$ \\
\multicolumn{1}{c|}{} &
  \multicolumn{1}{c|}{\cellcolor[HTML]{C0C0C0}BGP} &
  \cellcolor[HTML]{C0C0C0}$0.752 \pm 0.107$ &
  \cellcolor[HTML]{C0C0C0}$0.738 \pm 0.045$ &
  \cellcolor[HTML]{C0C0C0}$0.738 \pm 0.045$ &
  \cellcolor[HTML]{C0C0C0}$0.738 \pm 0.045$ &
  \cellcolor[HTML]{C0C0C0}$0.938 \pm 0.016$ \\
\multicolumn{1}{c|}{} &
  \multicolumn{1}{c|}{Friedman's P-value} &
  $\mathbf{8.55e-04}$ &
  $\mathbf{3.85e-04}$ &
  $\mathbf{3.85e-04}$ &
  $\mathbf{3.85e-04}$ &
  $\mathbf{3.20e-04}$ \\
\multicolumn{1}{c|}{\multirow{-8}{*}{GSAD}} &
  \multicolumn{1}{c|}{\cellcolor[HTML]{C0C0C0}Adjusted P-value} &
  \cellcolor[HTML]{C0C0C0}$\mathbf{4.27e-03}$ &
  \cellcolor[HTML]{C0C0C0}$\mathbf{1.93e-03}$ &
  \cellcolor[HTML]{C0C0C0}$\mathbf{1.93e-03}$ &
  \cellcolor[HTML]{C0C0C0}$\mathbf{1.93e-03}$ &
  \cellcolor[HTML]{C0C0C0}$\mathbf{1.60e-03}$ \\ \hline
\multicolumn{1}{c|}{\cellcolor[HTML]{C0C0C0}} &
  \multicolumn{1}{c|}{$\mathrm{MEGP}_{0}$} &
  $0.667 \pm 0.036$ &
  $0.776 \pm 0.017$ &
  $0.776 \pm 0.017$ &
  $0.776 \pm 0.017$ &
  $0.963 \pm 0.005$ \\
\rowcolor[HTML]{C0C0C0} 
\multicolumn{1}{c|}{\cellcolor[HTML]{C0C0C0}} &
  \multicolumn{1}{c|}{\cellcolor[HTML]{C0C0C0}$\mathrm{MEGP}_{25}$} &
  $0.635 \pm 0.043$ &
  $0.783 \pm 0.017$ &
  $0.783 \pm 0.017$ &
  $0.783 \pm 0.017$ &
  $0.962 \pm 0.006$ \\
\multicolumn{1}{c|}{\cellcolor[HTML]{C0C0C0}} &
  \multicolumn{1}{c|}{$\mathrm{MEGP}_{50}$} &
  $0.623 \pm 0.039$ &
  $0.787 \pm 0.016$ &
  $0.787 \pm 0.016$ &
  $0.787 \pm 0.016$ &
  $0.963 \pm 0.006$ \\
\rowcolor[HTML]{C0C0C0} 
\multicolumn{1}{c|}{\cellcolor[HTML]{C0C0C0}} &
  \multicolumn{1}{c|}{\cellcolor[HTML]{C0C0C0}$\mathrm{MEGP}_{75}$} &
  $0.628 \pm 0.036$ &
  $0.783 \pm 0.017$ &
  $0.783 \pm 0.017$ &
  $0.783 \pm 0.017$ &
  $0.963 \pm 0.006$ \\
\multicolumn{1}{c|}{\cellcolor[HTML]{C0C0C0}} &
  \multicolumn{1}{c|}{$\mathrm{MEGP}_{100}$} &
  $0.64 \pm 0.049$ &
  $0.778 \pm 0.019$ &
  $0.778 \pm 0.019$ &
  $0.778 \pm 0.019$ &
  $0.958 \pm 0.012$ \\
\rowcolor[HTML]{C0C0C0} 
\multicolumn{1}{c|}{\cellcolor[HTML]{C0C0C0}} &
  \multicolumn{1}{c|}{\cellcolor[HTML]{C0C0C0}BGP} &
  $0.763 \pm 0.065$ &
  $0.735 \pm 0.036$ &
  $0.735 \pm 0.036$ &
  $0.735 \pm 0.036$ &
  $0.95 \pm 0.01$ \\
\multicolumn{1}{c|}{\cellcolor[HTML]{C0C0C0}} &
  \multicolumn{1}{c|}{Friedman's P-value} &
  $\mathbf{8.85e-12}$ &
  $\mathbf{6.19e-09}$ &
  $\mathbf{6.19e-09}$ &
  $\mathbf{6.19e-09}$ &
  $\mathbf{5.81e-07}$ \\
\rowcolor[HTML]{C0C0C0} 
\multicolumn{1}{c|}{\multirow{-8}{*}{\cellcolor[HTML]{C0C0C0}HAPT}} &
  \multicolumn{1}{c|}{\cellcolor[HTML]{C0C0C0}Adjusted P-value} &
  $\mathbf{4.42e-11}$ &
  $\mathbf{3.09e-08}$ &
  $\mathbf{3.09e-08}$ &
  $\mathbf{3.09e-08}$ &
  $\mathbf{2.90e-06}$ \\ \hline
\multicolumn{1}{c|}{} &
  \multicolumn{1}{c|}{$\mathrm{MEGP}_{0}$} &
  $1.611 \pm 0.048$ &
  $0.514 \pm 0.02$ &
  $0.514 \pm 0.02$ &
  $0.514 \pm 0.02$ &
  $0.956 \pm 0.003$ \\
\multicolumn{1}{c|}{} &
  \multicolumn{1}{c|}{\cellcolor[HTML]{C0C0C0}$\mathrm{MEGP}_{25}$} &
  \cellcolor[HTML]{C0C0C0}$1.601 \pm 0.048$ &
  \cellcolor[HTML]{C0C0C0}$0.506 \pm 0.021$ &
  \cellcolor[HTML]{C0C0C0}$0.506 \pm 0.021$ &
  \cellcolor[HTML]{C0C0C0}$0.506 \pm 0.021$ &
  \cellcolor[HTML]{C0C0C0}$0.954 \pm 0.003$ \\
\multicolumn{1}{c|}{} &
  \multicolumn{1}{c|}{$\mathrm{MEGP}_{50}$} &
  $1.586 \pm 0.056$ &
  $0.504 \pm 0.021$ &
  $0.504 \pm 0.021$ &
  $0.504 \pm 0.021$ &
  $0.954 \pm 0.004$ \\
\multicolumn{1}{c|}{} &
  \multicolumn{1}{c|}{\cellcolor[HTML]{C0C0C0}$\mathrm{MEGP}_{75}$} &
  \cellcolor[HTML]{C0C0C0}$1.593 \pm 0.042$ &
  \cellcolor[HTML]{C0C0C0}$0.503 \pm 0.021$ &
  \cellcolor[HTML]{C0C0C0}$0.503 \pm 0.021$ &
  \cellcolor[HTML]{C0C0C0}$0.503 \pm 0.021$ &
  \cellcolor[HTML]{C0C0C0}$0.954 \pm 0.004$ \\
\multicolumn{1}{c|}{} &
  \multicolumn{1}{c|}{$\mathrm{MEGP}_{100}$} &
  $1.617 \pm 0.058$ &
  $0.494 \pm 0.021$ &
  $0.494 \pm 0.021$ &
  $0.494 \pm 0.021$ &
  $0.95 \pm 0.004$ \\
\multicolumn{1}{c|}{} &
  \multicolumn{1}{c|}{\cellcolor[HTML]{C0C0C0}BGP} &
  \cellcolor[HTML]{C0C0C0}$1.703 \pm 0.05$ &
  \cellcolor[HTML]{C0C0C0}$0.429 \pm 0.017$ &
  \cellcolor[HTML]{C0C0C0}$0.429 \pm 0.017$ &
  \cellcolor[HTML]{C0C0C0}$0.429 \pm 0.017$ &
  \cellcolor[HTML]{C0C0C0}$0.941 \pm 0.006$ \\
\multicolumn{1}{c|}{} &
  \multicolumn{1}{c|}{Friedman's P-value} &
  $\mathbf{1.26e-09}$ &
  $\mathbf{2.34e-14}$ &
  $\mathbf{2.34e-14}$ &
  $\mathbf{2.34e-14}$ &
  $\mathbf{3.54e-15}$ \\
\multicolumn{1}{c|}{\multirow{-8}{*}{ISOLET}} &
  \multicolumn{1}{c|}{\cellcolor[HTML]{C0C0C0}Adjusted P-value} &
  \cellcolor[HTML]{C0C0C0}$\mathbf{6.28e-09}$ &
  \cellcolor[HTML]{C0C0C0}$\mathbf{1.17e-13}$ &
  \cellcolor[HTML]{C0C0C0}$\mathbf{1.17e-13}$ &
  \cellcolor[HTML]{C0C0C0}$\mathbf{1.17e-13}$ &
  \cellcolor[HTML]{C0C0C0}$\mathbf{1.77e-14}$ \\ \hline
\multicolumn{1}{c|}{\cellcolor[HTML]{C0C0C0}} &
  \multicolumn{1}{c|}{$\mathrm{MEGP}_{0}$} &
  $0.443 \pm 0.044$ &
  $0.804 \pm 0.024$ &
  $0.804 \pm 0.024$ &
  $0.804 \pm 0.024$ &
  $0.818 \pm 0.04$ \\
\rowcolor[HTML]{C0C0C0} 
\multicolumn{1}{c|}{\cellcolor[HTML]{C0C0C0}} &
  \multicolumn{1}{c|}{\cellcolor[HTML]{C0C0C0}$\mathrm{MEGP}_{25}$} &
  $0.436 \pm 0.047$ &
  $0.8 \pm 0.023$ &
  $0.8 \pm 0.023$ &
  $0.8 \pm 0.023$ &
  $0.818 \pm 0.034$ \\
\multicolumn{1}{c|}{\cellcolor[HTML]{C0C0C0}} &
  \multicolumn{1}{c|}{$\mathrm{MEGP}_{50}$} &
  $0.427 \pm 0.06$ &
  $0.796 \pm 0.029$ &
  $0.796 \pm 0.029$ &
  $0.796 \pm 0.029$ &
  $0.818 \pm 0.027$ \\
\rowcolor[HTML]{C0C0C0} 
\multicolumn{1}{c|}{\cellcolor[HTML]{C0C0C0}} &
  \multicolumn{1}{c|}{\cellcolor[HTML]{C0C0C0}$\mathrm{MEGP}_{75}$} &
  $0.439 \pm 0.041$ &
  $0.798 \pm 0.031$ &
  $0.798 \pm 0.031$ &
  $0.798 \pm 0.031$ &
  $0.813 \pm 0.031$ \\
\multicolumn{1}{c|}{\cellcolor[HTML]{C0C0C0}} &
  \multicolumn{1}{c|}{$\mathrm{MEGP}_{100}$} &
  $0.442 \pm 0.046$ &
  $0.796 \pm 0.031$ &
  $0.796 \pm 0.031$ &
  $0.796 \pm 0.031$ &
  $0.82 \pm 0.037$ \\
\rowcolor[HTML]{C0C0C0} 
\multicolumn{1}{c|}{\cellcolor[HTML]{C0C0C0}} &
  \multicolumn{1}{c|}{\cellcolor[HTML]{C0C0C0}BGP} &
  $0.46 \pm 0.053$ &
  $0.792 \pm 0.025$ &
  $0.792 \pm 0.025$ &
  $0.792 \pm 0.025$ &
  $0.813 \pm 0.042$ \\
\multicolumn{1}{c|}{\cellcolor[HTML]{C0C0C0}} &
  \multicolumn{1}{c|}{Friedman's P-value} &
  $0.23$ &
  $0.35$ &
  $0.35$ &
  $0.35$ &
  $0.97$ \\
\rowcolor[HTML]{C0C0C0} 
\multicolumn{1}{c|}{\multirow{-8}{*}{\cellcolor[HTML]{C0C0C0}PD}} &
  \multicolumn{1}{c|}{\cellcolor[HTML]{C0C0C0}Adjusted P-value} &
  $1.00$ &
  $1.00$ &
  $1.00$ &
  $1.00$ &
  $1.00$ \\ \hline
\end{tabular}%
}
\end{table*}

%% file: box_loss.tex
\begin{figure*}[ht] 
\centering
\includegraphics[width=\textwidth]{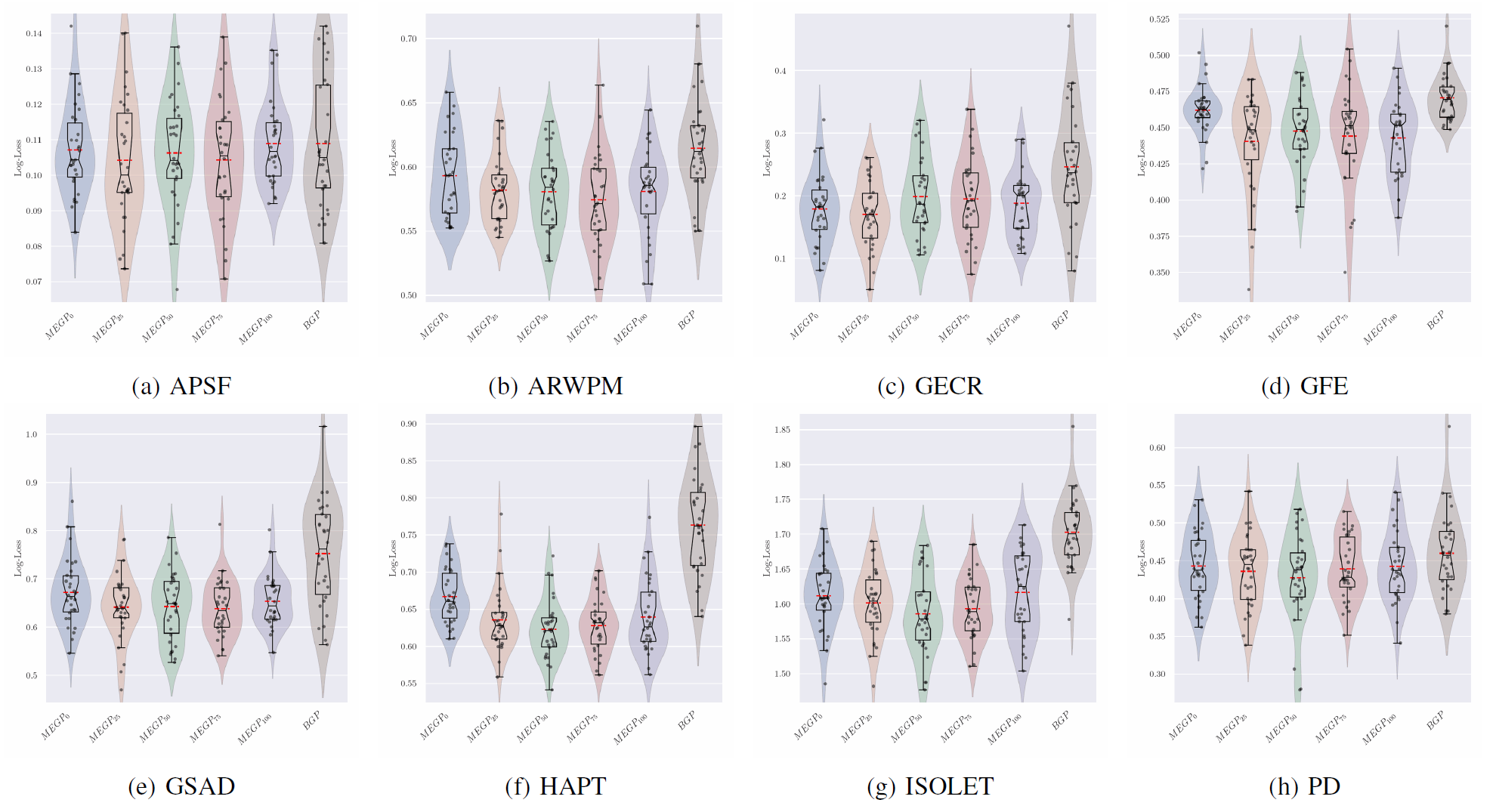}\label{fig:apsf_box_loss}%

\caption[The distribution of Log-Loss across BGP and MEGP models over 30 runs.]{Raincloud plots showing the distribution of Log-Loss score across 30 runs for the best model obtained from BGP and MEGP models with different ensemble selection probabilities (0\%, 25\%, 50\%, 75\%, 100\%). }

\label{fig:box_loss}
\end{figure*}

%% file: con_loss.tex
\begin{figure*}[ht] 
\centering
\includegraphics[width=\textwidth]{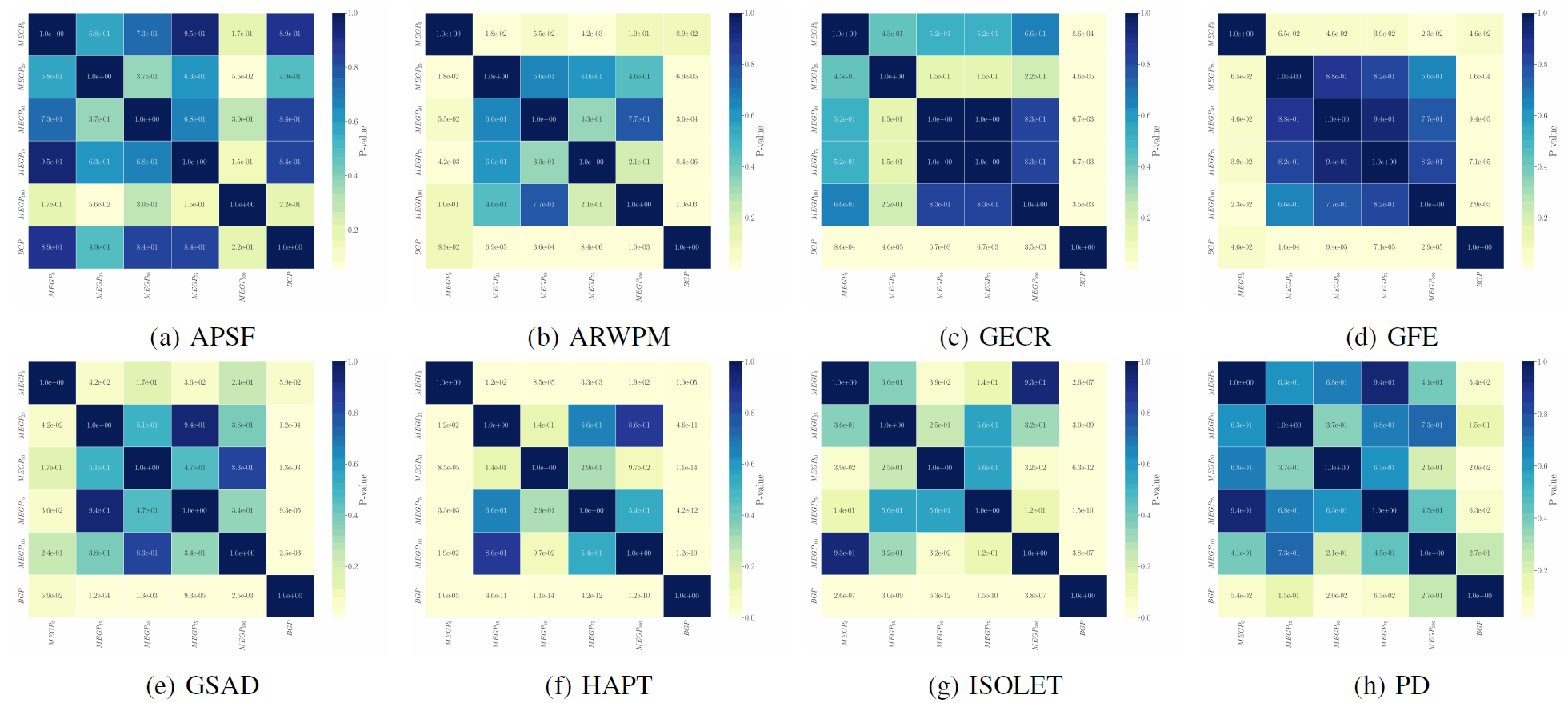}%

\caption[Heatmap of adjusted pairwise significance from Conover post-hoc test for Log-Loss.]{Heatmap of adjusted p-values from the pairwise Conover post-hoc test for Log-Loss, corrected using the Benjamini-Hochberg method. The heatmap highlights the statistical significance of pairwise comparisons between BGP and MEGP models with varying ensemble selection probabilities (0\%, 25\%, 50\%, 75\%, 100\%).}

\label{fig:con_loss}
\end{figure*}

%% file: cliff_loss.tex
\begin{figure*}[htbp] 
\centering
\includegraphics[width=\textwidth]{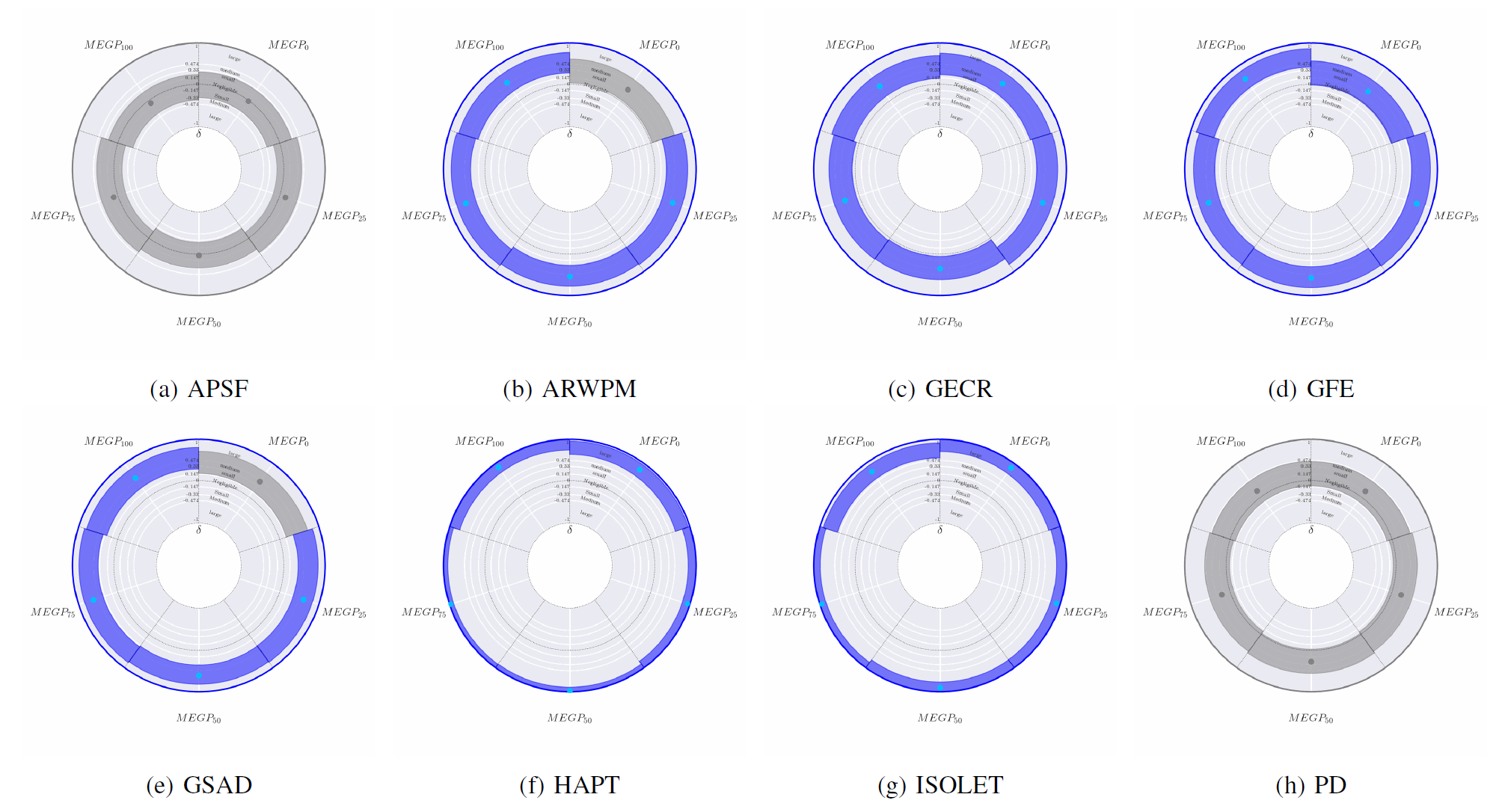}%

\caption[The Cliff's $\delta$ effect size measure and its 95\% confidence intervals for Log-Loss obtained from the best model of the 30 BGP and MEGP runs.]{Effect size analysis of Log-Loss across 30 runs for BGP and MEGP models using Cliff's $\delta$. Each point represents the actual Log-Loss value obtained, with segments denoting 95\% confidence intervals based on 10,000 bootstrap resamplings. The outer ring color visualizes statistical significance: grey illustrates no significant difference (adjusted Friedman's P-value $>0.05$), while color indicates significant differences; blue indicates that at least one MEGP configuration outperforms BGP (adjusted Conover's p-value $< 0.05$, Cliff's $\delta > 0$), and red signifies that all MEGP configurations underperform relative to BGP (adjusted Conover's p-value $< 0.05$, Cliff's $\delta < 0$). Segment colors show performance differences against BGP: grey for no significant difference (adjusted Conover's p-value $> 0.05$), blue for better performance (Cliff's $\delta > 0$), and red for worse performance (Cliff's $\delta < 0$).}

\label{fig:cliff_loss}
\end{figure*}

%% file: wtl_loss.tex
\begin{table*}
\centering
\caption[The results of Friedman and Conover tests and Cliff's $\delta$ analysis for the Log-Loss obtained from MEGP and BGP runs.]{Statistical comparison of Log-Loss results for testing data obtained from MEGP and BGP Runs. W, T, and L denote win, tie, and loss based on adjusted Friedman and Conover's p-values. Effect sizes are calculated using Cliff's Delta method and are categorized as negligible, small, medium, or large.}
\label{tab:wtl_loss}
\begin{tabular}{cccccc}
\hline
\multicolumn{6}{c}{Log-Loss} \\
\hline
Dataset & $MEGP_{0}$ & $MEGP_{25}$ & $MEGP_{50}$ & $MEGP_{75}$ & $MEGP_{100}$ \\
\hline
APSF & T (negligible) & T (negligible) & T (negligible) & T (negligible) & T (negligible) \\
ARWPM & T (small) & W (large) & W (large) & W (large) & W (large) \\
GECR & W (large) & W (large) & W (medium) & W (medium) & W (medium) \\
GFE & W (small) & W (large) & W (large) & W (large) & W (large) \\
GSAD & T (medium) & W (large) & W (large) & W (large) & W (large) \\
HAPT & W (large) & W (large) & W (large) & W (large) & W (large) \\
ISOLET & W (large) & W (large) & W (large) & W (large) & W (large) \\
PD & T (small) & T (small) & T (small) & T (small) & T (small) \\
\hline
W - T - L & 4 - 4 - 0 & 6 - 2 - 0 & 6 - 2 - 0 & 6 - 2 - 0 & 6 - 2 - 0 \\
\hline
\end{tabular}
\end{table*}

%% file: box_precision.tex
\begin{figure*}[ht] 
\centering
\includegraphics[width=\textwidth]{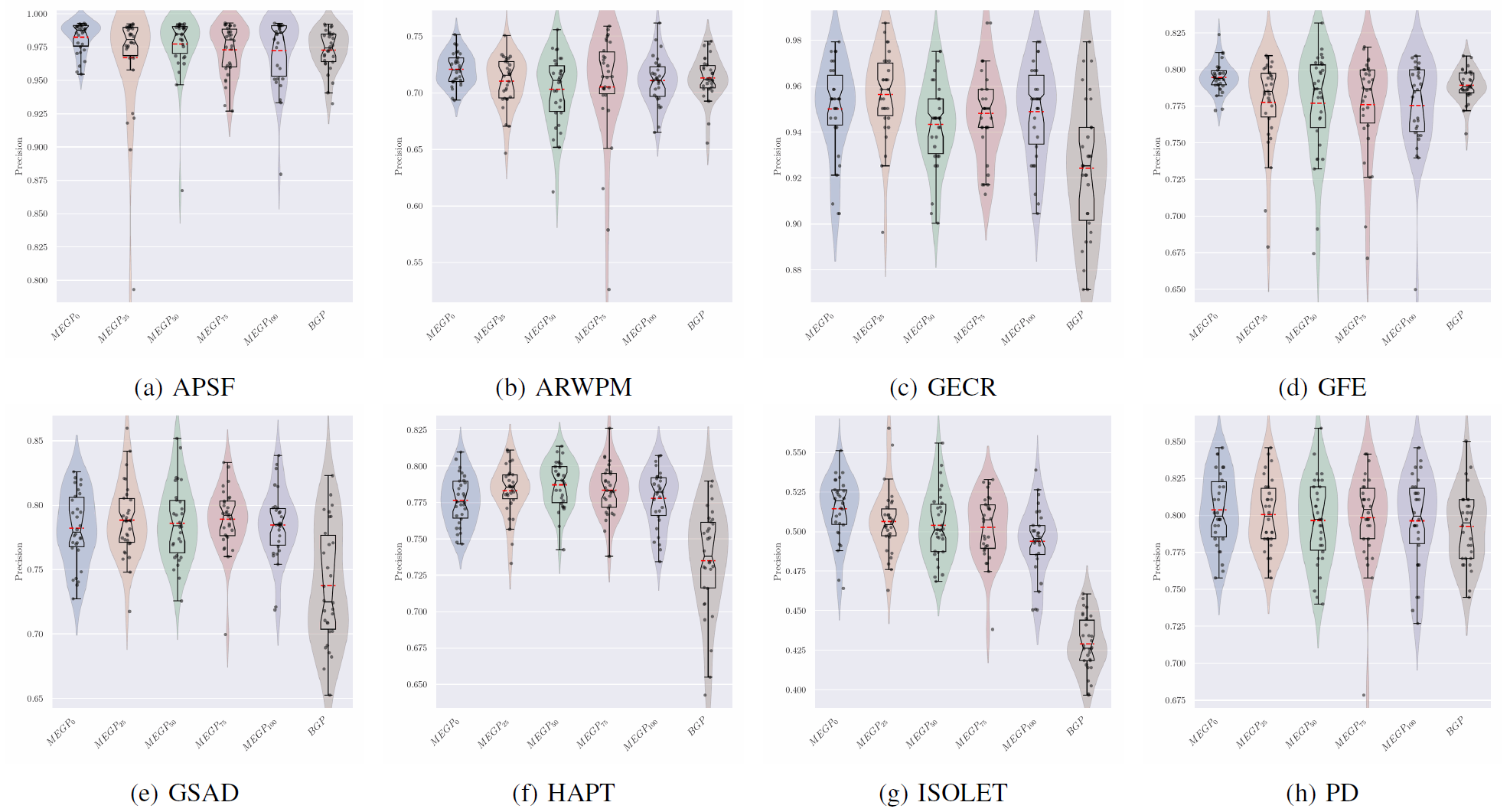}
\caption[The distribution of Precision across BGP and MEGP models over 30 runs.]{Raincloud plots showing the distribution of Precision across 30 runs for the best model obtained from BGP and MEGP models with different ensemble selection probabilities (0\%, 25\%, 50\%, 75\%, 100\%). }

\label{fig:box_precision}
\end{figure*}

%% file: con_precision.tex
\begin{figure*}[ht] 
\centering
\includegraphics[width=\textwidth]{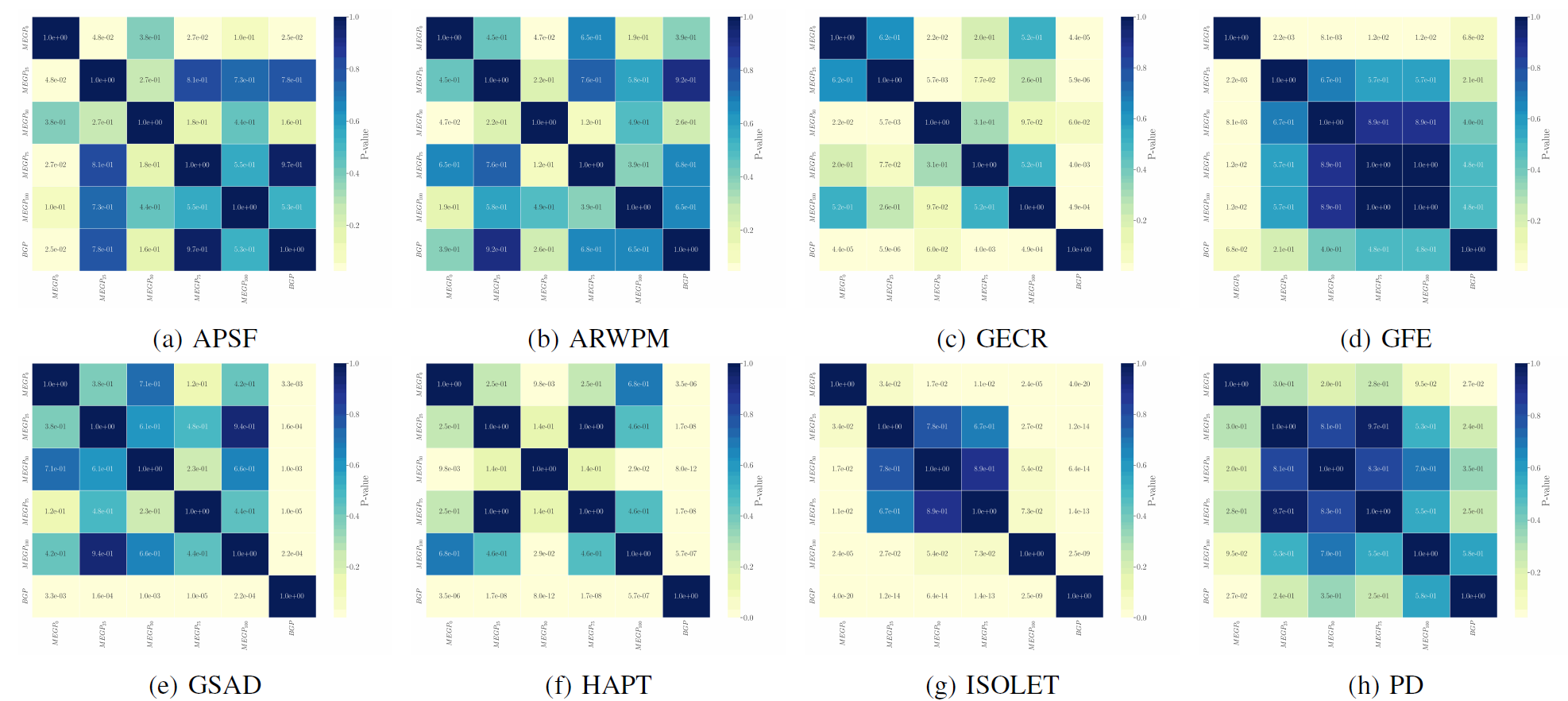}
\caption[Heatmap of adjusted pairwise significance from Conover post-hoc test for Precision.]{Heatmap of adjusted p-values from the pairwise Conover post-hoc test for Precision, corrected using the Benjamini-Hochberg method. The heatmap highlights the statistical significance of pairwise comparisons between BGP and MEGP models with varying ensemble selection probabilities (0\%, 25\%, 50\%, 75\%, 100\%).}

\label{fig:con_precision}
\end{figure*}

%% file: cliff_precision.tex
\begin{figure*}[htbp] 
\centering
\includegraphics[width=\textwidth]{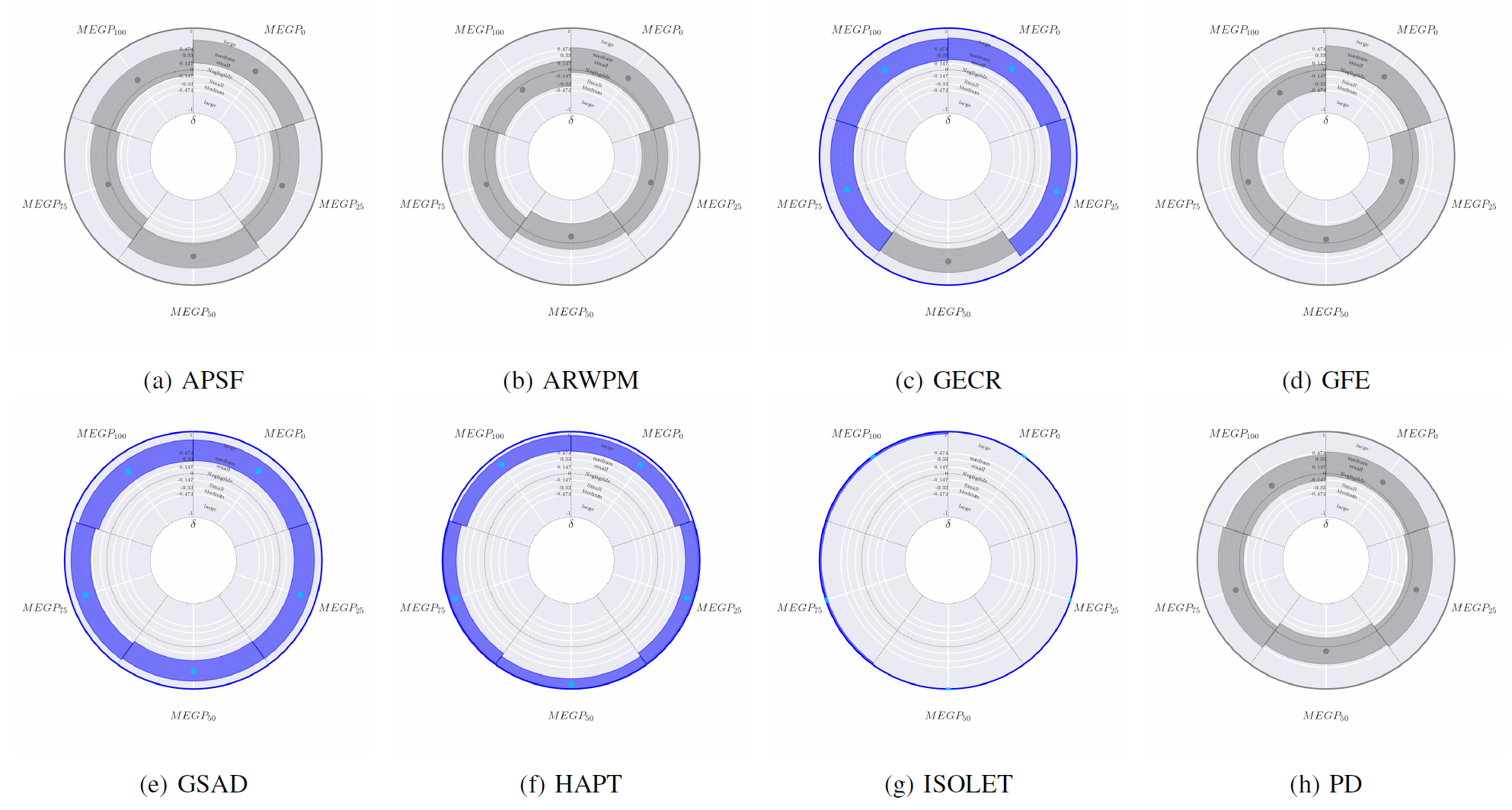}
\caption[The Cliff's $\delta$ effect size measure and its 95\% confidence intervals for Precision obtained from the best model of the 30 BGP and MEGP runs.]{Effect size analysis of Precision across 30 runs for BGP and MEGP models using Cliff's $\delta$. Each point represents the actual Precision value obtained, with segments denoting 95\% confidence intervals based on 10,000 bootstrap resamplings. The outer ring color visualizes statistical significance: grey illustrates no significant difference (adjusted Friedman's P-value $>0.05$), while color indicates significant differences; blue indicates that at least one MEGP configuration outperforms BGP (adjusted Conover's p-value $< 0.05$, Cliff's $\delta > 0$), and red signifies that all MEGP configurations underperform relative to BGP (adjusted Conover's p-value $< 0.05$, Cliff's $\delta < 0$). Segment colors show performance differences against BGP: grey for no significant difference (adjusted Conover's p-value $> 0.05$), blue for better performance (Cliff's $\delta > 0$), and red for worse performance (Cliff's $\delta < 0$).}

\label{fig:cliff_precision}
\end{figure*}

%% file: wtl_precision.tex
\begin{table*}
\centering
\caption[The results of Friedman and Conover tests and Cliff's $\delta$ analysis for the Precision obtained from MEGP and BGP runs.]{Statistical comparison of Precision results for testing data obtained from MEGP and BGP Runs. W, T, and L denote win, tie, and loss based on adjusted Friedman and Conover's p-values. Effect sizes are calculated using Cliff's Delta method and are categorized as negligible, small, medium, or large.}
\label{tab:wtl_precision}
\begin{tabular}{cccccc}
\hline
\multicolumn{6}{c}{Precision} \\
\hline
Dataset & $MEGP_{0}$ & $MEGP_{25}$ & $MEGP_{50}$ & $MEGP_{75}$ & $MEGP_{100}$ \\
\hline
APSF & T (medium) & T (small) & T (small) & T (negligible) & T (small) \\
ARWPM & T (small) & T (negligible) & T (small) & T (negligible) & T (negligible) \\
GECR & W (large) & W (large) & T (medium) & W (medium) & W (large) \\
GFE & T (small) & T (small) & T (negligible) & T (negligible) & T (small) \\
GSAD & W (large) & W (large) & W (large) & W (large) & W (large) \\
HAPT & W (large) & W (large) & W (large) & W (large) & W (large) \\
ISOLET & W (large) & W (large) & W (large) & W (large) & W (large) \\
PD & T (small) & T (small) & T (negligible) & T (small) & T (negligible) \\
\hline
W - T - L & 4 - 4 - 0 & 4 - 4 - 0 & 3 - 5 - 0 & 4 - 4 - 0 & 4 - 4 - 0 \\
\hline
\end{tabular}
\end{table*}

%% file: box_recall.tex
\begin{figure*}[ht] 
\centering
\includegraphics[width=\textwidth]{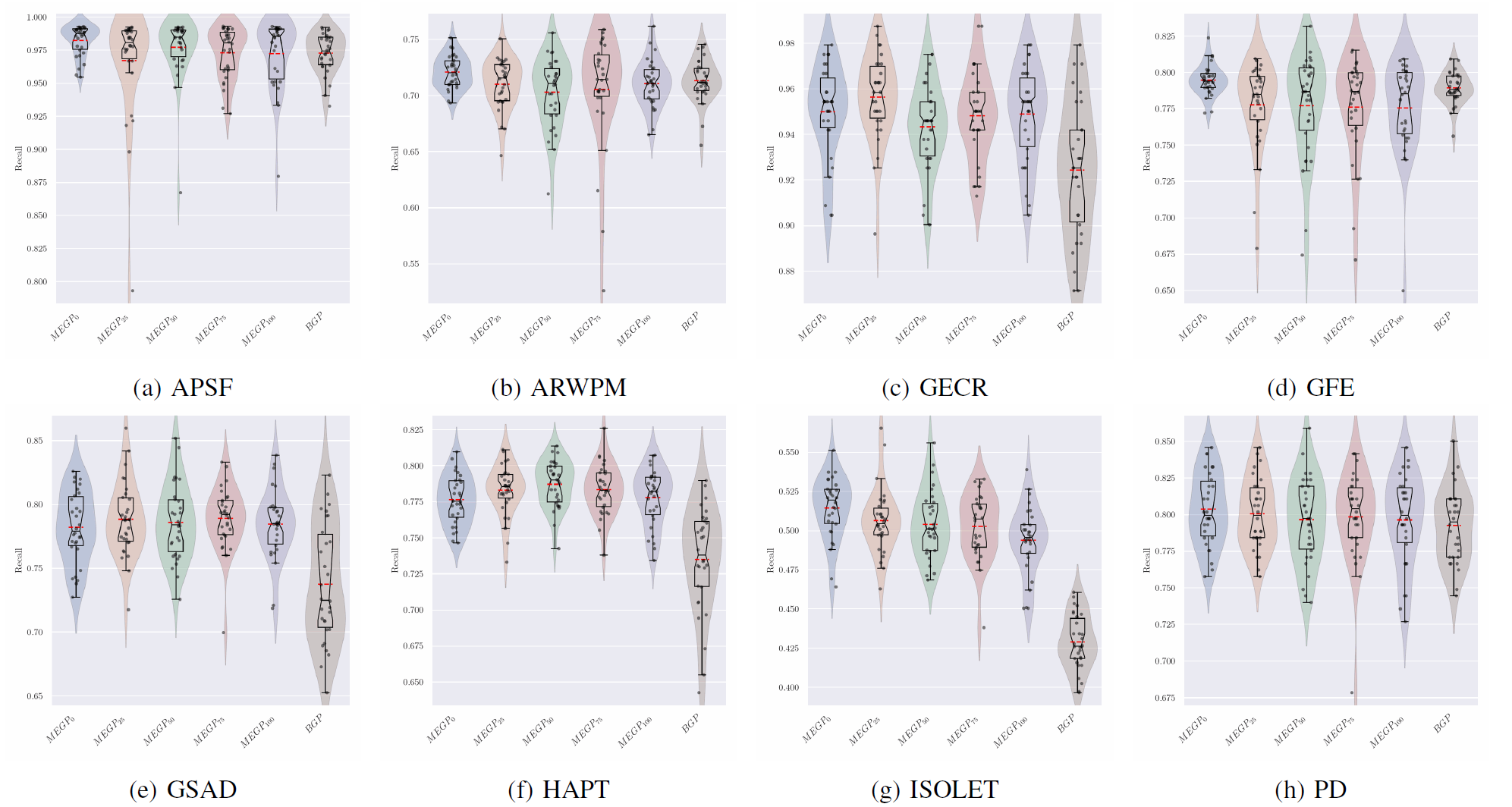}
\caption[The distribution of Recall across BGP and MEGP models over 30 runs.]{Raincloud plots showing the distribution of Recall across 30 runs for the best model obtained from BGP and MEGP models with different ensemble selection probabilities (0\%, 25\%, 50\%, 75\%, 100\%). }

\label{fig:box_recall}
\end{figure*}

%% file: con_recall.tex
\begin{figure*}[ht] 
\centering
\includegraphics[width=\textwidth]{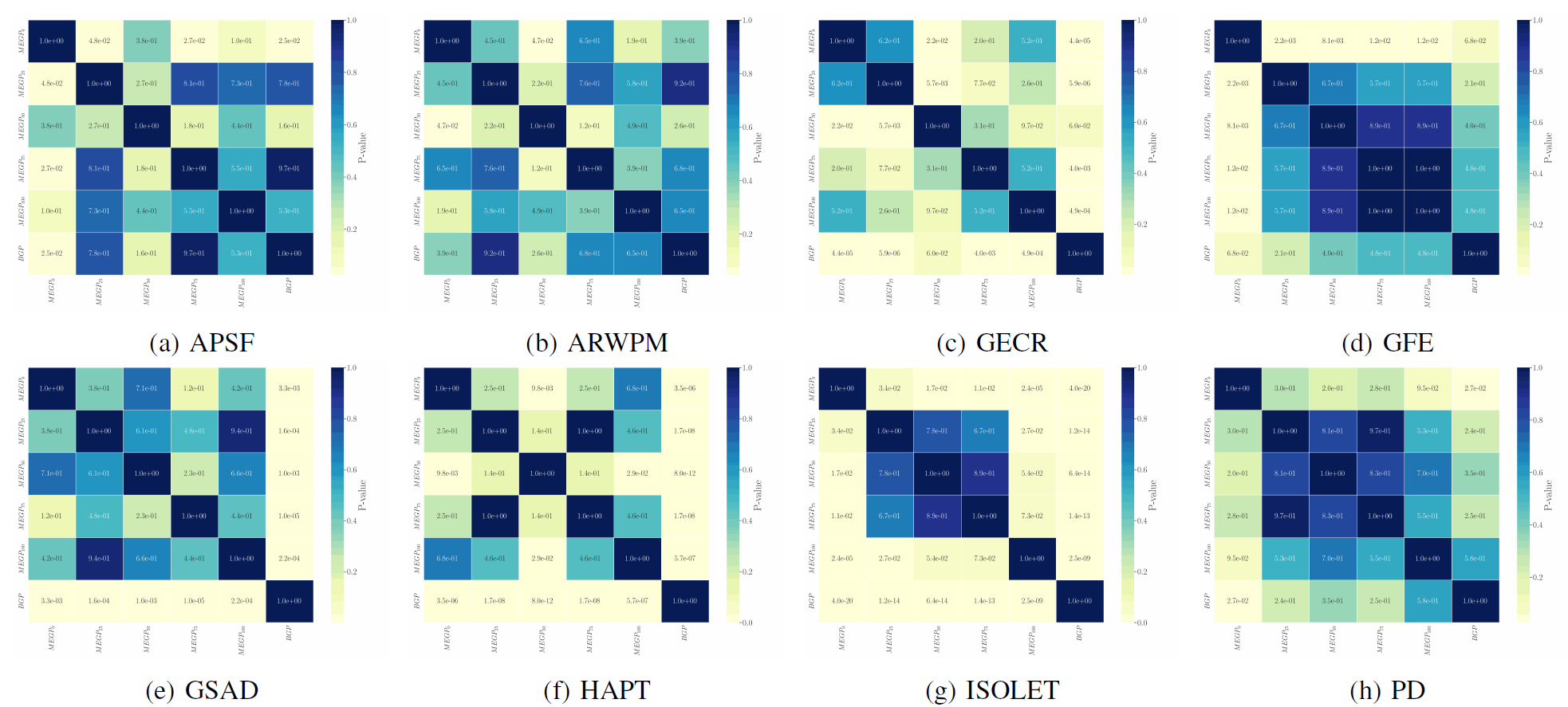}

\caption[Heatmap of adjusted pairwise significance from Conover post-hoc test for Recall.]{Heatmap of adjusted p-values from the pairwise Conover post-hoc test for Recall, corrected using the Benjamini-Hochberg method. The heatmap highlights the statistical significance of pairwise comparisons between BGP and MEGP models with varying ensemble selection probabilities (0\%, 25\%, 50\%, 75\%, 100\%).}

\label{fig:con_recall}
\end{figure*}

%% file: cliff_recall.tex
\begin{figure*}[htbp] 
\centering
\includegraphics[width=\textwidth]{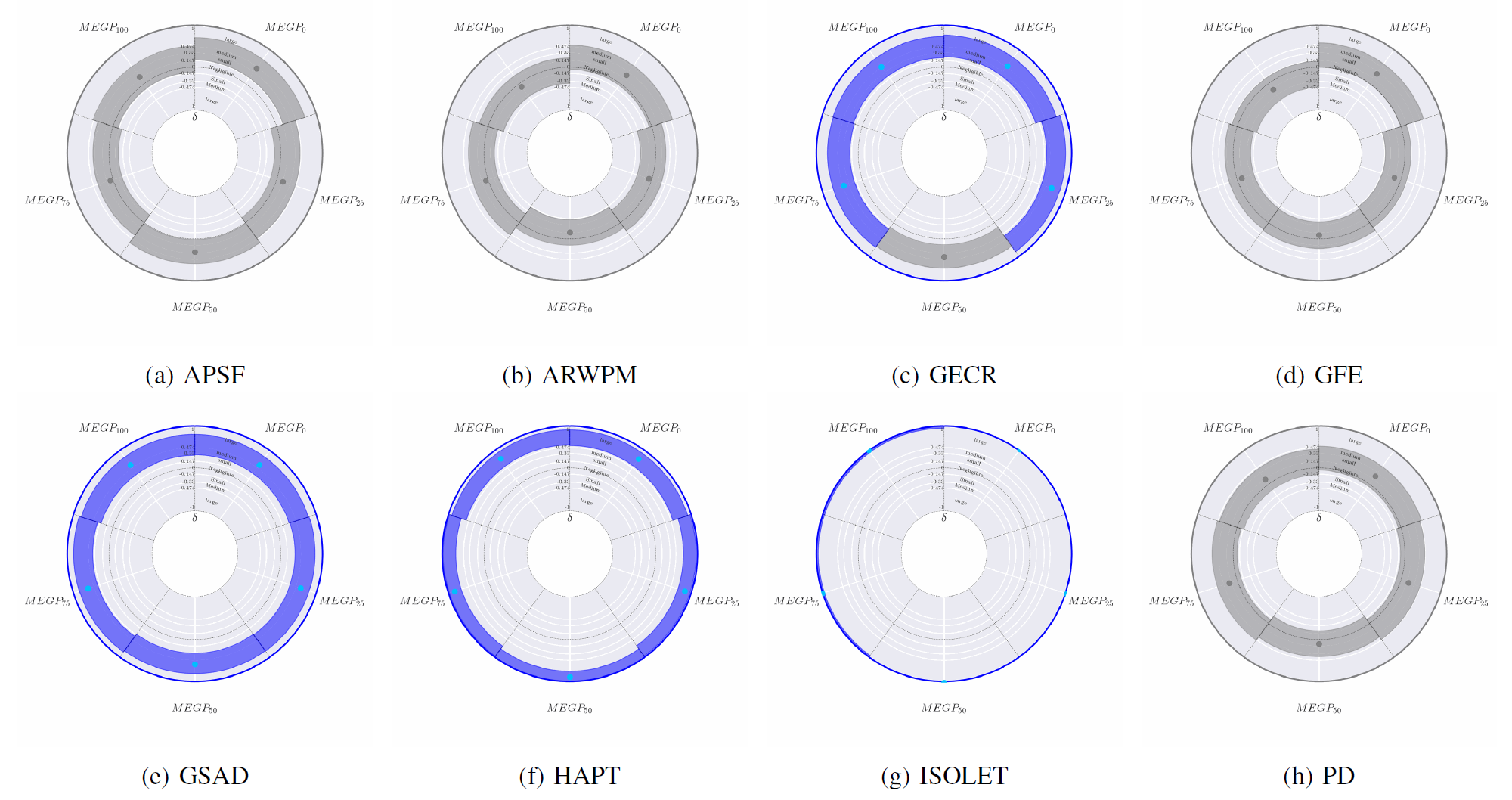}
\caption[The Cliff's $\delta$ effect size measure and its 95\% confidence intervals for Recall obtained from the best model of the 30 BGP and MEGP runs.]{Effect size analysis of Recall across 30 runs for BGP and MEGP models using Cliff's $\delta$. Each point represents the actual Recall value obtained, with segments denoting 95\% confidence intervals based on 10,000 bootstrap resamplings. The outer ring color visualizes statistical significance: grey illustrates no significant difference (adjusted Friedman's P-value $>0.05$), while color indicates significant differences; blue indicates that at least one MEGP configuration outperforms BGP (adjusted Conover's p-value $< 0.05$, Cliff's $\delta > 0$), and red signifies that all MEGP configurations underperform relative to BGP (adjusted Conover's p-value $< 0.05$, Cliff's $\delta < 0$). Segment colors show performance differences against BGP: grey for no significant difference (adjusted Conover's p-value $> 0.05$), blue for better performance (Cliff's $\delta > 0$), and red for worse performance (Cliff's $\delta < 0$).}

\label{fig:cliff_recall}
\end{figure*}

%% file: wtl_recall.tex
\begin{table*}
\centering
\caption[The results of Friedman and Conover tests and Cliff's $\delta$ analysis for the Recall obtained from MEGP and BGP runs.]{Statistical comparison of Recall results for testing data obtained from MEGP and BGP Runs. W, T, and L denote win, tie, and loss based on adjusted Friedman and Conover's p-values. Effect sizes are calculated using Cliff's Delta method and are categorized as negligible, small, medium, or large.}
\label{tab:wtl_recall}
\begin{tabular}{cccccc}
\hline
\multicolumn{6}{c}{Recall} \\
\hline
Dataset & $MEGP_{0}$ & $MEGP_{25}$ & $MEGP_{50}$ & $MEGP_{75}$ & $MEGP_{100}$ \\
\hline
APSF & T (medium) & T (small) & T (small) & T (negligible) & T (small) \\
ARWPM & T (small) & T (negligible) & T (small) & T (negligible) & T (negligible) \\
GECR & W (large) & W (large) & T (medium) & W (medium) & W (large) \\
GFE & T (small) & T (small) & T (negligible) & T (negligible) & T (small) \\
GSAD & W (large) & W (large) & W (large) & W (large) & W (large) \\
HAPT & W (large) & W (large) & W (large) & W (large) & W (large) \\
ISOLET & W (large) & W (large) & W (large) & W (large) & W (large) \\
PD & T (small) & T (small) & T (negligible) & T (small) & T (negligible) \\
\hline
W - T - L & 4 - 4 - 0 & 4 - 4 - 0 & 3 - 5 - 0 & 4 - 4 - 0 & 4 - 4 - 0 \\
\hline
\end{tabular}
\end{table*}

%% file: box_f1.tex
\begin{figure*}[ht] 
\centering
\includegraphics[width=\textwidth]{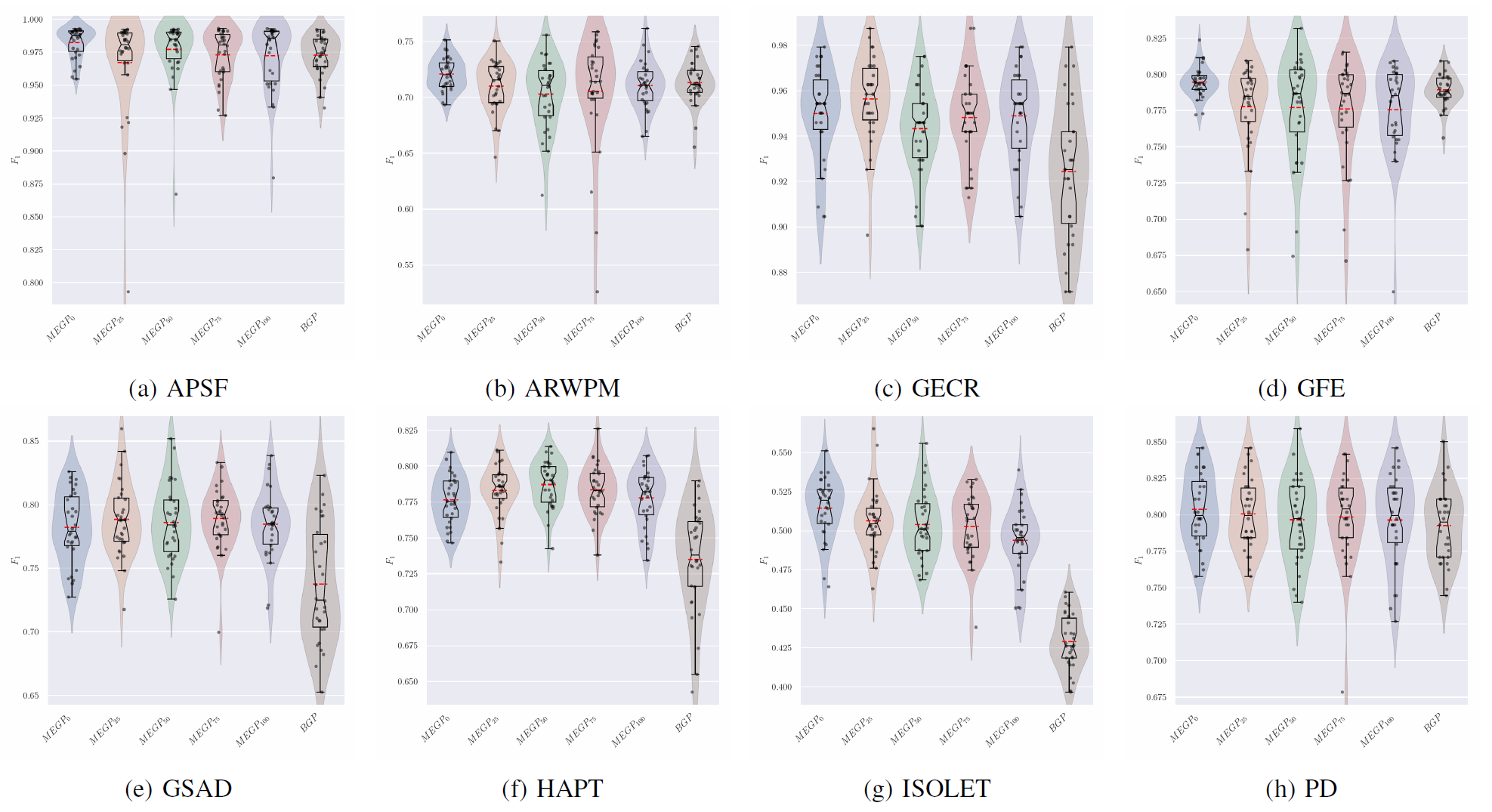}
\caption[The distribution of $F_{1}$ score across BGP and MEGP models over 30 runs.]{Raincloud plots showing the distribution of $F_{1}$ score across 30 runs for the best model obtained from BGP and MEGP models with different ensemble selection probabilities (0\%, 25\%, 50\%, 75\%, 100\%). }

\label{fig:box_f1}
\end{figure*}

%% file: con_f1.tex
\begin{figure*}[ht] 
\centering
\includegraphics[width=\textwidth]{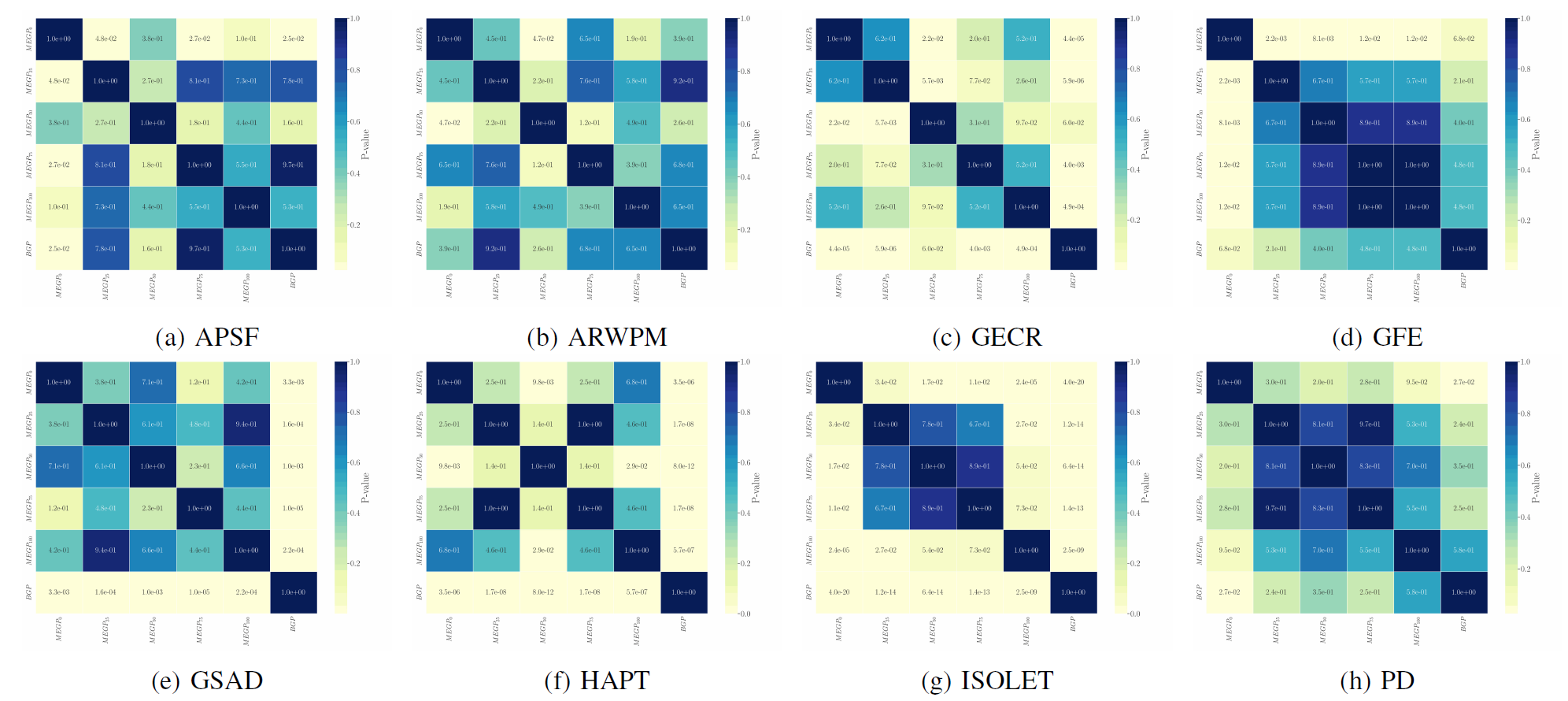}
\caption[Heatmap of adjusted pairwise significance from Conover post-hoc test for $F_{1}$ score.]{Heatmap of adjusted p-values from the pairwise Conover post-hoc test for $F_{1}$ score, corrected using the Benjamini-Hochberg method. The heatmap highlights the statistical significance of pairwise comparisons between BGP and MEGP models with varying ensemble selection probabilities (0\%, 25\%, 50\%, 75\%, 100\%).}

\label{fig:con_f1}
\end{figure*}

%% file: cliff_f1.tex
\begin{figure*}[htbp] 
\centering
\includegraphics[width=\textwidth]{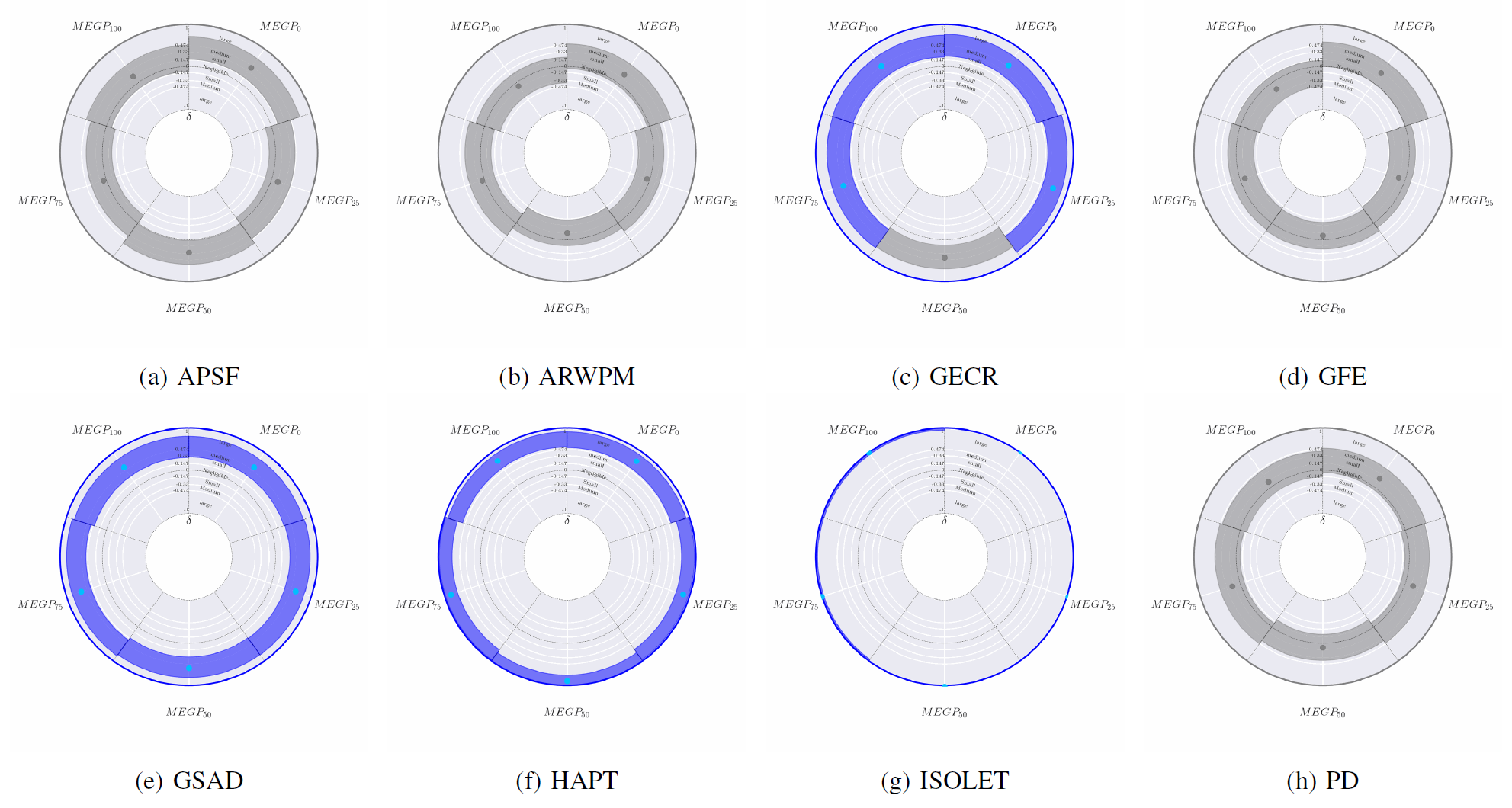}
\caption[The Cliff's $\delta$ effect size measure and its 95\% confidence intervals for $F_{1}$ score obtained from the best model of the 30 BGP and MEGP runs.]{Effect size analysis of $F_{1}$ score across 30 runs for BGP and MEGP models using Cliff's $\delta$. Each point represents the actual $F_{1}$ score value obtained, with segments denoting 95\% confidence intervals based on 10,000 bootstrap resamplings. The outer ring color visualizes statistical significance: grey illustrates no significant difference (adjusted Friedman's P-value $>0.05$), while color indicates significant differences; blue indicates that at least one MEGP configuration outperforms BGP (adjusted Conover's p-value $< 0.05$, Cliff's $\delta > 0$), and red signifies that all MEGP configurations underperform relative to BGP (adjusted Conover's p-value $< 0.05$, Cliff's $\delta < 0$). Segment colors show performance differences against BGP: grey for no significant difference (adjusted Conover's p-value $> 0.05$), blue for better performance (Cliff's $\delta > 0$), and red for worse performance (Cliff's $\delta < 0$).}

\label{fig:cliff_f1}
\end{figure*}

%% file: wtl_f1.tex
\begin{table*}
\centering
\caption[The results of Friedman and Conover tests and Cliff's $\delta$ analysis for the $F_{1}$ obtained from MEGP and BGP runs.]{Statistical comparison of $F_{1}$ results for testing data obtained from MEGP and BGP Runs. W, T, and L denote win, tie, and loss based on adjusted Friedman and Conover's p-values. Effect sizes are calculated using Cliff's Delta method and are categorized as negligible, small, medium, or large.}
\label{tab:wtl_f1}
\begin{tabular}{cccccc}
\hline
\multicolumn{6}{c}{$F_{1}$} \\
\hline
Dataset & $MEGP_{0}$ & $MEGP_{25}$ & $MEGP_{50}$ & $MEGP_{75}$ & $MEGP_{100}$ \\
\hline
APSF & T (medium) & T (small) & T (small) & T (negligible) & T (small) \\
ARWPM & T (small) & T (negligible) & T (small) & T (negligible) & T (negligible) \\
GECR & W (large) & W (large) & T (medium) & W (medium) & W (large) \\
GFE & T (small) & T (small) & T (negligible) & T (negligible) & T (small) \\
GSAD & W (large) & W (large) & W (large) & W (large) & W (large) \\
HAPT & W (large) & W (large) & W (large) & W (large) & W (large) \\
ISOLET & W (large) & W (large) & W (large) & W (large) & W (large) \\
PD & T (small) & T (small) & T (negligible) & T (small) & T (negligible) \\
\hline
W - T - L & 4 - 4 - 0 & 4 - 4 - 0 & 3 - 5 - 0 & 4 - 4 - 0 & 4 - 4 - 0 \\
\hline
\end{tabular}
\end{table*}

%% file: box_auc.tex
\begin{figure*}[ht] 
\centering
\includegraphics[width=\textwidth]{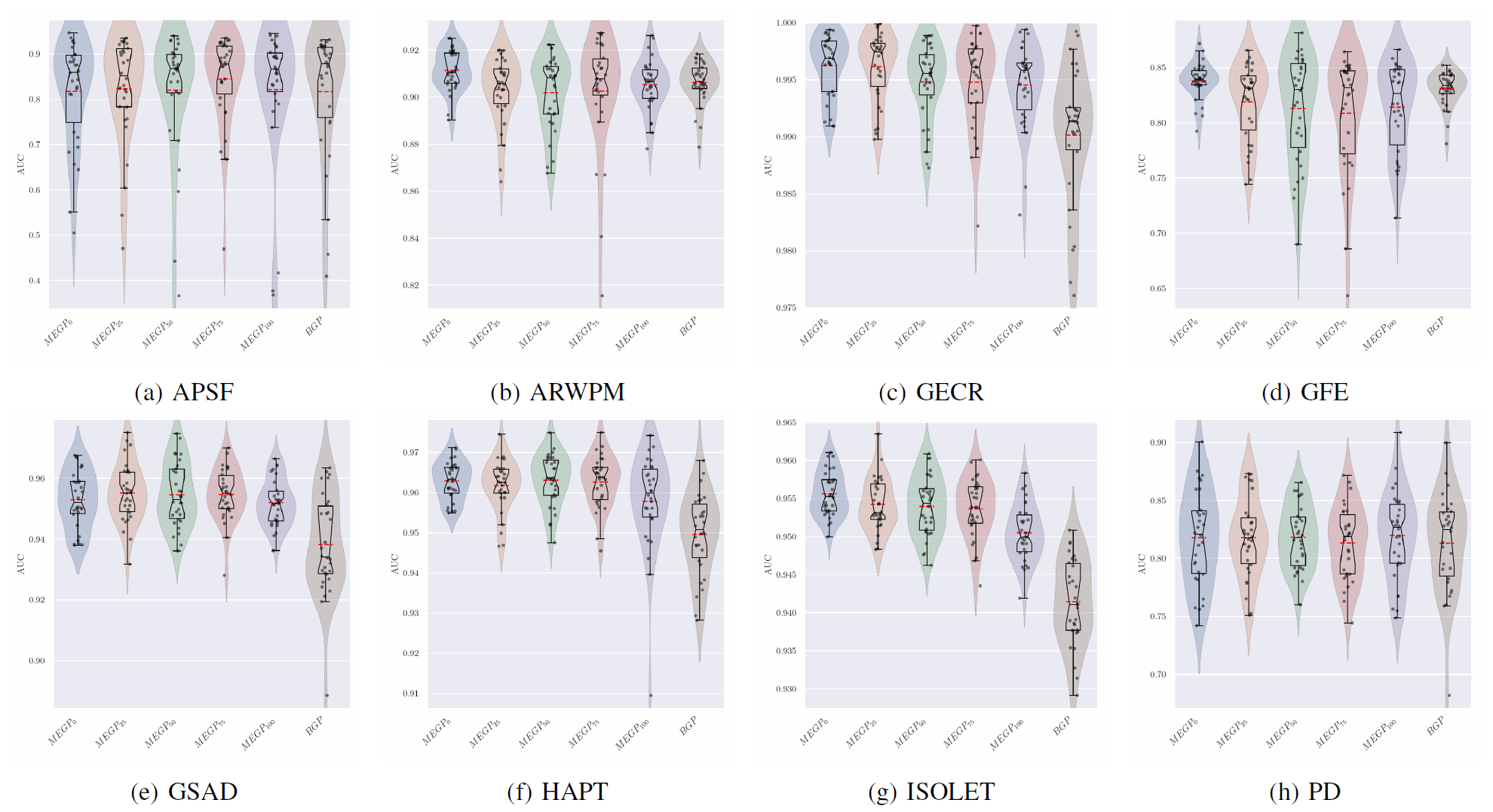}
\caption[The distribution of AUC across BGP and MEGP models over 30 runs.]{Raincloud plots showing the distribution of AUC across 30 runs for the best model obtained from BGP and MEGP models with different ensemble selection probabilities (0\%, 25\%, 50\%, 75\%, 100\%). }

\label{fig:box_auc}
\end{figure*}

%% file: con_auc.tex
\begin{figure*}[ht] 
\centering
\includegraphics[width=\textwidth]{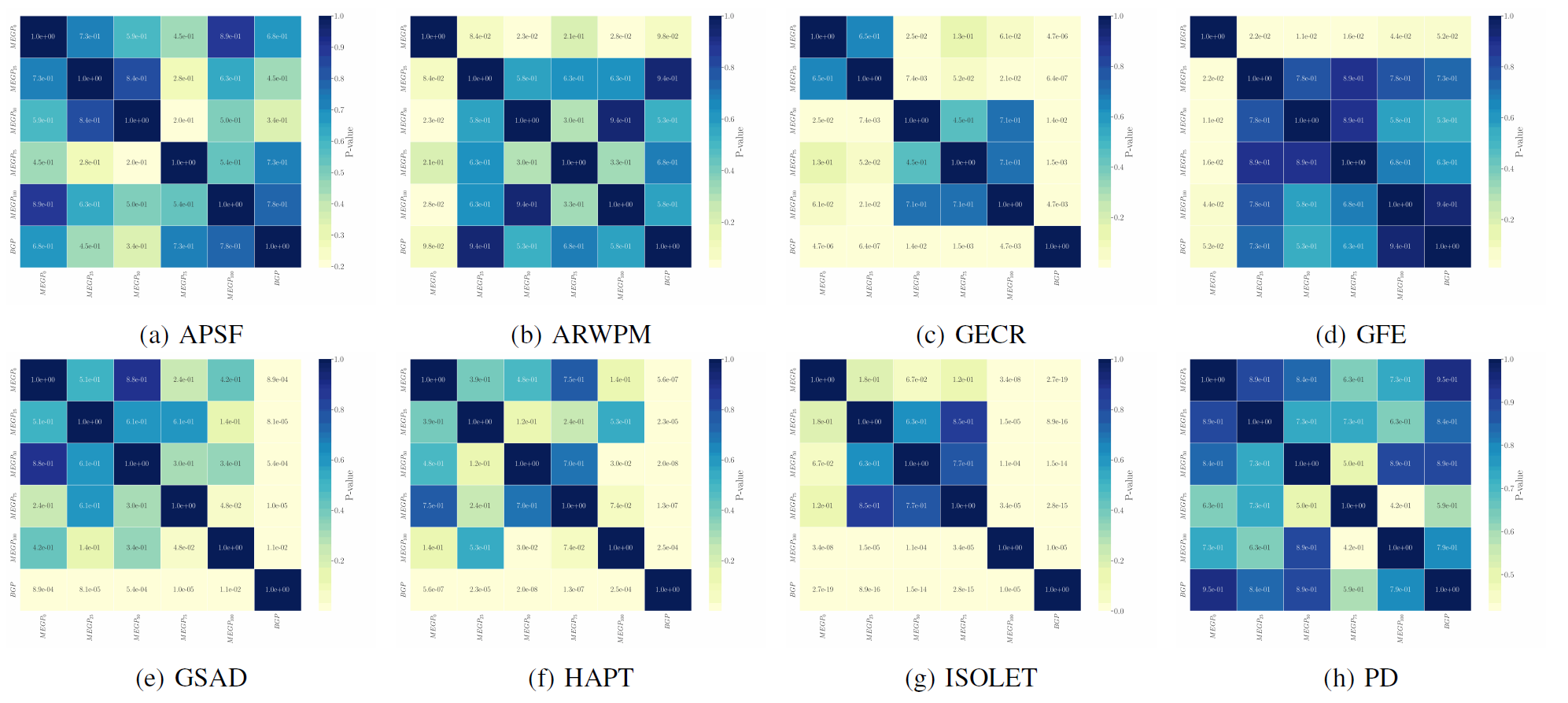}
\caption[Heatmap of adjusted pairwise significance from Conover post-hoc test for AUC.]{Heatmap of adjusted p-values from the pairwise Conover post-hoc test for AUC, corrected using the Benjamini-Hochberg method. The heatmap highlights the statistical significance of pairwise comparisons between BGP and MEGP models with varying ensemble selection probabilities (0\%, 25\%, 50\%, 75\%, 100\%).}

\label{fig:con_auc}
\end{figure*}

%% file: cliff_auc.tex
\begin{figure*}[htbp] 
\centering
\includegraphics[width=\textwidth]{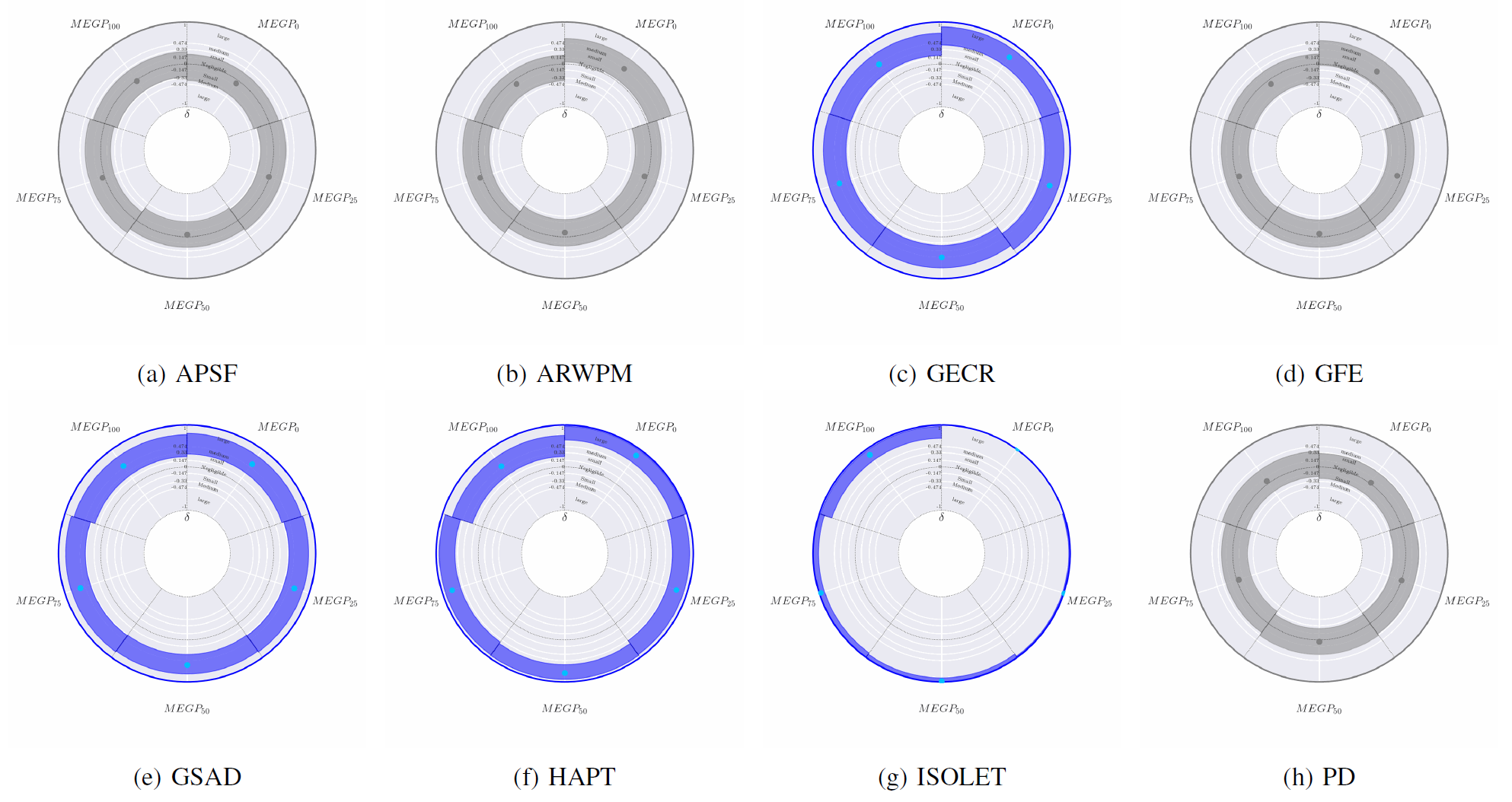}
\caption[The Cliff's $\delta$ effect size measure and its 95\% confidence intervals for AUC obtained from the best model of the 30 BGP and MEGP runs.]{Effect size analysis of AUC across 30 runs for BGP and MEGP models using Cliff's $\delta$. Each point represents the actual AUC value obtained, with segments denoting 95\% confidence intervals based on 10,000 bootstrap resamplings. The outer ring color visualizes statistical significance: grey illustrates no significant difference (adjusted Friedman's P-value $>0.05$), while color indicates significant differences; blue indicates that at least one MEGP configuration outperforms BGP (adjusted Conover's p-value $< 0.05$, Cliff's $\delta > 0$), and red signifies that all MEGP configurations underperform relative to BGP (adjusted Conover's p-value $< 0.05$, Cliff's $\delta < 0$). Segment colors show performance differences against BGP: grey for no significant difference (adjusted Conover's p-value $> 0.05$), blue for better performance (Cliff's $\delta > 0$), and red for worse performance (Cliff's $\delta < 0$).}

\label{fig:cliff_auc}
\end{figure*}

%% file: wtl_auc.tex
\begin{table*}
\centering
\caption[The results of Friedman and Conover tests and Cliff's $\delta$ analysis for the AUC obtained from MEGP and BGP runs.]{Statistical comparison of AUC results for testing data obtained from MEGP and BGP Runs. W, T, and L denote win, tie, and loss based on adjusted Friedman and Conover's p-values. Effect sizes are calculated using Cliff's Delta method and are categorized as negligible, small, medium, or large.}
\label{tab:wtl_auc}
\begin{tabular}{cccccc}
\hline
\multicolumn{6}{c}{AUC} \\
\hline
Dataset & $MEGP_{0}$ & $MEGP_{25}$ & $MEGP_{50}$ & $MEGP_{75}$ & $MEGP_{100}$ \\
\hline
APSF & T (negligible) & T (negligible) & T (negligible) & T (negligible) & T (negligible) \\
ARWPM & T (medium) & T (negligible) & T (negligible) & T (negligible) & T (negligible) \\
GECR & W (large) & W (large) & W (medium) & W (large) & W (medium) \\
GFE & T (small) & T (negligible) & T (negligible) & T (negligible) & T (negligible) \\
GSAD & W (large) & W (large) & W (large) & W (large) & W (large) \\
HAPT & W (large) & W (large) & W (large) & W (large) & W (large) \\
ISOLET & W (large) & W (large) & W (large) & W (large) & W (large) \\
PD & T (negligible) & T (negligible) & T (negligible) & T (negligible) & T (negligible) \\
\hline
W - T - L & 4 - 4 - 0 & 4 - 4 - 0 & 4 - 4 - 0 & 4 - 4 - 0 & 4 - 4 - 0 \\
\hline
\end{tabular}
\end{table*}